\documentclass[10pt,twocolumn,letterpaper]{article}

\usepackage{cvpr}              %

\usepackage{arydshln}

\renewcommand{\paragraph}[1]{\vspace{.5em}\noindent\textbf{#1}}

\definecolor{cvprblue}{rgb}{0.21,0.49,0.74}
\usepackage[pagebackref,breaklinks,colorlinks,allcolors=cvprblue]{hyperref}
\usepackage{multirow}
\usepackage{adjustbox}

\usepackage{xspace}
\makeatletter 
\DeclareRobustCommand\onedot{\futurelet\@let@token\@onedot} 
\def\@onedot{\ifx\@let@token.\else.\null\fi\xspace}  
\def\eg{\emph{e.g}\onedot} 
 
\def\ie{\emph{i.e}\onedot}

\makeatother

\def\paperID{39020} %
\def\confName{CVPR}
\def\confYear{2026}

\title{Lighting-grounded Video Generation with Renderer-based Agent Reasoning}

\author{
Ziqi Cai\textsuperscript{1,2,4}~~~
Taoyu Yang\textsuperscript{1,2}~~~
Zheng Chang\textsuperscript{5}~~~
Si Li\textsuperscript{5}~~~
Han Jiang\textsuperscript{4}~~~
Shuchen Weng\textsuperscript{3,1}~~~
Boxin Shi\textsuperscript{1,2,}\footnotemark
\vspace{2mm}
\\
\parbox{\textwidth}{\centering \small
{\textsuperscript{1}State Key Laboratory for Multimedia Information Processing, School of Computer Science, Peking University}\\
{\textsuperscript{2}National Engineering Research Center of Visual Technology, School of Computer Science, Peking University}\\
{\textsuperscript{3}Beijing Academy of Artificial Intelligence}~~ 
{\textsuperscript{4}OpenBayes Information Technology Co., Ltd.}\\
{\textsuperscript{5}School of Artificial Intelligence, Beijing University of Posts and Telecommunications}
}
\vspace{2.5mm}
\\
{\tt\small \{czq,yangty1031\}@stu.pku.edu.cn},~~
{\tt\small \{zhengchang98,lisi\}@bupt.edu.cn},\\
{\tt\small hahn@openbayes.com},~~
{\tt\small \{shuchenweng,shiboxin\}@pku.edu.cn}
\vspace{-4.5mm}
}

\begin{document}
\twocolumn[{%
\renewcommand\twocolumn[1][]{#1}%
\maketitle
\begin{center}
    \centering
    \captionsetup{type=figure}
    \includegraphics[width=\textwidth]{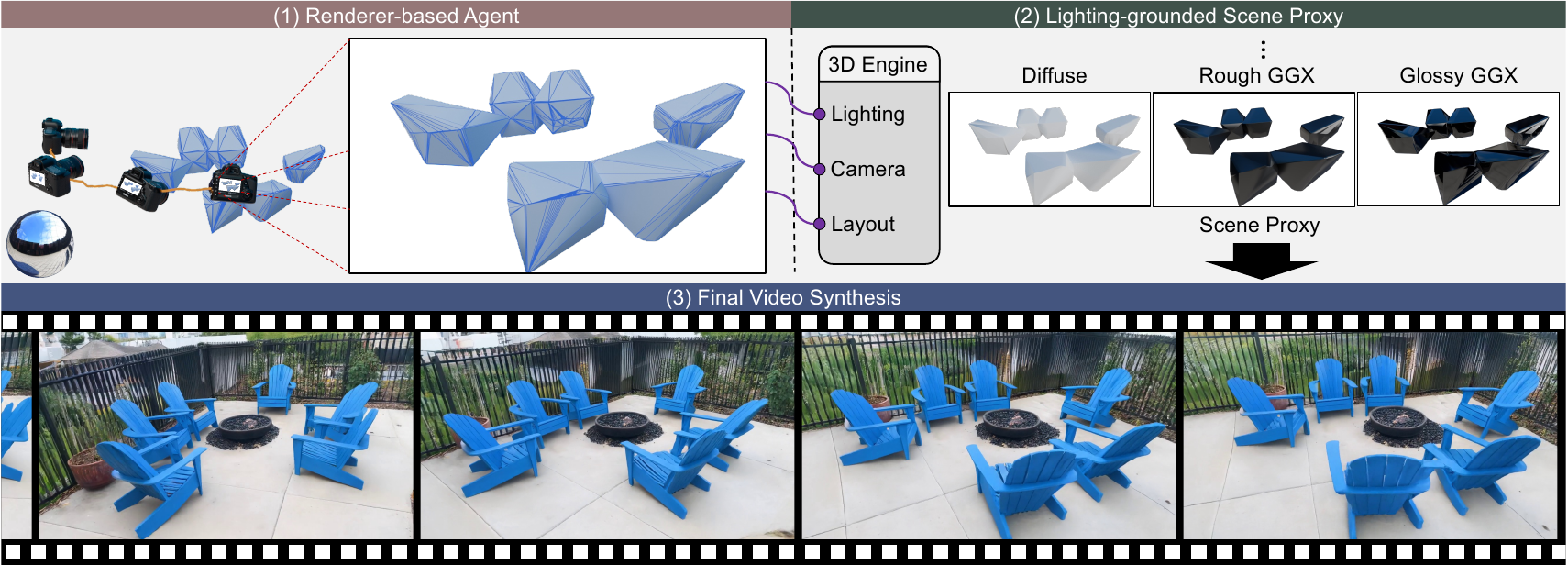}
    \captionof{figure}{Overall framework. (1) A renderer-based agent produces a coarse geometric layout, camera trajectory, and a High Dynamic Range (HDR) environment map. (2) Physically-based rendering generates a lighting-grounded scene proxy containing diffuse, rough, and glossy materials with shading signals. (3) These physical cues are injected into a video diffusion model to synthesize photorealistic sequences with accurate lighting behavior, faithful scene layout, and precisely aligned camera trajectory.}
    \label{fig:teaser}
\end{center}
}]

\maketitle

\renewcommand{\thefootnote}{\fnsymbol{footnote}}
\footnotetext[1]{Corresponding author.}
\renewcommand{\thefootnote}{\arabic{footnote}}
\begin{abstract}
Diffusion models have achieved remarkable progress in video generation, but their controllability remains a major limitation. Key scene factors such as layout, lighting, and camera trajectory are often entangled or only weakly modeled, restricting their applicability in domains like filmmaking and virtual production where explicit scene control is essential. We present LiVER, a diffusion-based framework for scene-controllable video generation. To achieve this, we introduce a novel framework that conditions video synthesis on explicit 3D scene properties, supported by a new large-scale dataset with dense annotations of object layout, lighting, and camera parameters. Our method disentangles these properties by rendering control signals from a unified 3D representation. We propose a lightweight conditioning module and a progressive training strategy to integrate these signals into a foundational video diffusion model, ensuring stable convergence and high fidelity. Our framework enables a wide range of applications, including image-to-video and video-to-video synthesis where the underlying 3D scene is fully editable. To further enhance usability, we develop a scene agent that automatically translates high-level user instructions into the required 3D control signals. Experiments show that LiVER achieves state-of-the-art photorealism and temporal consistency while enabling precise, disentangled control over scene factors, setting a new standard for controllable video generation.
\end{abstract}

\section{Introduction}

Recent video generation models~\cite{blattmann2023stable, yang2024cogvideox, wan2025wan} have demonstrated impressive visual quality, temporal consistency, and diverse scenarios.
Towards physically realistic video generation, researchers have paid great attention on data curation~\cite{gu2025phyworldbench}, prompt enhancement~\cite{thinkbeforediffuse}, and architecture improvement~\cite{waver}. Despite these improvements, these data-driven approaches still struggle to model complex physical interactions (\eg, occlusion relationships between objects in dynamic scenarios).

Introducing grounded references to explicitly model the realistic world has proven to be effective at better controllability in motion~\cite{physctrl}, layout~\cite{kim2025videofrom3d,liu2025sketch3dve}, and camera~\cite{ren2025gen3c,he2025cameractrl,motionprompting}. Although these 3D-aware conditions can provide a strong geometric foundation, existing works largely overlook their potential for computing physically-accurate lighting  (\eg, taking BRDF into consideration). Consequently, mismatching lighting effects (\eg, shadows, reflections, and ambient occlusion) are still produced in realistic material representation (\eg, skin, metals, and glass) in generated videos.

In this paper, we propose \textbf{LiVER}, a framework for \textbf{Li}ghting-grounded \textbf{V}ideo gen\textbf{ER}ation with renderer-based agent reasoning.
As illustrated in \cref{fig:teaser}, LiVER effectively generates videos with diverse and physically realistic lighting effects from text descriptions (\eg, soft daylight). %
We first construct a renderer-based agent framework to retrieve and generate scene lighting, layout, and camera trajectory as a coarse scene representation to guide video generation.
Instead of directly adopting full 3D representations, we represent the scene using a stack of 2D render passes (\eg, diffuse, glossy GGX, and rough GGX) generated by a 3D engine~\cite{blender}.
Formulated as stacked image sequences, this proxy preserves physically realistic lighting cues from the 3D scene while providing the scene layout information.
Leveraging the generative priors of a pretrained text-to-video model (\ie, Wan2.2-5B~\cite{wan2025wan}), we then design a lightweight conditional encoder and adapter to achieve this alignment and translate this proxy into the final visually appealing video.
Finally, to effectively optimize for these lighting effects, we propose a three-stage training scheme to enhance lighting diversity while preserving the base model’s visual quality.

To facilitate model training and evaluation, we collect and render the \textbf{LiVERSet}, a \textbf{Li}ghting-grounded \textbf{V}ideo gen\textbf{ER}ation data\textbf{Set}. 
This dataset comprises two complementary subsets: \textit{(i)} a real-world subset \textbf{LiVER-Real} that captures complex and realistic lighting phenomena and \textit{(ii)} a synthetic subset \textbf{LiVER-Syn} that features diverse and controllable physically-based rendered lighting. 
We provide comprehensive annotations for both subsets, including scene geometry, environment maps, camera poses, and text descriptions. 
In total, \textbf{LiVERSet} contains over 11K videos (81 frames each at $720 \times 1280$ resolution), split into 10K for training and 1K for evaluation.

We summarize our contributions as follows:
\begin{itemize}

\item We propose a framework for lighting-grounded video generation with physically realistic lighting effects, and introduce a lighting-aware dataset with comprehensive annotations for both real-world and synthetic videos.

\item We construct a renderer-based agent to reason a structured scene graph and render the lighting-aware scene proxy and a lightweight encoder as well as adapter to effectively align this proxy with the video latent space.

\item We design a three-stage training scheme to improve lighting diversity and preserve visual quality. Extensive experiments demonstrate our method achieves state-of-the-art performance in physically-accurate lighting phenomena.

\end{itemize}

\section{Related Work}

\subsection{Text-to-Video Generation} 
Recent years have witnessed remarkable progress in Text-to-Video (T2V) generation, largely driven by the success of diffusion models~\cite{ddpm, diffusiontransformer}. Video foundation models~\cite{wan2025wan,hunyuan,yang2024cogvideox} have demonstrated their ability to generate visually appealing and temporally coherent video clips from a single text description. Based on this advancement, researchers have paid attention to long video generation~\cite{causalvid, selfforcing} by introducing causal reasoning, improved video controllability by receiving multi-modal conditions~\cite{vace, FullDiT}, and constructed multimodal datasets and benchmarks~\cite{huang2024vbench, sun2025t2v} to evaluate shortcomings and improve performance with curated data. However, most of these approaches are data-driven, inherently struggling to model complex physical interactions for scenarios with multiple instances. We propose to model physical properties as explicit conditions to guide the model follow underlying physical principles.

\begin{figure*}
    \centering
    \includegraphics[width=1\linewidth]{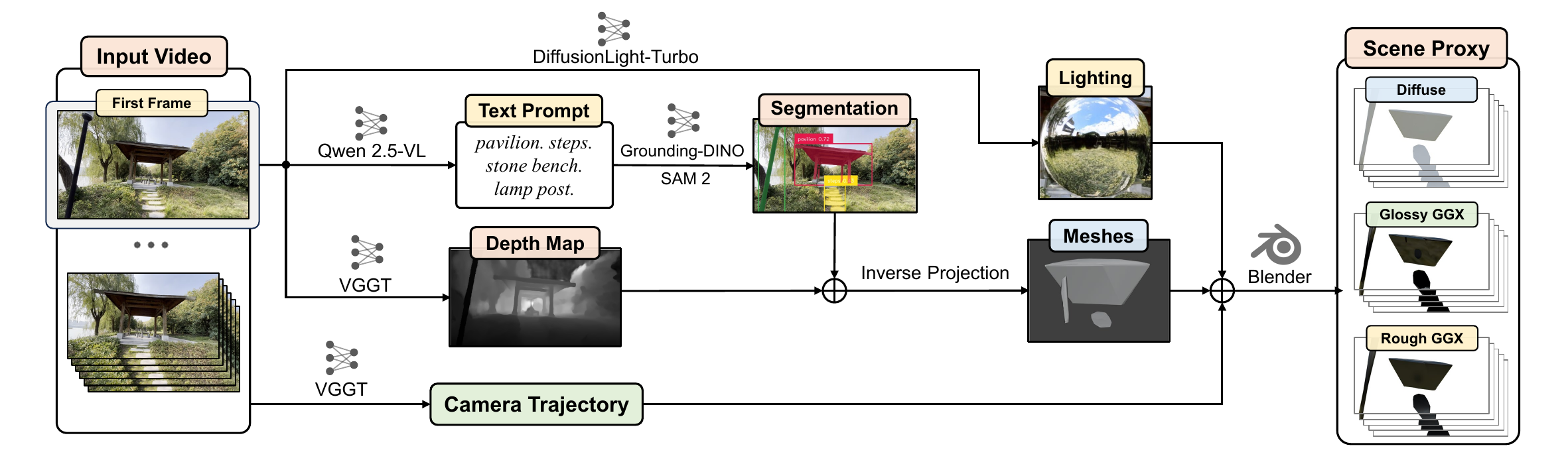}
    \caption{Our data annotation pipeline for \textbf{LiVER-Real}. We process each video to reconstruct its 3D geometry and estimate its HDR environment map. These are then used to render three pixel-aligned lighting representations (Diffuse, Glossy GGX, Rough GGX), which are concatenated to form the final conditioning input.}
    \label{fig:data-pipeline}
\end{figure*}

\subsection{3D-grounded Video Generation} 
A prominent direction for improving physical realism is to ground the generation process in explicit 3D spatial information.

Early efforts used 2D proxies to improve physical realism. For instance, Boximator~\cite{wang2024boximator} and TrailBlazer~\cite{ma2024trailblazer} allow users to define bounding-box trajectories, while methods like Ctrl-V~\cite{luo2025ctrlv} and TrackDiffusion~\cite{li2025trackdiffusion} condition generation on pre-defined object tracklets. MotionPrompting~\cite{motionprompting} employs sparse point trajectories to direct object motion. Recent models have introduced direct trajectory conditioning to model physical properties like camera. CameraCtrl~\cite{he2025cameractrl} uses a plug-in module for explicit camera pose control, while Collaborative Video Diffusion (CVD)~\cite{cvd} synchronizes views along different camera paths using cross-video attention. MotionCtrl~\cite{motionctrl} provides unified control of camera motion and object motion by conditioning on camera poses and sparse object trajectories. More integrated frameworks like CineMaster~\cite{wang2025cinemaster} combine 3D box and camera conditioning. 

While these methods introduce 3D-aware conditions to provide a strong geometric foundation for video generation and improve scene consistency, they largely still ignore the physically-accurate lighting for generated videos, leading to unrealistic artifacts. To fill in this gap, we propose to model lighting as a unified part of physical properties.

\subsection{Image and Video Relighting}
Image and video relighting task aims to modify the illumination conditions of a scene after it has been captured, by extracting and understanding explicit lighting representations~\cite{zhang2025iclight,Cai2024CVPR,Cai2025CVPR,pandey2021total}.

DiLightNet~\cite{zeng2024dilightnet} augments image diffusion models with explicit radiance hints for detailed lighting edits. GenLit~\cite{bharadwaj2024genlit} reframes single-image relighting as a video diffusion task, achieving realistic results. For video, Light-A-Video~\cite{zhou2025light} uses a training-free fusion pipeline for relighting, while LumiSculpt~\cite{zhang2024lumisculpt} introduces a plug-in network to control light properties and motion. However, they often entangle the lighting with other physical properties like camera and scene layout. 

Inspired by these works, we design our LiVER model to integrate lighting as the primary condition, rendered from the 3D scene proxy to generate general videos with physically accurate lighting while preserving controllability over scene layout and camera.

\begin{figure*}
    \centering
    \includegraphics[width=1\linewidth]{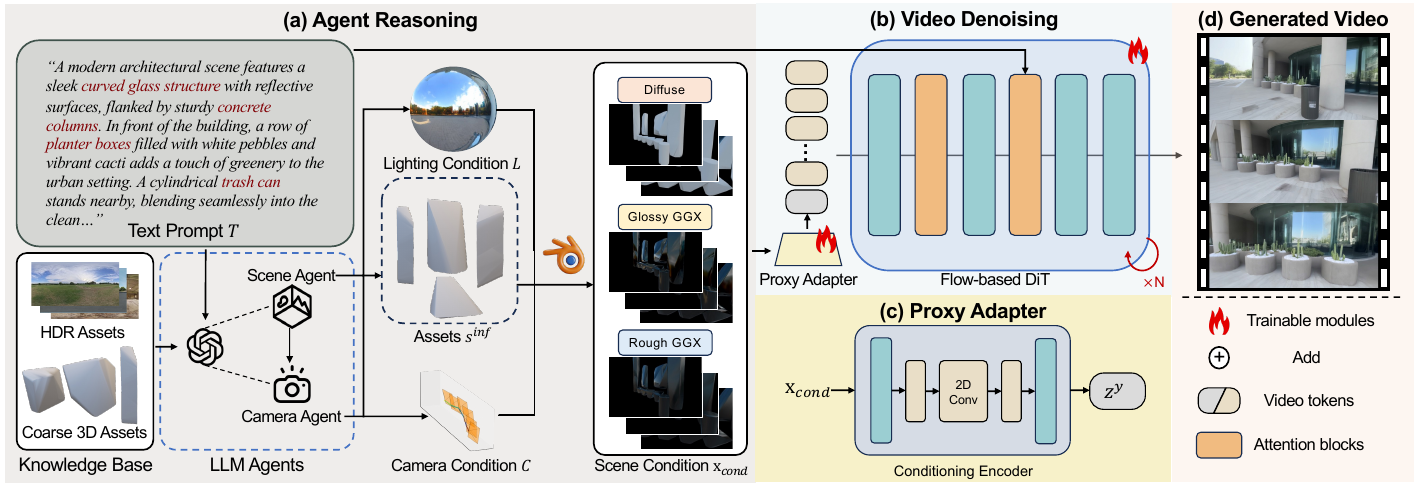}
    \caption{
    Pipeline of LiVER. Given a text prompt $T$, our Scene Agent parses object categories, spatial relations, and coarse geometry to construct an initial 3D scene. The Camera Agent infers a camera trajectory consistent with the described viewpoint and scene semantics, producing the camera condition $C$. The 3D scene is then rendered through a physically-based renderer to obtain the lighting-grounded scene proxy, including diffuse, glossy GGX, and rough GGX components, forming the scene condition $X_{\text{cond}}$. These components encode physically meaningful cues such as material response, shading, and reflections. The Lighting condition $L$, represented by an environment map, provides global illumination cues. The DiT-based video diffusion model integrates all conditions $X_{\text{cond}}$, $C$, $L$ and generates a photorealistic video that preserves the scene layout, camera trajectory, and physically accurate lighting behavior.}
    \label{fig:pipeline}
\end{figure*}

\section{Dataset}

We now describe the proposed \textbf{LiVERSet} in detail. The dataset contains two complementary components: \textit{(i)} \textbf{LiVER-Real} contains real videos exhibiting complex, naturally occurring lighting together with detailed physical annotations. \textit{(ii)} \textbf{LiVER-Syn} is a physically based rendered collection with dynamic illumination, providing broader lighting variability than what is captured in the real data. For both subsets, we derive a unified scene proxy as the primary control condition for video generation, paired with text descriptions to provide semantic guidance.

\paragraph{LiVER-Real Video Annotation.}
Existing real-world videos $x^\mathrm{real}$ lack the explicit physical annotations required for precise lighting control. Therefore, we develop an annotation pipeline to reconstruct a dynamic 3D scene and global lighting from video, as shown in \cref{fig:data-pipeline}. We first estimate per-frame camera poses $c^\mathrm{real}$ using VGGT~\cite{wang2025vggt}, and extract only the first-frame depth map from the same model, complemented by an initial-frame object segmentation from Grounding-DINO~\cite{liu2023gdino} and SAM 2~\cite{ravi2024sam}. This combination allows us to lift the 2D segmented objects into a coarse 3D scene mesh $s^\mathrm{real}$. For lighting modeling, we employ DiffusionLight-Turbo~\cite{chinchuthakun2025diffusionlightturbo} to estimate a single HDR environment map, which serves as a robust approximation of the scene's global lighting representation $l^\mathrm{real}$.

\paragraph{LiVER-Syn Data Rendering.}
To supplement the diverse lighting phenomena limited in real-world data and to learn fine-grained control, we construct a complementary synthetic dataset.
We first carefully curate a subset of Objaverse-XL~\cite{objaverseXL} to include high-quality PBR materials. A 3D scene $s^\mathrm{syn}$ is then procedurally generated by randomly sampling several objects from this subset.
To ensure the diverse lighting representation $l^\mathrm{syn}$, we illuminate scenes using a diverse set of HDR environment maps from the Poly Haven~\cite{polyhaven} library.
We then introduce dynamic effects by rotating the HDR environment map over the video's duration horizontally around the scene's vertical axis (\ie, yaw rotation). The total rotation angle for each clip is sampled uniformly from the range $[180^\circ,240^\circ]$ to ensure a visually significant change in lighting direction (\eg, sun moves to the opposite side of the scene). By procedurally moving the camera position $c^\mathrm{syn}$, we render the final photorealistic target video $x^\mathrm{syn}$.

\paragraph{Scene Proxy Construction.}
The scene proxy is designed to provide realistic lighting cues by decomposing the scene's complex illumination into fundamental lighting components and providing scene layout information.
    For the two subsets, we use a physically-based renderer (Blender~\cite{blender} in our implementation) to render a scene proxy $y \in \mathbb{R}^{F\times 9 \times H \times W}$ based on the 3D scene mesh $s^i$, lighting representation $l^i$, and camera trajectory $c^i$:
\begin{equation}
y=[x^\mathrm{DIFF}, x^\mathrm{GGX1}, x^\mathrm{GGX2}]=R(s^i,l^i,c^i),
\label{eq:condition}
\end{equation}
where $i \in \{\mathrm{real},\mathrm{syn}\} $, $F$ is the number of frames and $R$ is the renderer~\cite{blender}. The scene proxy is a stack of 2D render passes, including the purely diffuse $x^\mathrm{DIFF}$, rough GGX (high roughness) $x^\mathrm{GGX1}$, and glossy GGX (low roughness) $x^\mathrm{GGX2}$ to capture low-frequency ambient lighting, medium-frequency broad reflections, and high-frequency specular highlights, respectively.

\paragraph{Caption Annotation.} 
To provide text descriptions to guide the scene semantics, we use Qwen 2.5-VL~\cite{qwen25vl} as the vision-language model to generate a caption for each video clip. Example prompts used for caption generation are detailed in the supplementary material.

\paragraph{Dataset Statistics.}
Our final composite dataset comprises approximately 11K video clips, split into a 10K training set and a 1K evaluation set, both equally split between real-world and synthetic data. Each video has a duration of 81 frames at a $720 \times 1280$ resolution.

\section{Methodology}
In this section, we present the details of our LiVER model. 
We begin by introducing the renderer-based agent reasoning, which interprets a user-provided text description into an explicit scene proxy (\cref{subsec:agentic_reasoning}). 
We then detail the approach to use this scene proxy to guide the lighting-grounded video generation, which includes the task formulation, the scene proxy encoder, and the adapter (\cref{subsec:video_generation}). 
Finally, we present the stage-wise training scheme designed to effectively train these components, optimizing for proxy translation, lighting control, and lighting diversity (\cref{subsec:training}).

\subsection{Renderer-based Agent Reasoning}
\label{subsec:agentic_reasoning}

To bridge high-level user intent with structured lighting control, we design an intelligent renderer-based agent to translate text descriptions into the scene proxy. As shown in \cref{fig:pipeline}, this process involves three components: scene building, lighting setup, and camera planning.

\paragraph{Scene Building.}
Given a text description, the agent first performs semantic decomposition to parse object categories and their spatial relationships, and organizes them into a structured scene graph $\mathcal{G} = (V, E)$.
In this graph, each node $v_i \in V$ is an instantiated object that encapsulates semantic categories and materials properties, while an edge $e_{i,j} \in E$ represents the spatial relationships between the $i$-th and $j$-th objects (\eg, ``in front of").
Then, the agent retrieves a suitable mesh asset from our curated library~\cite{objaverseXL} for each node $v_i$, and optimizes their poses to satisfy the relational constraints defined in $E$.

\paragraph{Lighting Setup.}
After constructing the scene geometry $s^\mathrm{geo}$ based on the scene graph, the agent parses the original text description for lighting cues (\eg, ``warm mood" and ``overcast sky") and selects an appropriate HDR environment map from the Poly Haven library~\cite{polyhaven} to configure a physically plausible illumination setup $l^\mathrm{inf}$ that matches the described mood.
When there is no suitable environment map, the agent generates one using pretrained generation models~\cite{wang2022stylelight}.

\paragraph{Camera Planning.} 
Given the static scene graph $\mathcal{G}$, the camera planner generates a dynamic camera trajectory $c^\mathrm{inf}$ for $F$ frames. It first parses cinematographic hints from the text description (\eg, ``orbit", ``dolly zoom", and ``crane shot") to establish a camera motion plan. This defines a set of keyframe poses to specify the camera's position and orientation. A temporally smooth trajectory $c^\mathrm{inf}$ is then generated by interpolating these keyframes using a spline.

\paragraph{Proxy Rendering.}
Finally, the agent assembles reasoned assets $[s^\mathrm{inf},l^\mathrm{inf},c^\mathrm{inf}]$ as the final scene representation. This representation is then fed into renderer to render the 2D scene proxy $y$ as \cref{eq:condition}, enabling physically consistent scene authoring for lighting-grounded video generation.

\subsection{Lighting-grounded Video Generation} 
\label{subsec:video_generation}
To translate the scene proxy into videos with physically realistic lighting, our LiVER model leverages the generative priors of a pretrained video model. We additionally introduce a proxy encoder to capture lighting cues and an adapter to align proxy tokens with the video latent space.

\paragraph{Video Generation Backbone.}
We build our video generation model upon Wan2.2-5B~\cite{wan2025wan} to leverage its priors for realistic video generation.
We use a spatiotemporal Variational Autoencoder (VAE) to map high-dimensional videos $x$ into a compact latent space as $z = \mathcal{E}(x)$.
During training, we follow the flow matching formulation~\cite{lipman2022flow} to sample a Gaussian noise $\epsilon \sim \mathcal{N}(0, I)$. This noise is then linearly interpolated with the latent code $z_t = t z + (1 - t) \epsilon$ for a random timestep $t \in [0, 1]$.
The model is trained to predict the ground truth velocity vector $v_t = \frac{\mathrm{d} z_t}{\mathrm{d}t} = z - \epsilon$:
\begin{equation}
\mathcal{L} = \mathbb{E}_{z, \epsilon, t} \left[ \left| u_\theta(z_t, y,c^\text{txt}, t) - v_t \right|^2 \right],
\end{equation}
where $y$ is the scene proxy, $c^\text{txt}$ is the text embedding of the caption, and $u_\theta$ is our LiVER model.

\paragraph{Scene Proxy Encoding.}
As the scene proxy is organized into a stack of 2D render passes $y\in \mathbb{R}^{F \times 9 \times H \times W}$, we can simply employ a lightweight 2D proxy encoder $\mathcal{E}_\text{proxy}$ to map it into a compact feature representation.
Architecturally, the proxy encoder is implemented by multiple 2D convolutional blocks, each containing a convolution, a GroupNorm~\cite{groupnorm} layer, and a SiLU~\cite{silu} non-linearity. 
The network progressively downsamples the spatial dimensions while mapping the input to higher-dimensional features $z^\mathrm{y}=\mathcal{E}_\text{proxy}(y)$ for each frame, where $z^y \in \mathbb{R}^{F \times C \times H' \times W'}$, $H',W' = H/16, W/16$ are the downsampled resolution, and $C$ is the dimensions of the features.
This design minimizes computational overhead while capturing lighting cues and providing the information about the scene.

\paragraph{Video Latent Integration.}
To ensure the video latent code semantically aligns with the scene proxy, we design a lightweight conditioning encoder to inject the encoded scene proxy features $z^\mathrm{y}$ directly into the video latent space. 
Specifically, we first stack the multiple RGB rendered images along the channel dimension to form a 9-channel input. Instead of using complex 3D convolutions, we employ a 2D convolutional network with a sequence of downsampling blocks to extract spatial features $z^\mathrm{y} \in \mathbb{R}^{F \times C \times H' \times W'}$. This explicitly aligns the spatial resolution and channel dimensions of the proxy features with the VAE latent code $z \in \mathbb{R}^{C \times H' \times W'}$ of the video, where $C=4$.
To guide the video generation direction, the encoder learns a spatial residual that is directly superimposed onto the video latent code. These encoded proxy features $z^\mathrm{y}$ are used to modulate the original video latents $z$:
\begin{equation}
z' = z + \alpha \cdot z^\mathrm{y},
\label{eq:residual_injection}
\end{equation}
where $z^\mathrm{y}$ is the output of the 2D conditioning encoder, and $\alpha$ is a learnable scalar weight initialized to zero. This strategy ensures the conditioning encoder has no initial impact on video generation at the start of training, allowing the proxy features to gradually guide the video latent space and ultimately enable lighting-grounded precise control.

\subsection{Stage-wise Training Scheme} \label{subsec:training} 
To effectively learn the conditioning pathway while preserving the video backbone's generative priors, we adopt a three-stage training scheme to optimize for proxy translation, lighting control, and lighting diversity.  

\paragraph{Conditional Pathway Training.} 
We first freeze the entire video diffusion backbone and train only the proxy encoder and adapter modules for 10 epochs. This initial stage aims to translate the scene proxy into coarse control signals over the generation process.

\paragraph{Joint LoRA Fine-tuning.} 
We unfreeze LoRA~\cite{lora} layers integrated into the video backbone, and jointly fine-tune these layers along with the proxy encoder and adapter for another 10 epochs. This stage refines the semantic alignment, effectively balancing proxy controllability and overall visual quality.

\paragraph{Lighting Diversity Expansion.} 
We continue the joint LoRA fine-tuning while mixing real videos with our synthetic data in a $1:1$ ratio. This final stage adapts the model to more general scenarios and enhances its ability to render diverse lighting phenomena.

\section{Experiments}

\subsection{Implementation Details}
\label{subsec:implementation}
Our model is built upon the Wan 2.2-5B-TI2V checkpoint~\cite{wan2025wan}, inheriting its robust generative priors. We use LoRA~\cite{lora} to reduce computational costs and prevent catastrophic forgetting of the base model's capabilities.

Training is conducted on our curated dataset for approximately 100K steps. We use 8 NVIDIA H100 GPUs with a per-GPU batch size of 2, resulting in a total batch size of 16. We employ the AdamW optimizer~\cite{adamw} with a constant learning rate of $1 \times 10^{-5}$. The model generates videos at a resolution of $704\times1280$. Further details on the network architecture and training hyperparameters are provided in the supplementary material.

\subsection{Baselines}

We conduct a comprehensive quantitative and qualitative comparison against several state-of-the-art methods capable of generating video from 3D-aware conditions:
\begin{itemize}
    \item CameraCtrl~\cite{he2025cameractrl}: Controls video generation by conditioning on explicit camera pose sequences to enforce camera-consistent motion.
    \item MotionCtrl~\cite{motionctrl}: Controls both camera and object motion by conditioning the diffusion model on camera poses and sparse object trajectories through lightweight temporal and spatial motion modules.
    \item VideoFrom3D~\cite{kim2025videofrom3d}: Generates 3D scene videos from coarse geometry by producing anchor views with an image diffusion model and interpolating them via a video diffusion model.
\end{itemize}
For a fair comparison, all methods are evaluated on a held-out test set derived from our dataset. Since CameraCtrl and MotionCtrl only generate 16-frame clips, we compare ours against these methods using the first 16 frames of our generated videos. For VideoFrom3D, comparisons follow our 81-frame evaluation protocol, and thus both methods are evaluated using full 81 frames. 
Since VideoFrom3D requires training a style-specific LoRA for each test sample ($\sim 40$ minutes on an NVIDIA H100 GPU), we train and evaluate VideoFrom3D on only 20 videos randomly selected from our test set (\ie, 20 LoRAs). Consequently, the quantitative results reported for VideoFrom3D are based on this 20-sample subset.

\subsection{Evaluation Metrics}
We assess performance using a suite of standard metrics targeting different aspects of video generation:

\paragraph{Quality \& Realism.} We use Fréchet Video Distance (FVD)~\cite{fvd} to assess distributional similarity between generated and real videos, and per-frame Fréchet Inception Distance (FID)~\cite{fid} to evaluate image quality. For semantic consistency with the text prompts, we report CLIP image–text similarity computed with the pretrained CLIP model~\cite{clip}.

\paragraph{Control Fidelity.} To evaluate how faithfully the models adhere to the input conditions, we measure:
    \begin{itemize}
        \item \textbf{Camera Pose Error:} We estimate the camera poses from the generated videos using VGGT~\cite{wang2025vggt}, align each predicted trajectory to the ground truth via a global Sim(3), and report absolute trajectory error (ATE), mean per-step translation error (RPEt) and mean per-step rotation error (RPEr). For readability, ATE and RPEt are scaled by $\times 100$, while RPEr is in degrees.
        \item \textbf{Lighting Error:} We use an existing lighting estimator~\cite{chinchuthakun2025diffusionlightturbo} to recover HDR environment maps from the generated video frames. We then compute the scale-invariant mean squared error (SI-MSE) between the predicted and ground-truth lighting. This metric provides both an overall lighting error and a lighting instability measure, defined as the standard deviation of the SI-MSE over time.
        \item \textbf{Layout Error:} Frame-wise object masks are obtained via a segmentation model~\cite{ravi2024sam}, and compared to ground-truth masks using mean Intersection-over-Union (mIoU) to assess how accurately the generated videos preserve scene layout and object shapes.
    \end{itemize}

\subsection{Quantitative Comparison}
We evaluate our approach against publicly available baseline models on the test split of LiVER-Real, which is composed of original videos sourced from the public dataset of Ling et al.~\cite{ling2024dl3dv}.
Table~\ref{tab:quantitative_results} summarizes the quantitative results. Our method attains the lowest FVD and FID scores and the highest CLIP score, indicating improvements in both video quality and overall realism. In addition, our model achieves the highest control fidelity, exhibiting reduced camera pose and lighting errors. It also delivers the highest mIoU, demonstrating more accurate preservation of object shapes and spatial structure throughout the sequence.

\begin{table*}[h]
    \centering
    \caption{Quantitative comparison with state-of-the-art methods. Our method consistently outperforms the baselines.  $^\dagger$Only compare first 16 frames.}
    \label{tab:quantitative_results}
    \setlength\tabcolsep{16pt}
    \resizebox{\linewidth}{!}{
        \begin{tabular}{l|ccc|ccc|cc|c}
            \toprule
            Method & FVD $\downarrow$ & FID $\downarrow$ &  CLIP $\uparrow$ & ATE $\downarrow$ & RPEt $\downarrow$ & RPEr $\downarrow$ & LE $\downarrow$ & LI $\downarrow$ & mIoU $\uparrow$\\
            \hline
            
            CameraCtrl~\cite{he2025cameractrl} & 48.03& 98.29 & 28.75 & 2.15 & 1.39 & 1.68 & 0.06 & 0.03 & 0.68 \\
            MotionCtrl~\cite{motionctrl} & 63.13 & 97.21 & 26.67 & 3.42 & 2.03 & 7.32  & 0.07 & 0.04 & 0.66 \\
            \textbf{Ours$^\dagger$} & \textbf{32.45} & \textbf{42.32} & \textbf{29.62} & \textbf{1.30} & \textbf{0.81} & \textbf{1.16} &\textbf{0.05} & \textbf{0.02} & \textbf{0.86} \\
            
            \hline
            VideoFrom3D~\cite{kim2025videofrom3d} & 36.94 & 157.89 & 24.51 & 17.55 & 3.85 & 3.12 & 0.05 & 0.03 &0.74  \\
            \textbf{Ours} & \textbf{32.56} & \textbf{129.56} & \textbf{30.97} & \textbf{2.48} & \textbf{0.71} & \textbf{0.50} & \textbf{0.04} &  \textbf{0.02} & \textbf{0.87}\\
            \bottomrule
        \end{tabular}
    }
    \vspace{-2.5mm}
\end{table*}

\newcommand{\lframe}[1]{%
  \includegraphics[width=\lw,height=\lh]{#1}\hspace{2pt}%
}

\newcommand{\rframe}[1]{%
  \hspace{2pt}\includegraphics[width=\lw,height=\lh]{#1}%
}

\newcommand{\rowtop}{\rule{0pt}{0.08\linewidth}}

\begin{figure*}[t] 
\def\lw{0.099\linewidth}
\def\lh{0.05445\linewidth}
\def\hlw{0.05\linewidth}
\def\ftsz{\normalsize}
\def\sskip{1pt}
\renewcommand\tabcolsep{0.0pt}
\renewcommand{\arraystretch}{0}
\centering \small
\begin{tabular}{c ccccc@{}: @{}ccccc}
\rowtop\rotatebox{90}{\footnotesize{CC~\cite{he2025cameractrl}}}\hspace{1mm} &
\includegraphics[width=\lw,height=\lh]{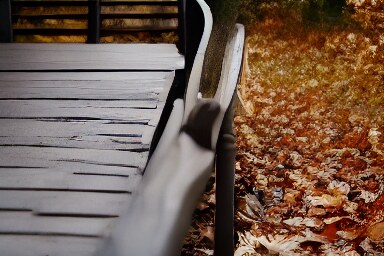} & 
\includegraphics[width=\lw,height=\lh]{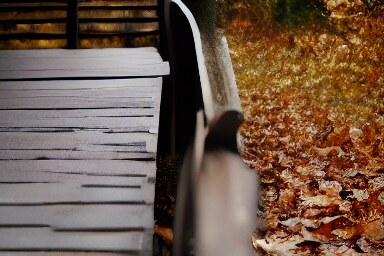} & 
\includegraphics[width=\lw,height=\lh]{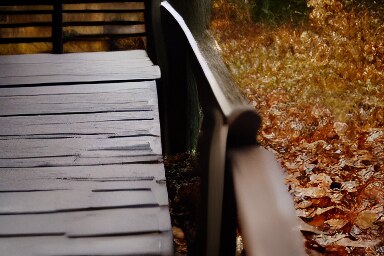} & 
\includegraphics[width=\lw,height=\lh]{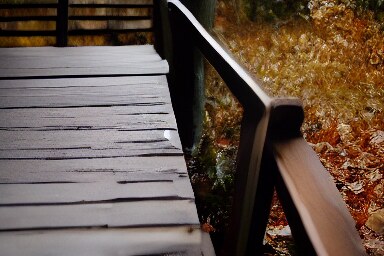} & 
\lframe{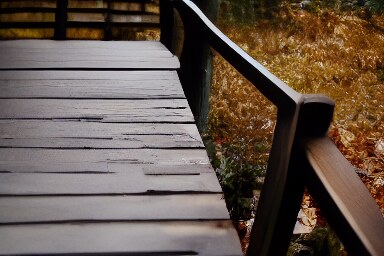} &
\rframe{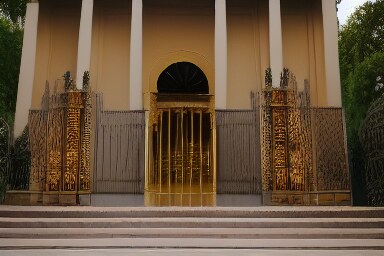} & 
\includegraphics[width=\lw,height=\lh]{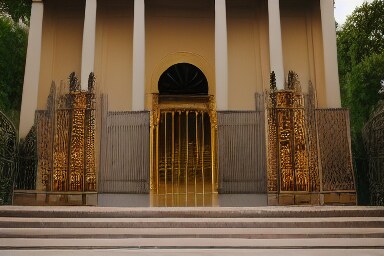} & 
\includegraphics[width=\lw,height=\lh]{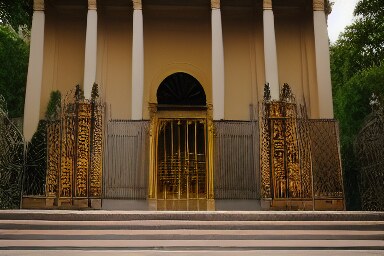} & 
\includegraphics[width=\lw,height=\lh]{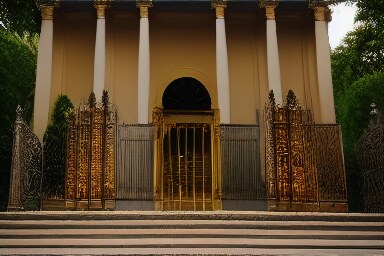} & 
\includegraphics[width=\lw,height=\lh]{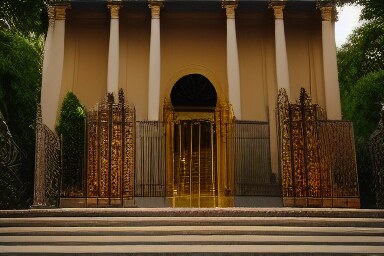} \\ \noalign{\vskip \sskip}
\rotatebox{90}{\footnotesize{MC~\cite{motionctrl}}}\hspace{1mm} &
\includegraphics[width=\lw,height=\lh]{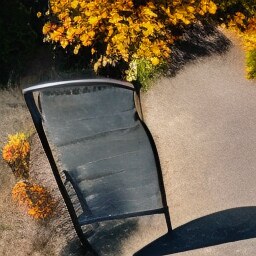} & 
\includegraphics[width=\lw,height=\lh]{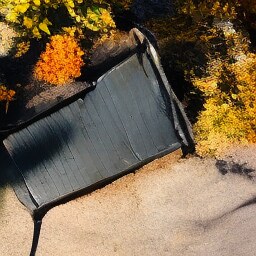} & 
\includegraphics[width=\lw,height=\lh]{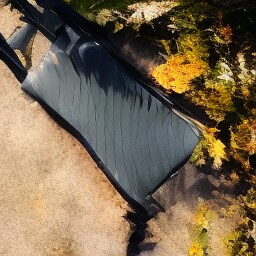} & 
\includegraphics[width=\lw,height=\lh]{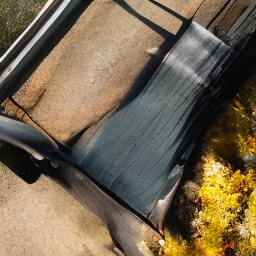} & 
\lframe{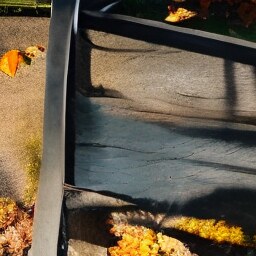} &
\rframe{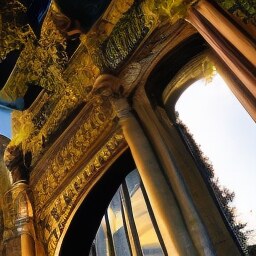} & 
\includegraphics[width=\lw,height=\lh]{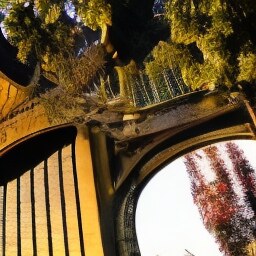} & 
\includegraphics[width=\lw,height=\lh]{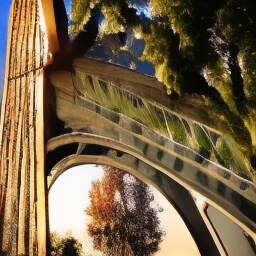} & 
\includegraphics[width=\lw,height=\lh]{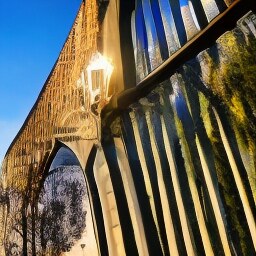} & 
\includegraphics[width=\lw,height=\lh]{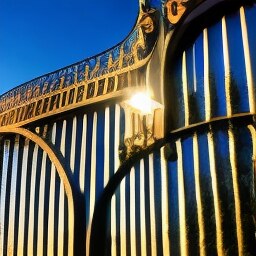} \\ \noalign{\vskip \sskip}
\rotatebox{90}{\footnotesize{VD~\cite{kim2025videofrom3d}}}\hspace{1mm} &
\includegraphics[width=\lw,height=\lh]{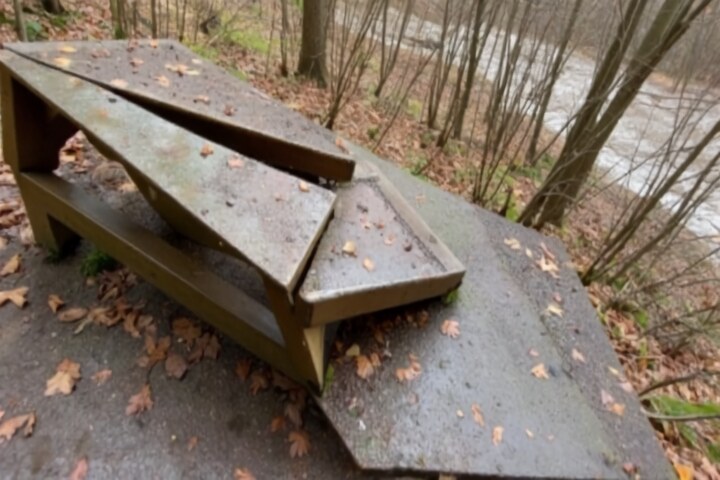} & 
\includegraphics[width=\lw,height=\lh]{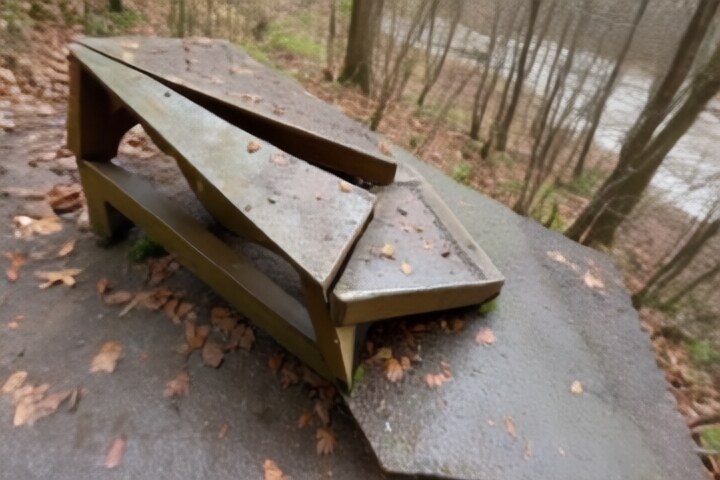} & 
\includegraphics[width=\lw,height=\lh]{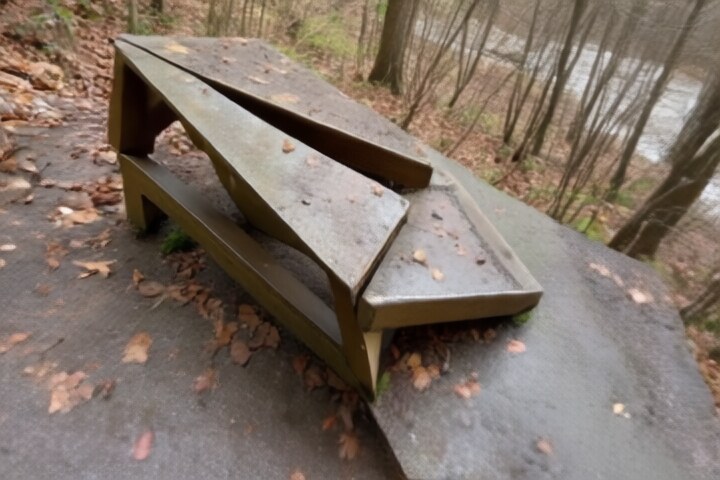} & 
\includegraphics[width=\lw,height=\lh]{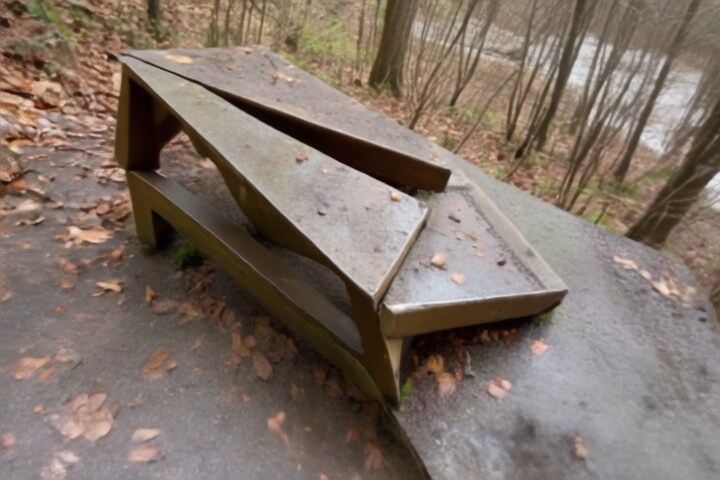} & 
\lframe{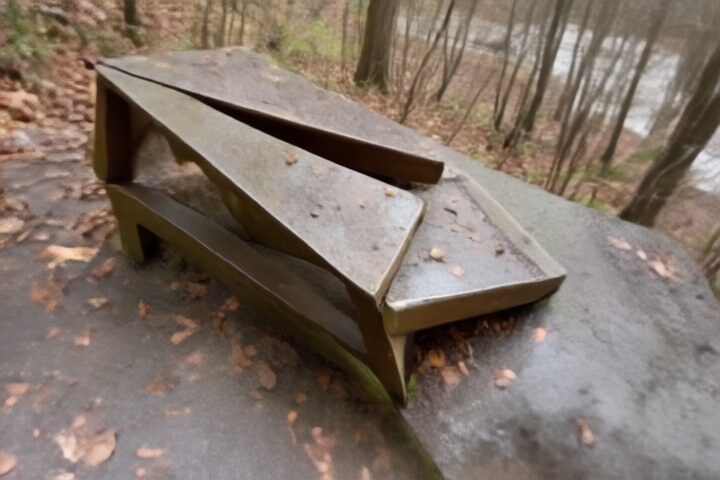} &
\rframe{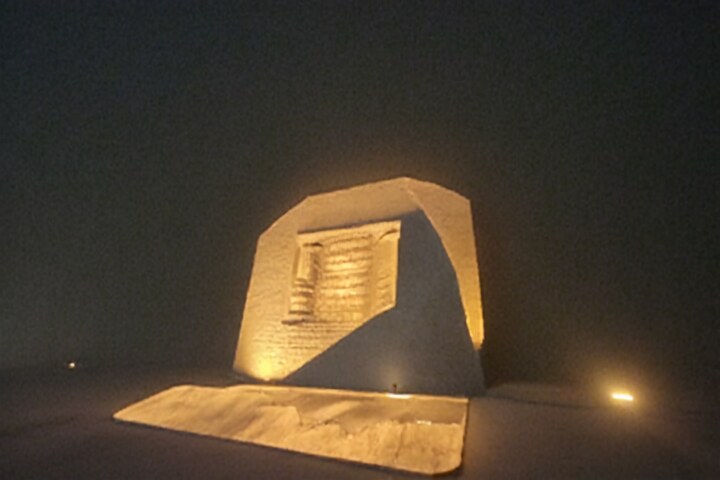} & 
\includegraphics[width=\lw,height=\lh]{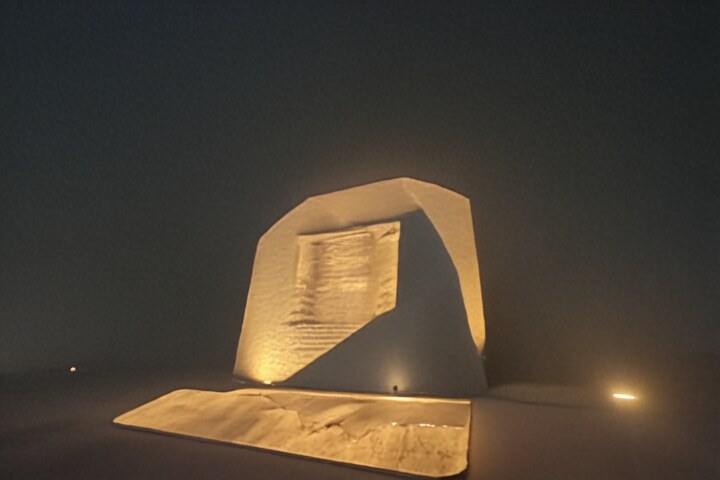} & 
\includegraphics[width=\lw,height=\lh]{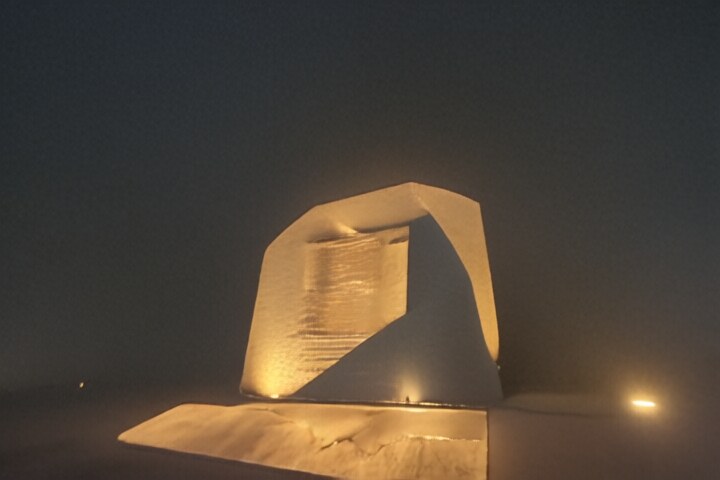} & 
\includegraphics[width=\lw,height=\lh]{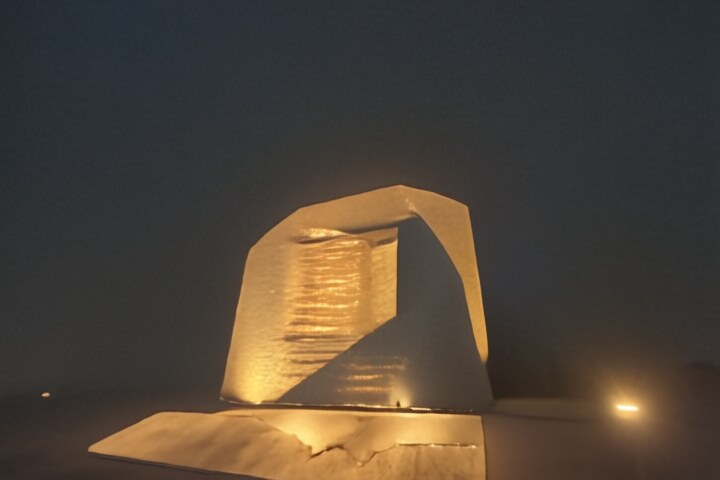} & 
\includegraphics[width=\lw,height=\lh]{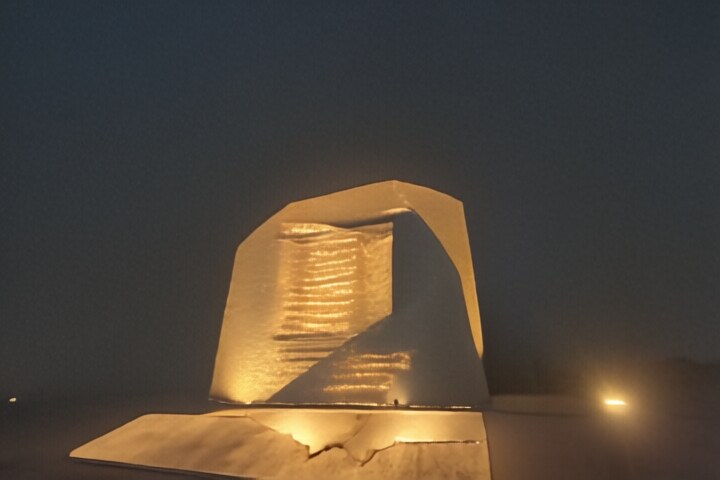} \\ \noalign{\vskip \sskip}
\rotatebox{90}{\hspace{2mm}\footnotesize{Ours}}\hspace{1mm} &
\includegraphics[width=\lw,height=\lh]{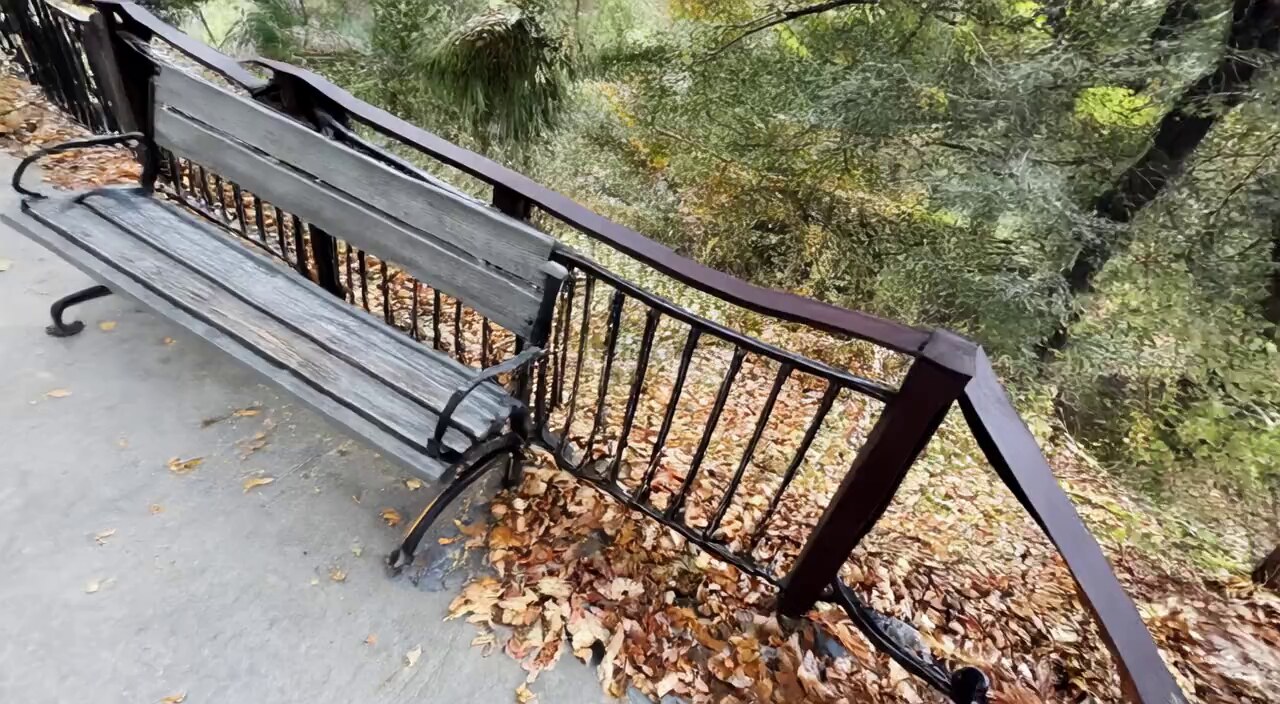} & 
\includegraphics[width=\lw,height=\lh]{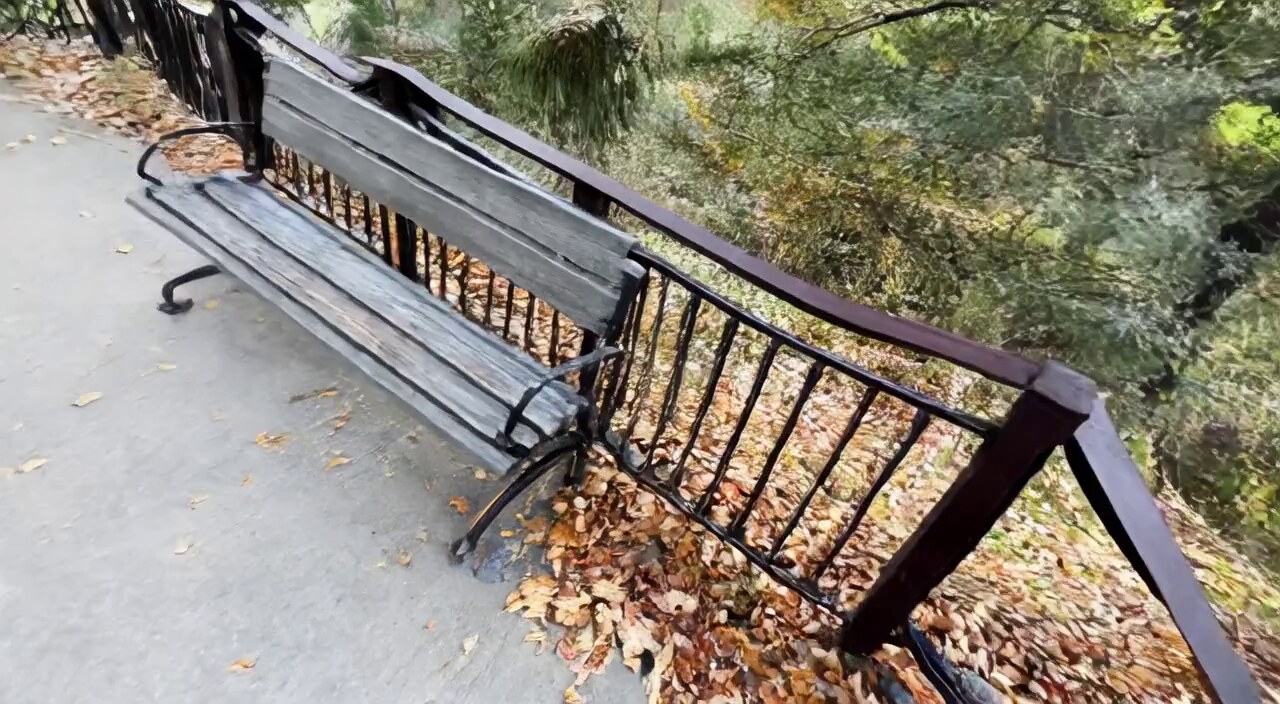} & 
\includegraphics[width=\lw,height=\lh]{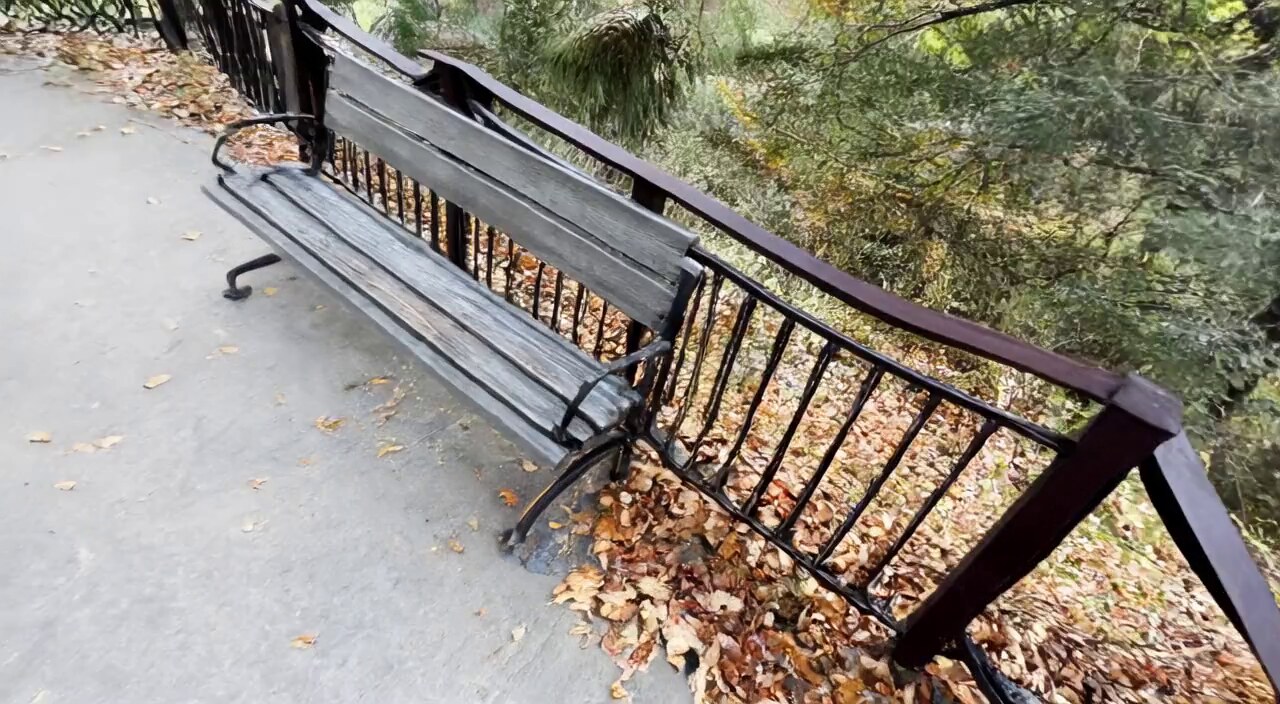} & 
\includegraphics[width=\lw,height=\lh]{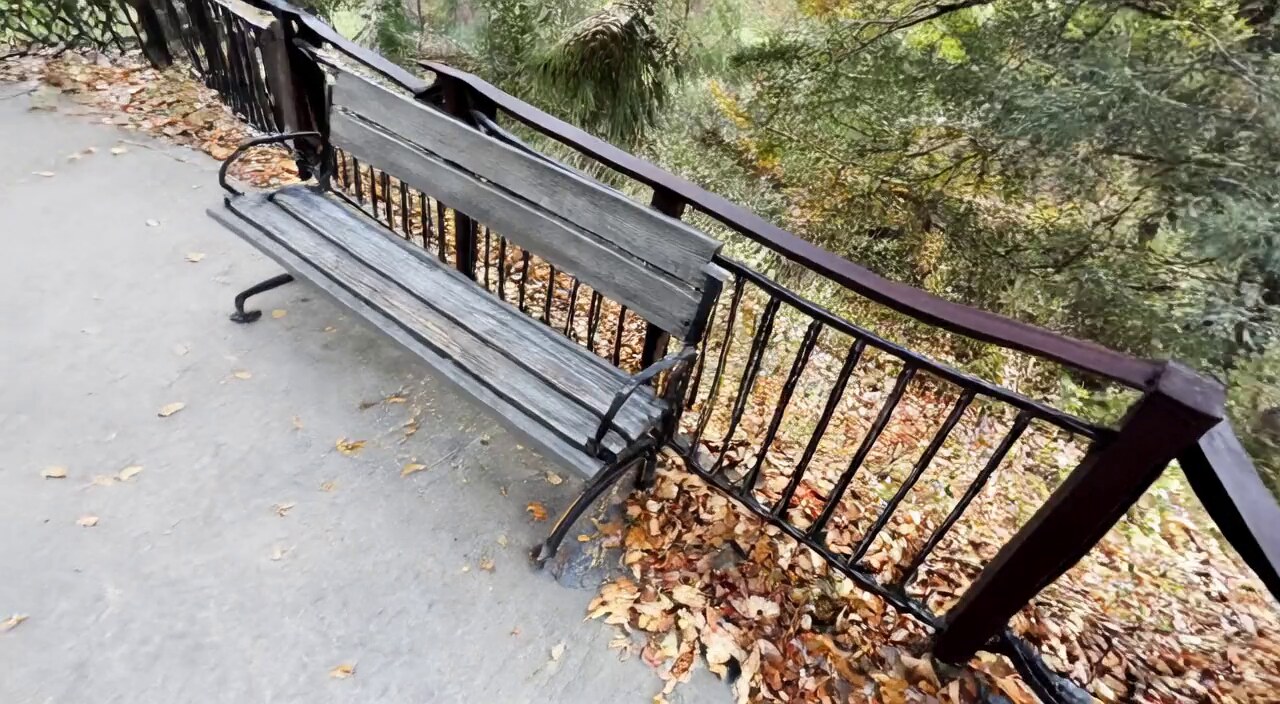} & 
\lframe{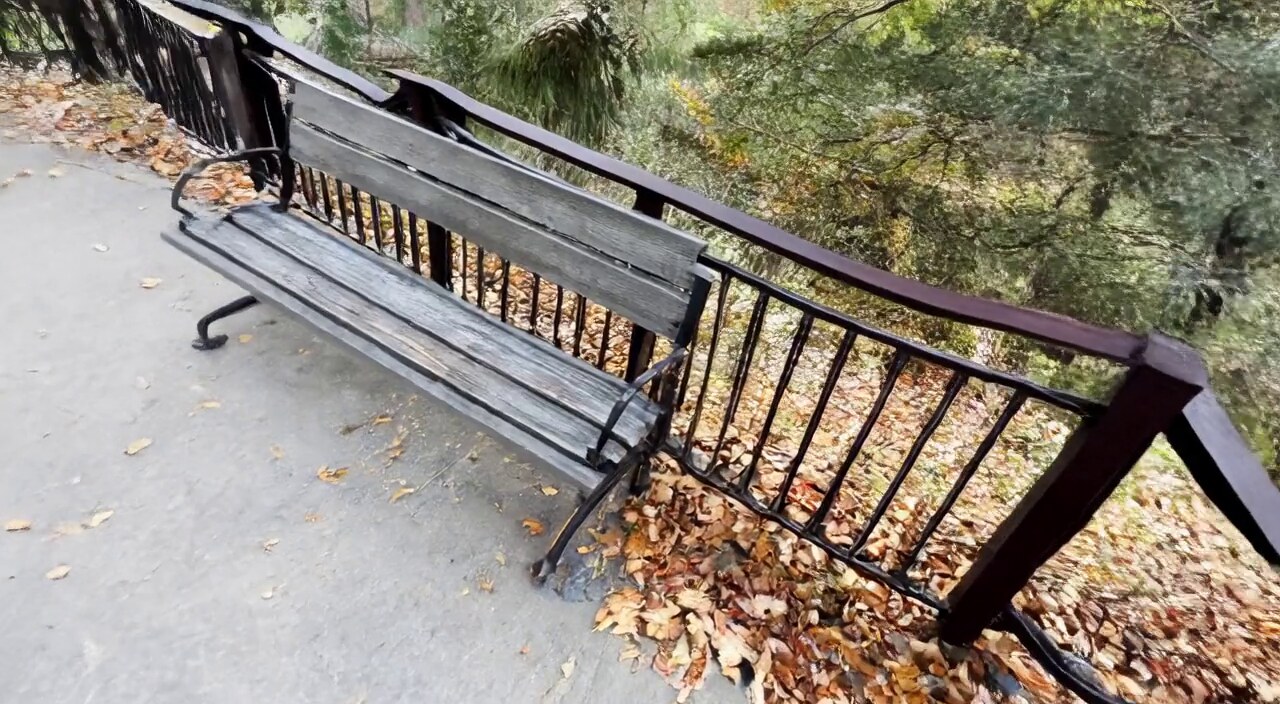} &
\rframe{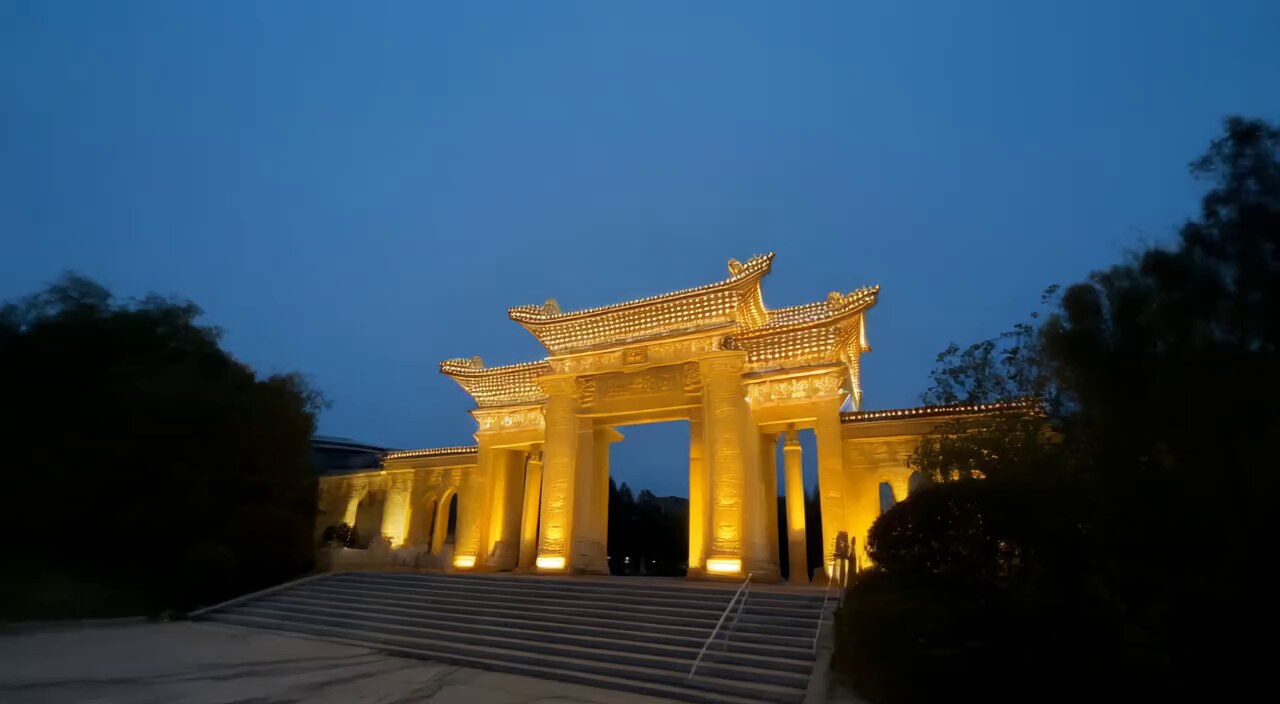} & 
\includegraphics[width=\lw,height=\lh]{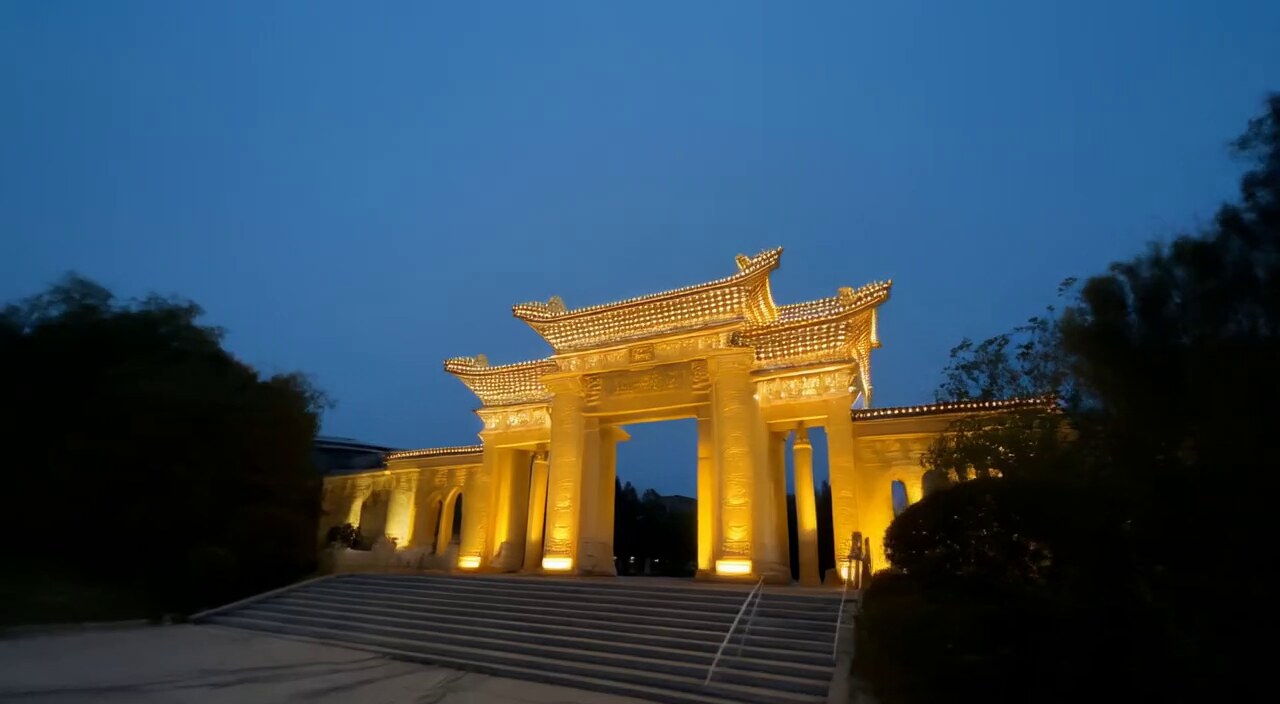} & 
\includegraphics[width=\lw,height=\lh]{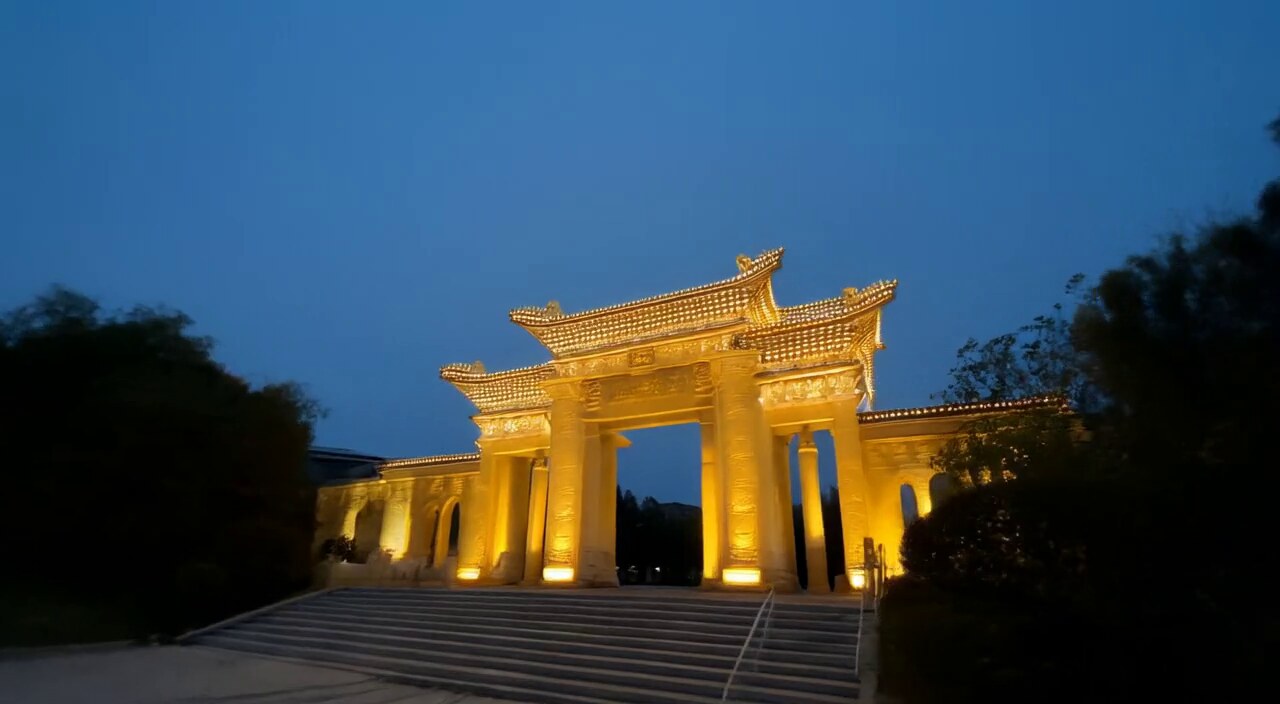} & 
\includegraphics[width=\lw,height=\lh]{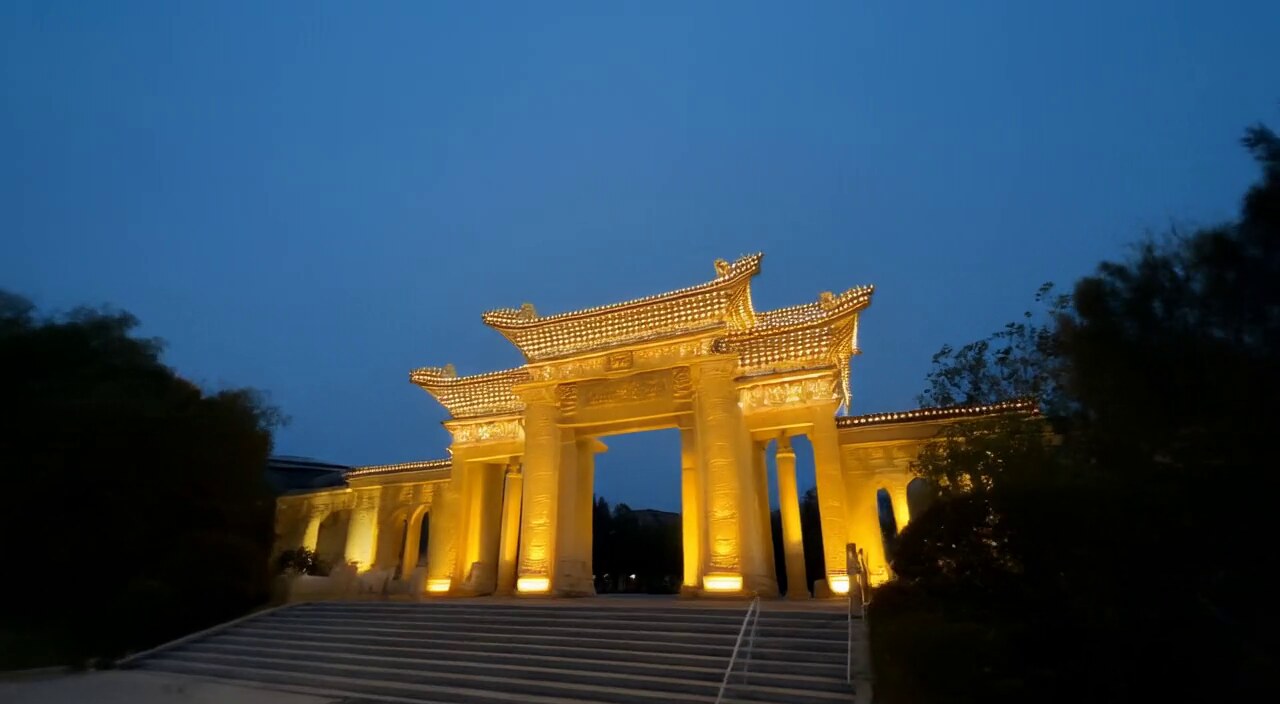} & 
\includegraphics[width=\lw,height=\lh]{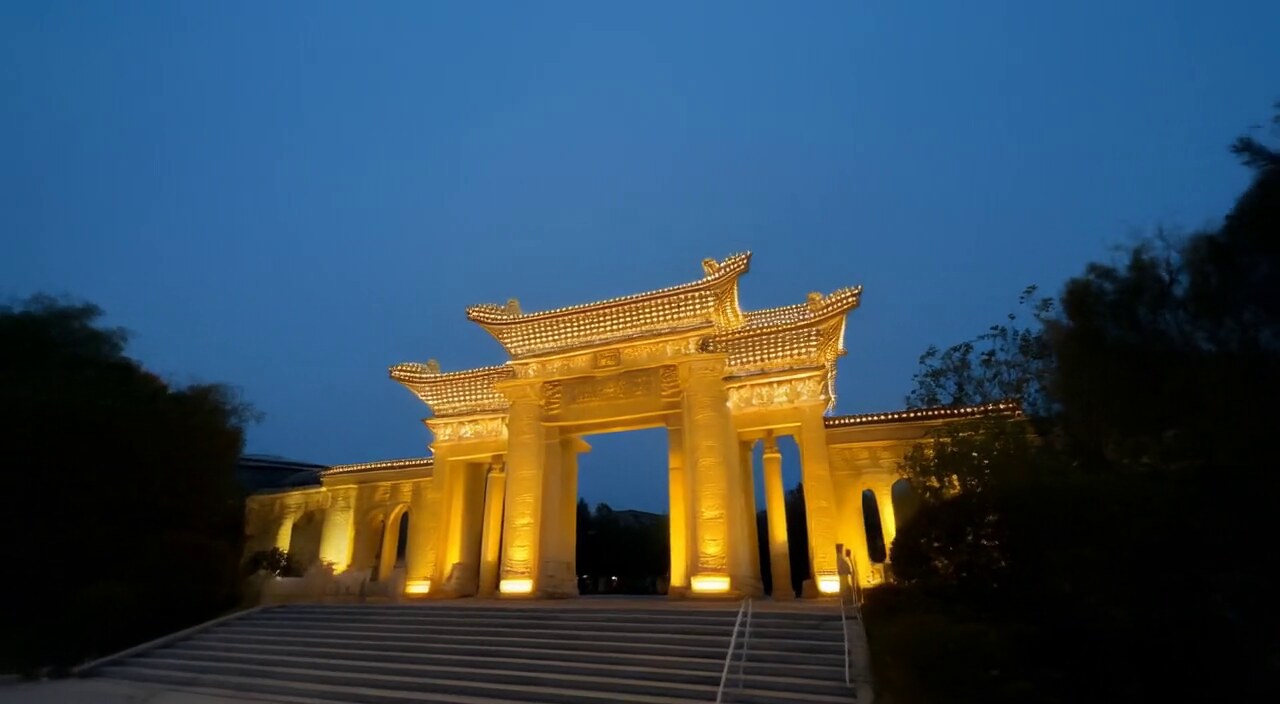} \\ \noalign{\vskip \sskip}
\rotatebox{90}{\hspace{2mm}\footnotesize{Ref}}\hspace{1mm} &
\includegraphics[width=\lw,height=\lh]{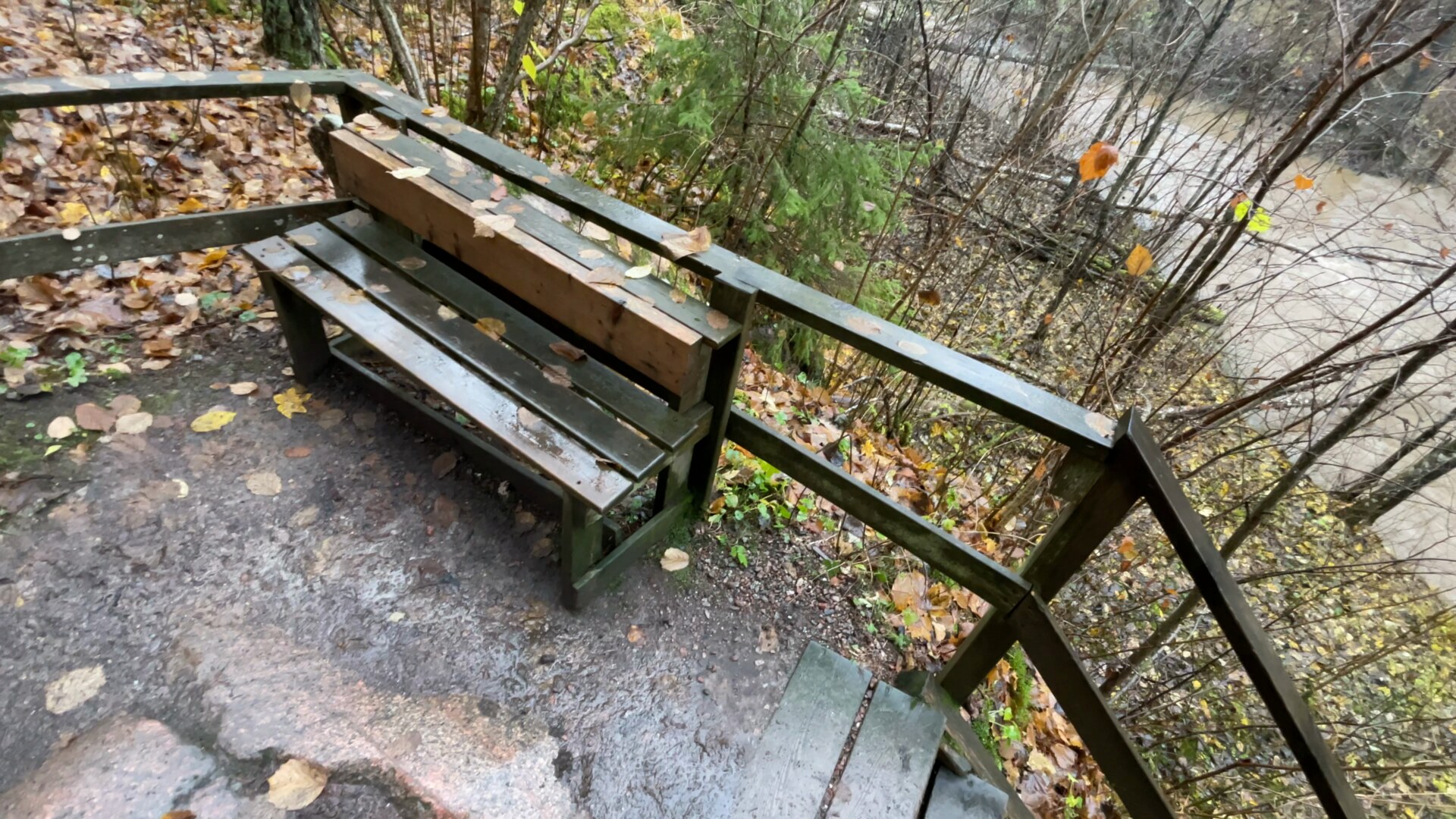} & 
\includegraphics[width=\lw,height=\lh]{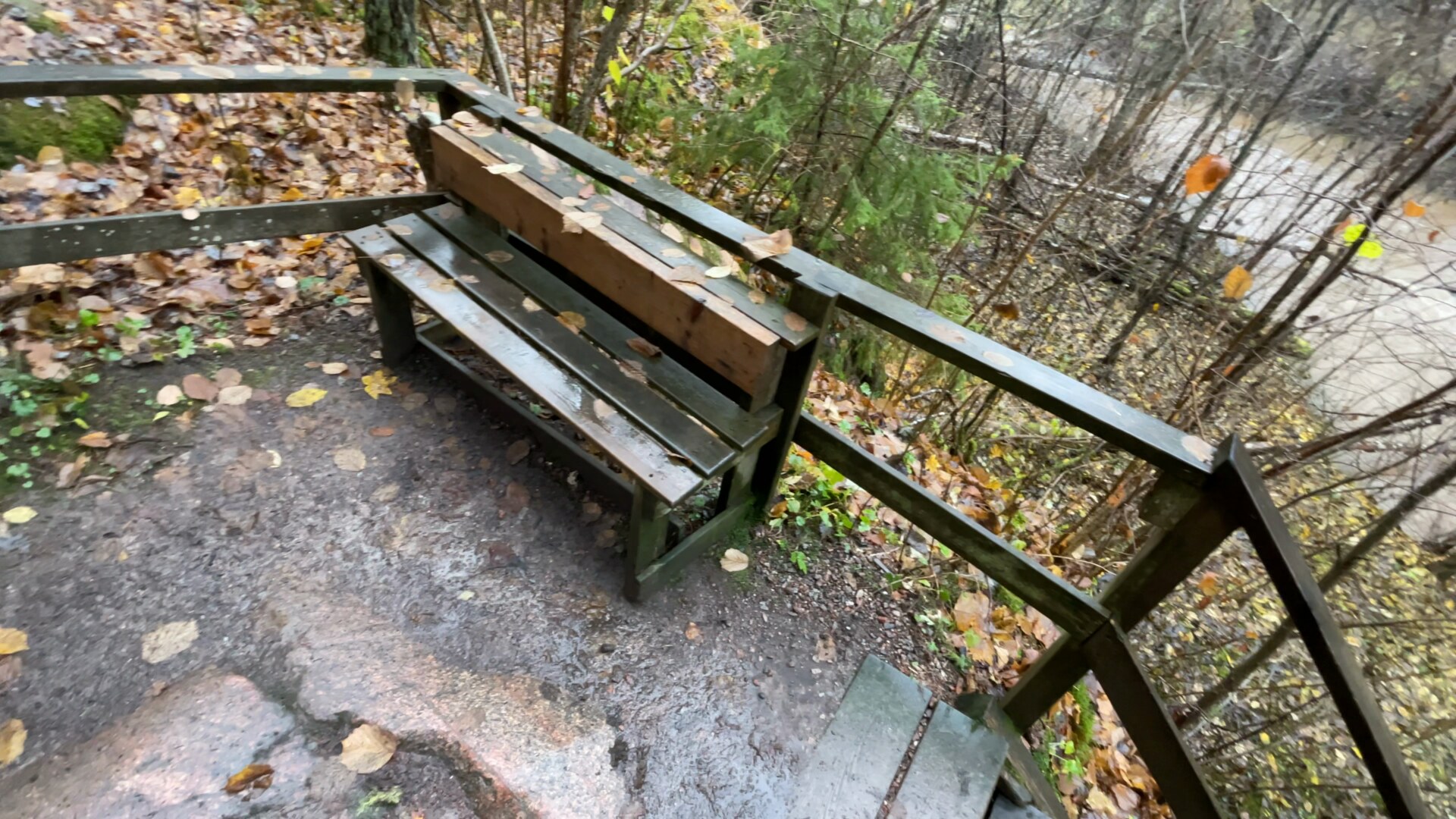} & 
\includegraphics[width=\lw,height=\lh]{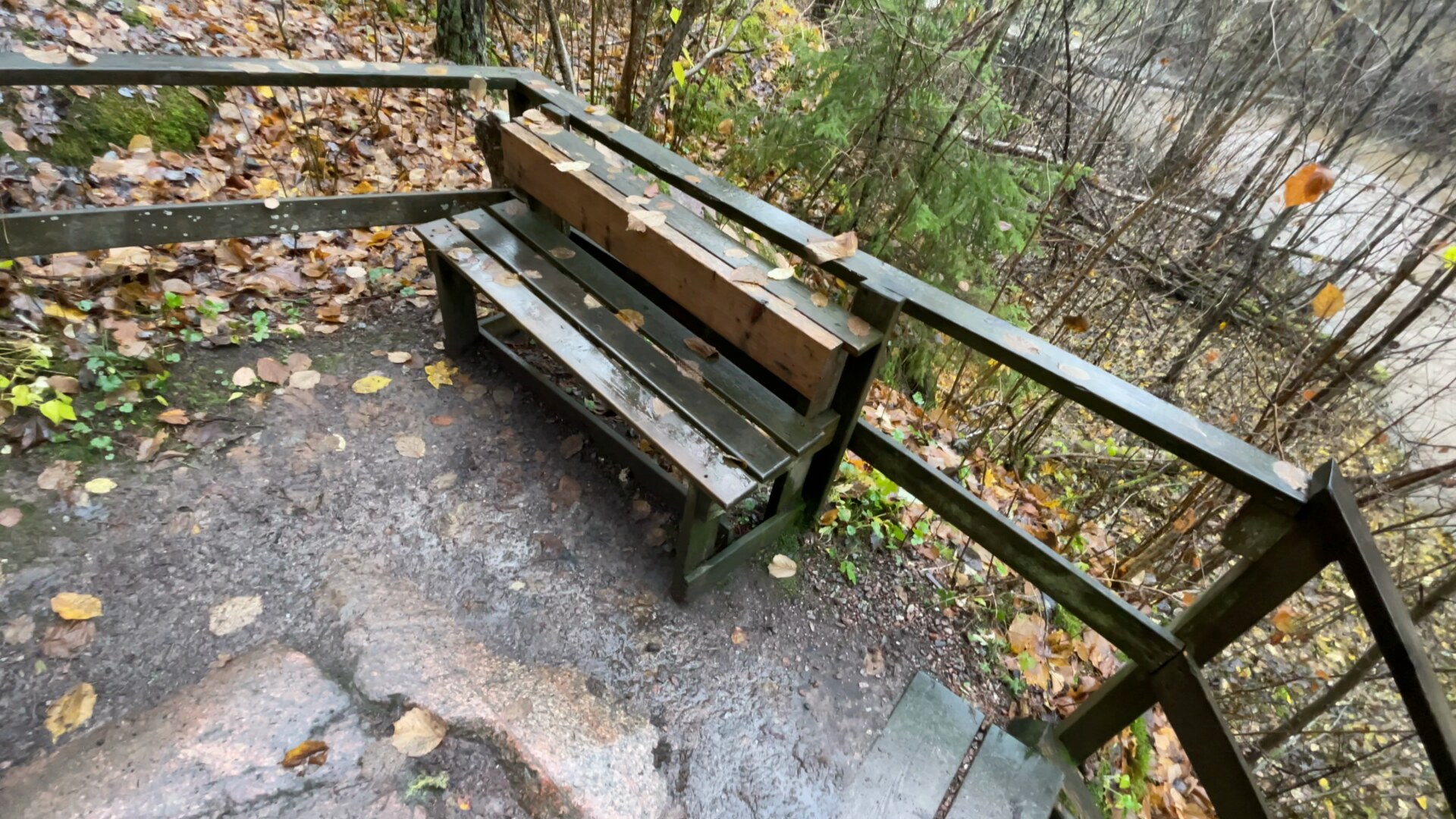} & 
\includegraphics[width=\lw,height=\lh]{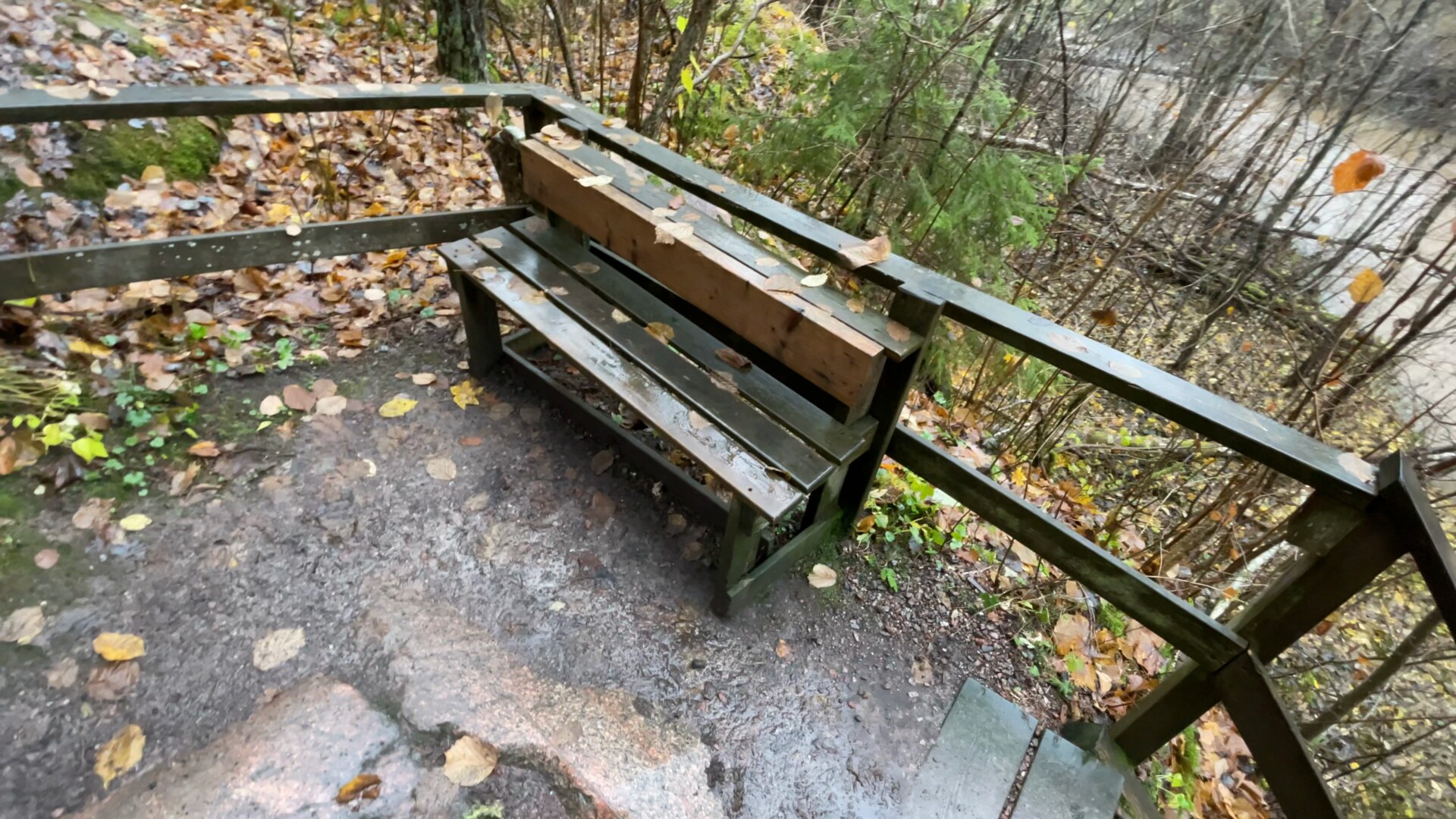} & 
\lframe{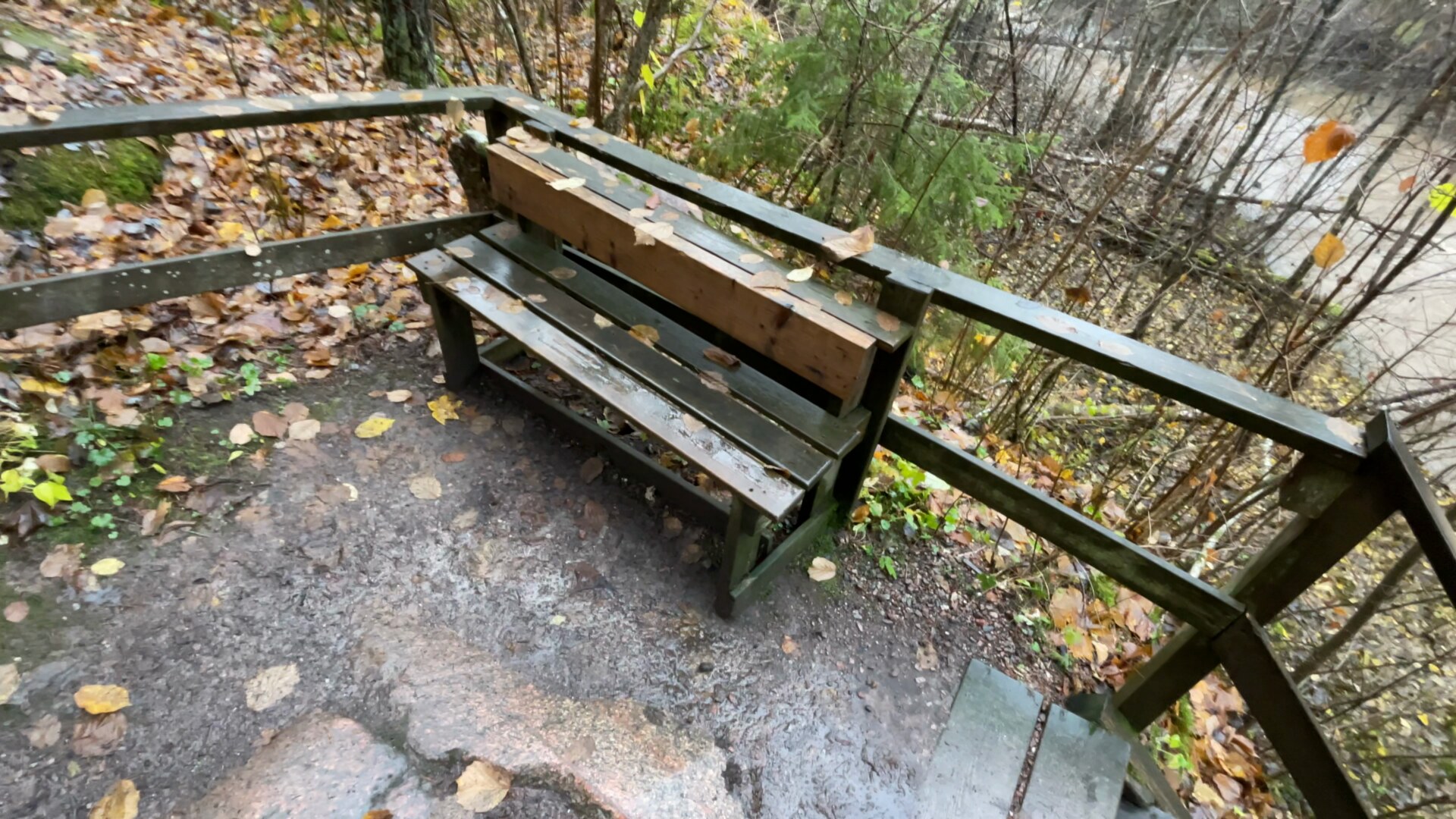} &
\rframe{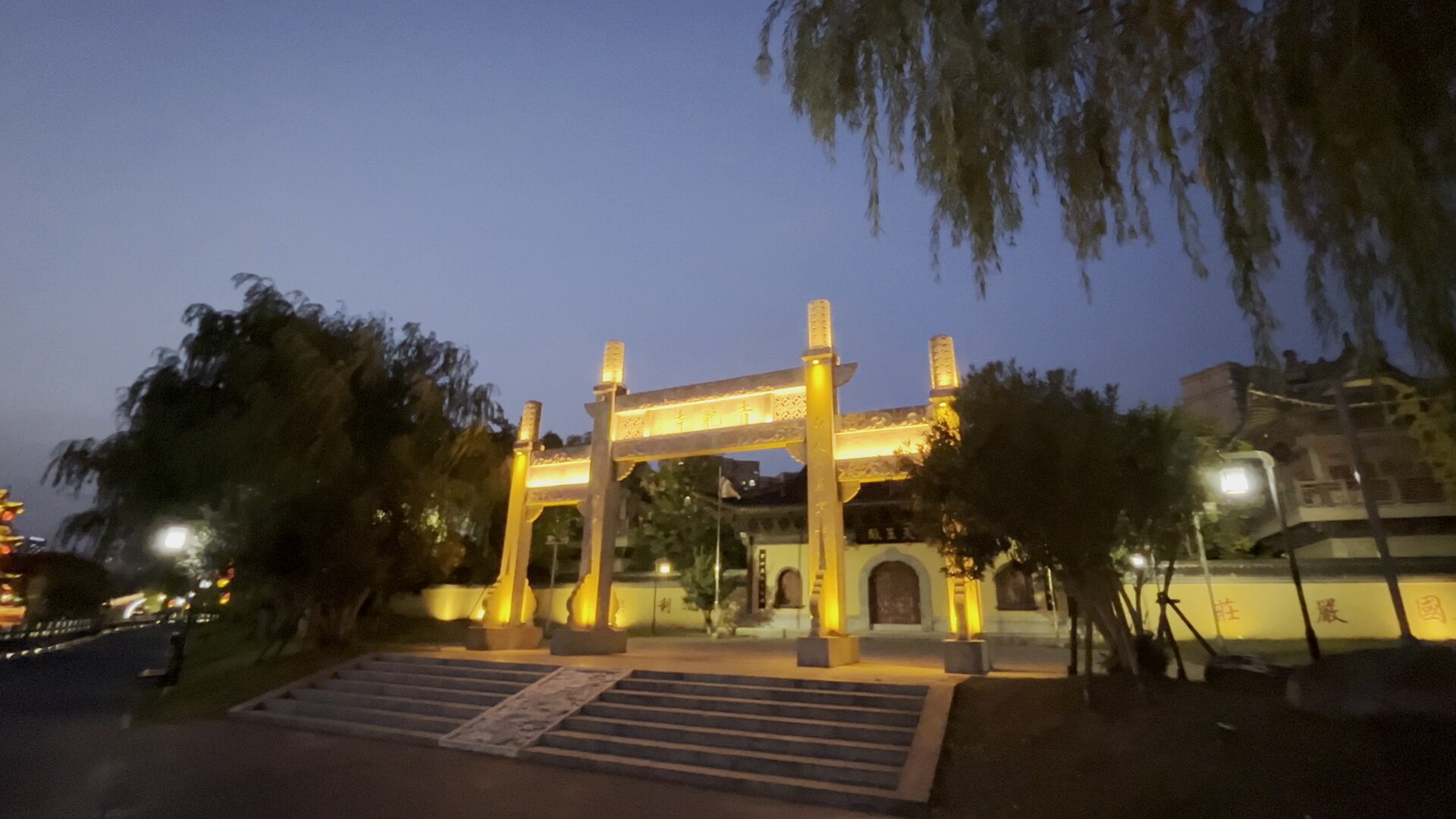} & 
\includegraphics[width=\lw,height=\lh]{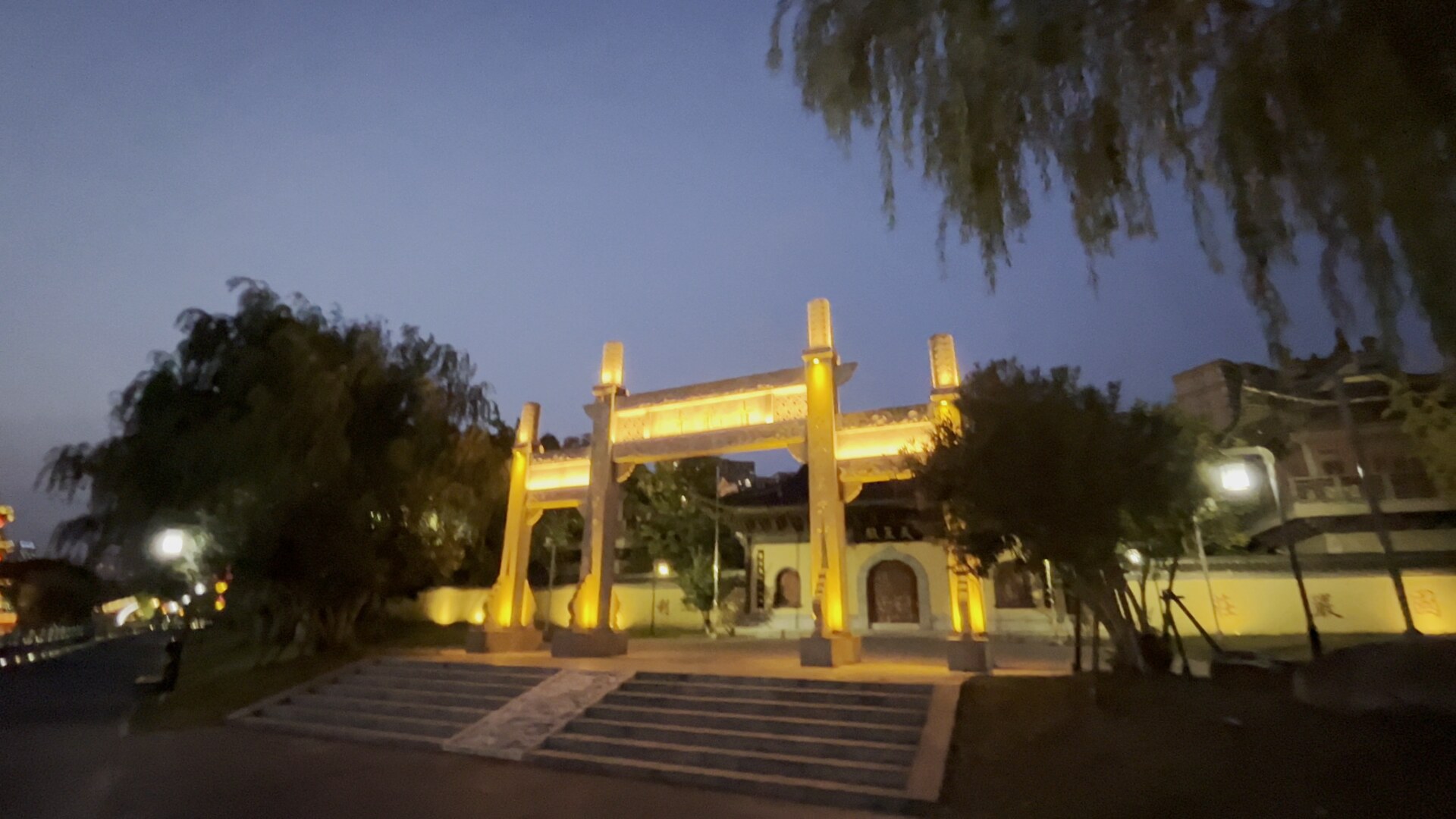} & 
\includegraphics[width=\lw,height=\lh]{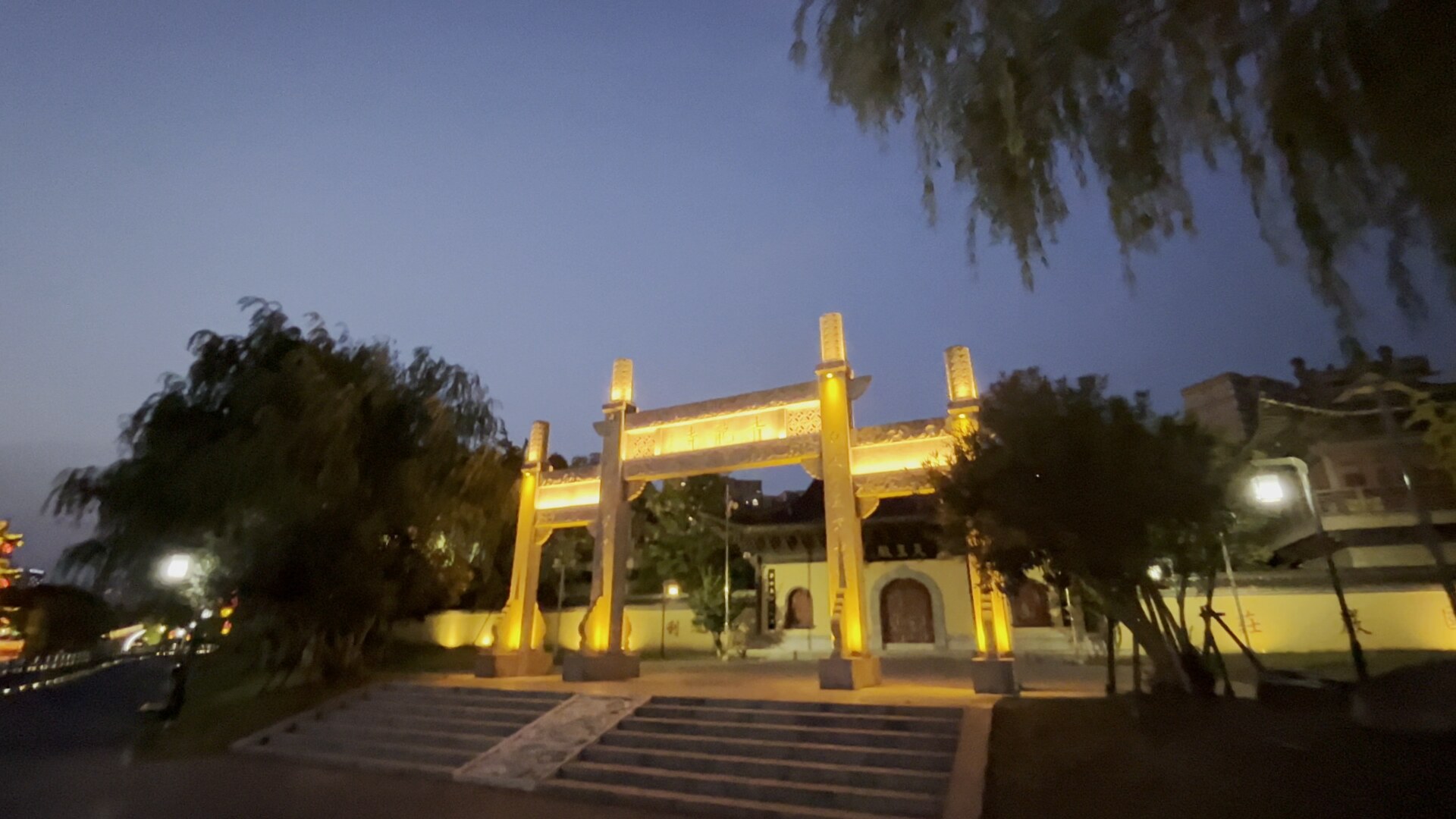} & 
\includegraphics[width=\lw,height=\lh]{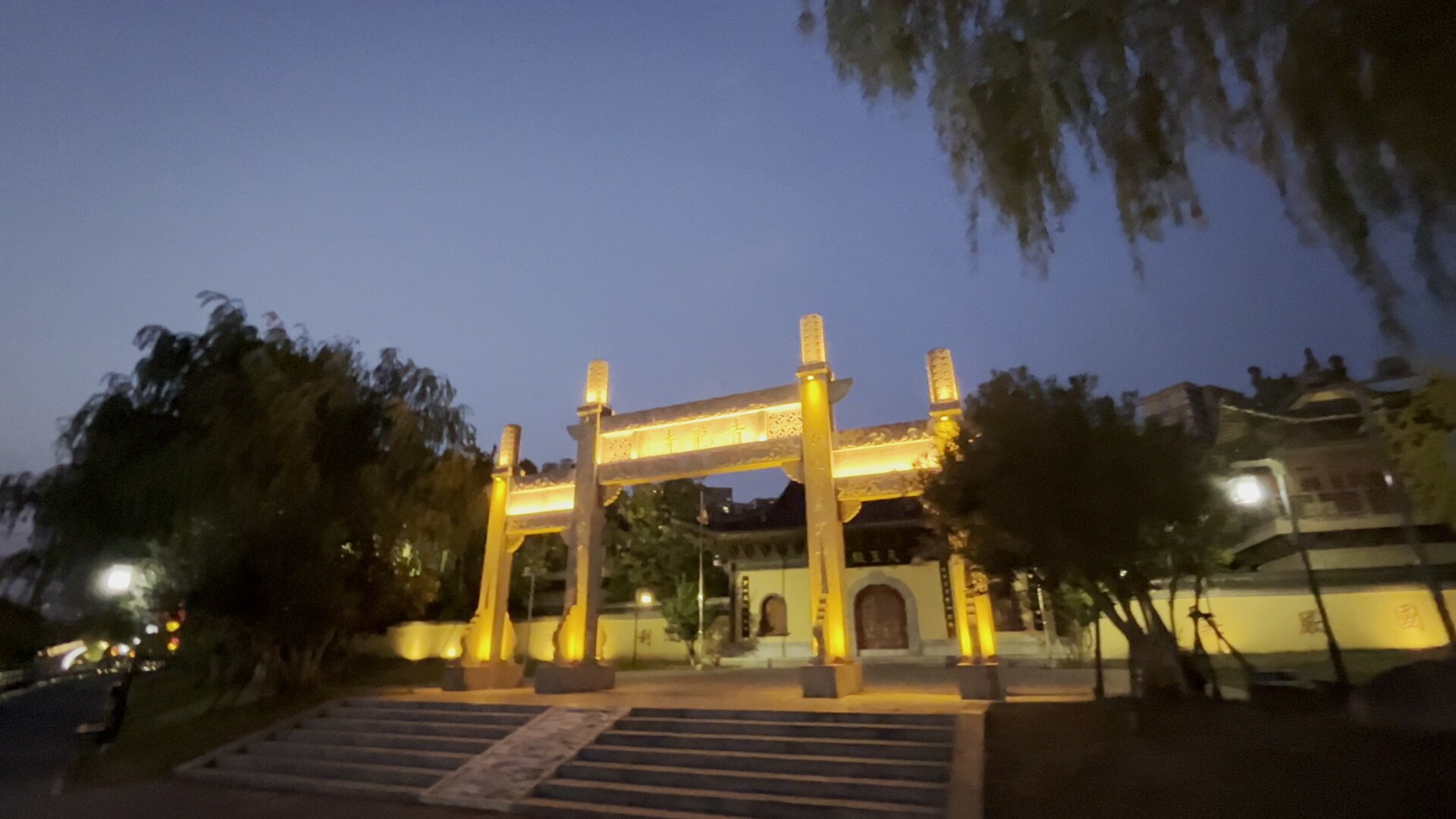} & 
\includegraphics[width=\lw,height=\lh]{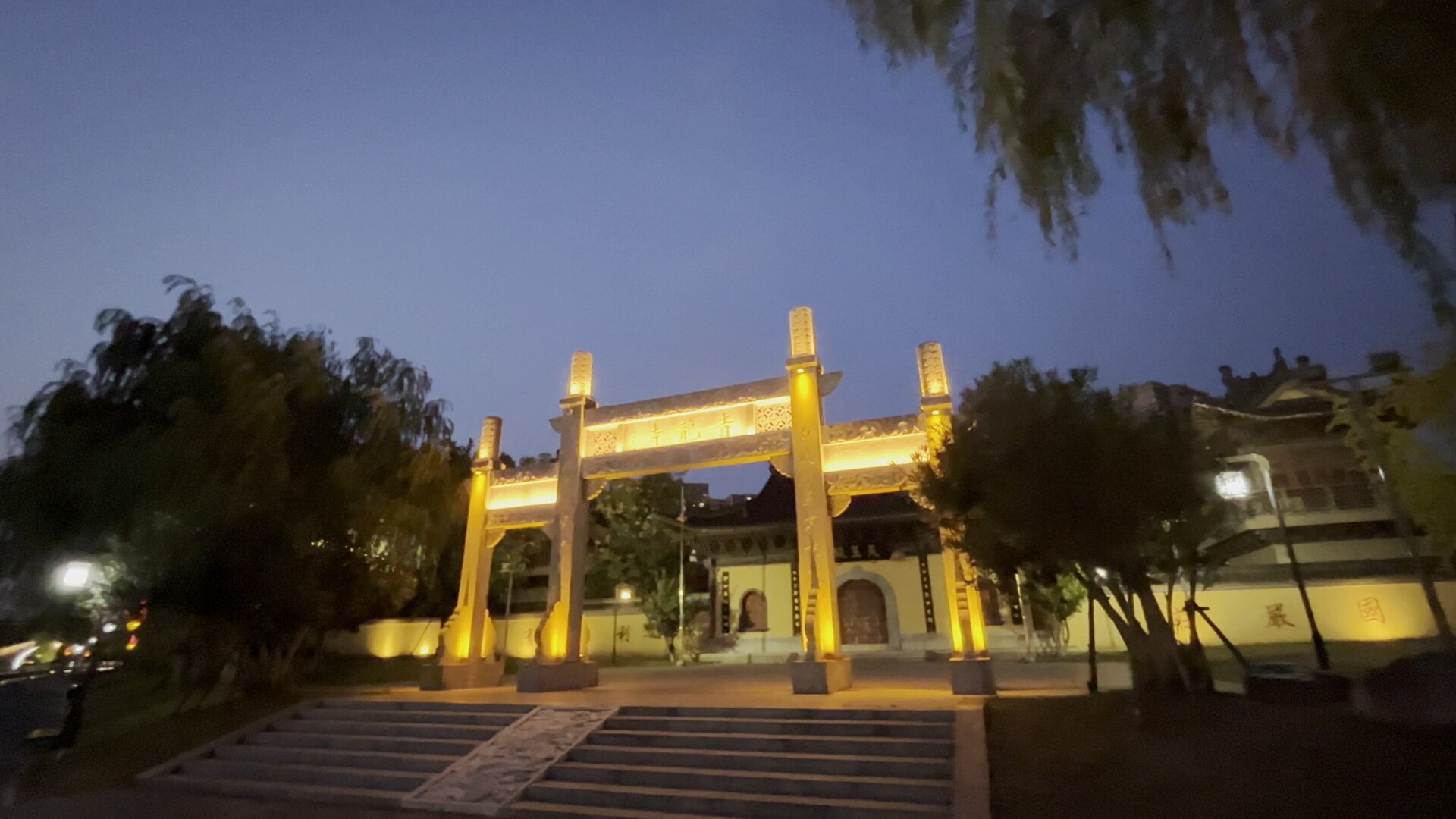} \\
\noalign{\vskip 2pt}
\hdashline
\noalign{\vskip 2pt}
\rotatebox{90}{\footnotesize{CC~\cite{he2025cameractrl}}}\hspace{1mm} &
\includegraphics[width=\lw,height=\lh]{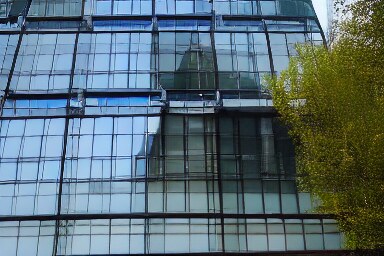} & 
\includegraphics[width=\lw,height=\lh]{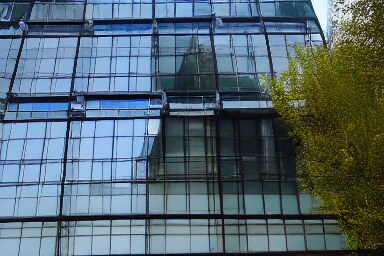} & 
\includegraphics[width=\lw,height=\lh]{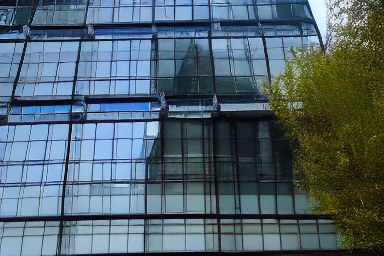} & 
\includegraphics[width=\lw,height=\lh]{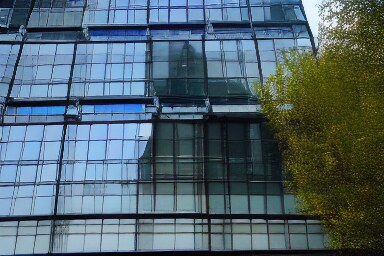} & 
\lframe{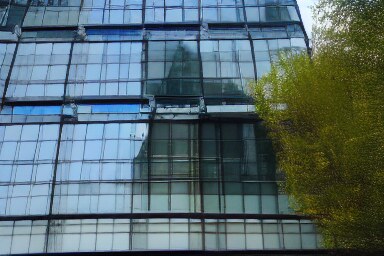} &
\rframe{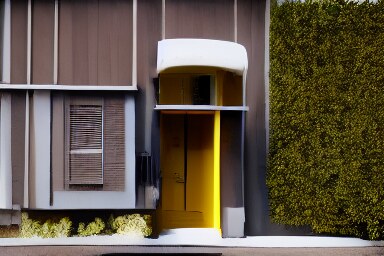} & 
\includegraphics[width=\lw,height=\lh]{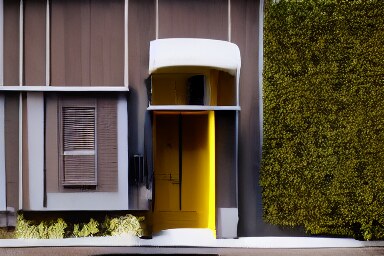} & 
\includegraphics[width=\lw,height=\lh]{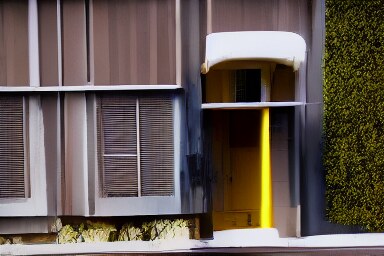} & 
\includegraphics[width=\lw,height=\lh]{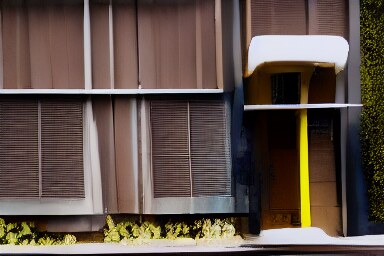} & 
\includegraphics[width=\lw,height=\lh]{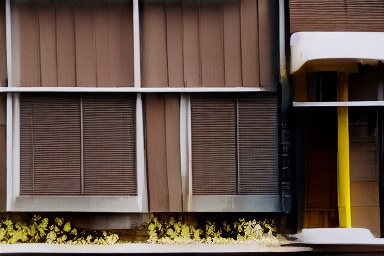} \\ \noalign{\vskip \sskip}
\rotatebox{90}{\footnotesize{MC~\cite{motionctrl}}}\hspace{1mm} &
\includegraphics[width=\lw,height=\lh]{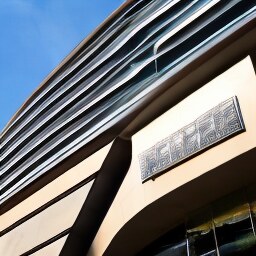} & 
\includegraphics[width=\lw,height=\lh]{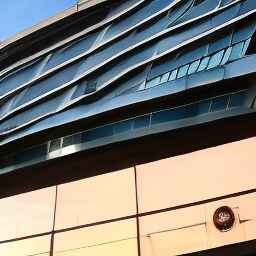} & 
\includegraphics[width=\lw,height=\lh]{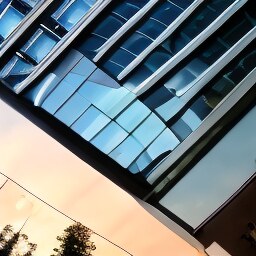} & 
\includegraphics[width=\lw,height=\lh]{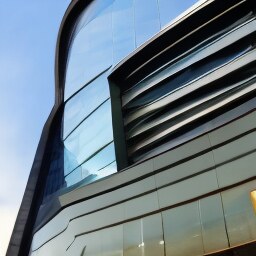} & 
\lframe{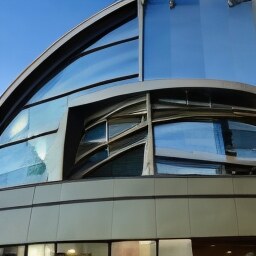} &
\rframe{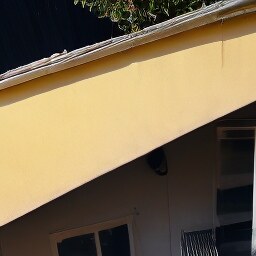} & 
\includegraphics[width=\lw,height=\lh]{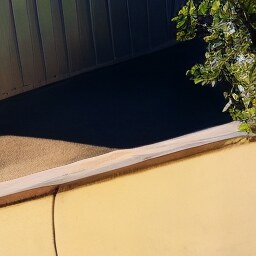} & 
\includegraphics[width=\lw,height=\lh]{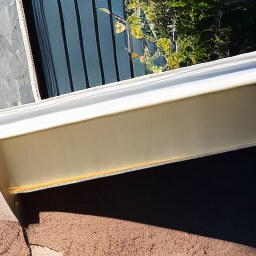} & 
\includegraphics[width=\lw,height=\lh]{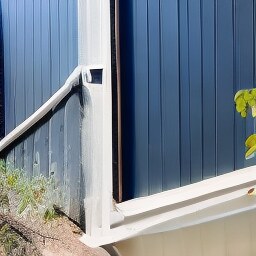} & 
\includegraphics[width=\lw,height=\lh]{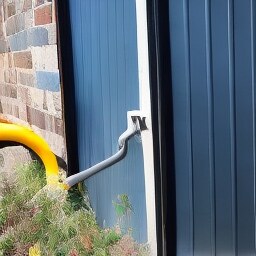} \\ \noalign{\vskip \sskip}
\rotatebox{90}{\footnotesize{VD~\cite{kim2025videofrom3d}}}\hspace{1mm} &
\includegraphics[width=\lw,height=\lh]{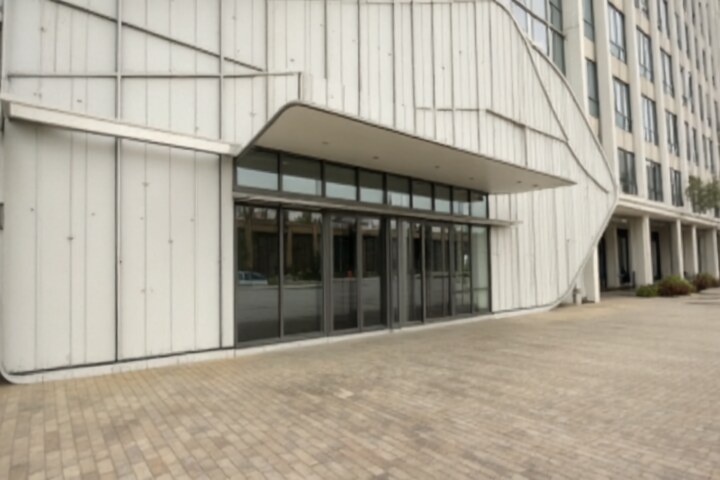} & 
\includegraphics[width=\lw,height=\lh]{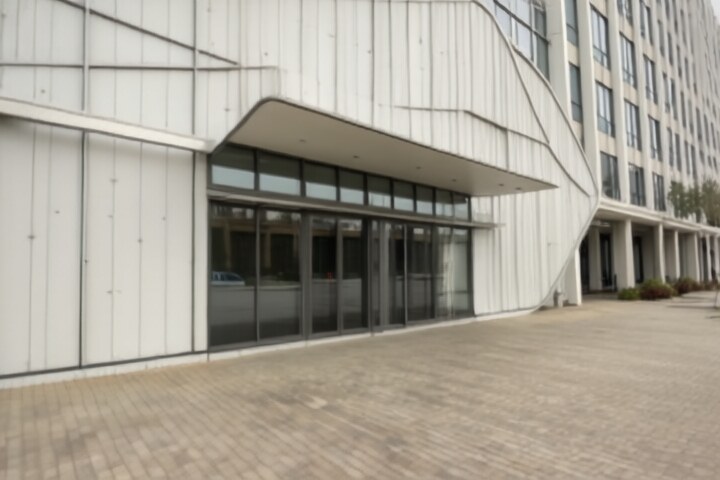} & 
\includegraphics[width=\lw,height=\lh]{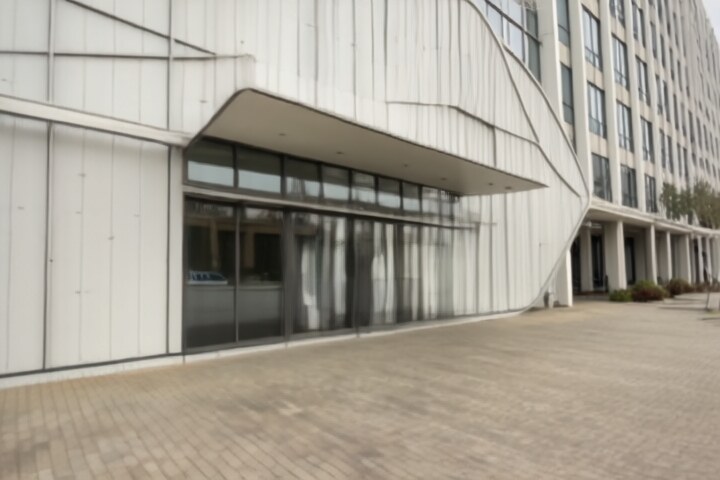} & 
\includegraphics[width=\lw,height=\lh]{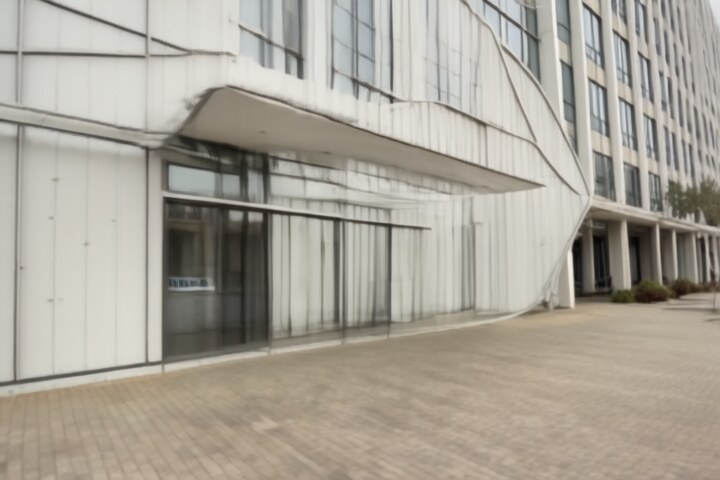} & 
\lframe{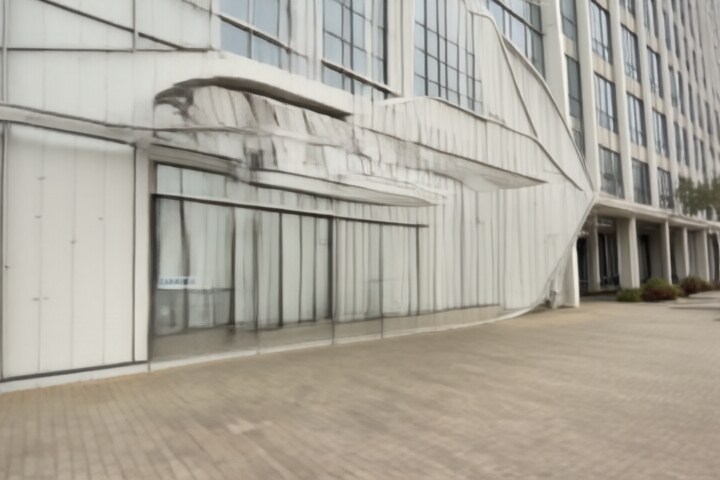} &
\rframe{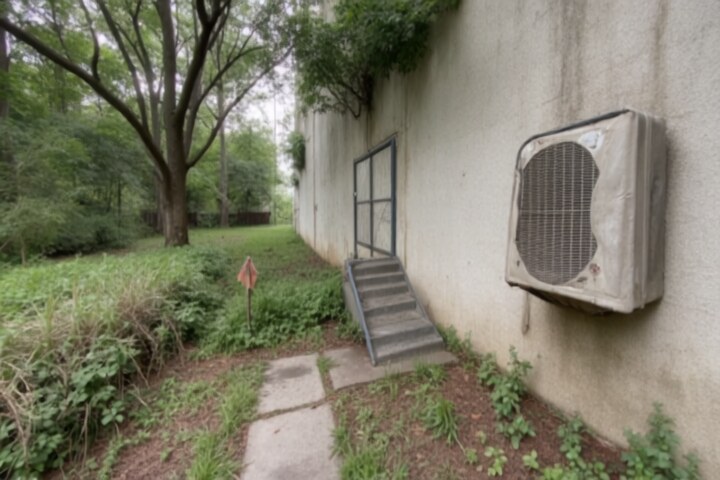} & 
\includegraphics[width=\lw,height=\lh]{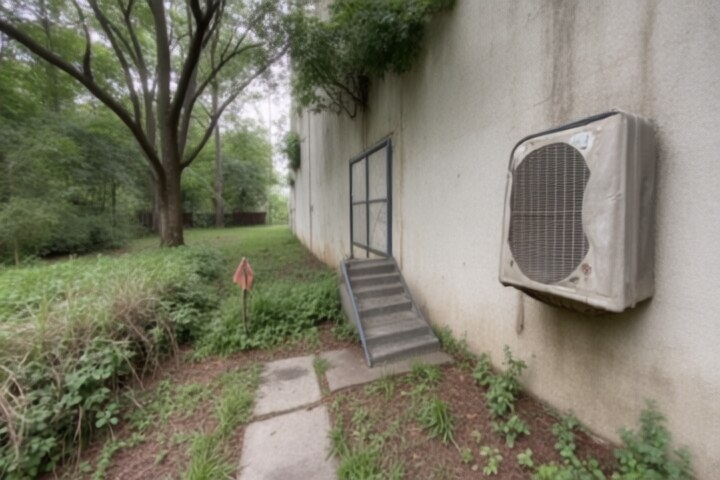} & 
\includegraphics[width=\lw,height=\lh]{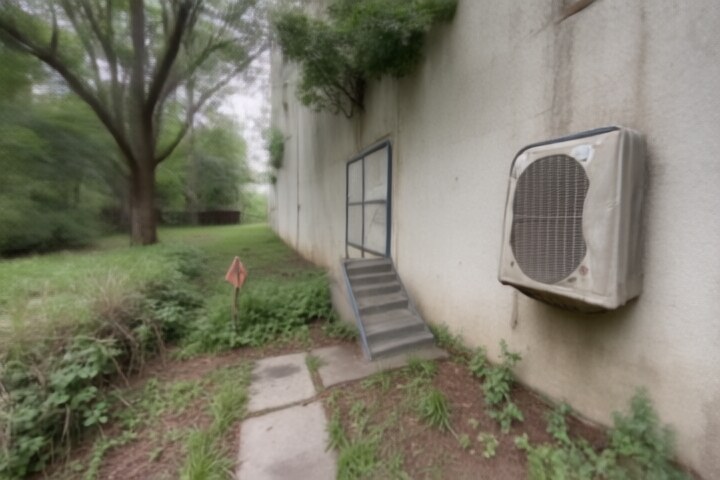} & 
\includegraphics[width=\lw,height=\lh]{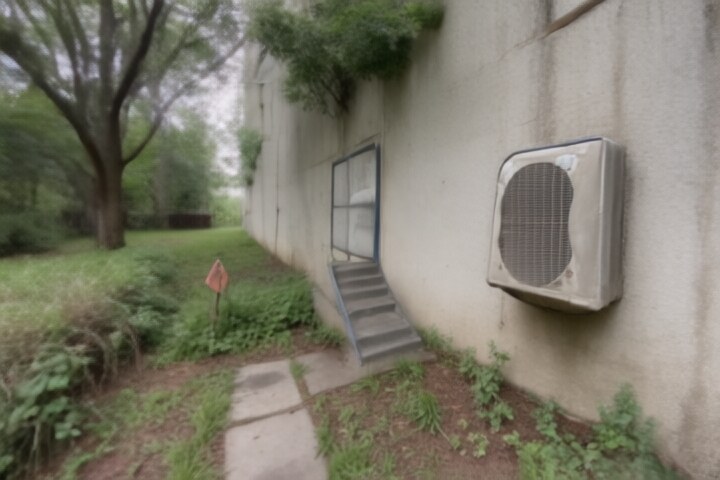} & 
\includegraphics[width=\lw,height=\lh]{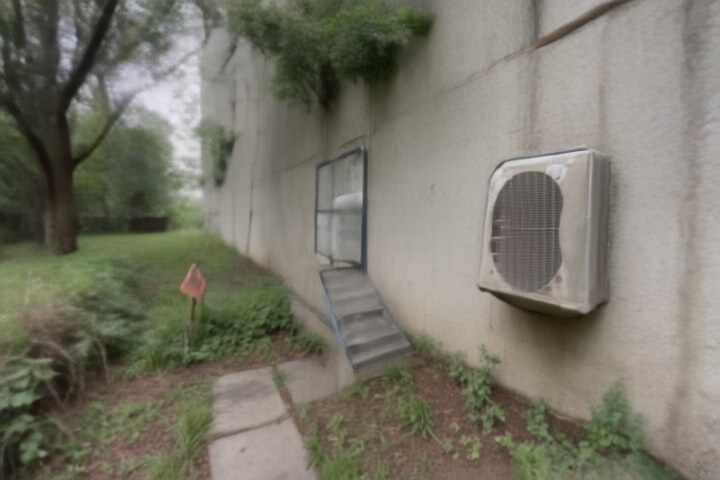} \\ \noalign{\vskip \sskip}
\rotatebox{90}{\hspace{2mm}\footnotesize{Ours}}\hspace{1mm} &
\includegraphics[width=\lw,height=\lh]{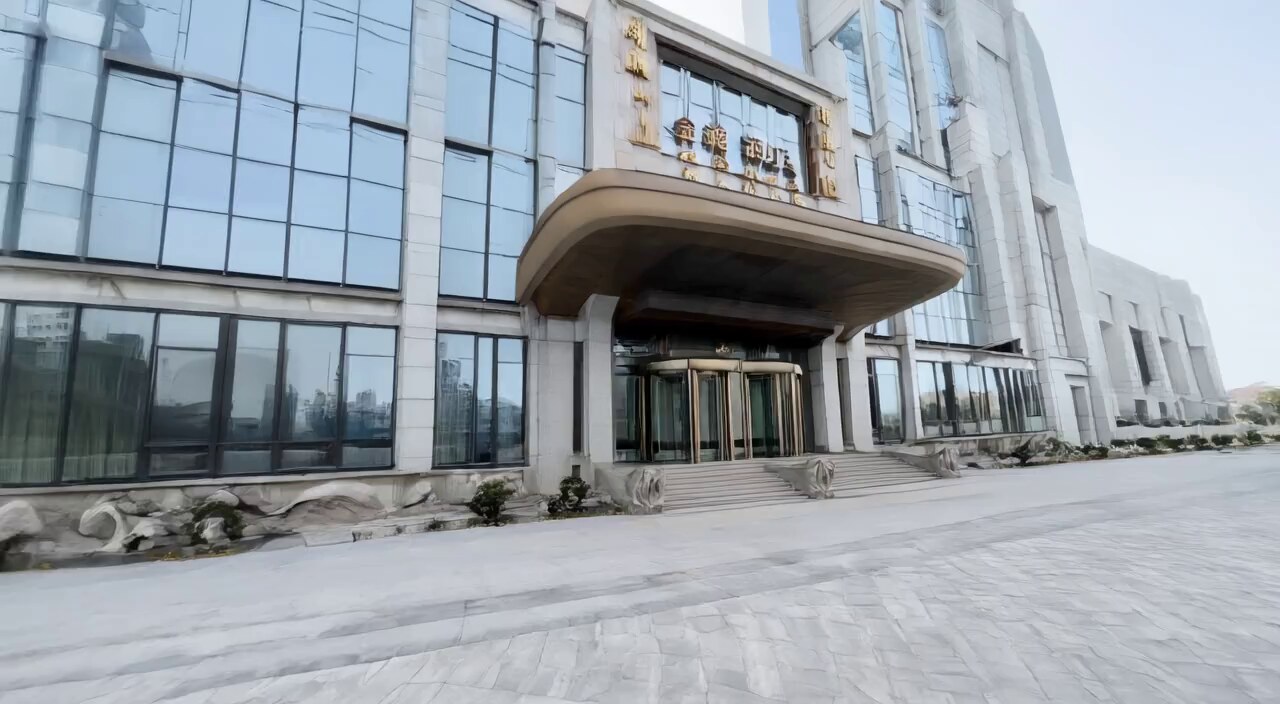} & 
\includegraphics[width=\lw,height=\lh]{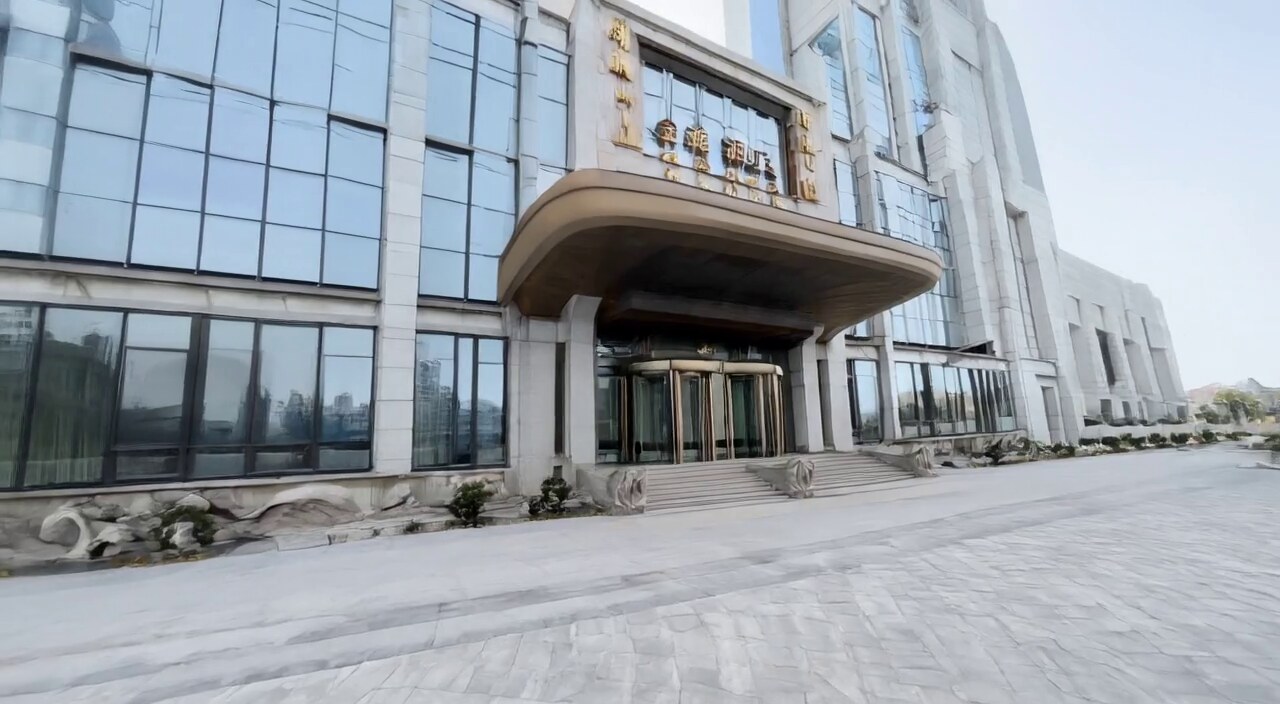} & 
\includegraphics[width=\lw,height=\lh]{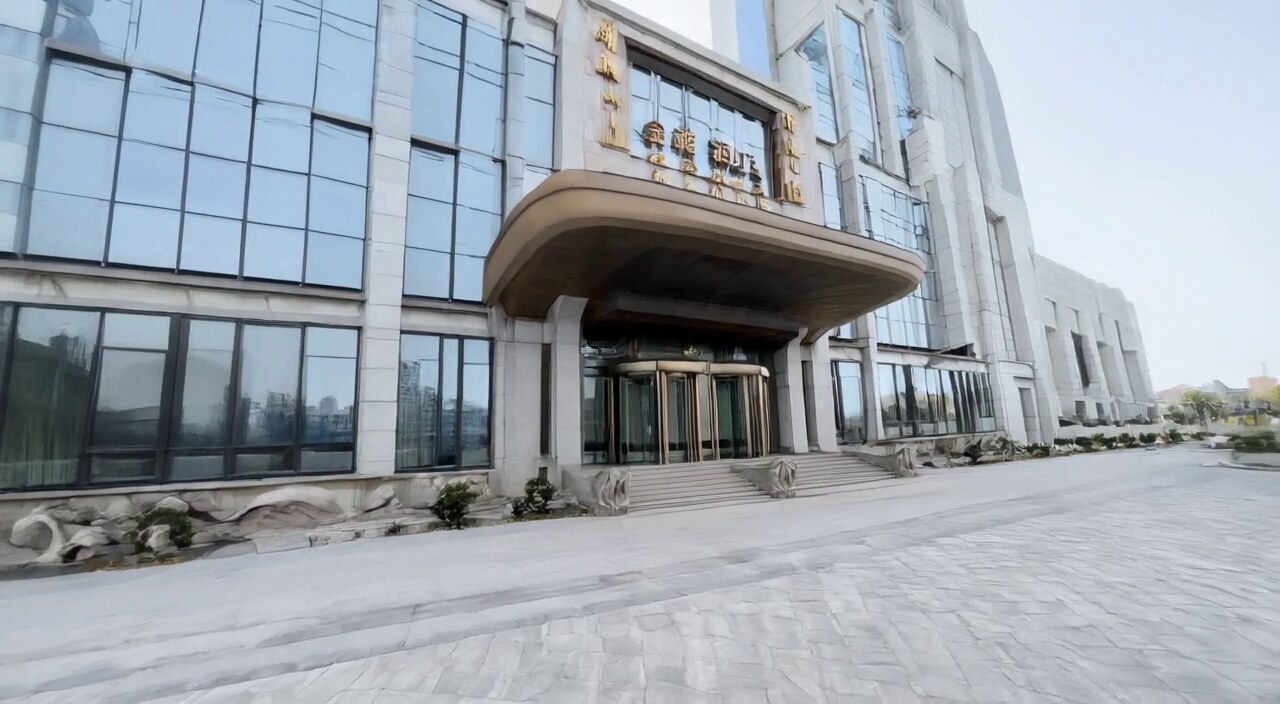} & 
\includegraphics[width=\lw,height=\lh]{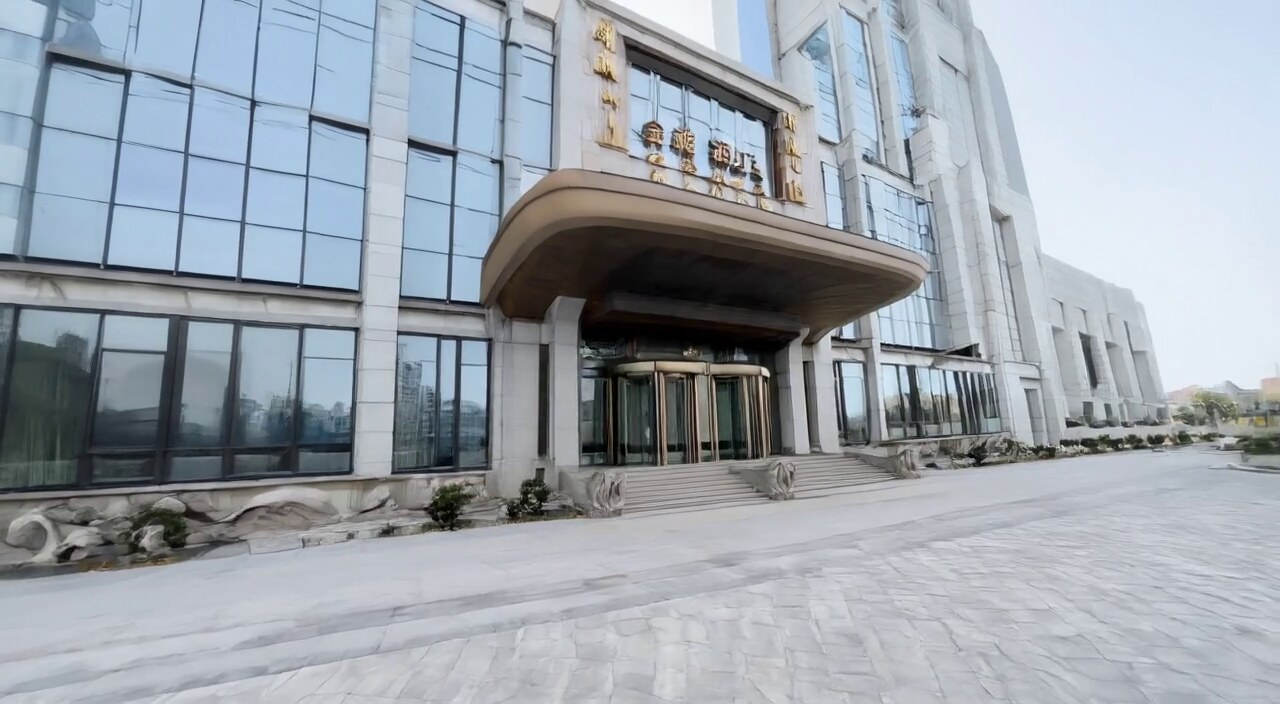} & 
\lframe{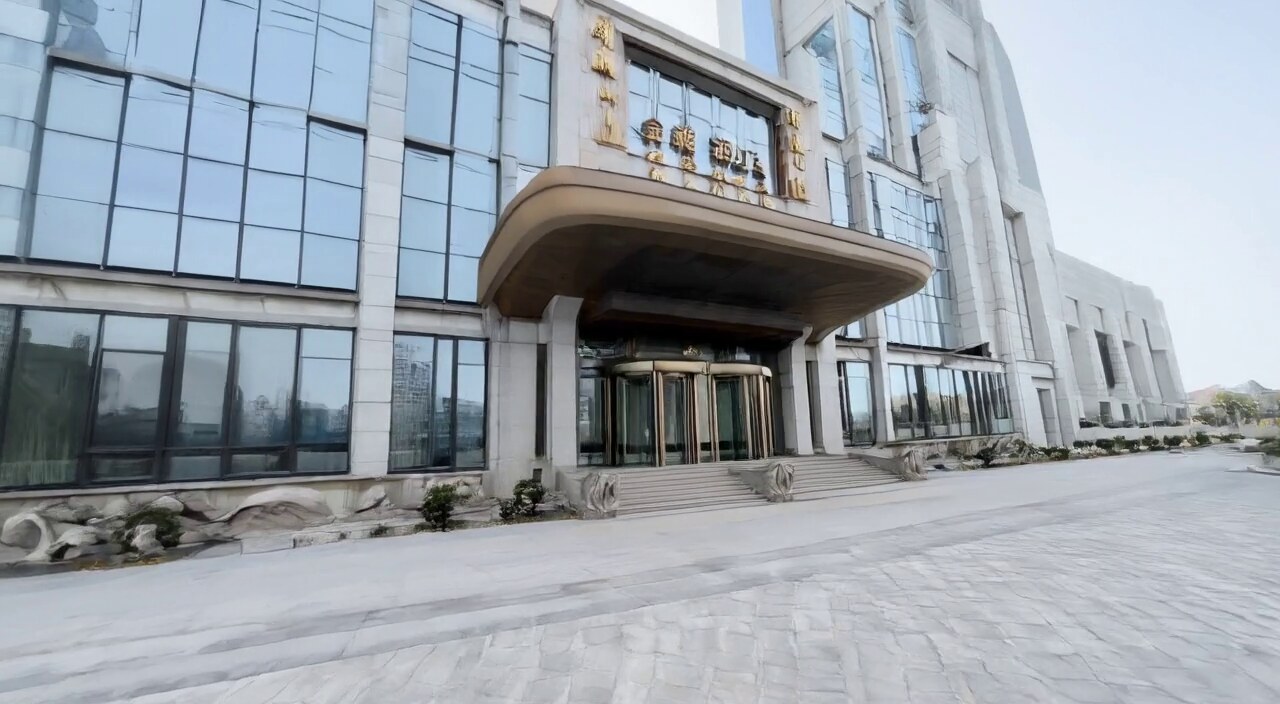} &
\rframe{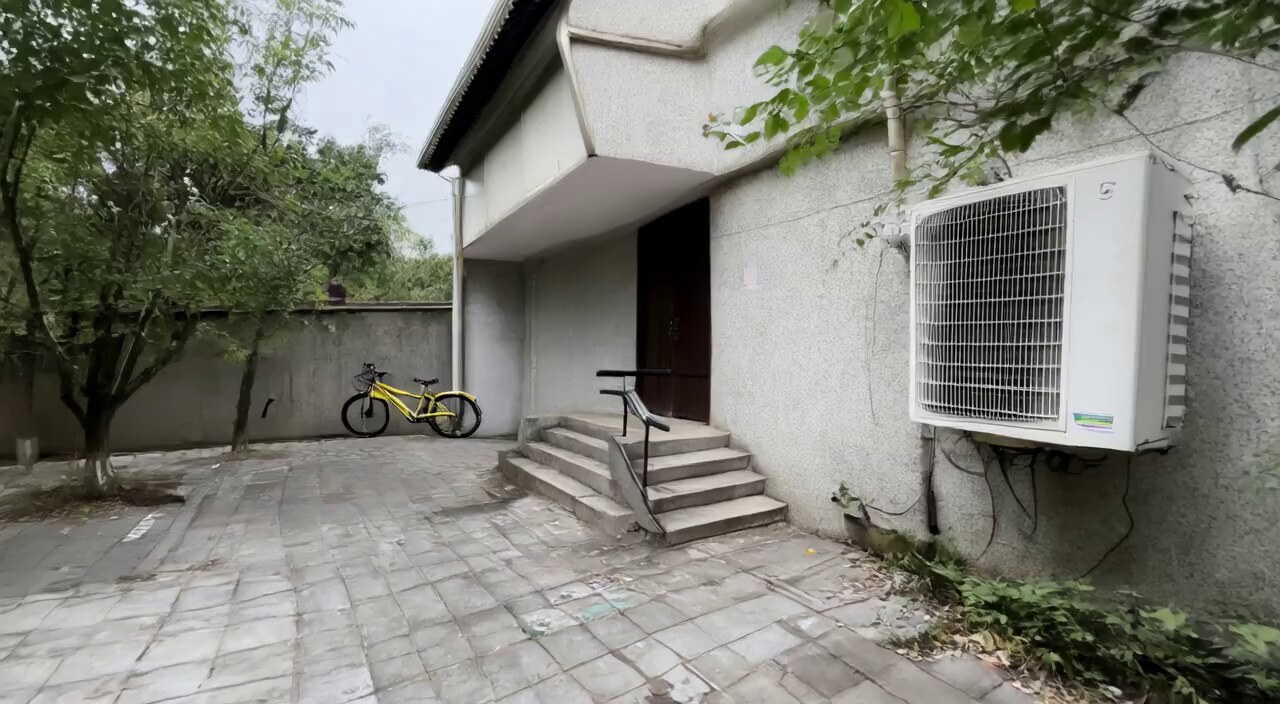} & 
\includegraphics[width=\lw,height=\lh]{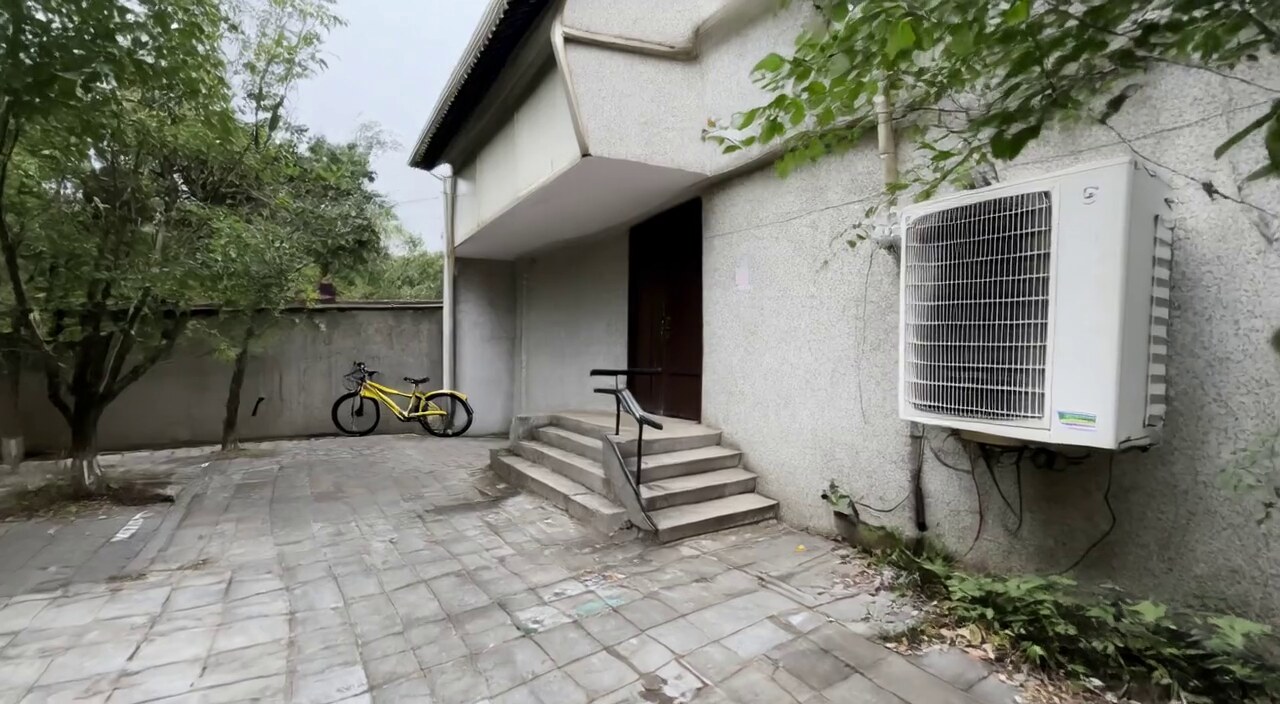} & 
\includegraphics[width=\lw,height=\lh]{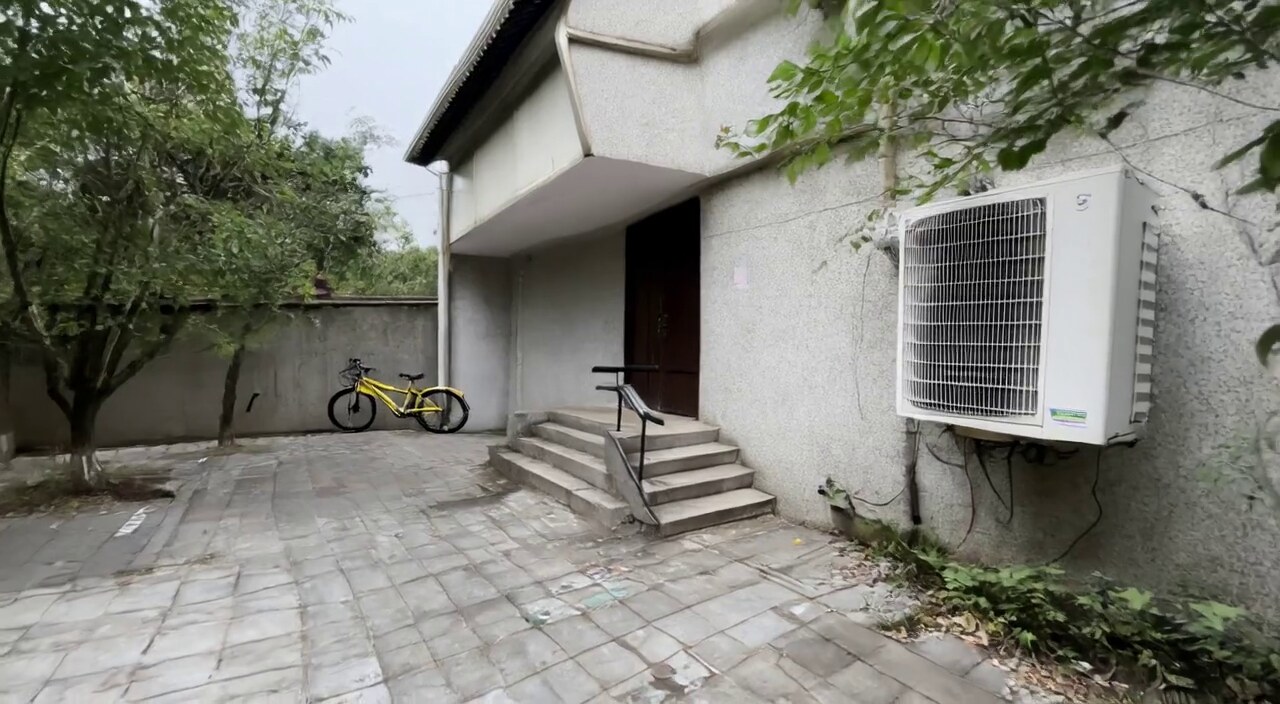} & 
\includegraphics[width=\lw,height=\lh]{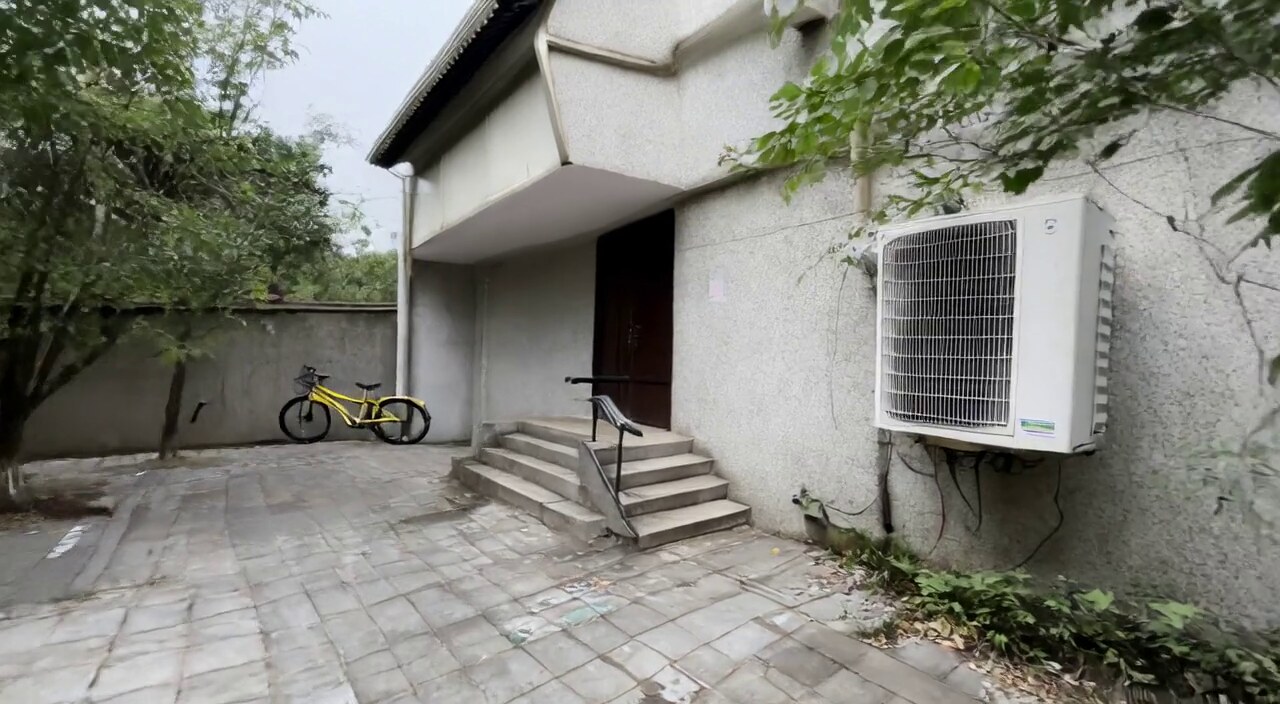} & 
\includegraphics[width=\lw,height=\lh]{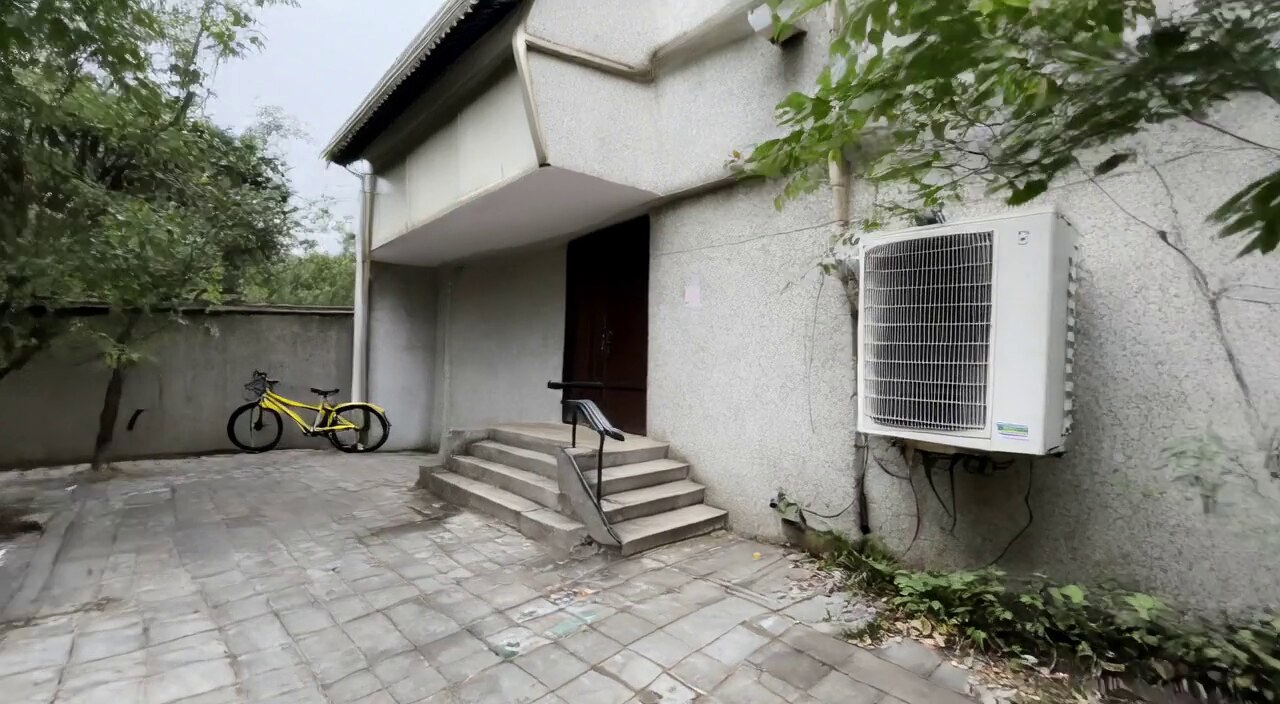} \\ \noalign{\vskip \sskip}

\rotatebox{90}{\hspace{2mm}\footnotesize{Ref}}\hspace{1mm} &
\includegraphics[width=\lw,height=\lh]{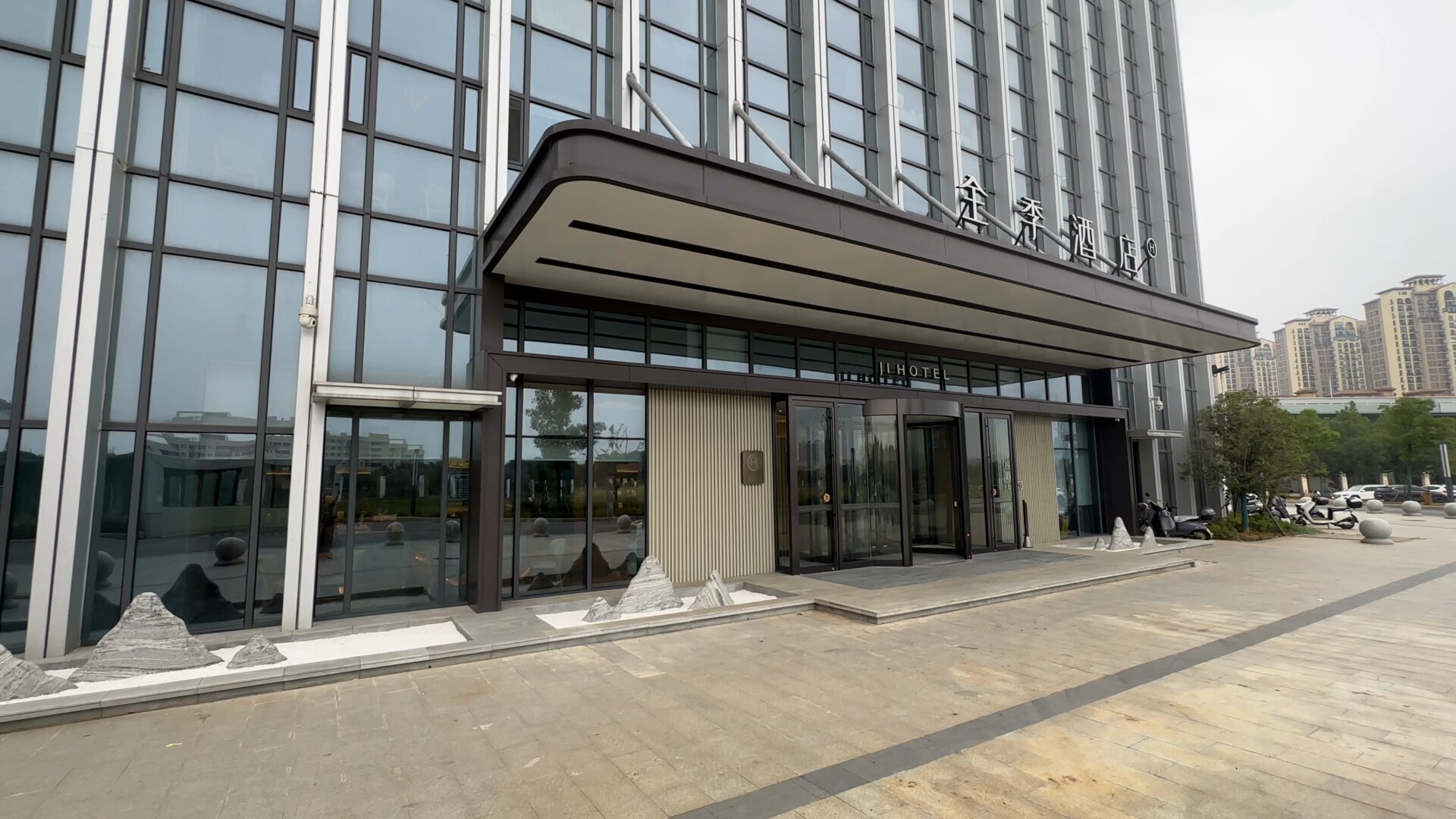} & 
\includegraphics[width=\lw,height=\lh]{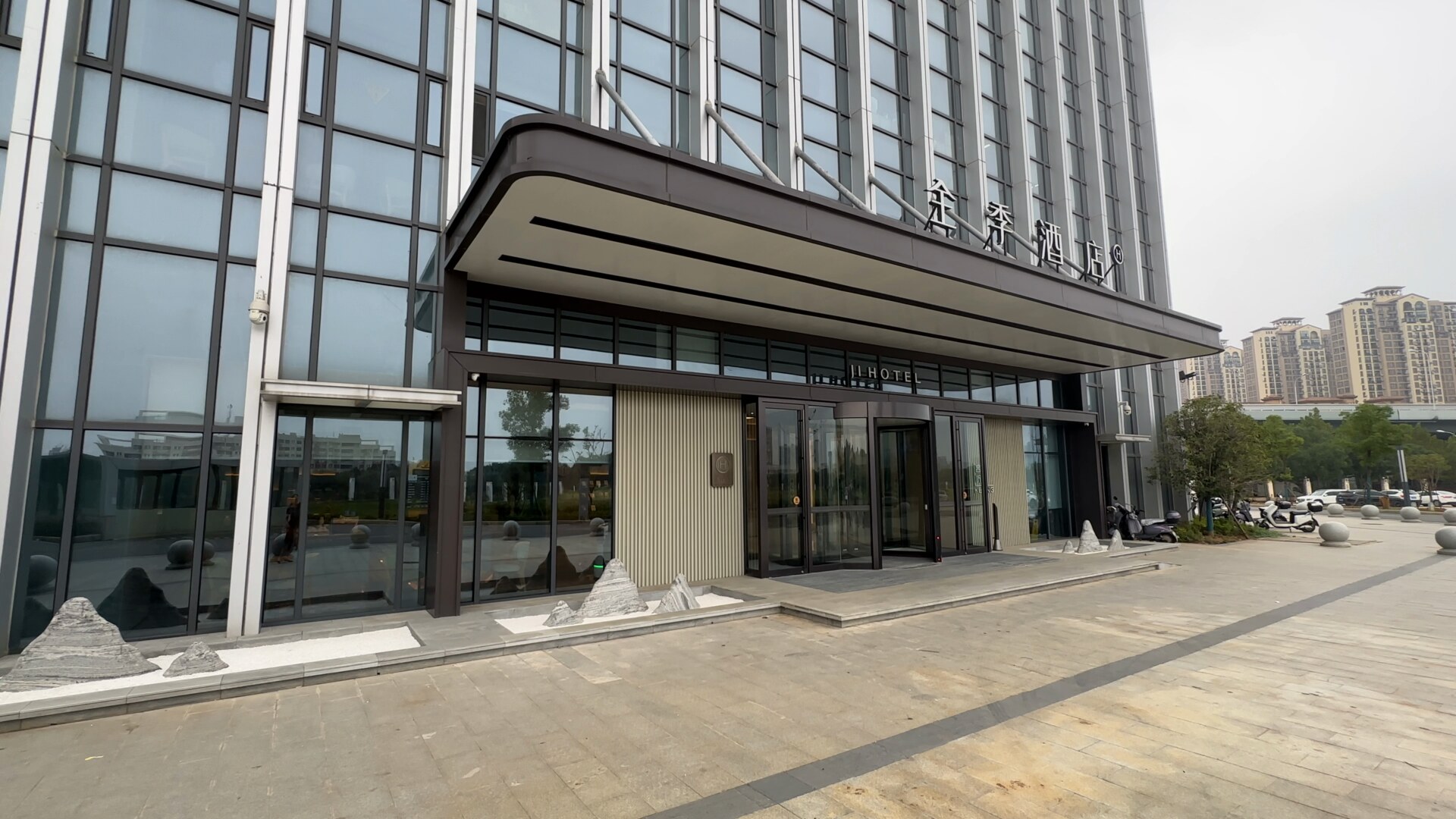} & 
\includegraphics[width=\lw,height=\lh]{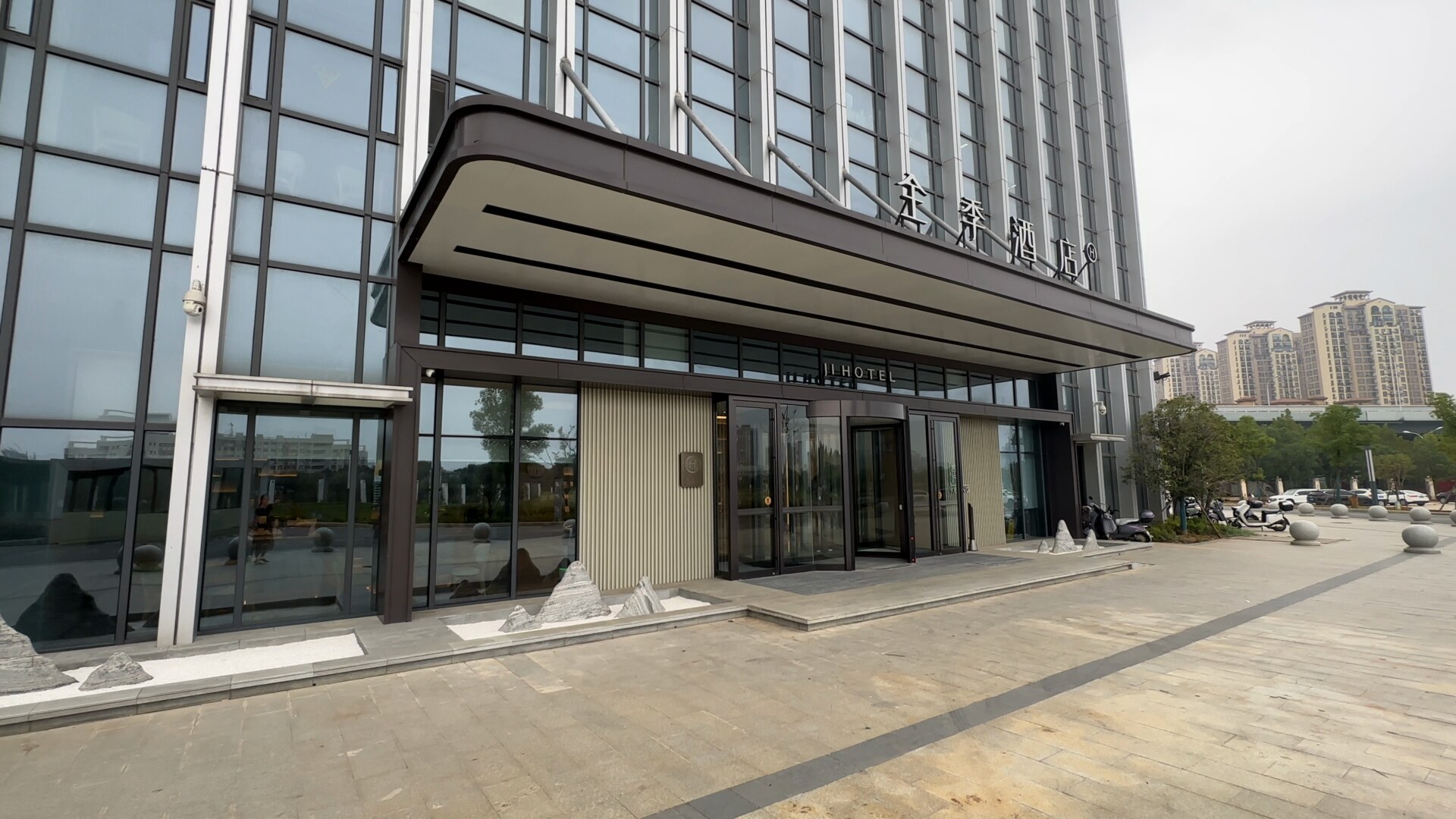} & 
\includegraphics[width=\lw,height=\lh]{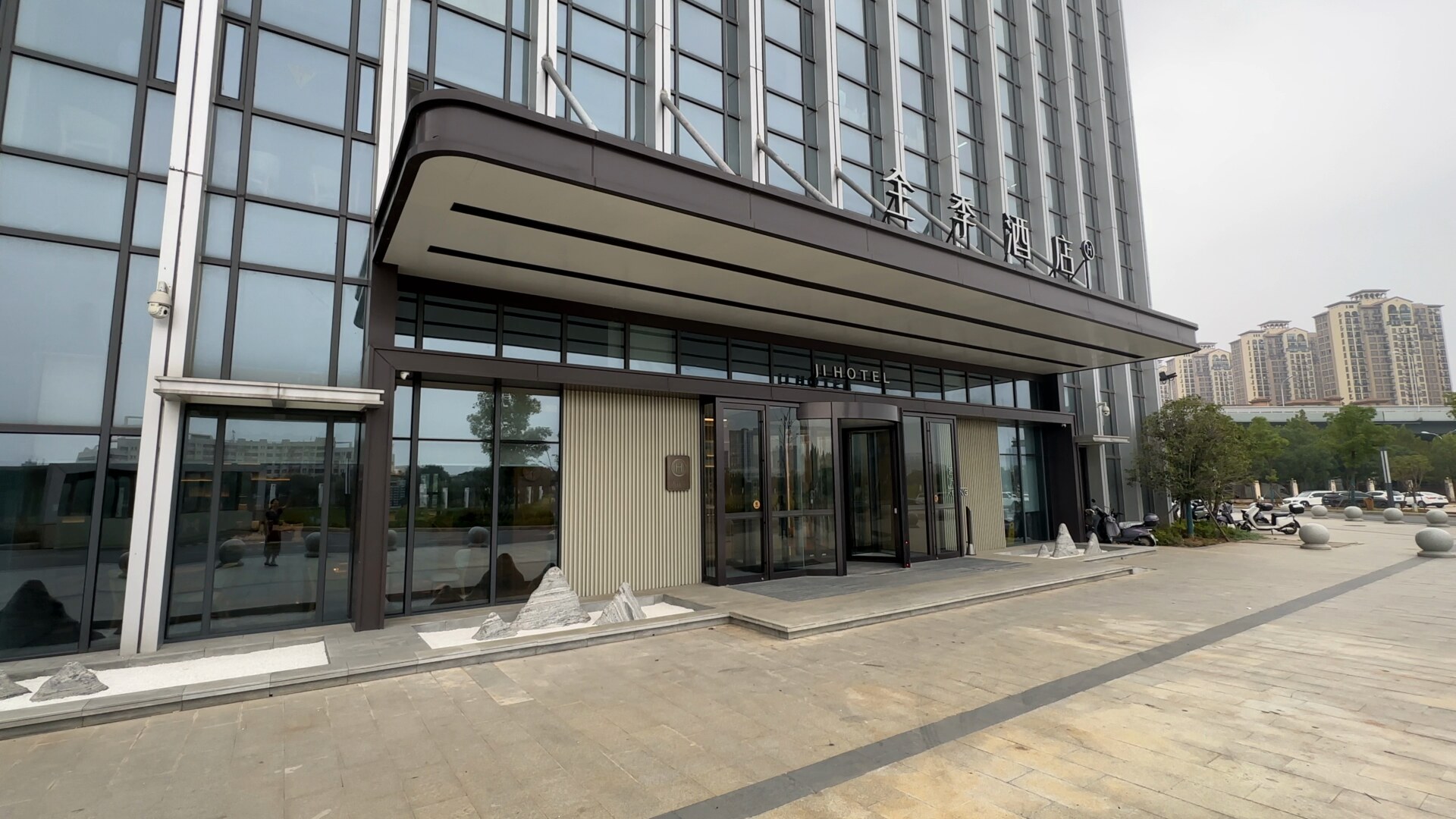} & 
\lframe{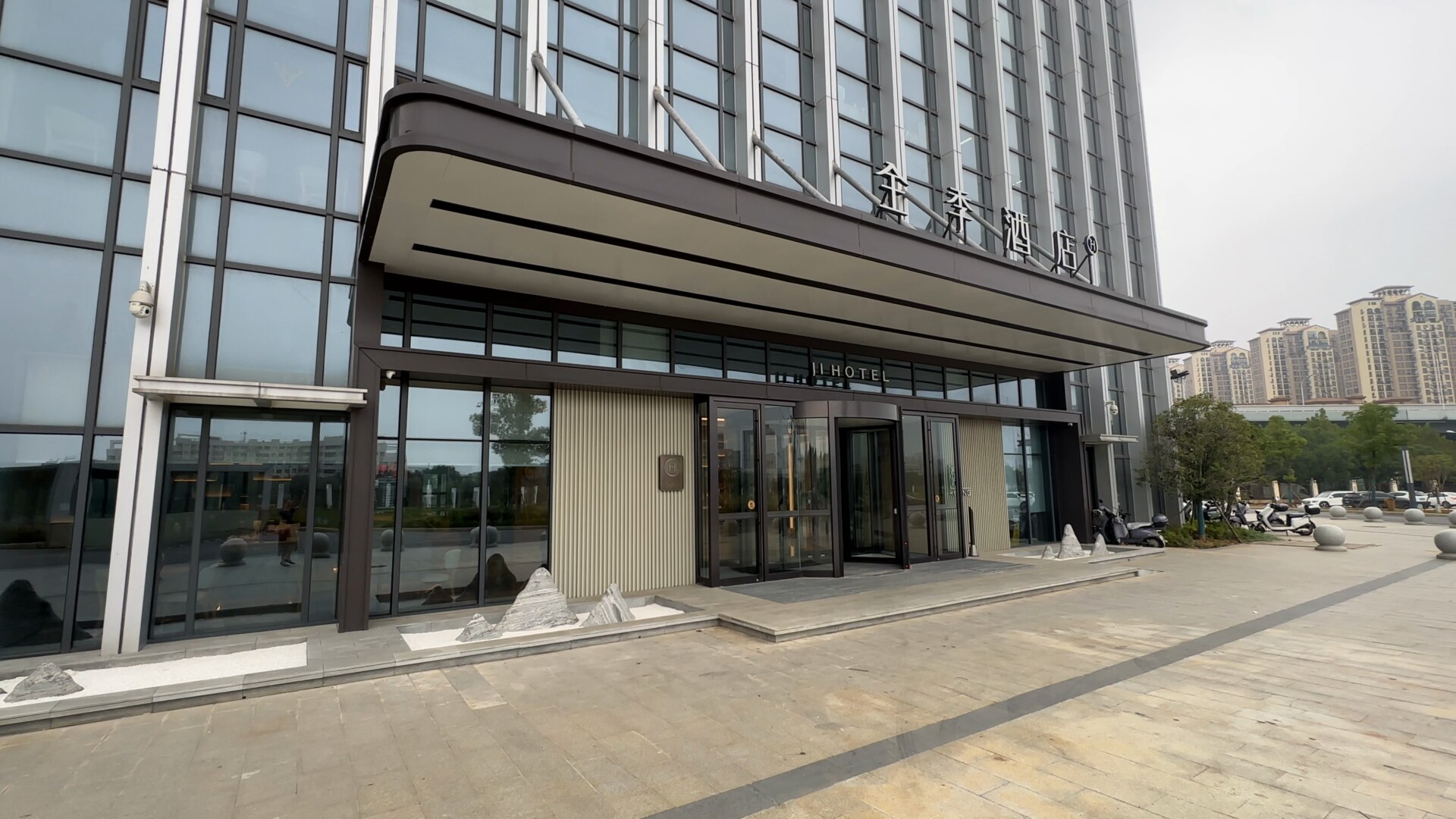} &
\rframe{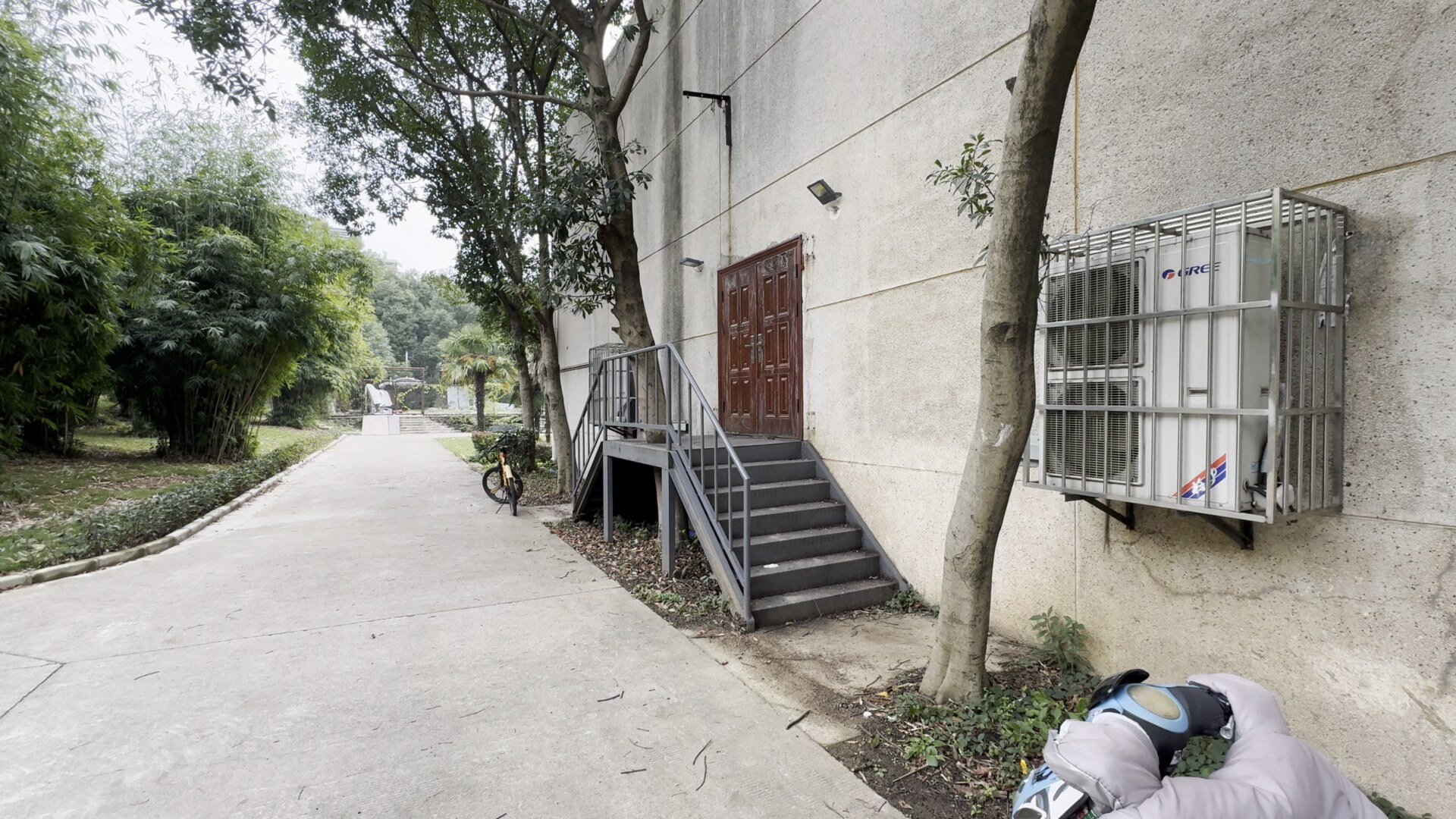} & 
\includegraphics[width=\lw,height=\lh]{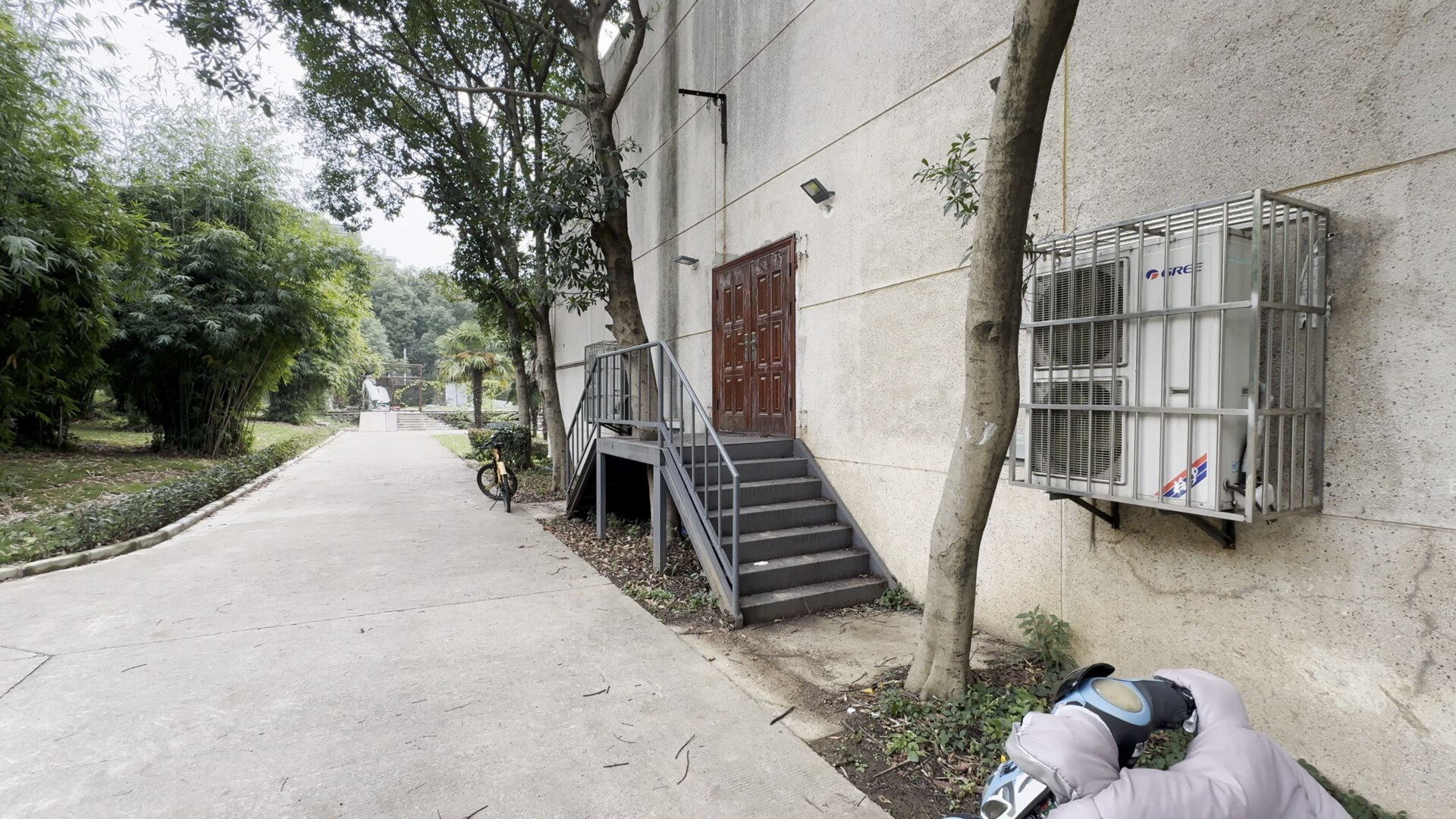} & 
\includegraphics[width=\lw,height=\lh]{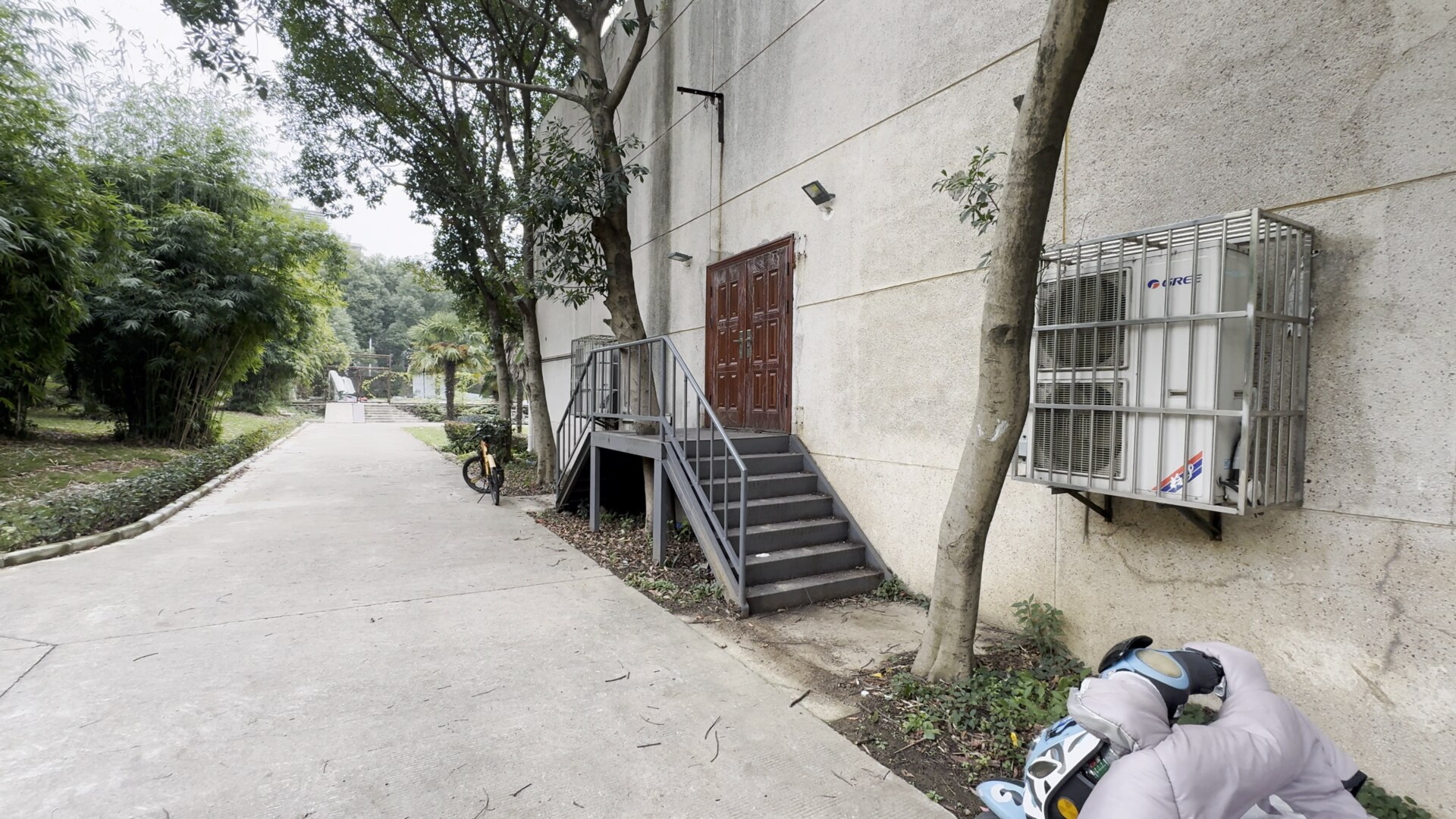} & 
\includegraphics[width=\lw,height=\lh]{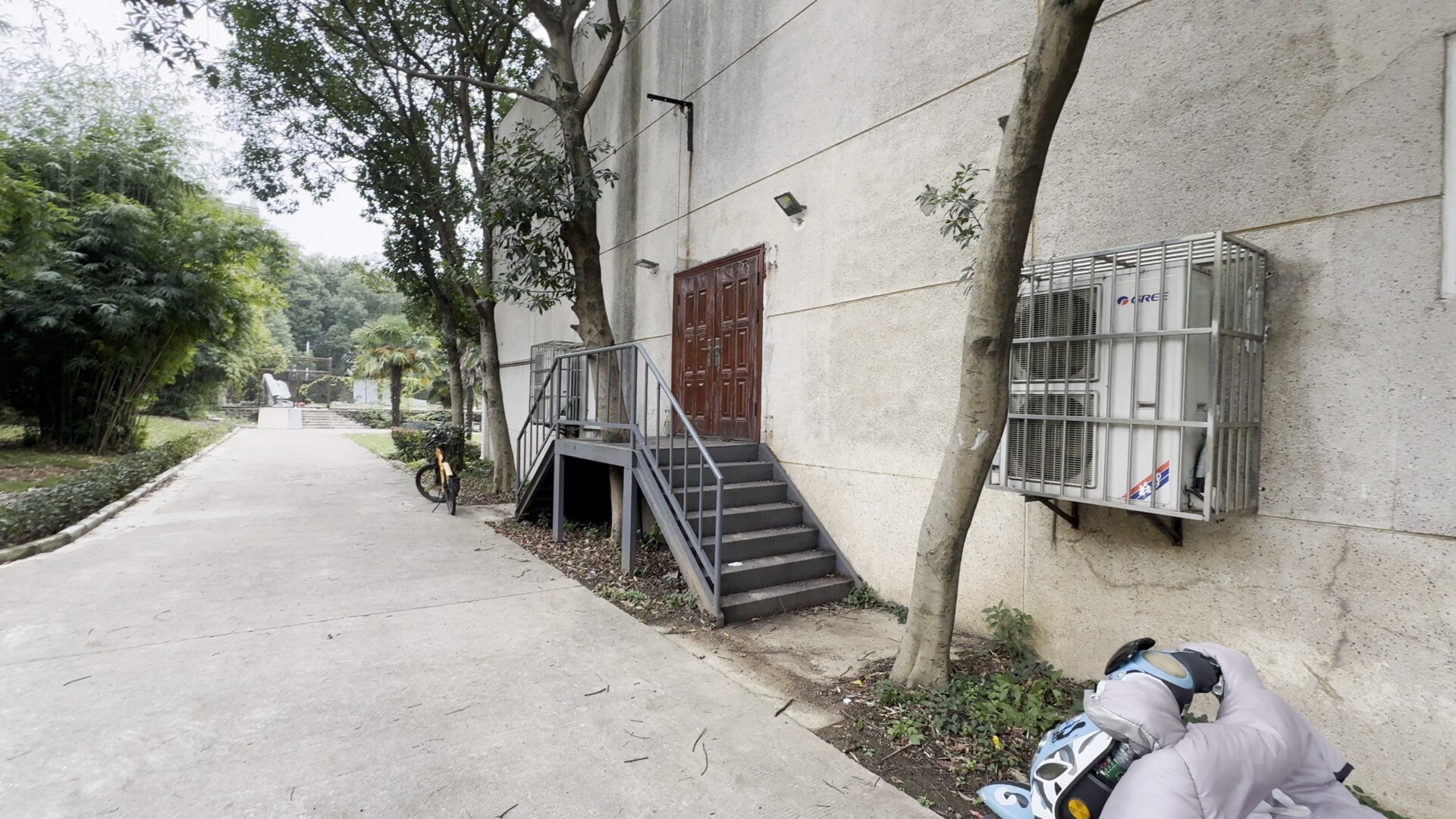} & 
\includegraphics[width=\lw,height=\lh]{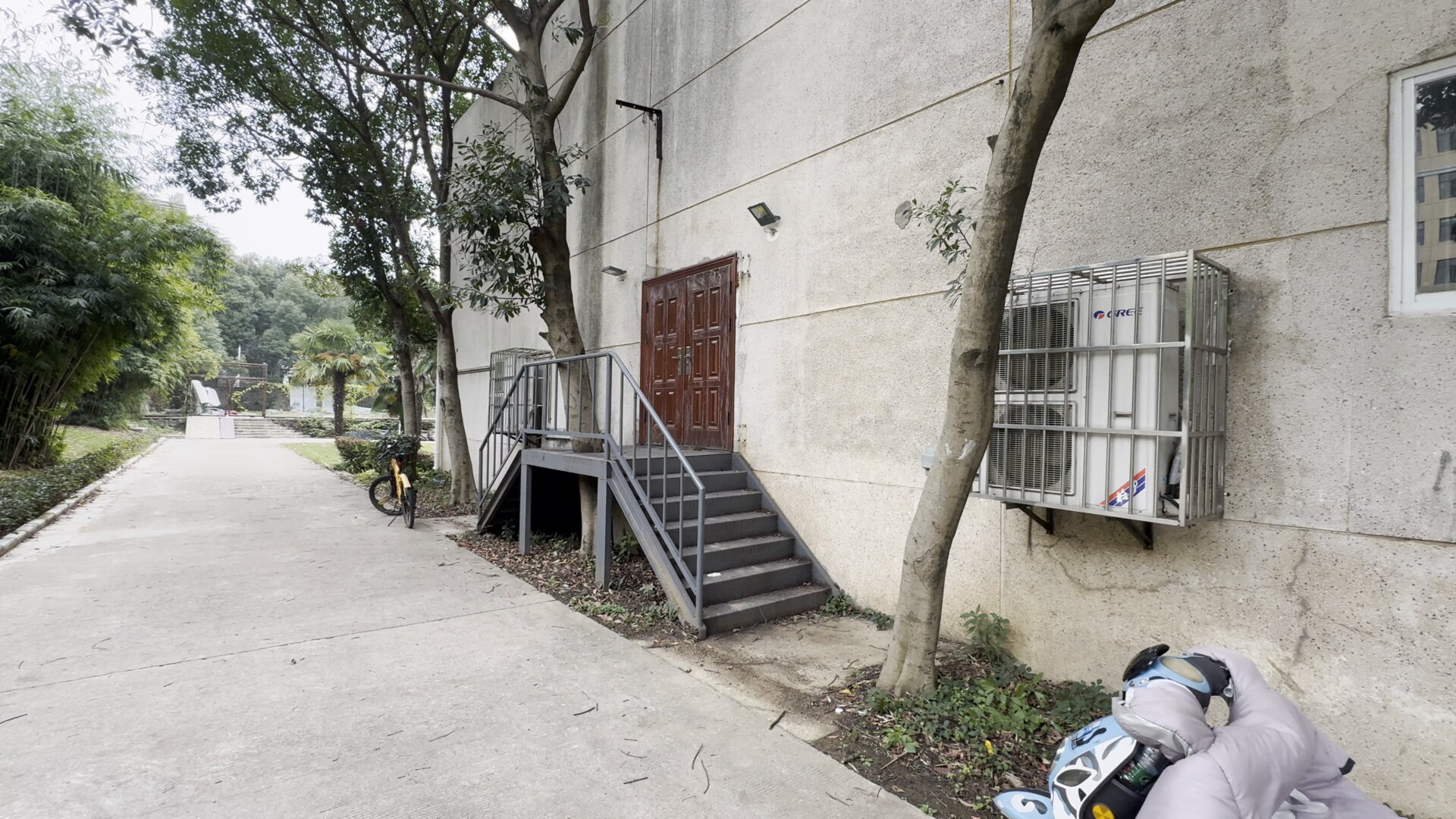} \\
\noalign{\vskip 2pt}
\hdashline
\end{tabular}

\def\lw{0.0999\linewidth}

\begin{tabular}{c ccccc@{\hspace{-2pt}} @{\hspace{-2pt}}ccccc}
\rotatebox{90}{\footnotesize{VD~\cite{kim2025videofrom3d}}}\hspace{1mm} &
\includegraphics[width=\lw,height=\lh]{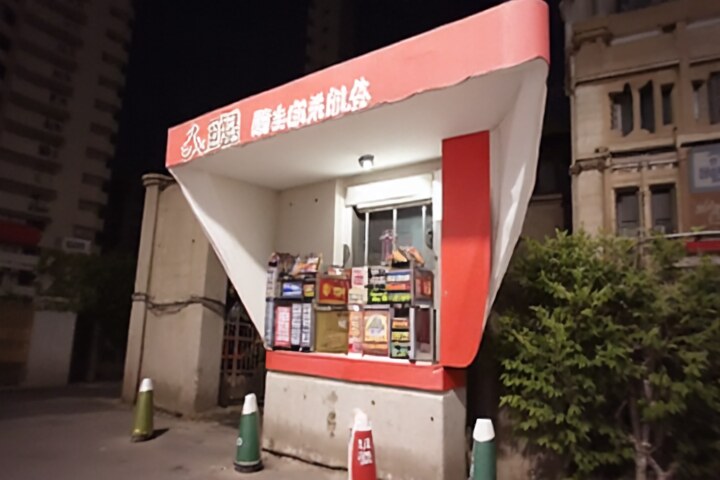} & 
\includegraphics[width=\lw,height=\lh]{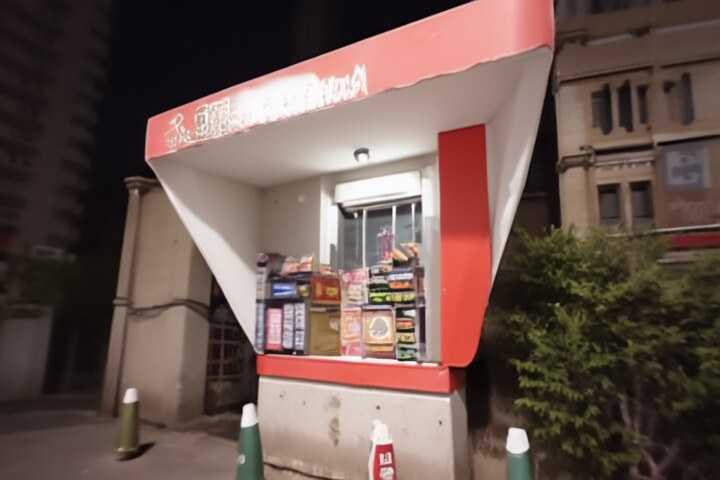} & 
\includegraphics[width=\lw,height=\lh]{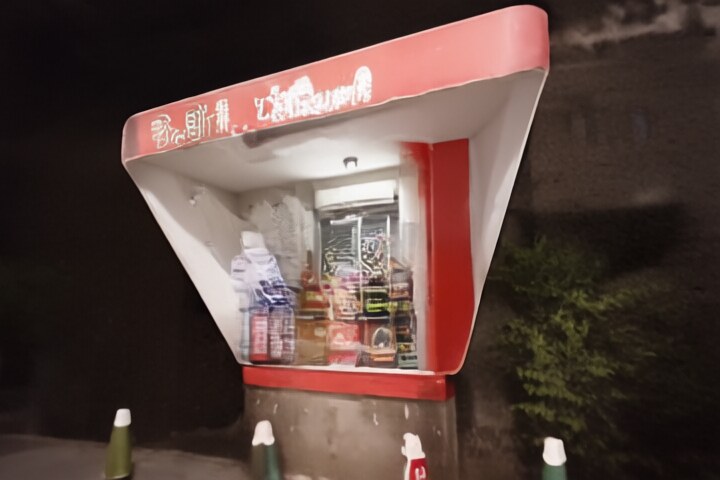} & 
\includegraphics[width=\lw,height=\lh]{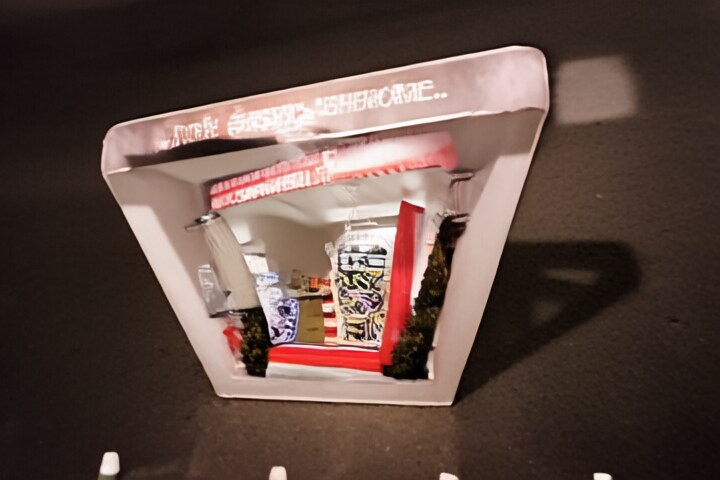} & 
\lframe{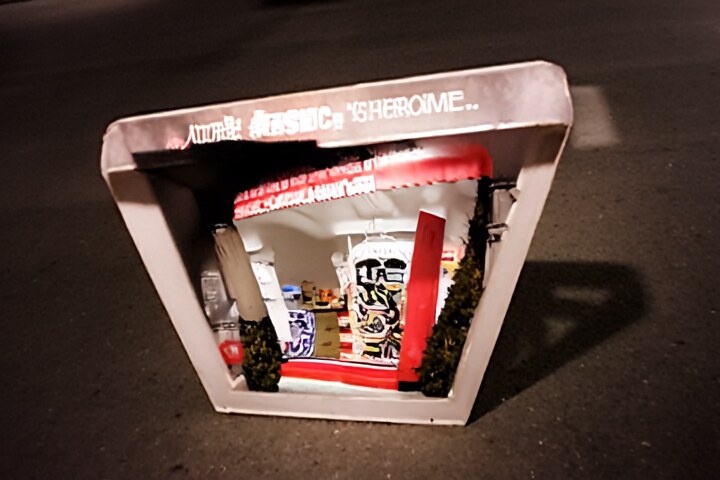} &
\rframe{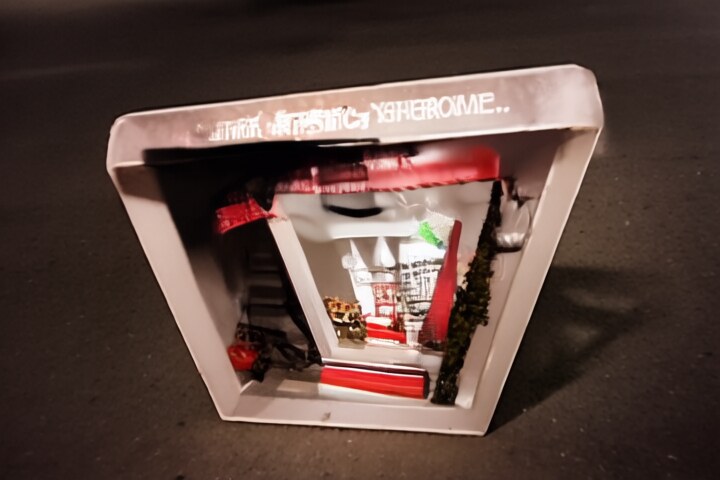} & 
\includegraphics[width=\lw,height=\lh]{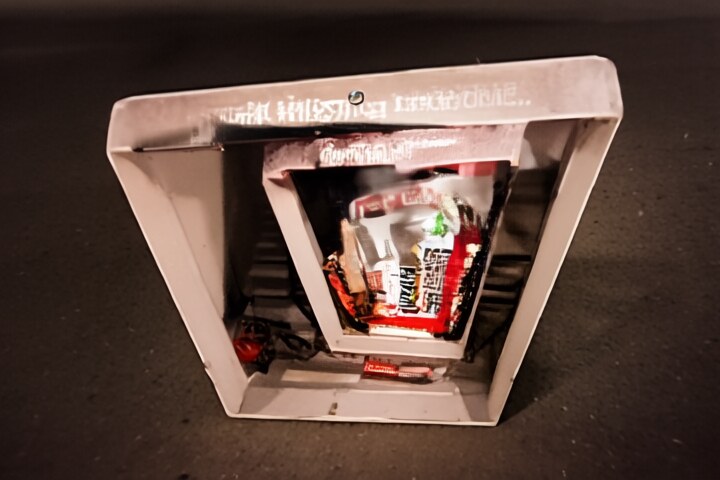} & 
\includegraphics[width=\lw,height=\lh]{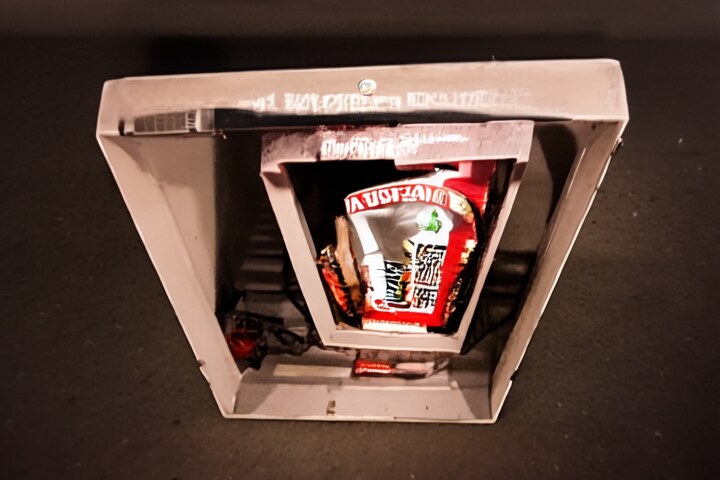} & 
\includegraphics[width=\lw,height=\lh]{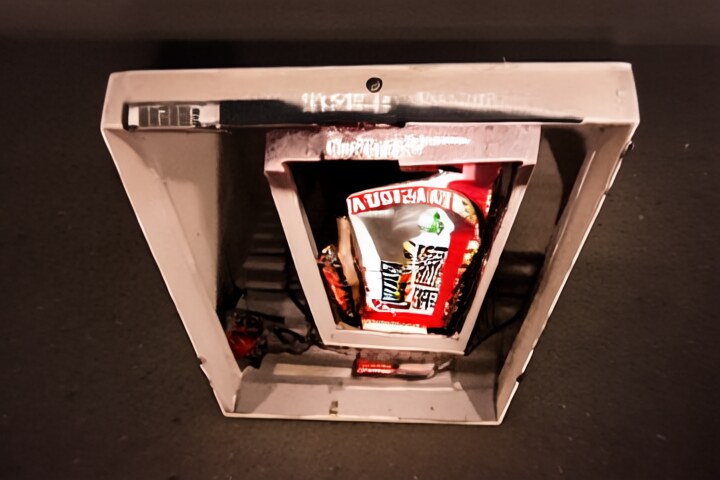} & 
\includegraphics[width=\lw,height=\lh]{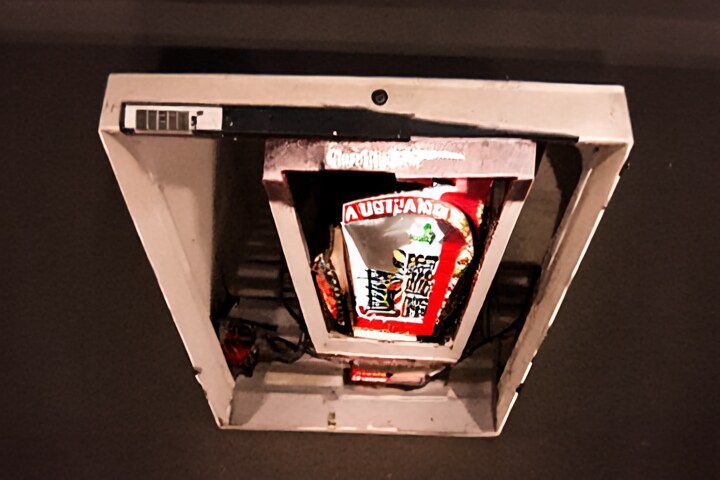} \\ \noalign{\vskip \sskip}
\rotatebox{90}{\hspace{2mm}\footnotesize{Ours}}\hspace{1mm} &
\includegraphics[width=\lw,height=\lh]{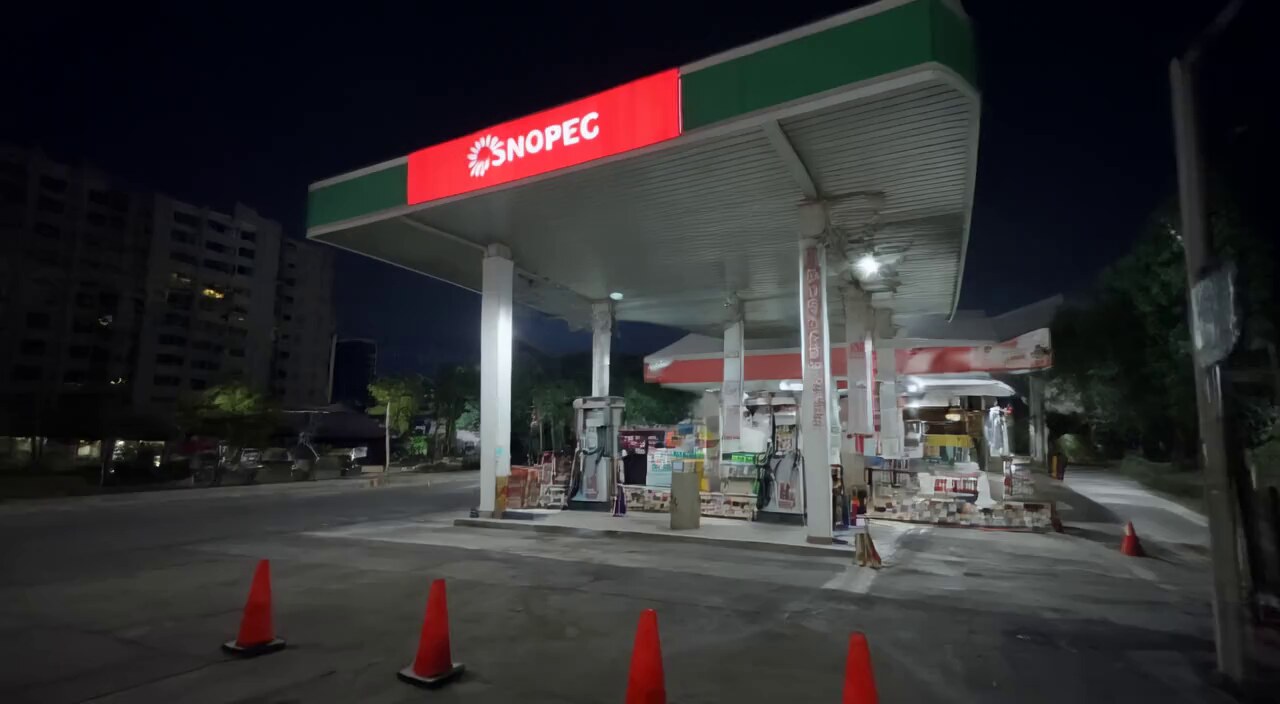} & 
\includegraphics[width=\lw,height=\lh]{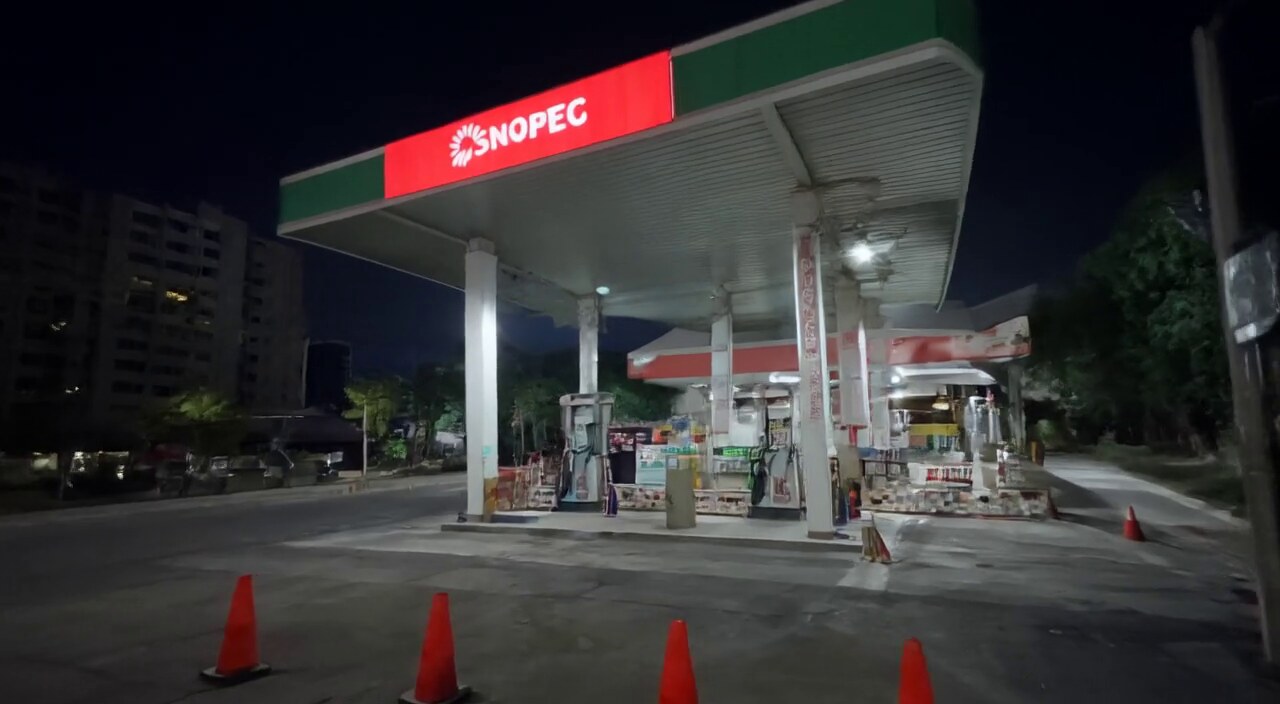} & 
\includegraphics[width=\lw,height=\lh]{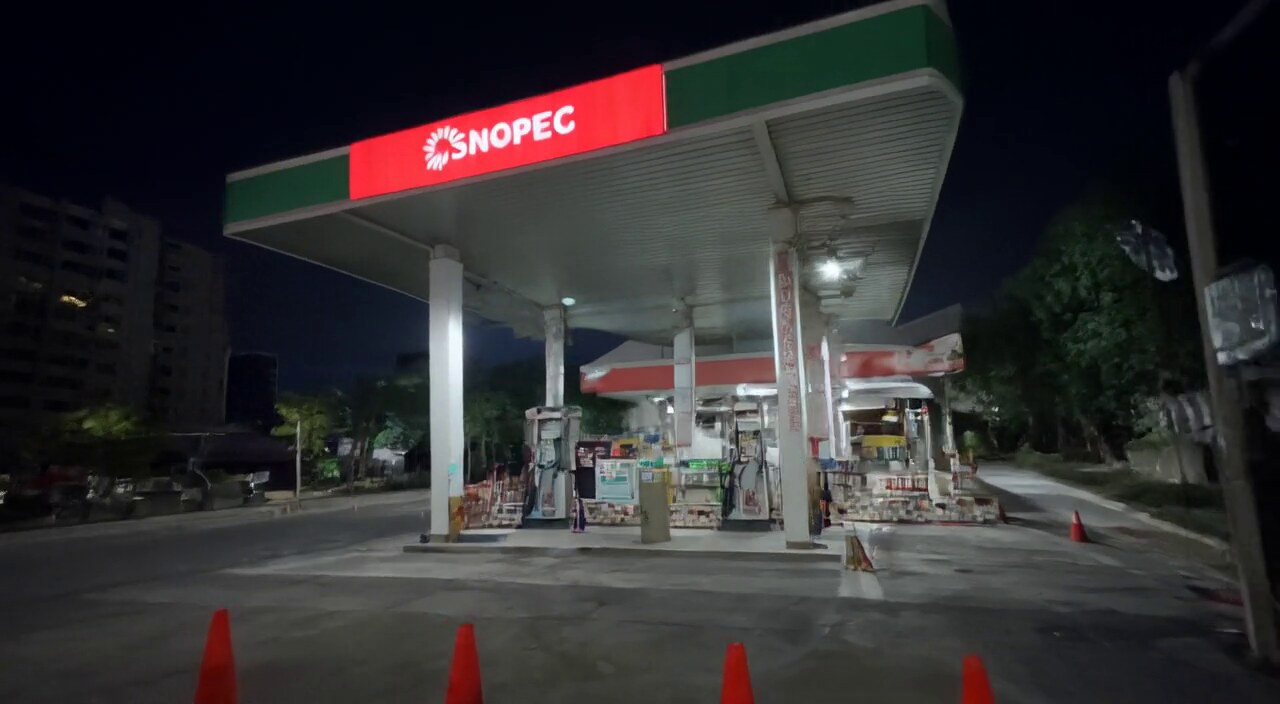} & 
\includegraphics[width=\lw,height=\lh]{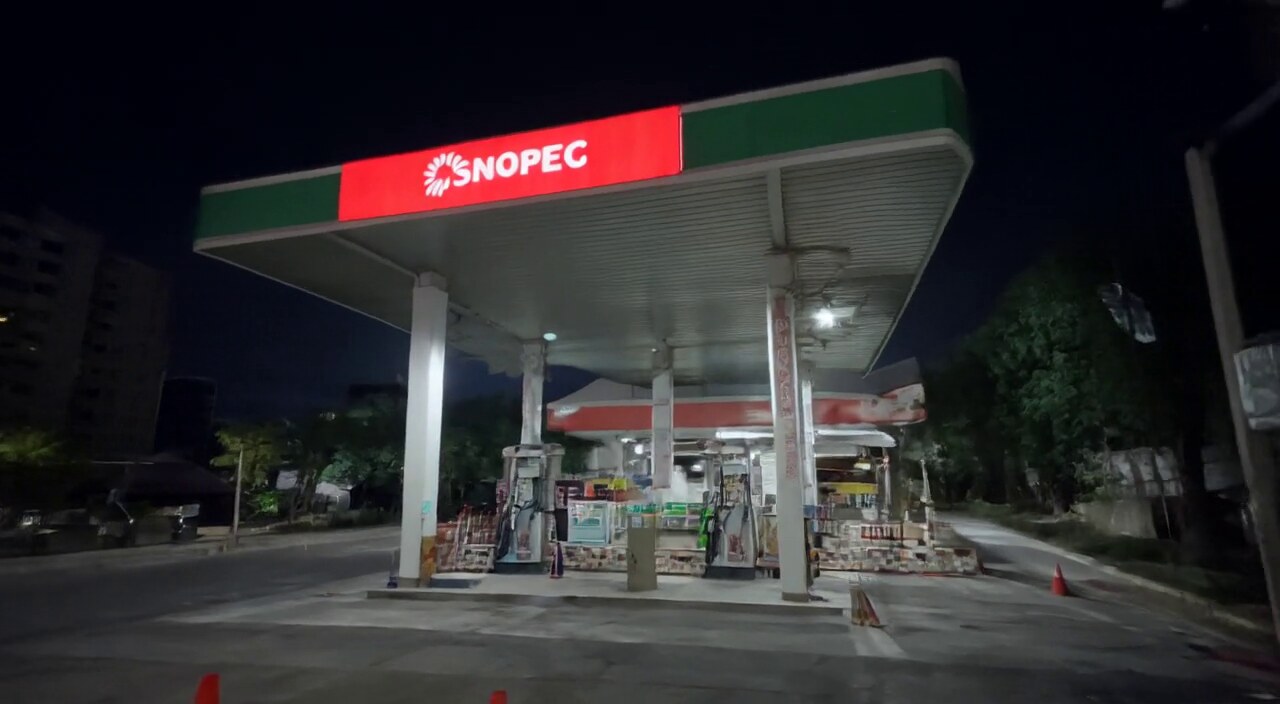} & 
\lframe{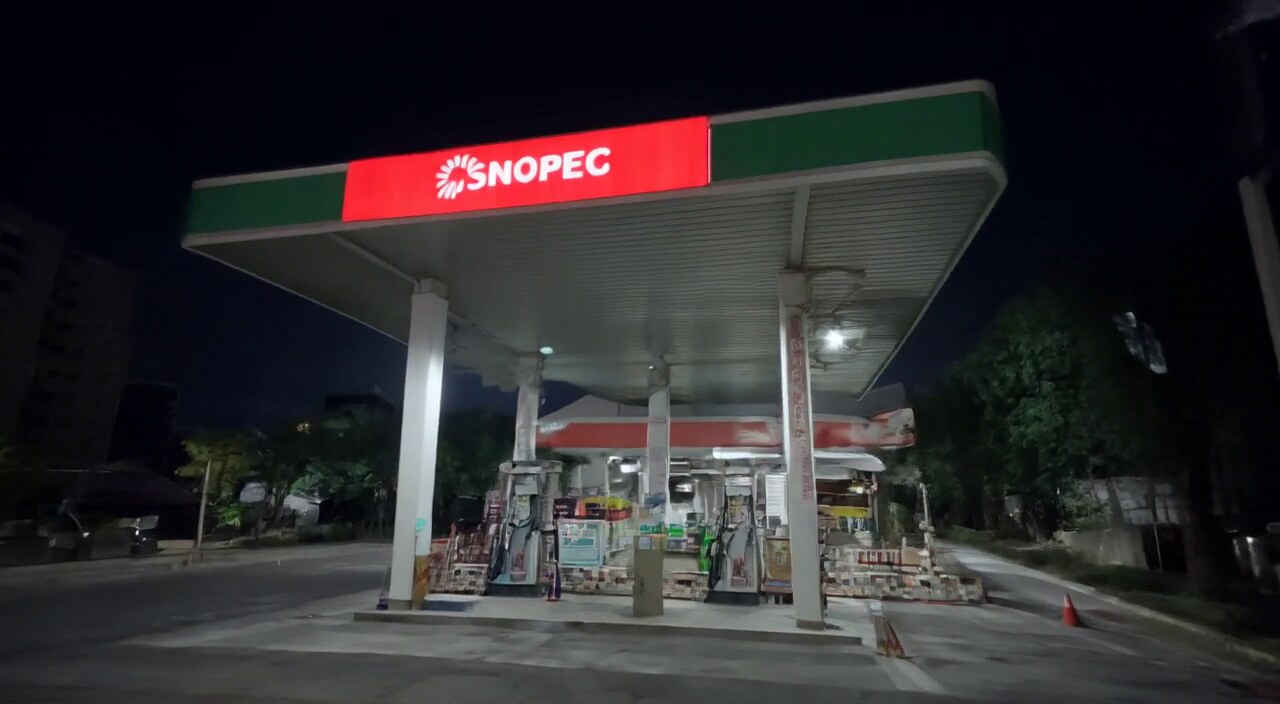} &
\rframe{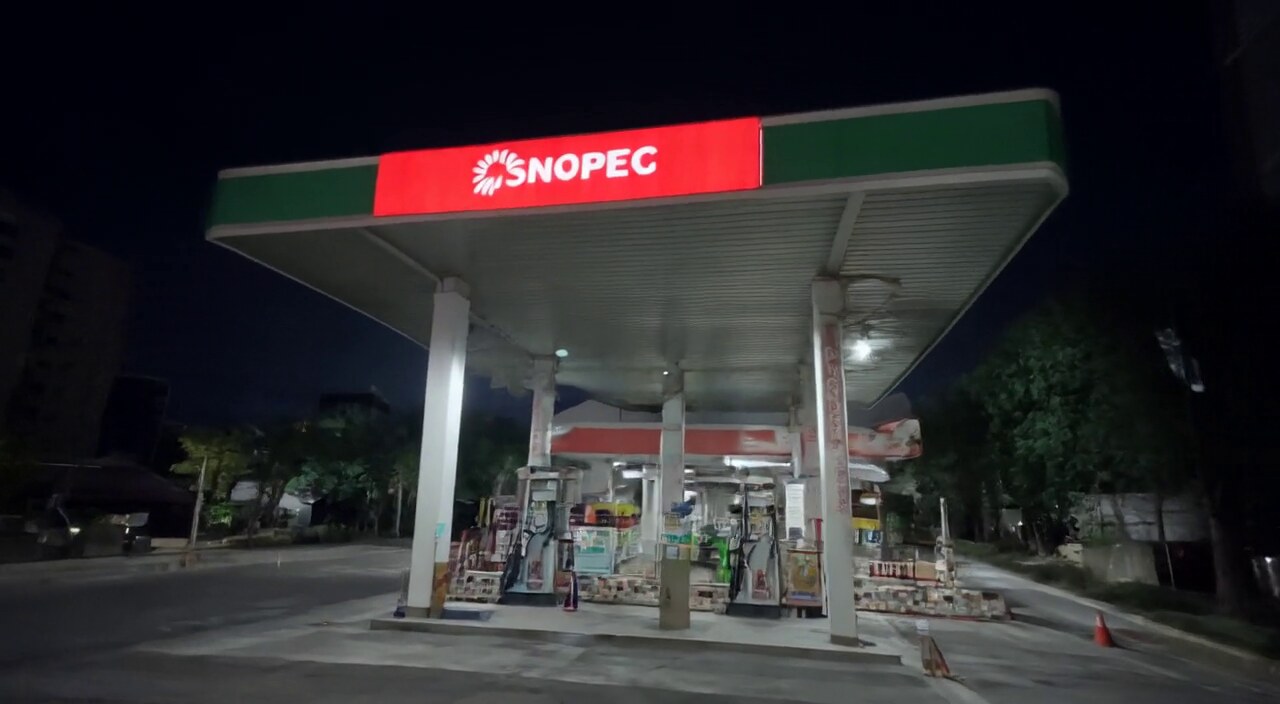} & 
\includegraphics[width=\lw,height=\lh]{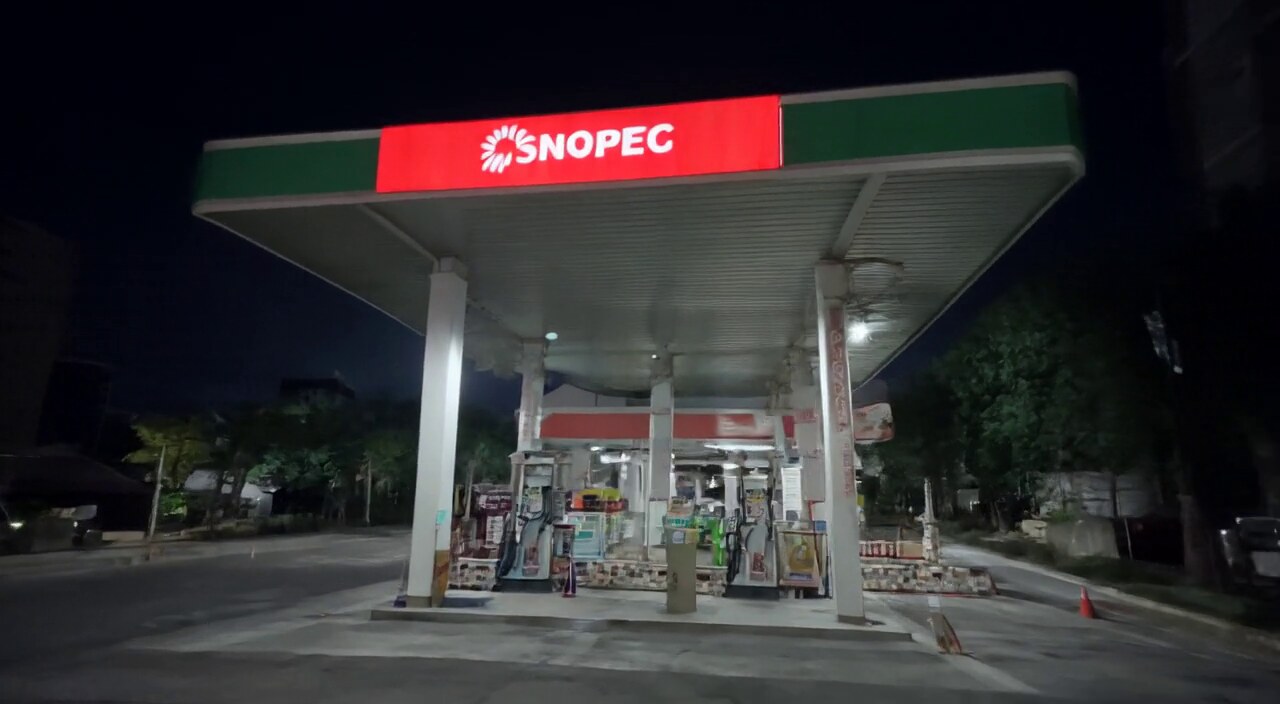} & 
\includegraphics[width=\lw,height=\lh]{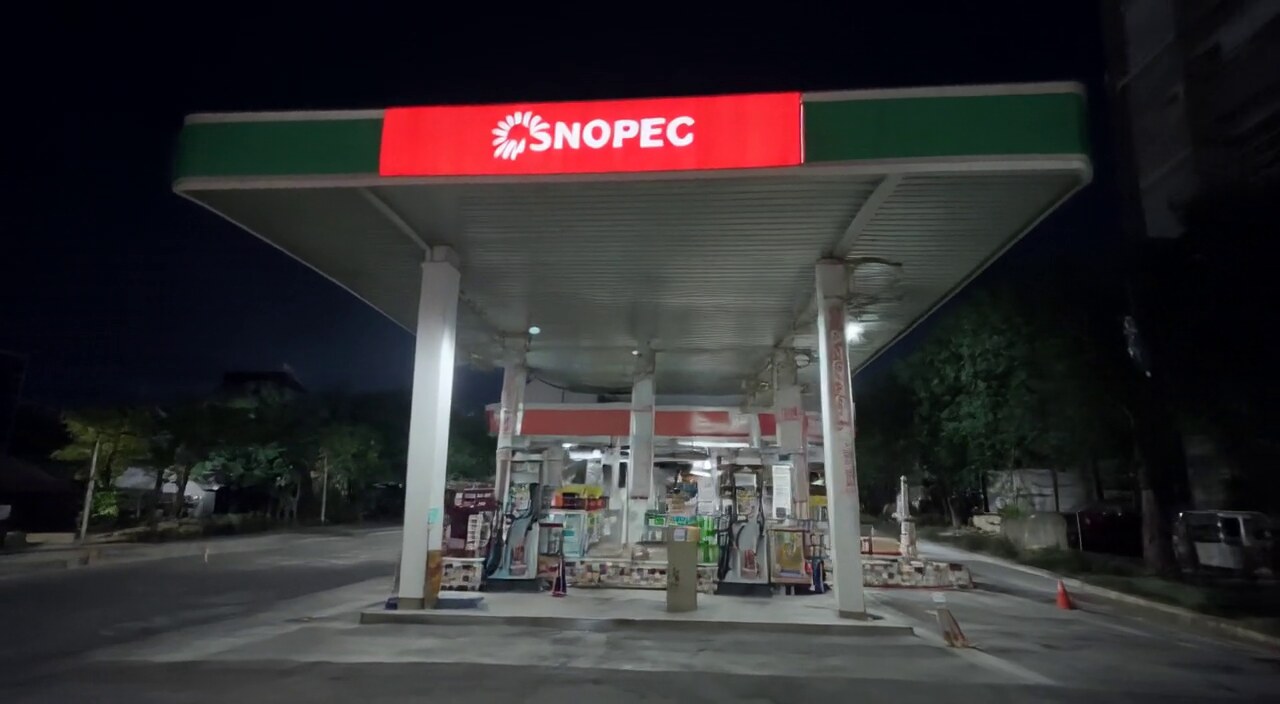} & 
\includegraphics[width=\lw,height=\lh]{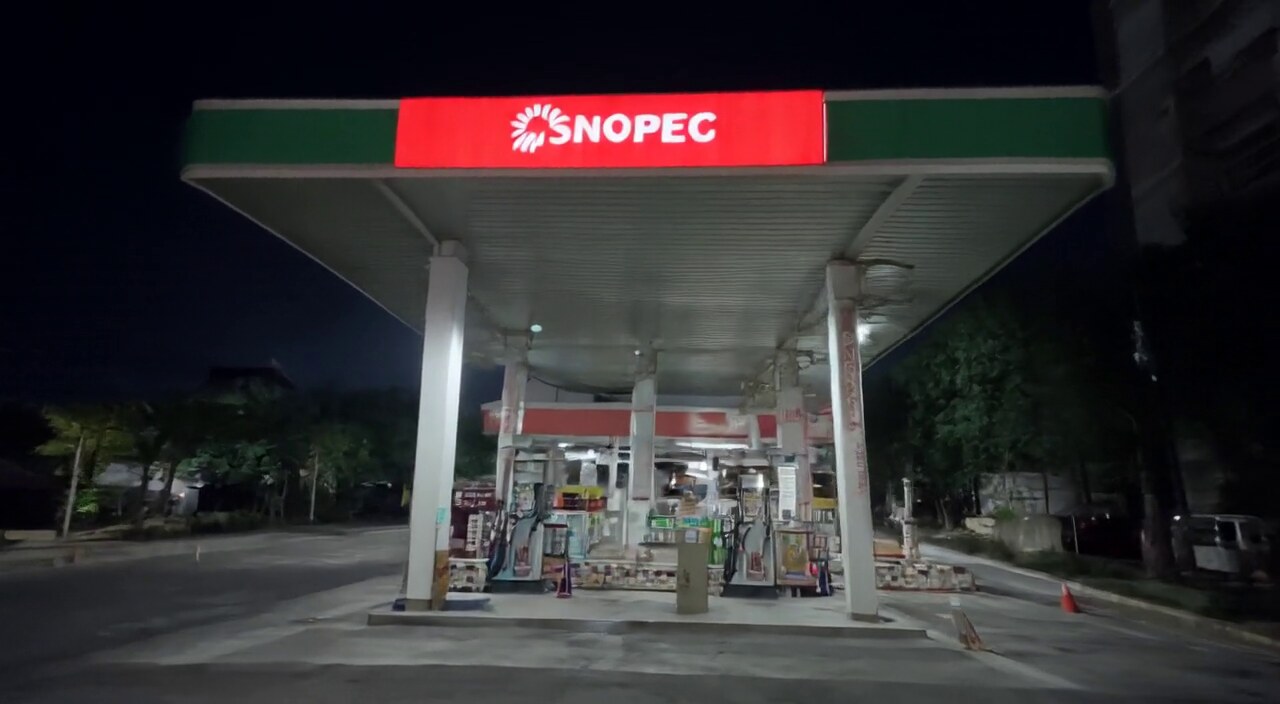} & 
\includegraphics[width=\lw,height=\lh]{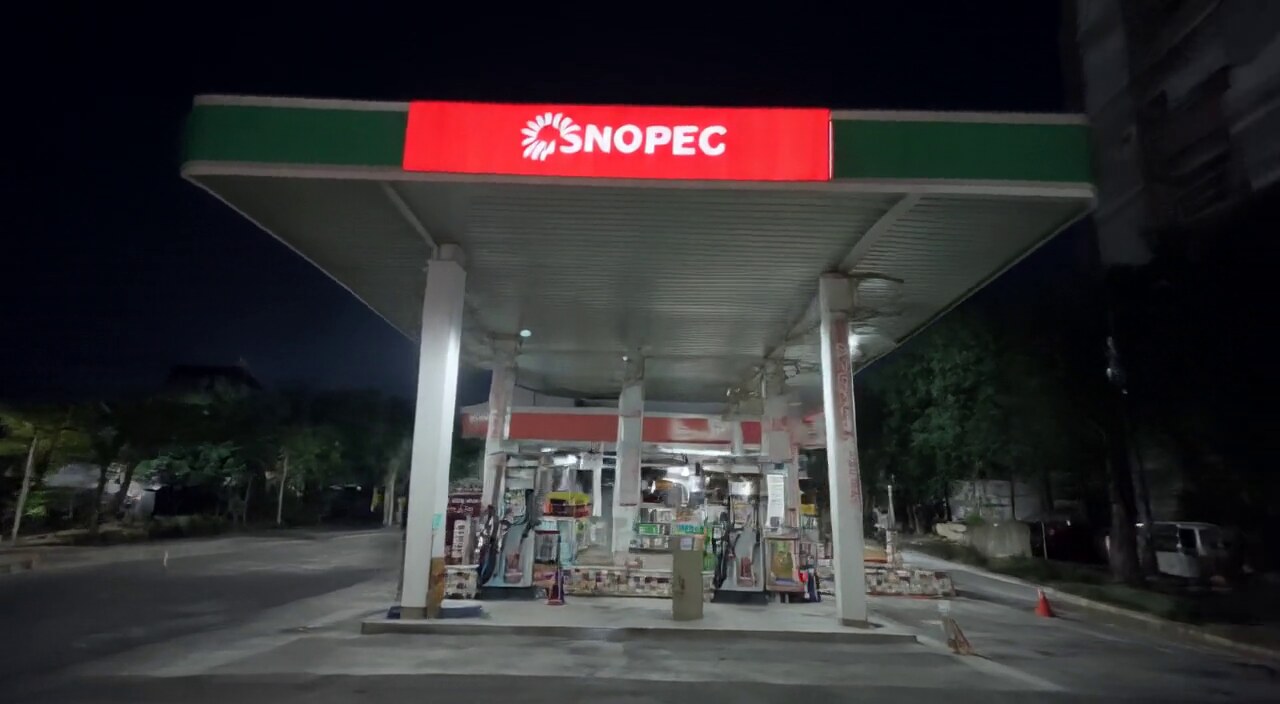} \\ \noalign{\vskip \sskip}

\rotatebox{90}{\hspace{2mm}\footnotesize{Ref}}\hspace{1mm} &
\includegraphics[width=\lw,height=\lh]{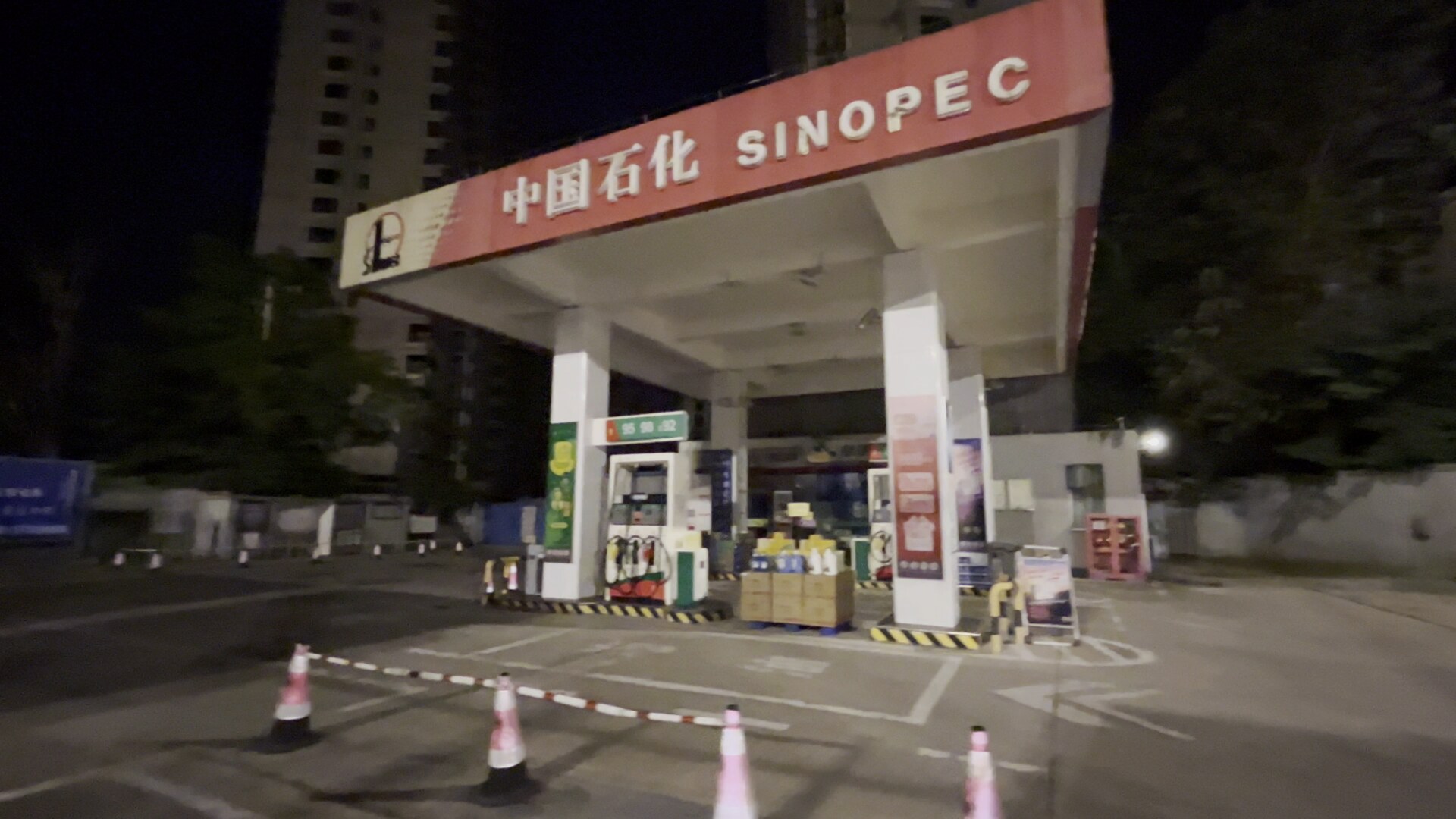} & 
\includegraphics[width=\lw,height=\lh]{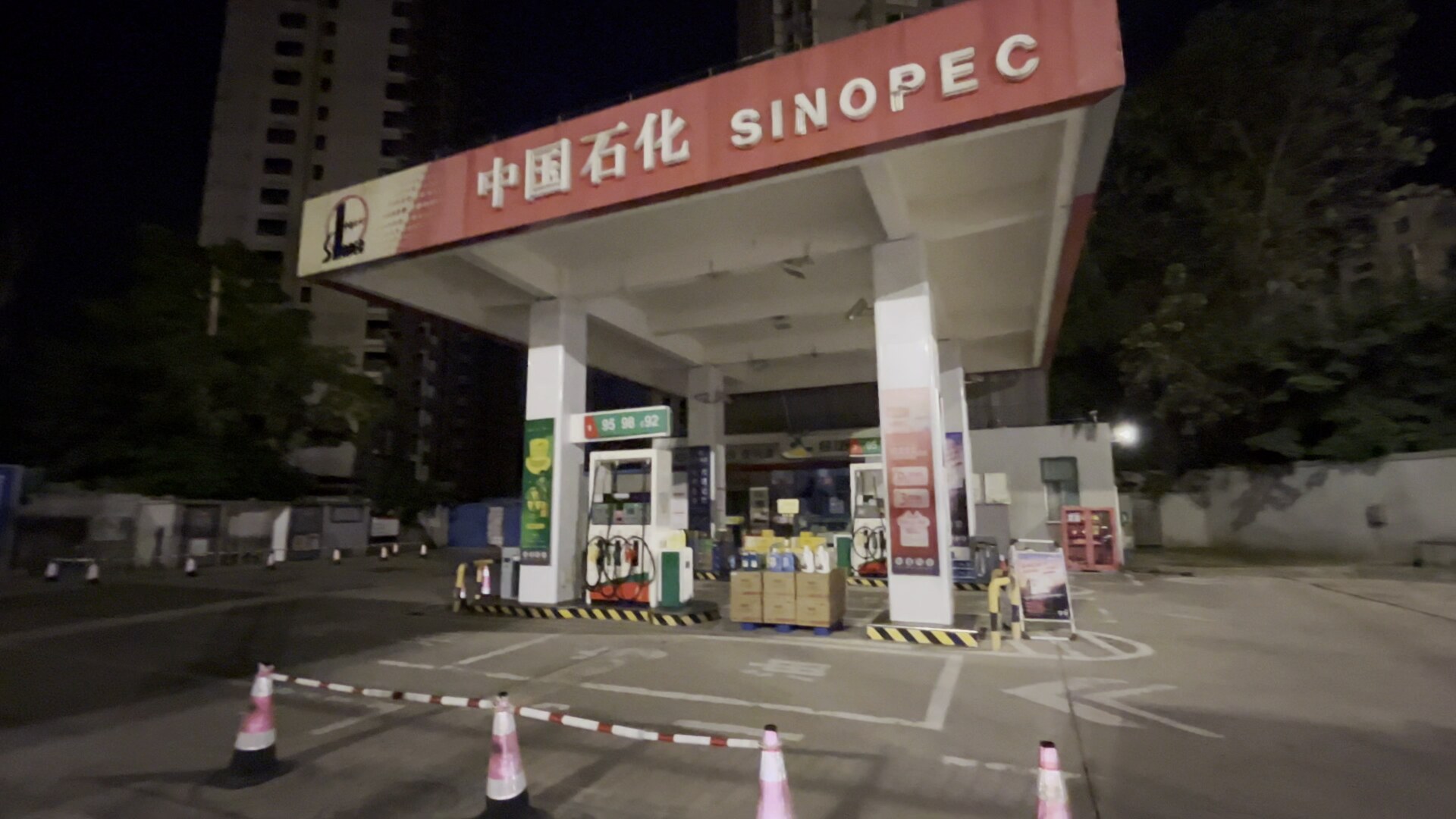} & 
\includegraphics[width=\lw,height=\lh]{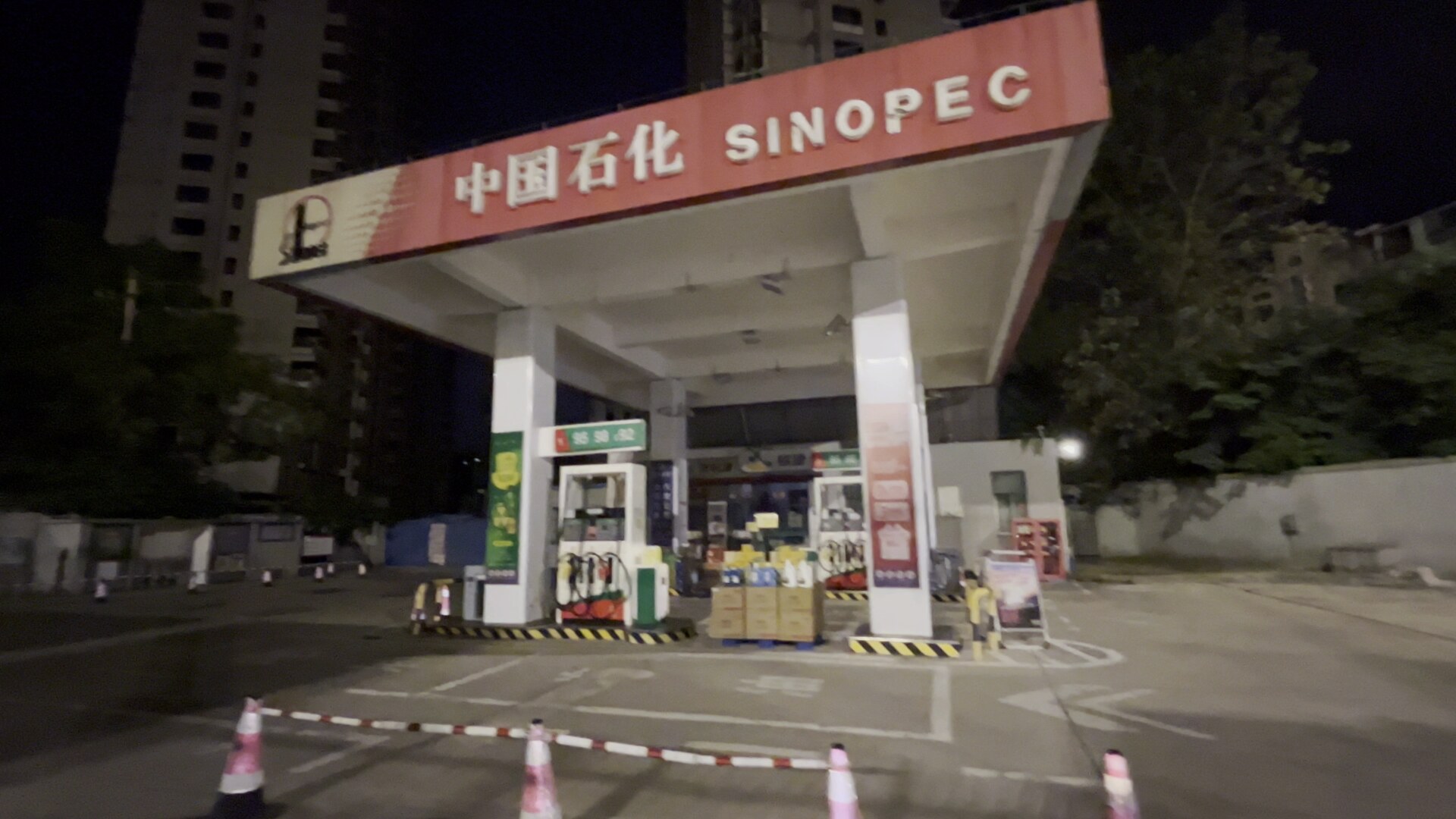} & 
\includegraphics[width=\lw,height=\lh]{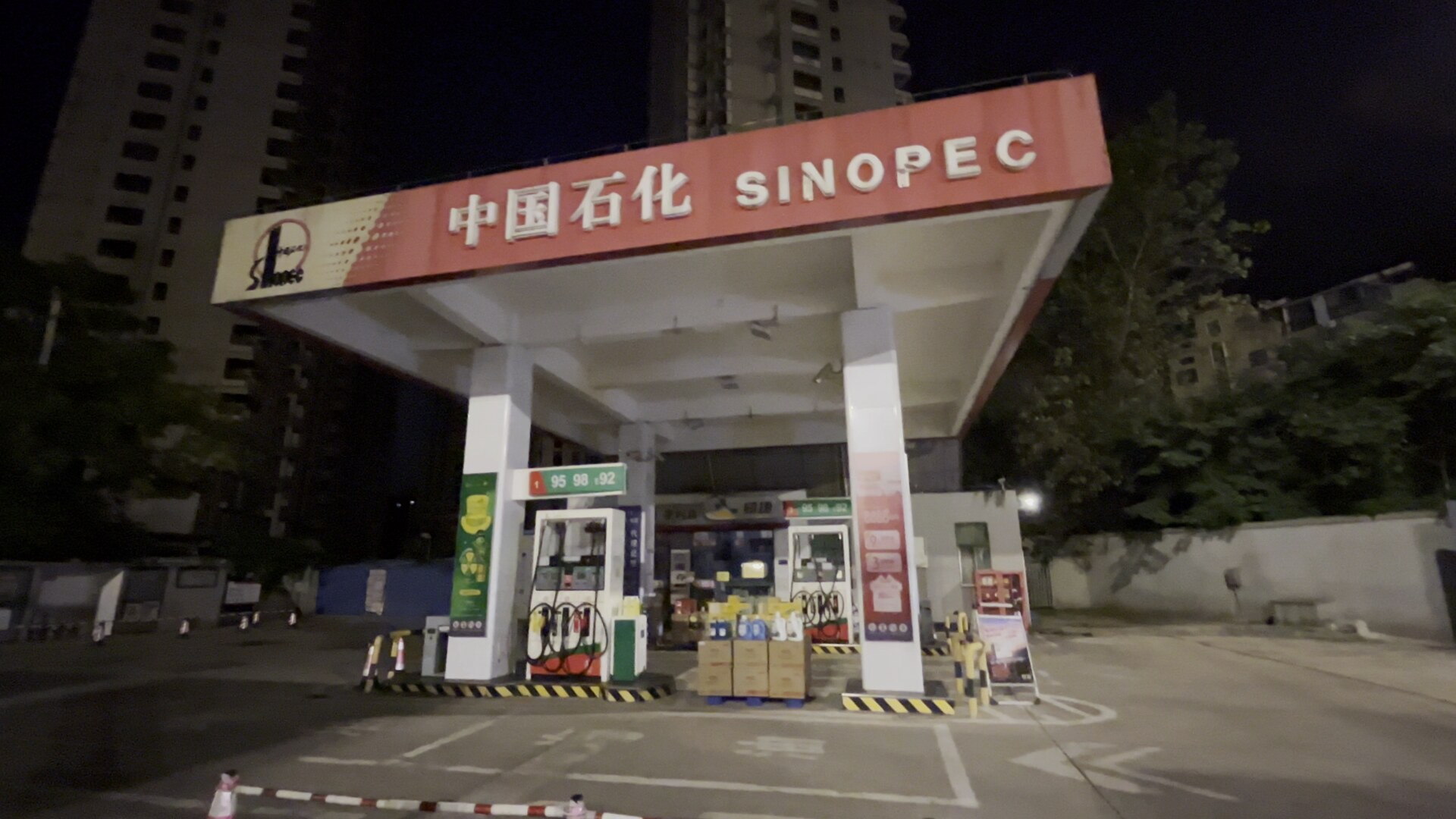} & 
\lframe{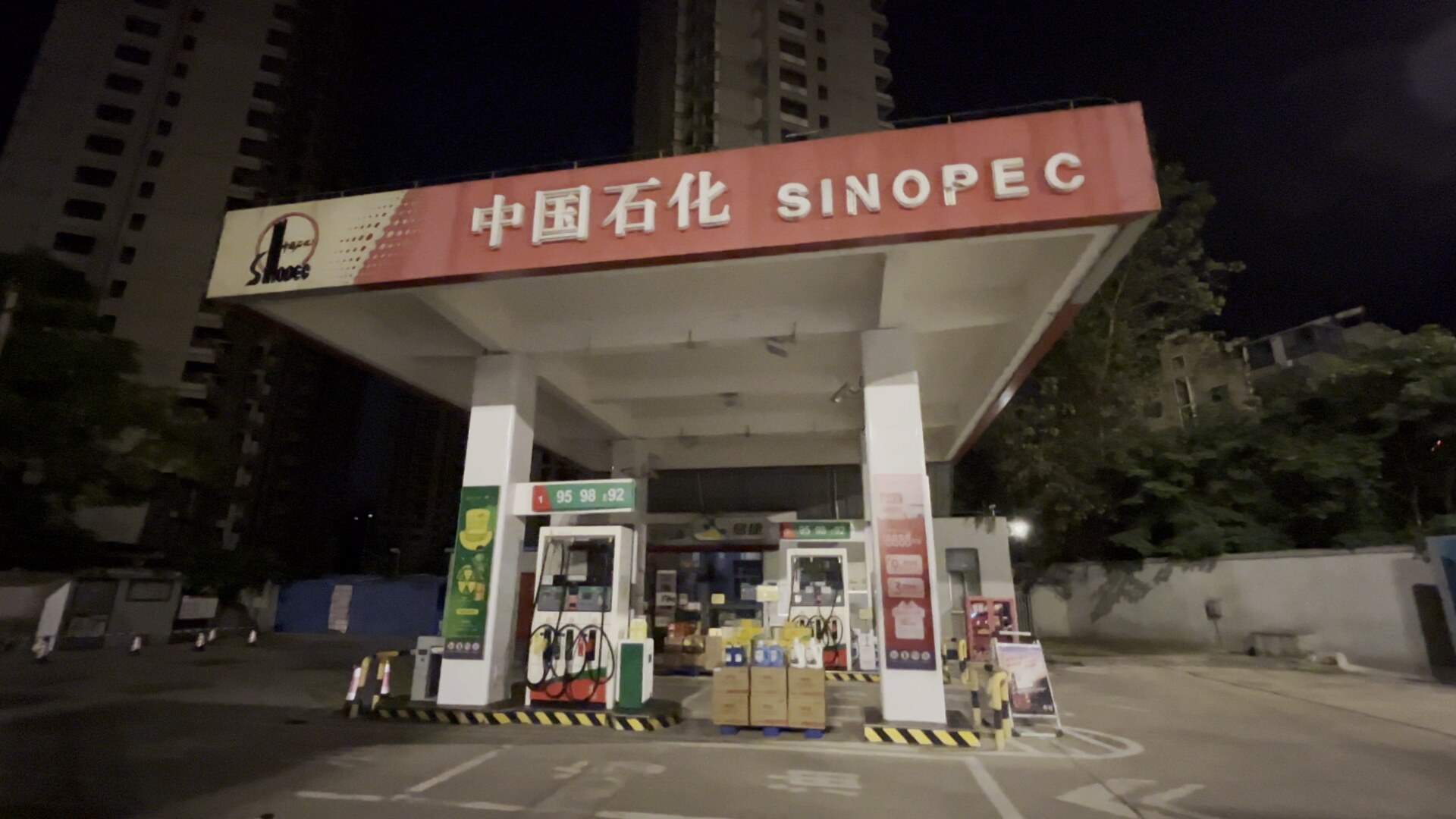} &
\rframe{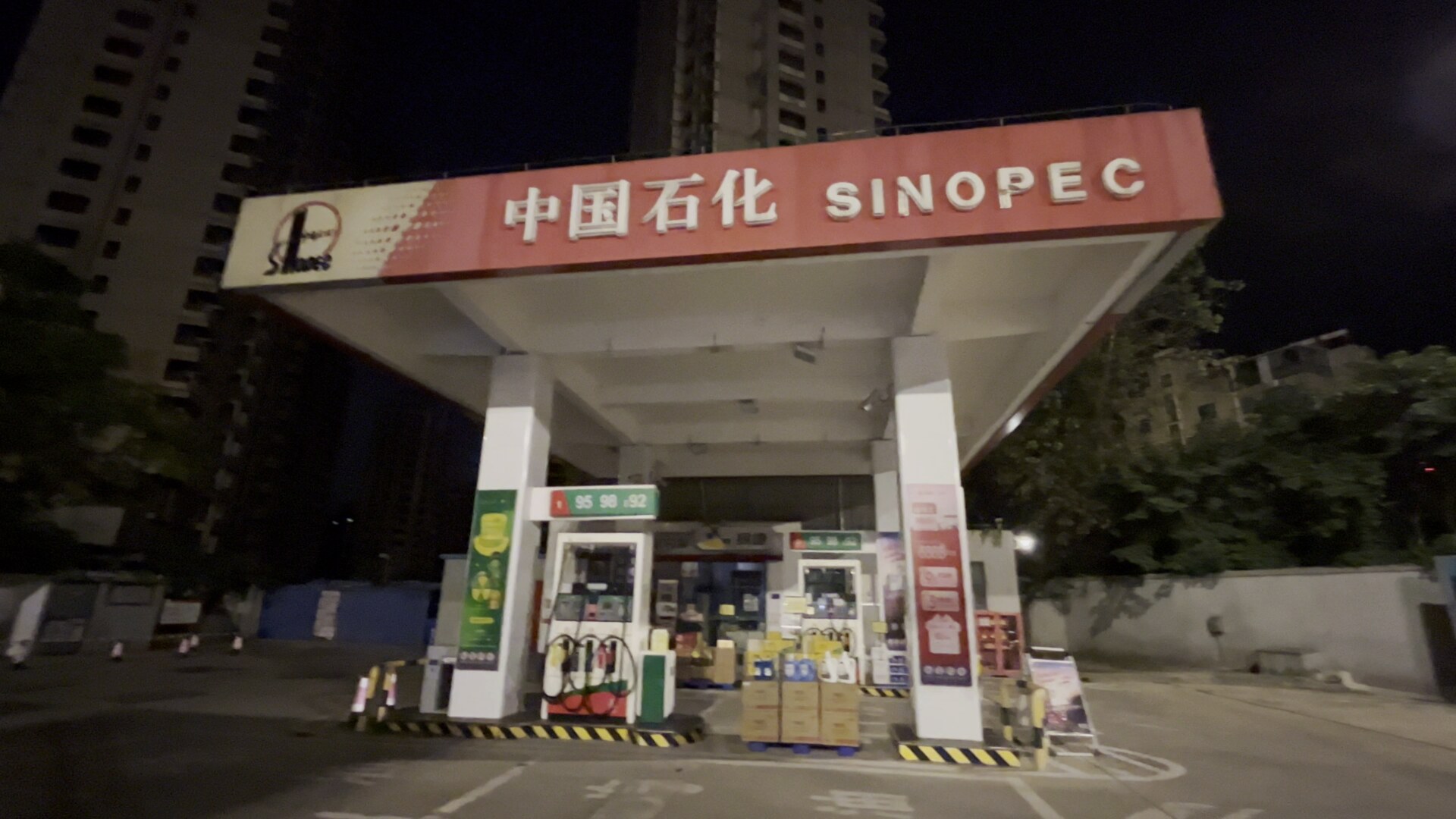} & 
\includegraphics[width=\lw,height=\lh]{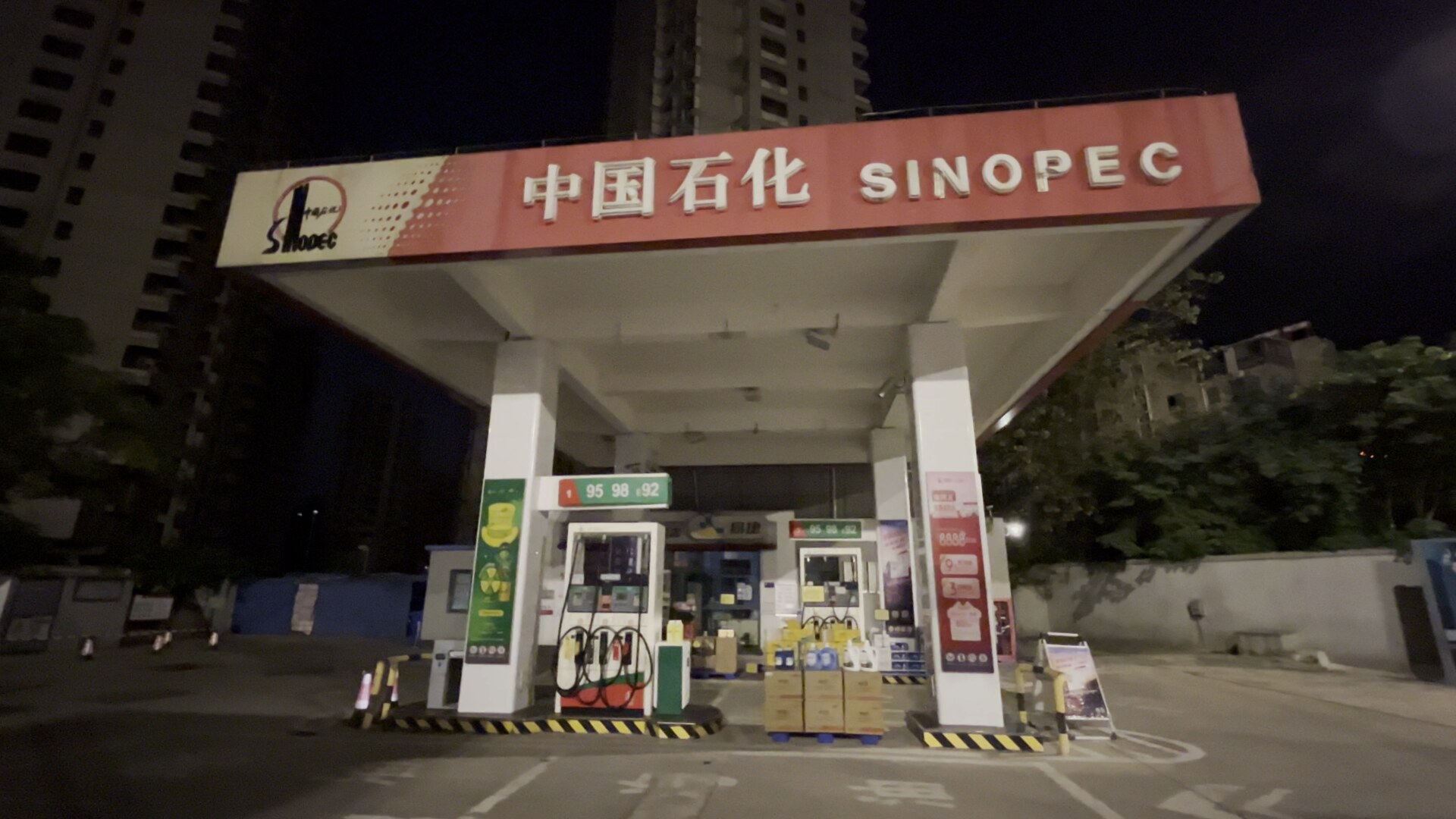} & 
\includegraphics[width=\lw,height=\lh]{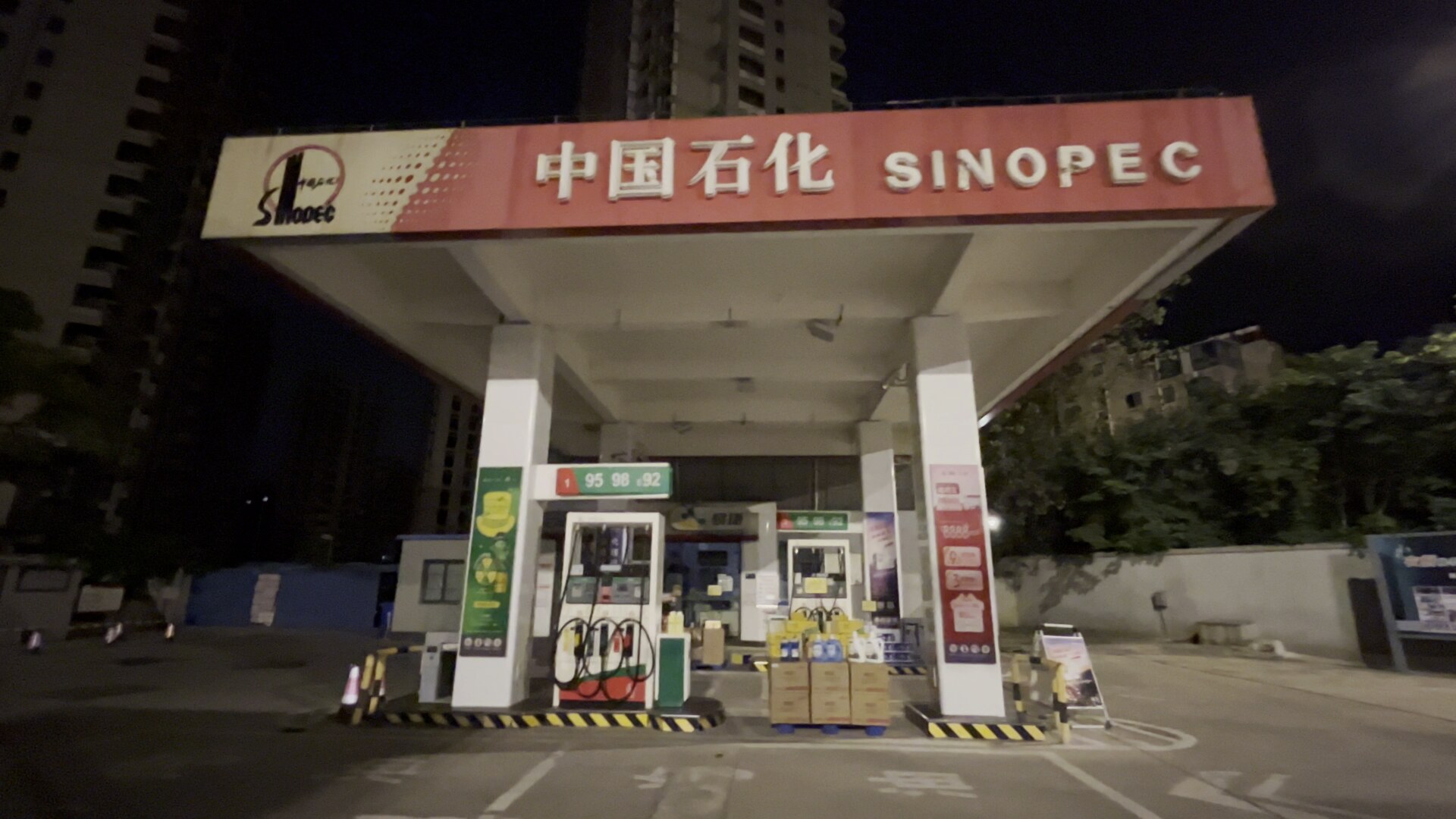} & 
\includegraphics[width=\lw,height=\lh]{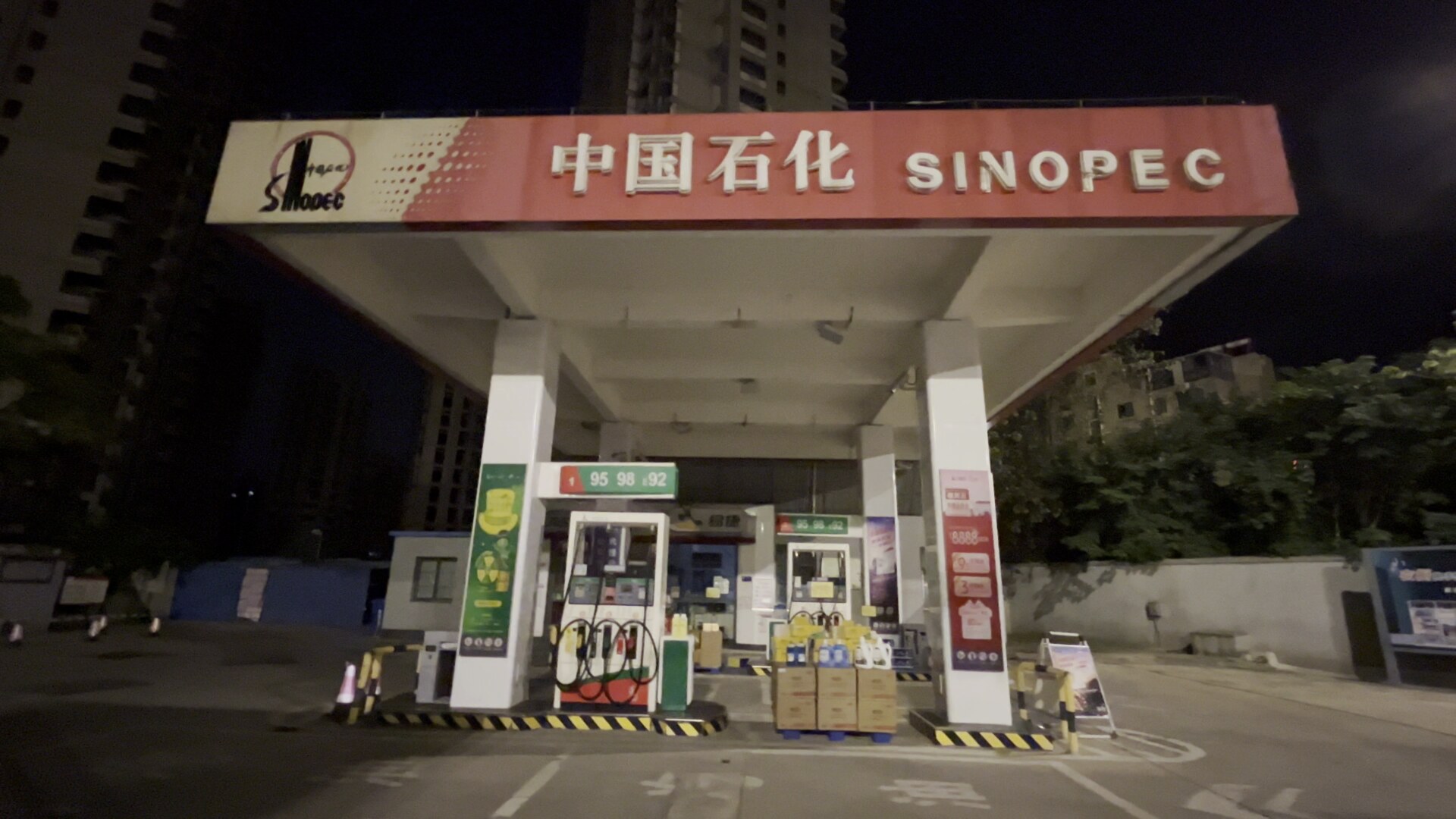} & 
\includegraphics[width=\lw,height=\lh]{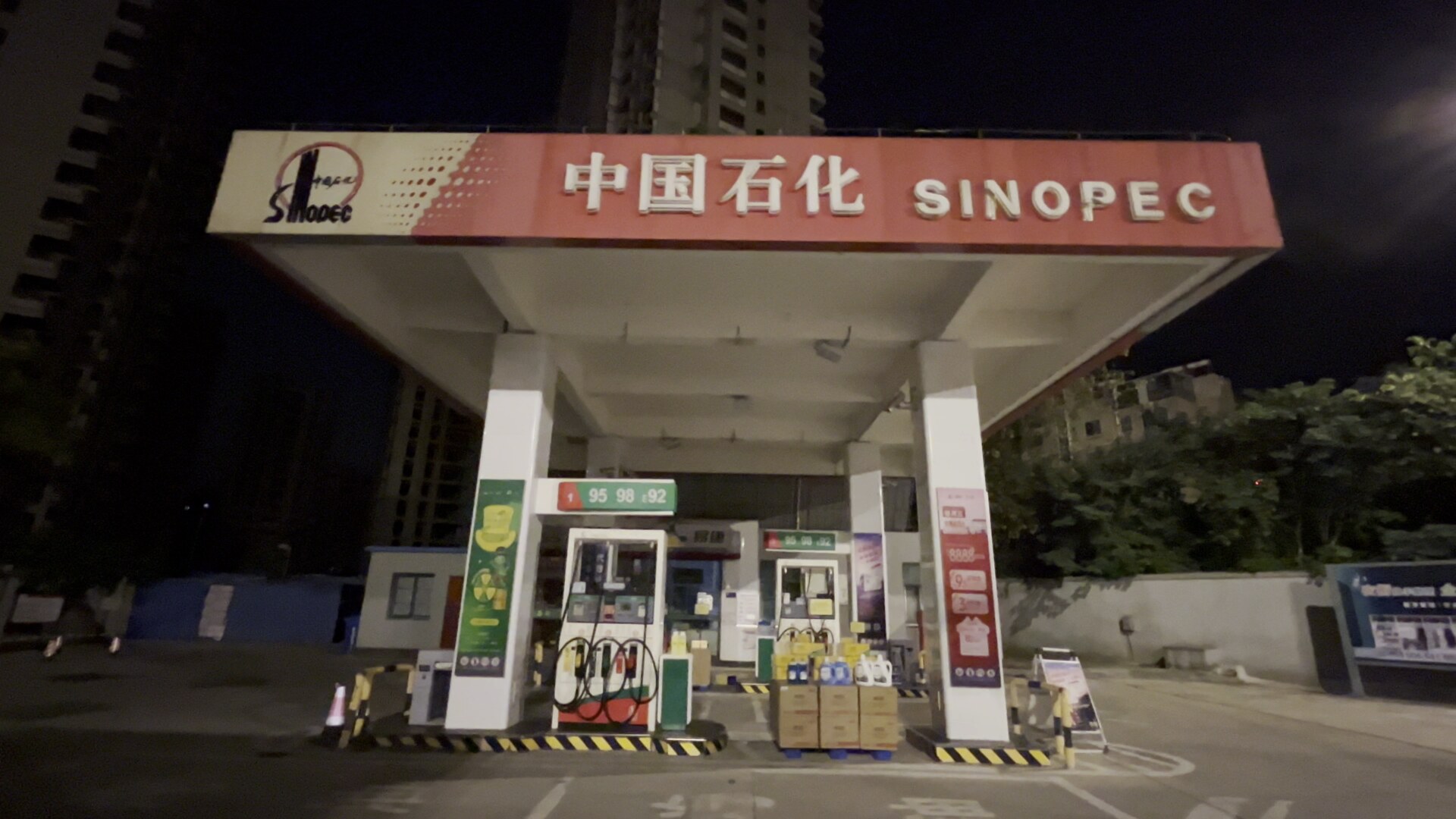} 
\end{tabular}
\caption{Qualitative comparison with state-of-the-art controllable video generation models. In each block, each row corresponds to one video, and frames are arranged from left to right in temporal order. The top row shows the results of each comparison method, followed by ours, with the ground truth (GT) shown in the final row.}
\label{fig:qualitative_comparison}
\end{figure*}

\begin{figure*}[tb] 
\def\lw{0.11111111\linewidth}
\def\lh{0.055\linewidth}
\def\hlw{0.05\linewidth}
\def\ftsz{\normalsize}
\renewcommand\tabcolsep{0.0pt}
\renewcommand{\arraystretch}{0}
\centering \small
\begin{tabular}{ccccccccccc}
\rotatebox{90}{\hspace{1mm}\footnotesize{Diffuse}}\hspace{1mm} & 
\includegraphics[width=\lw]{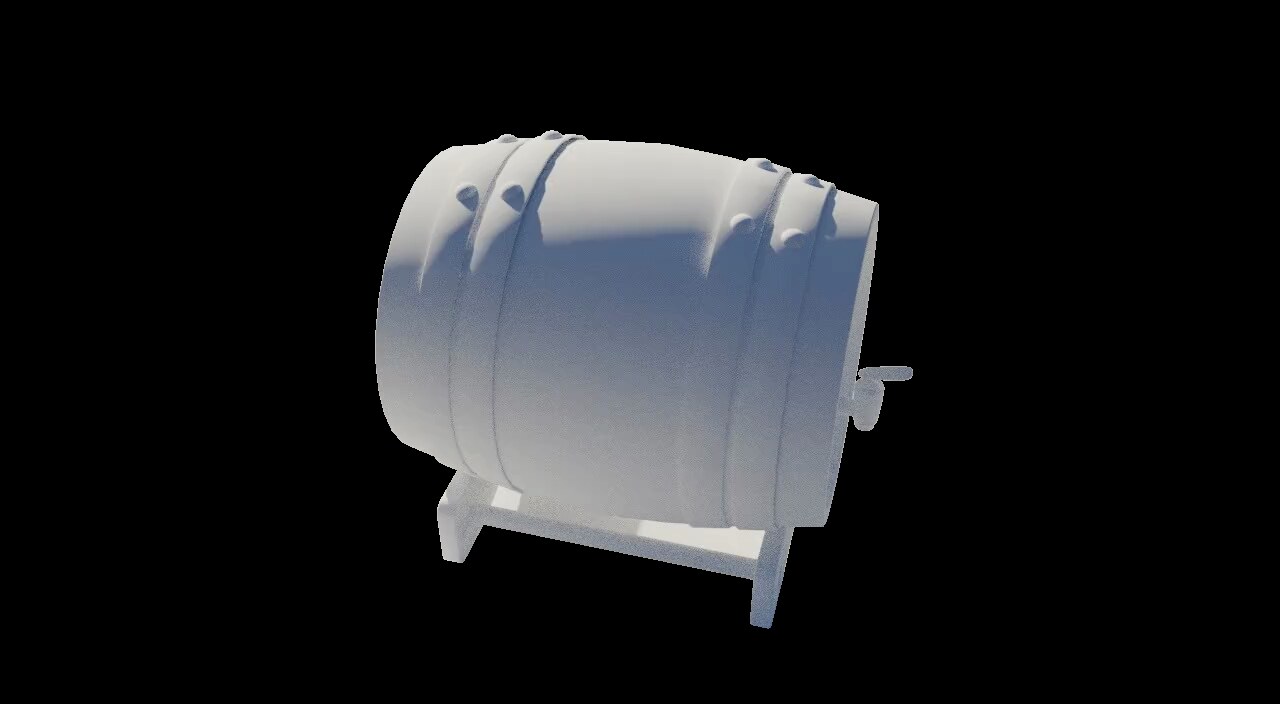} & 
\includegraphics[width=\lw]{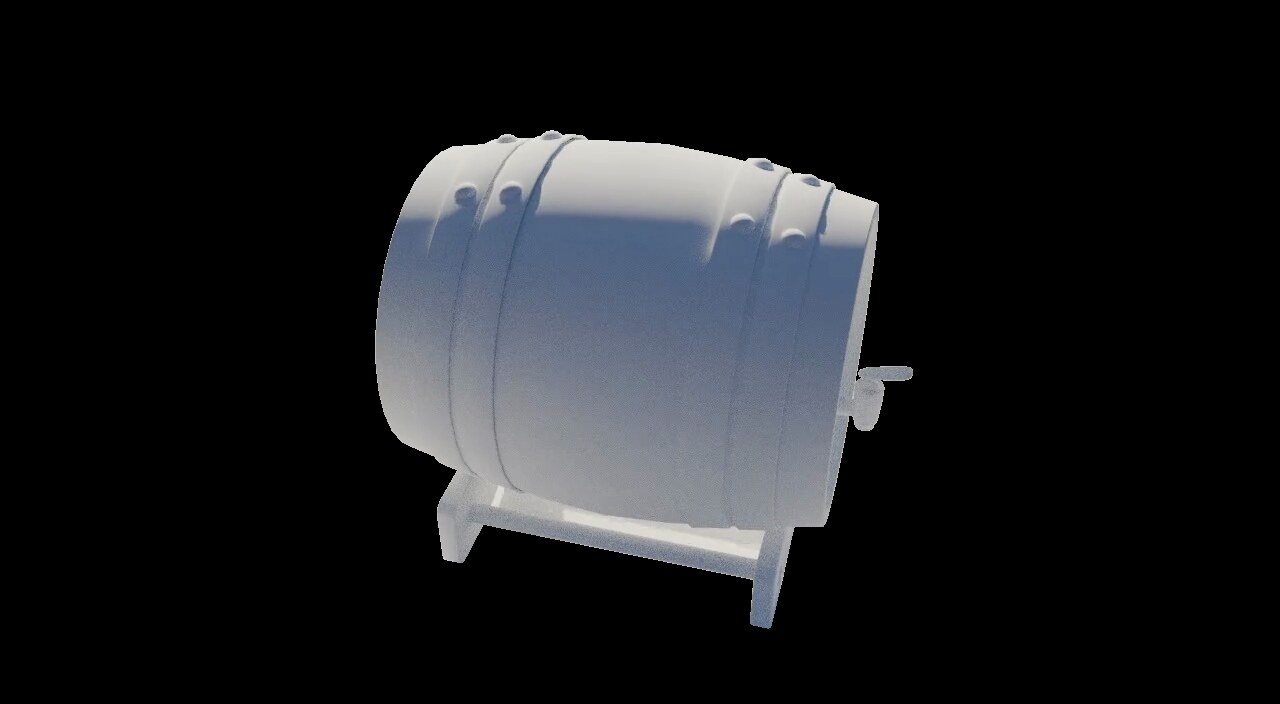} & 
\includegraphics[width=\lw]{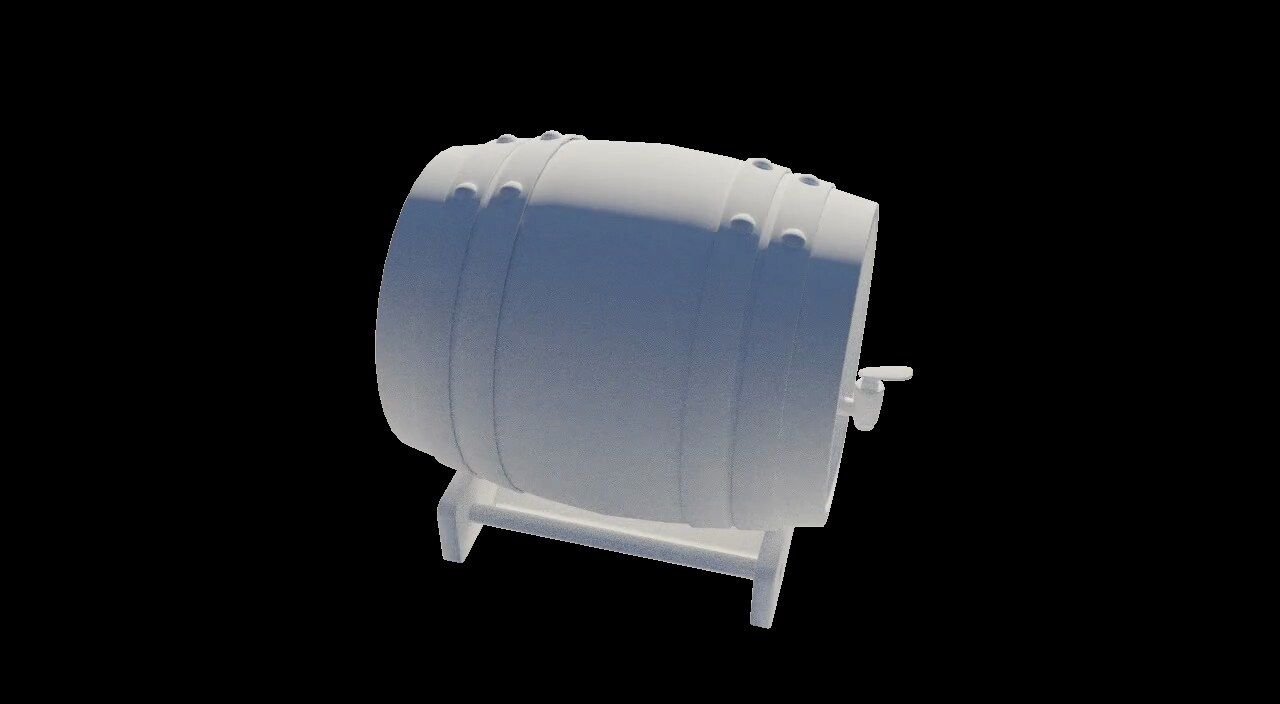} & 
\includegraphics[width=\lw]{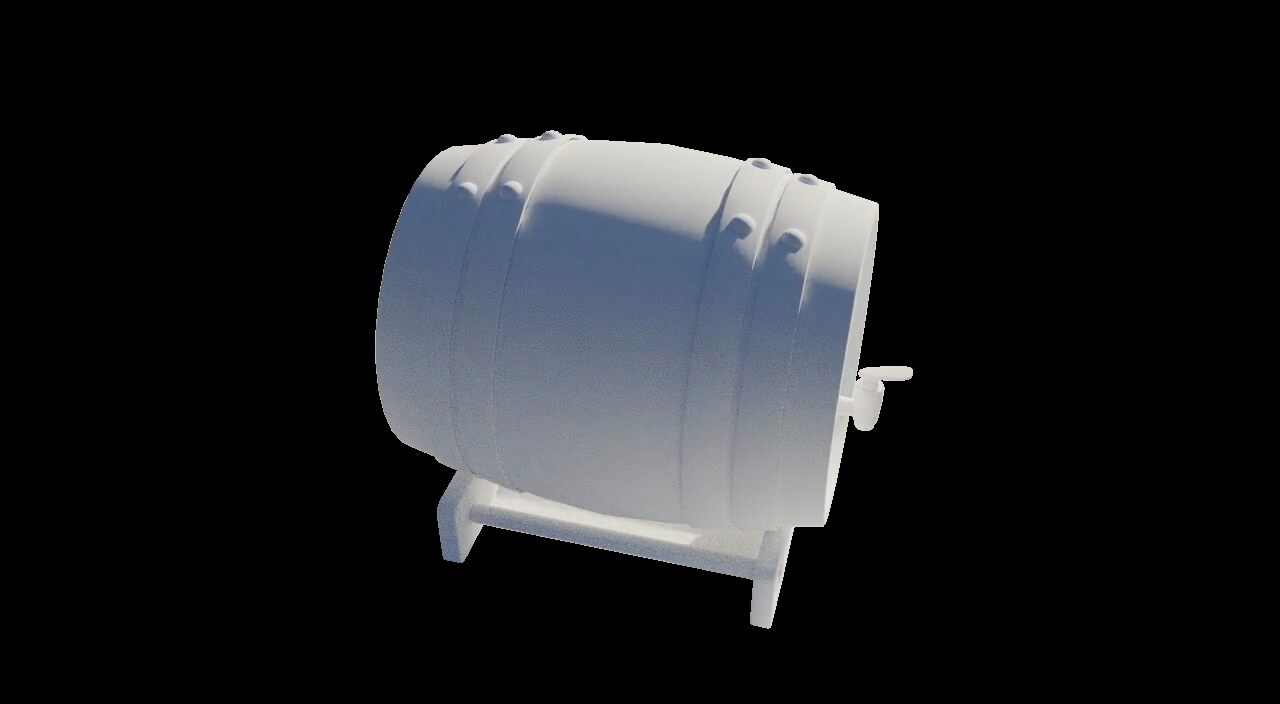} & 
\includegraphics[width=\lw]{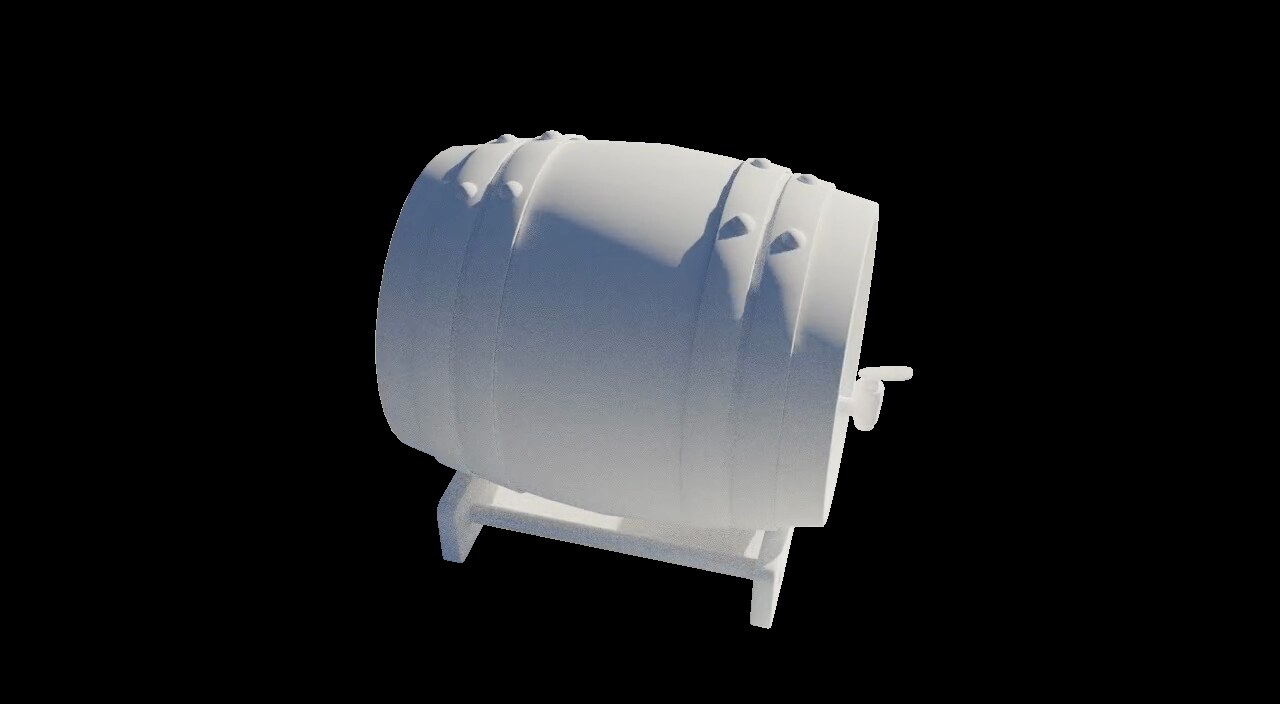} &
\includegraphics[width=\lw]{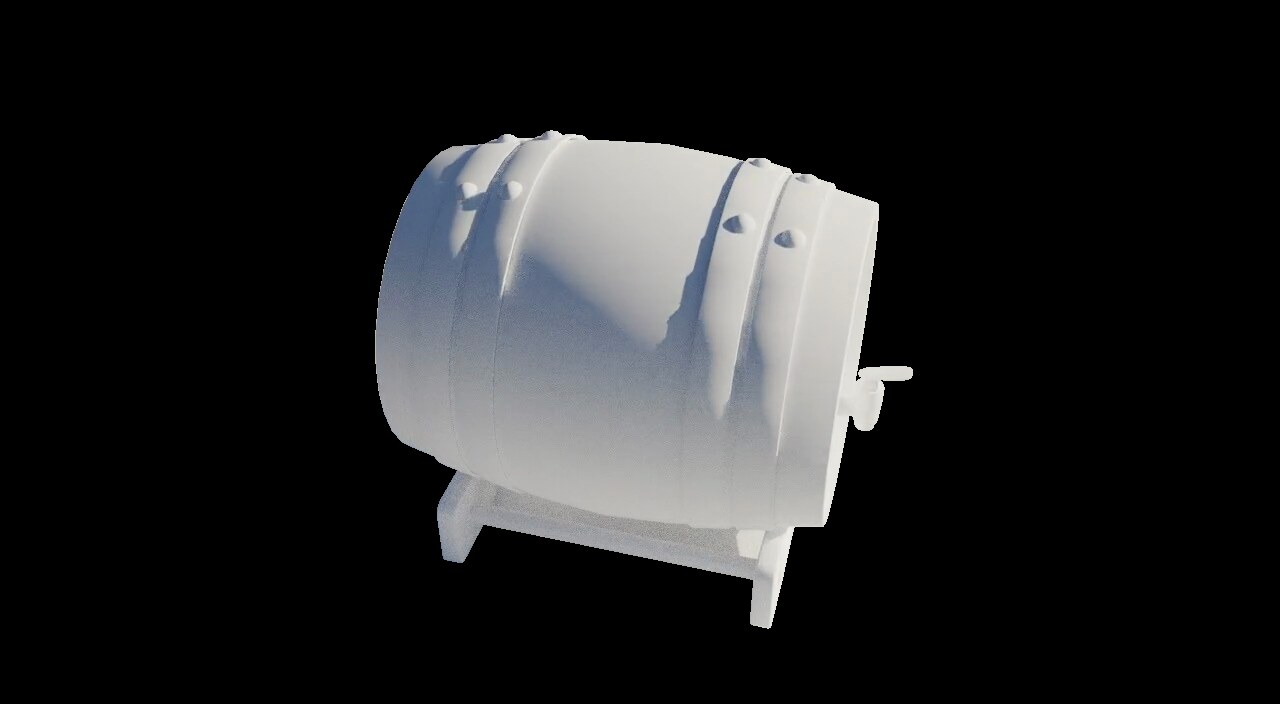} & 
\includegraphics[width=\lw]{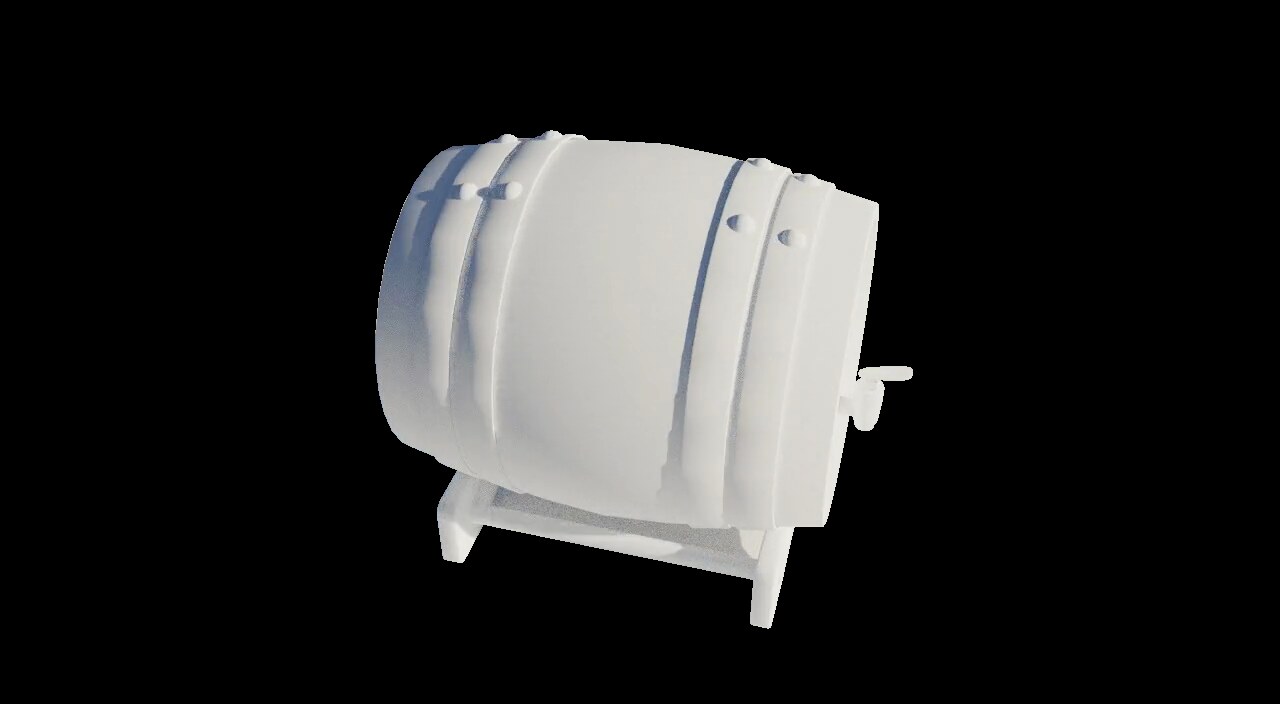} & 
\includegraphics[width=\lw]{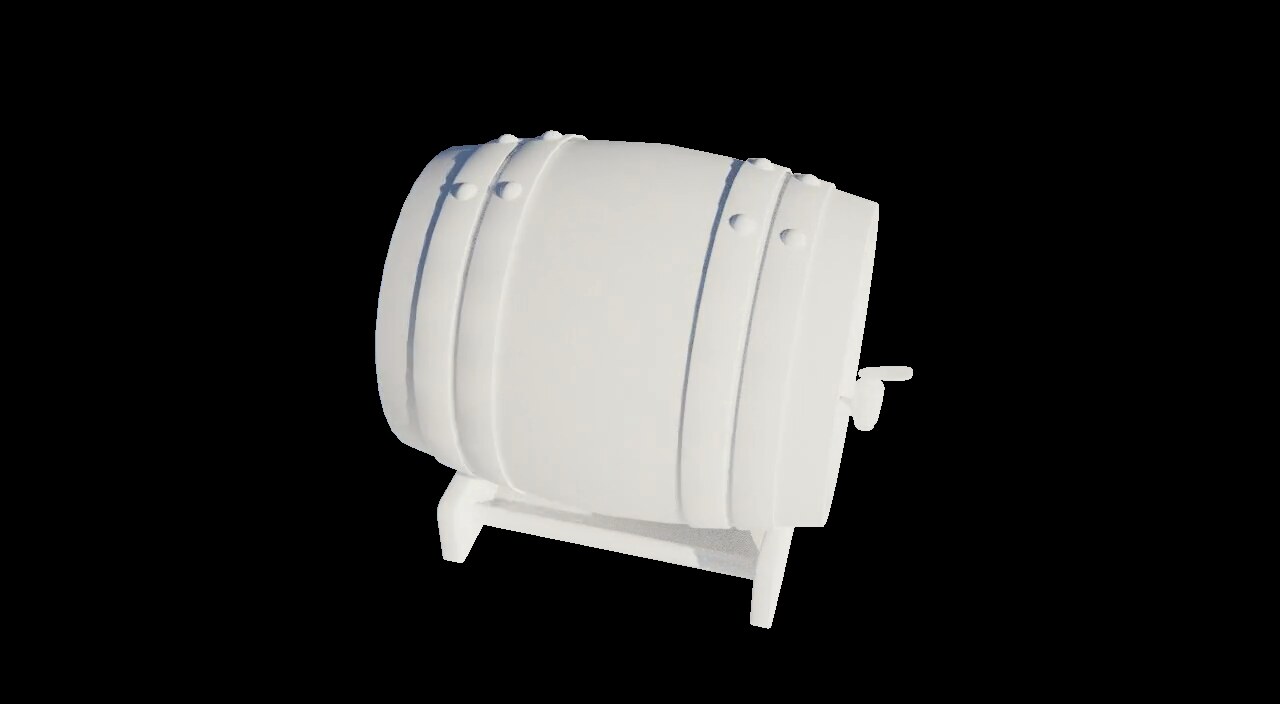} & 
\includegraphics[width=\lw]{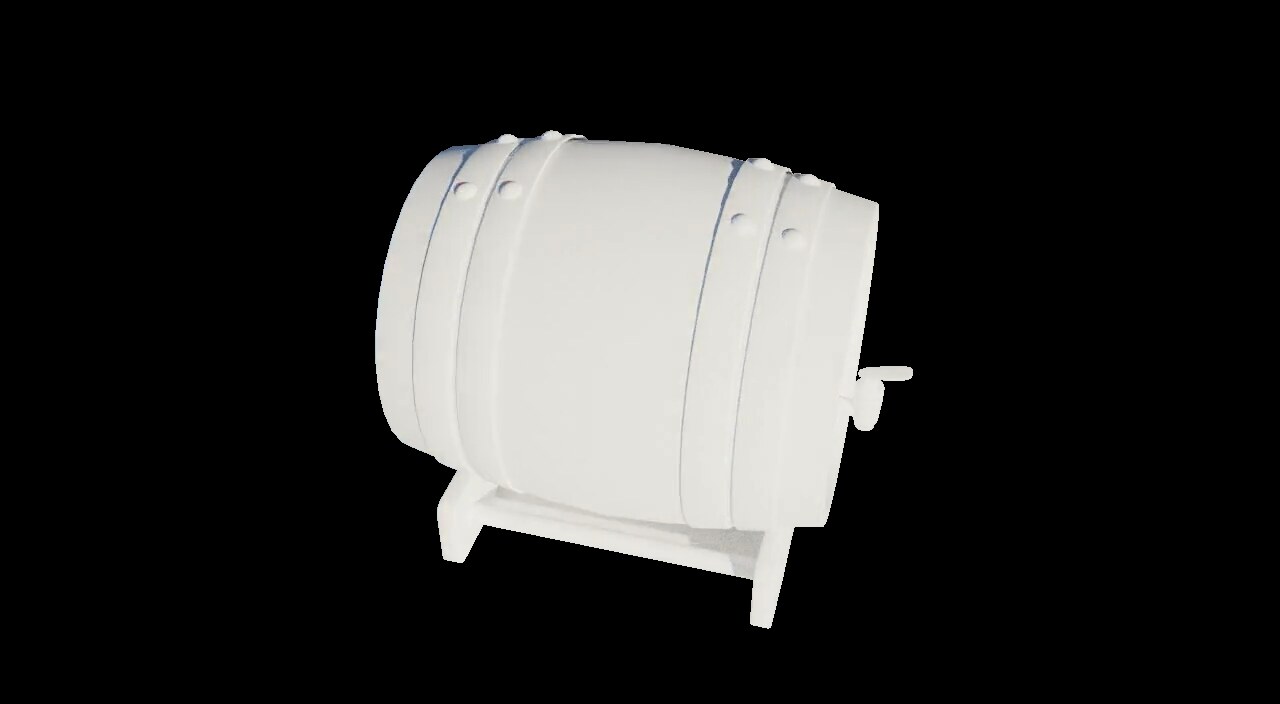} \\
\rotatebox{90}{\hspace{1mm}\footnotesize{Rough}}\hspace{1mm} & 
\includegraphics[width=\lw]{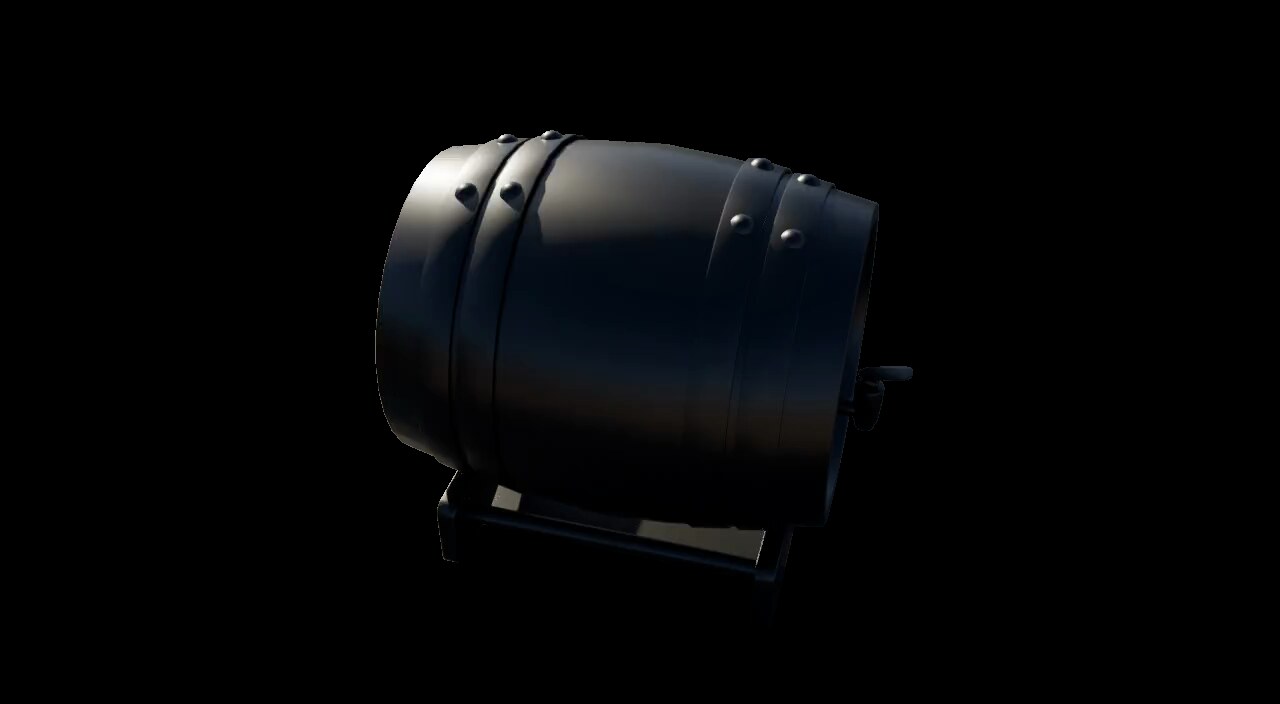} & 
\includegraphics[width=\lw]{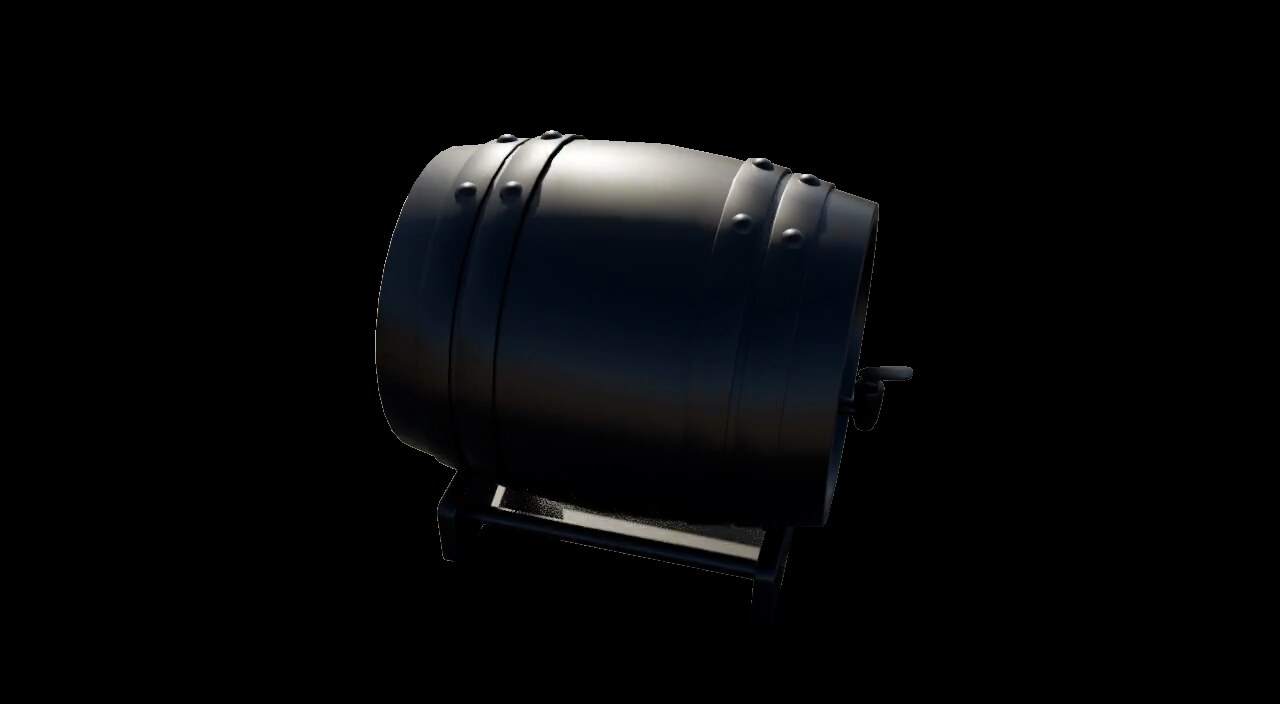} & 
\includegraphics[width=\lw]{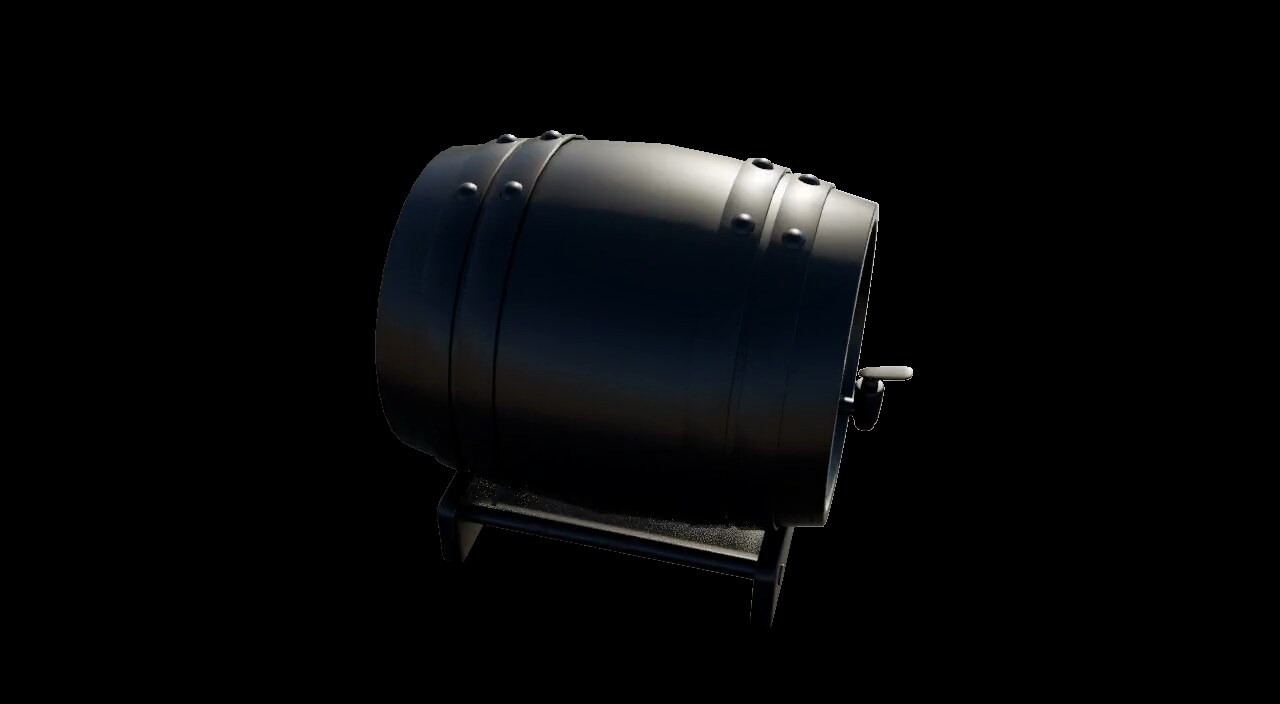} & 
\includegraphics[width=\lw]{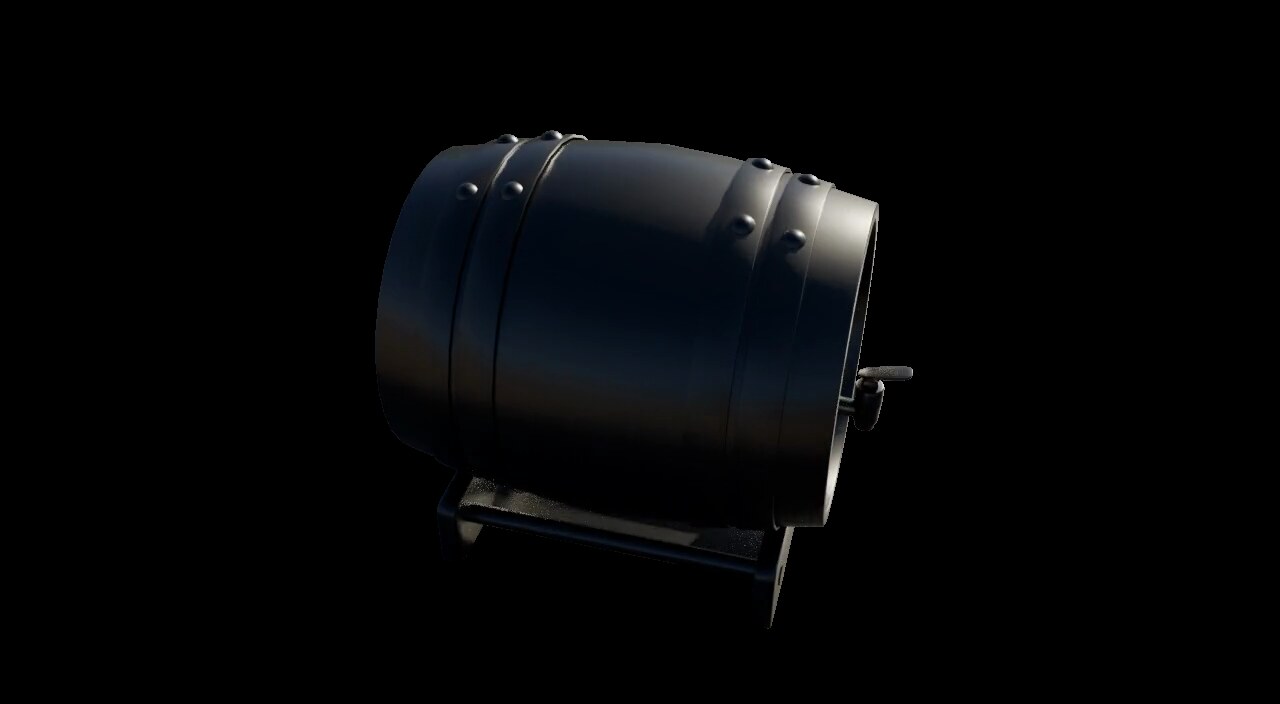} & 
\includegraphics[width=\lw]{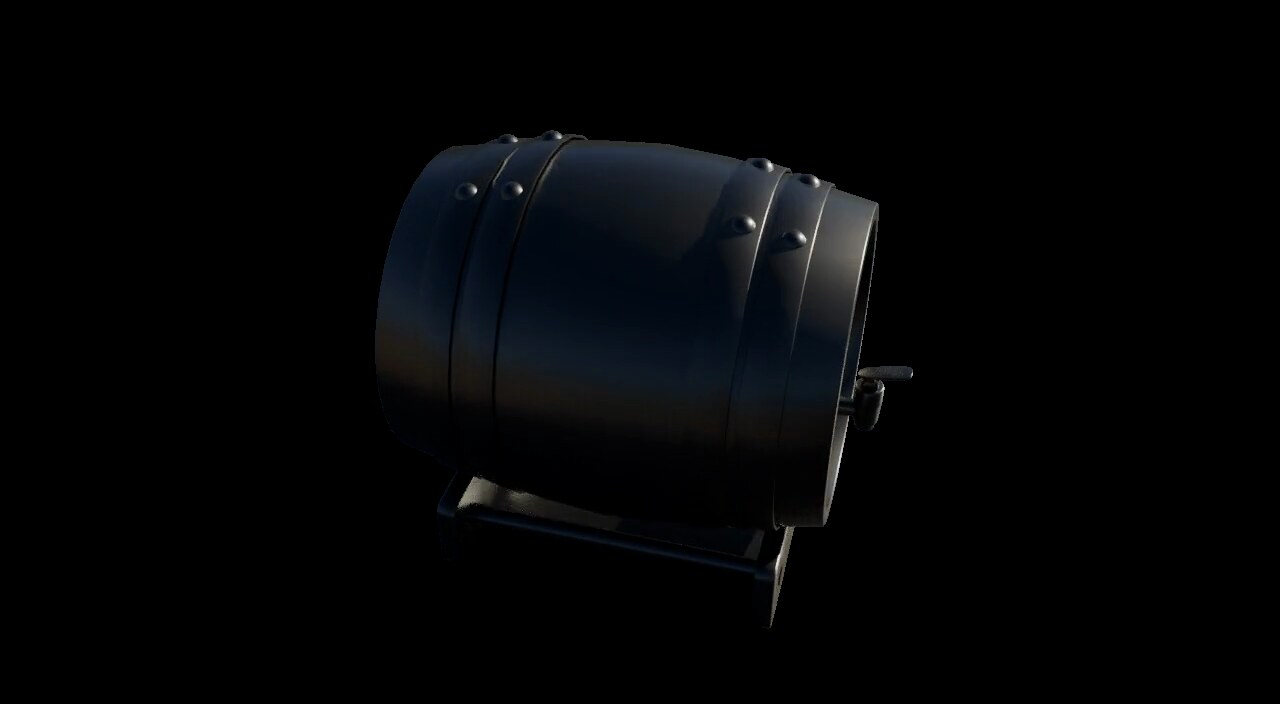} &
\includegraphics[width=\lw]{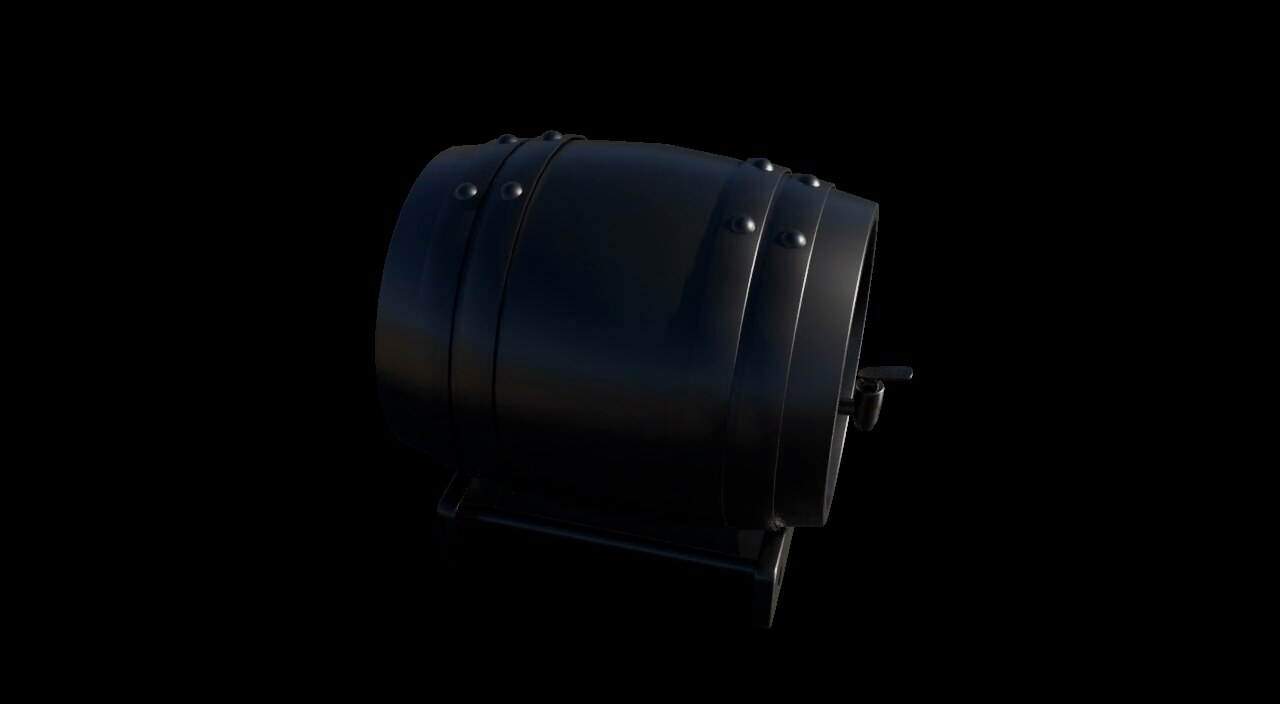} & 
\includegraphics[width=\lw]{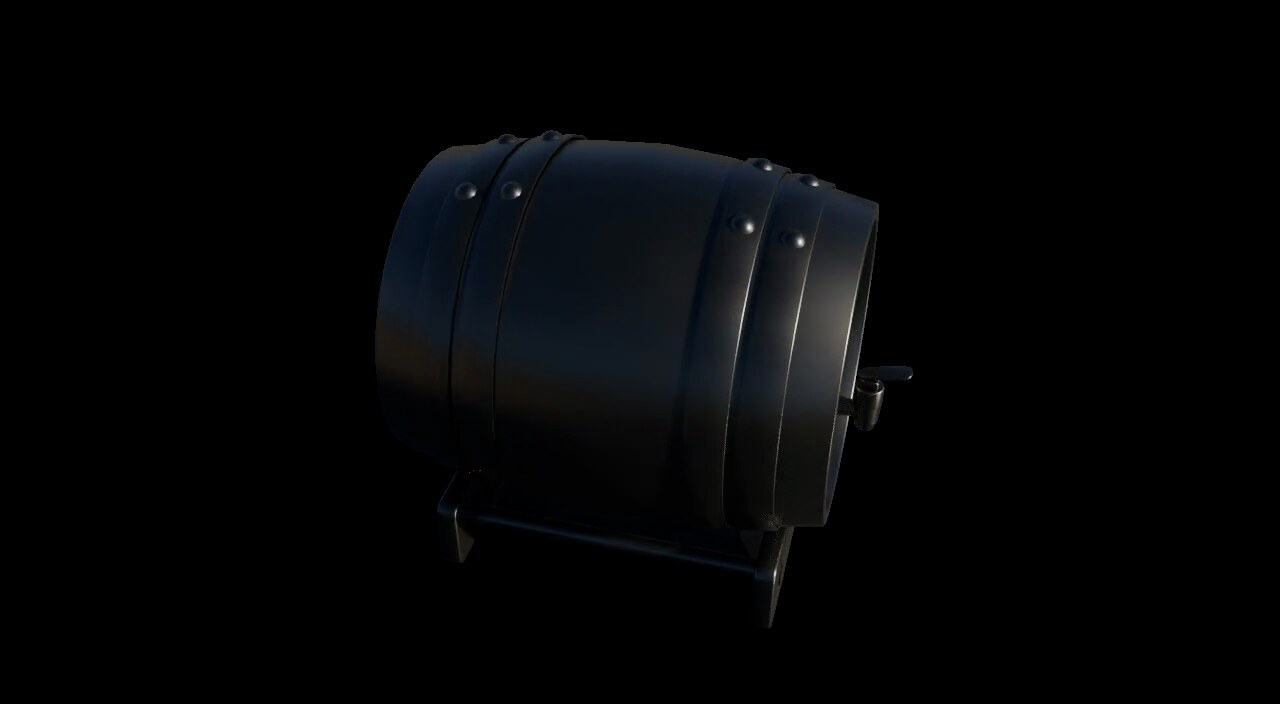} & 
\includegraphics[width=\lw]{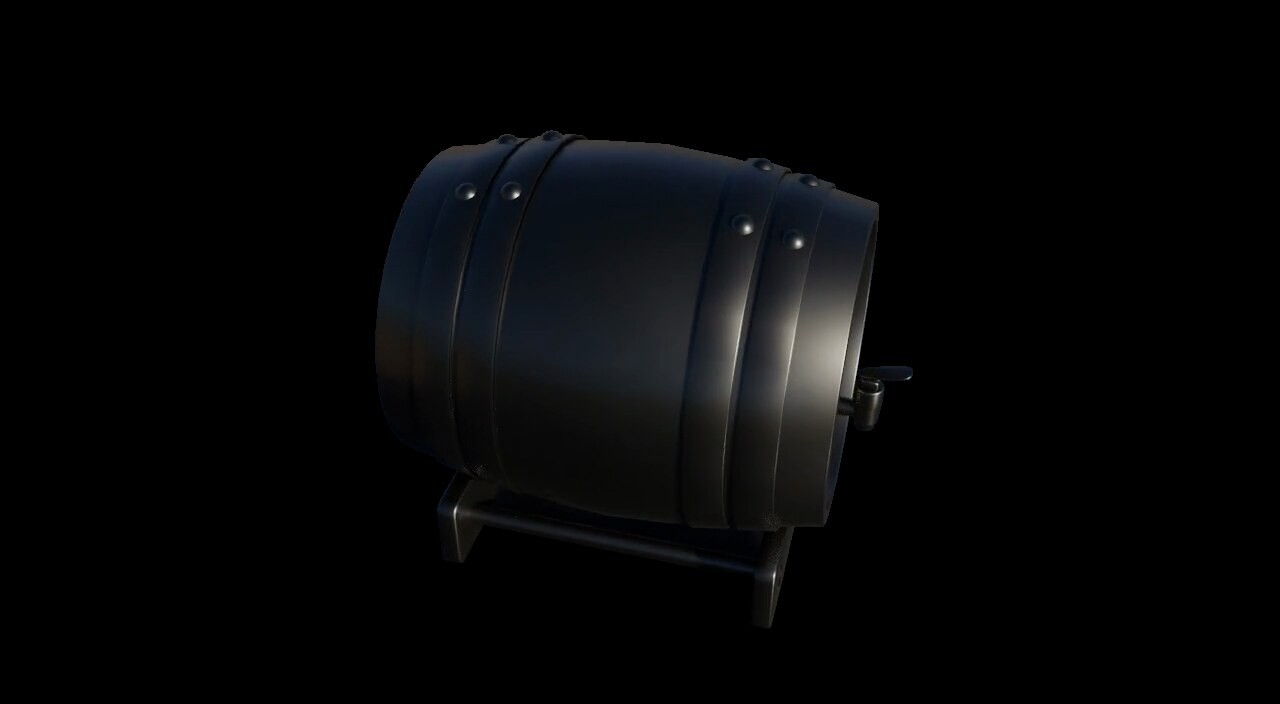} & 
\includegraphics[width=\lw]{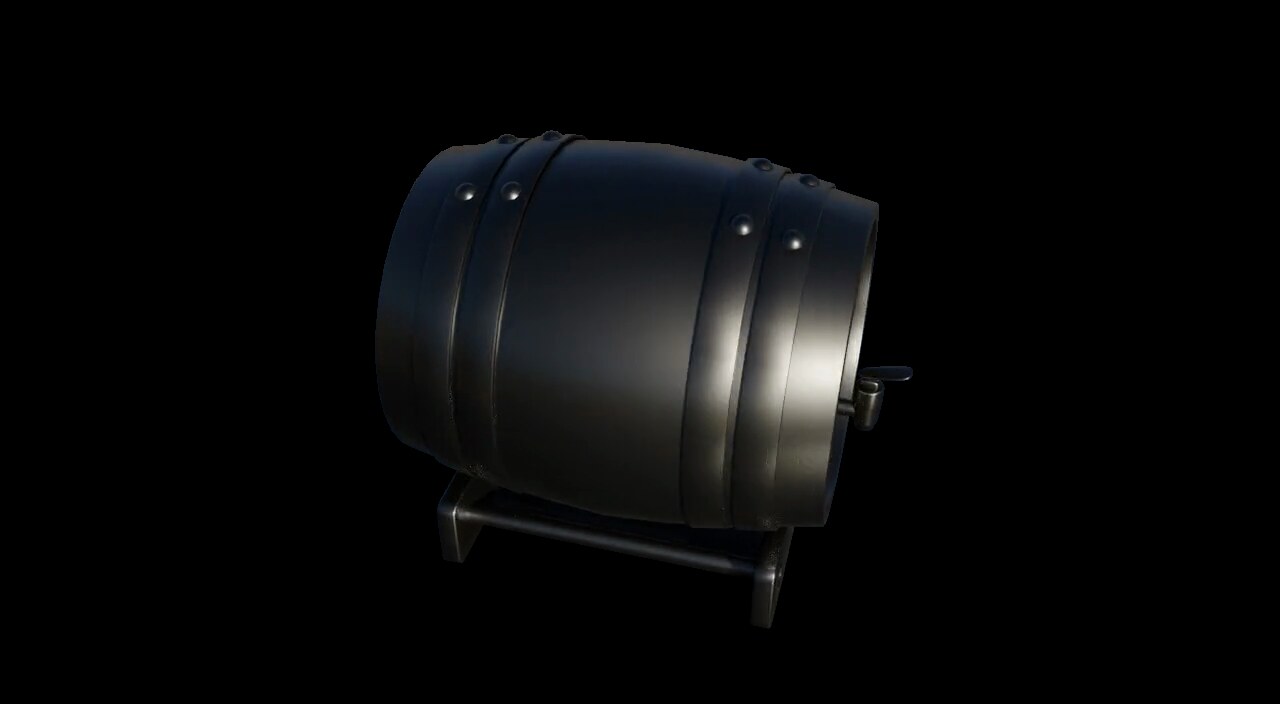}  \\
\rotatebox{90}{\hspace{2mm}\footnotesize{Glossy}}\hspace{1mm} & 
\includegraphics[width=\lw]{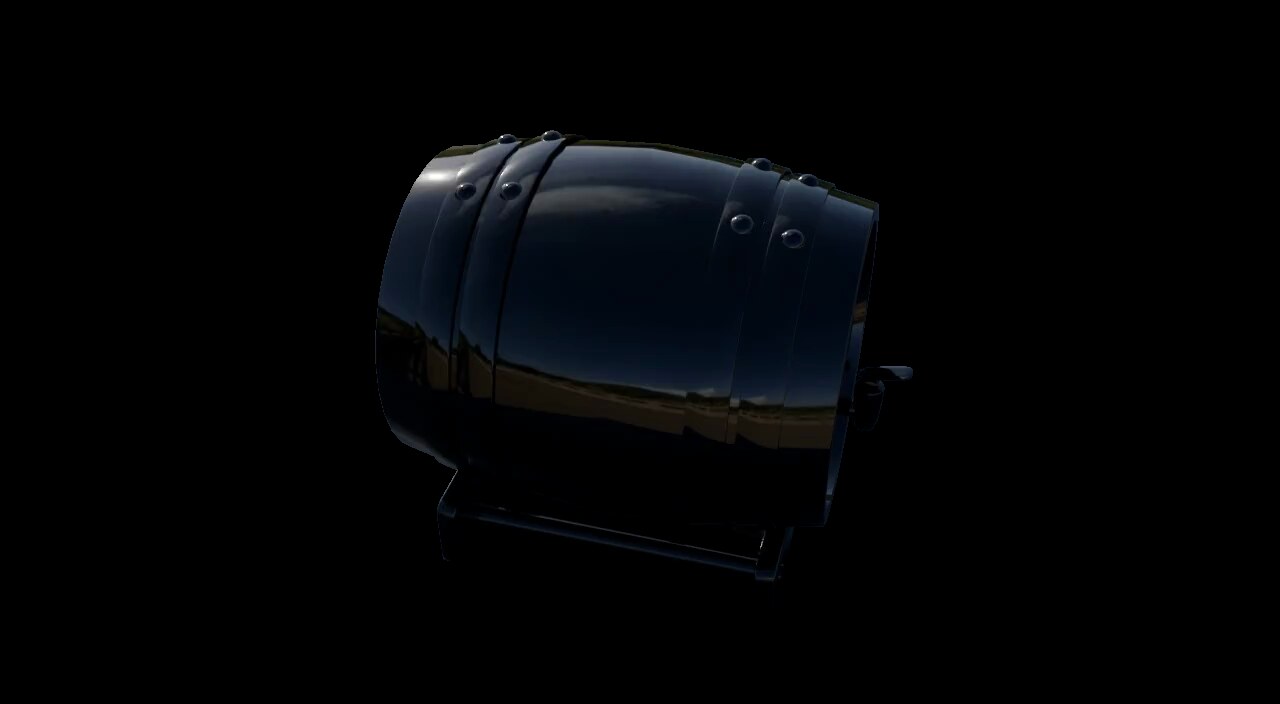} & 
\includegraphics[width=\lw]{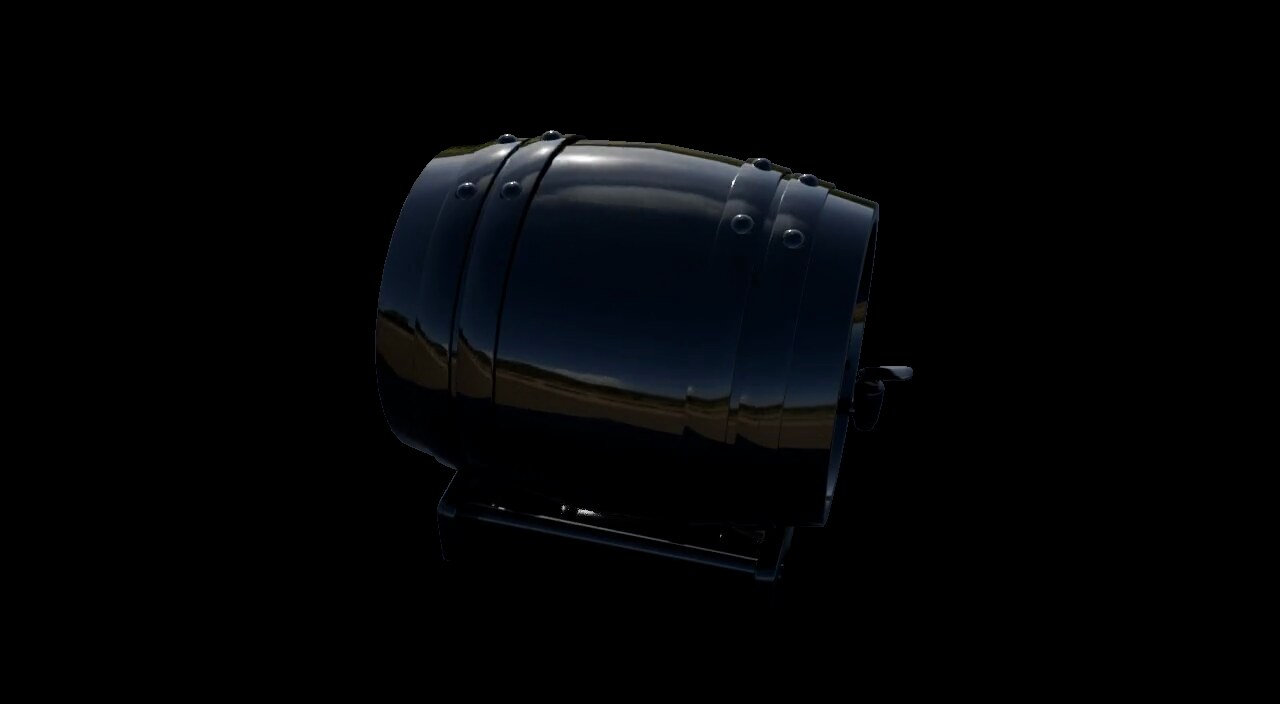} & 
\includegraphics[width=\lw]{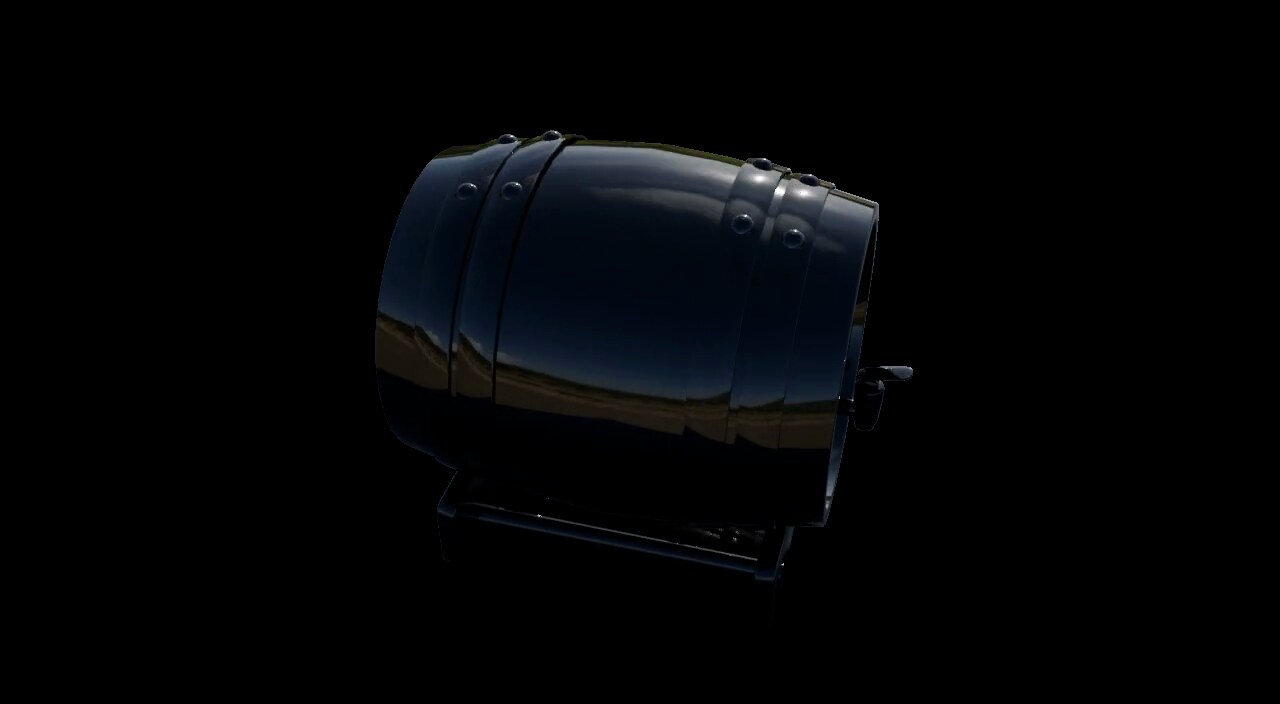} & 
\includegraphics[width=\lw]{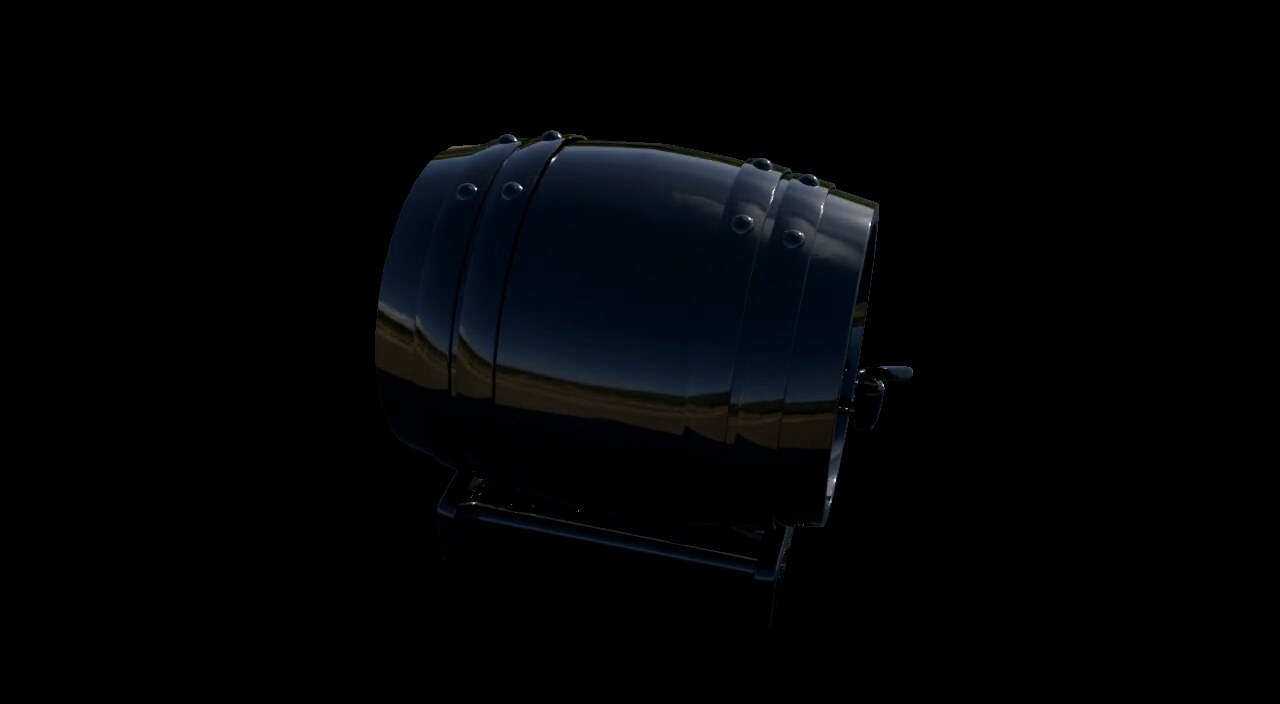} & 
\includegraphics[width=\lw]{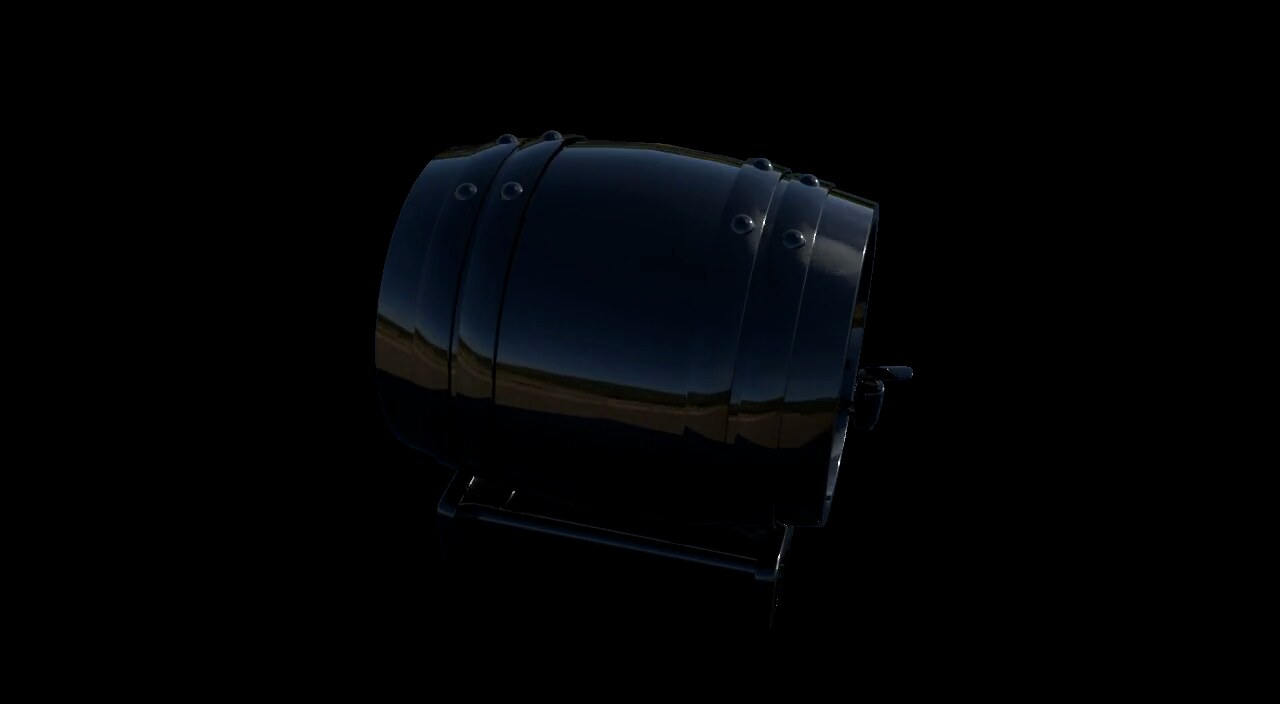} &
\includegraphics[width=\lw]{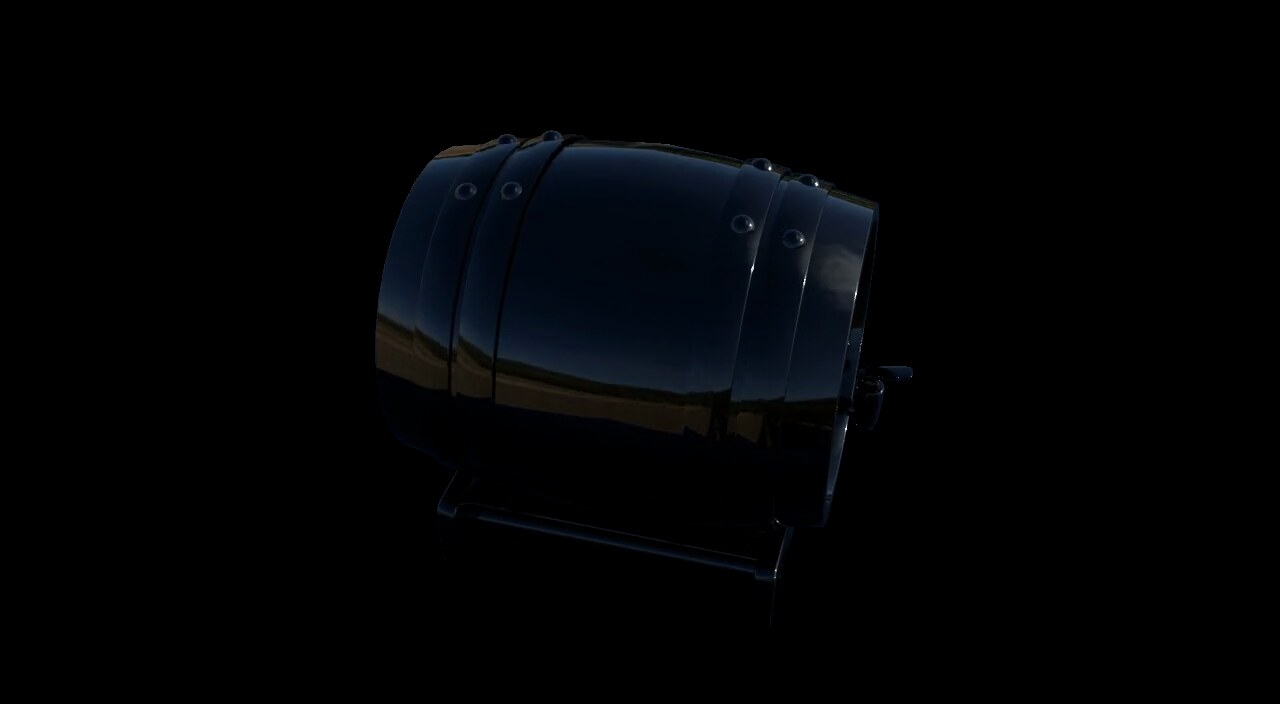} & 
\includegraphics[width=\lw]{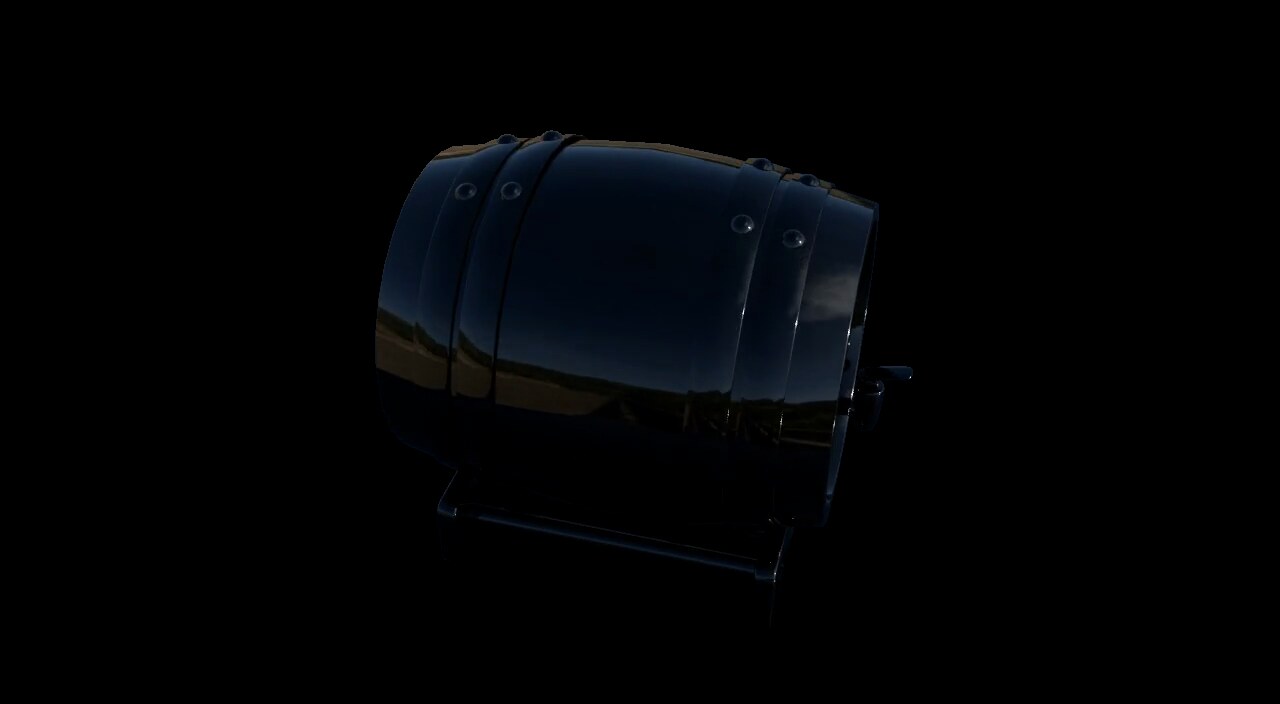} & 
\includegraphics[width=\lw]{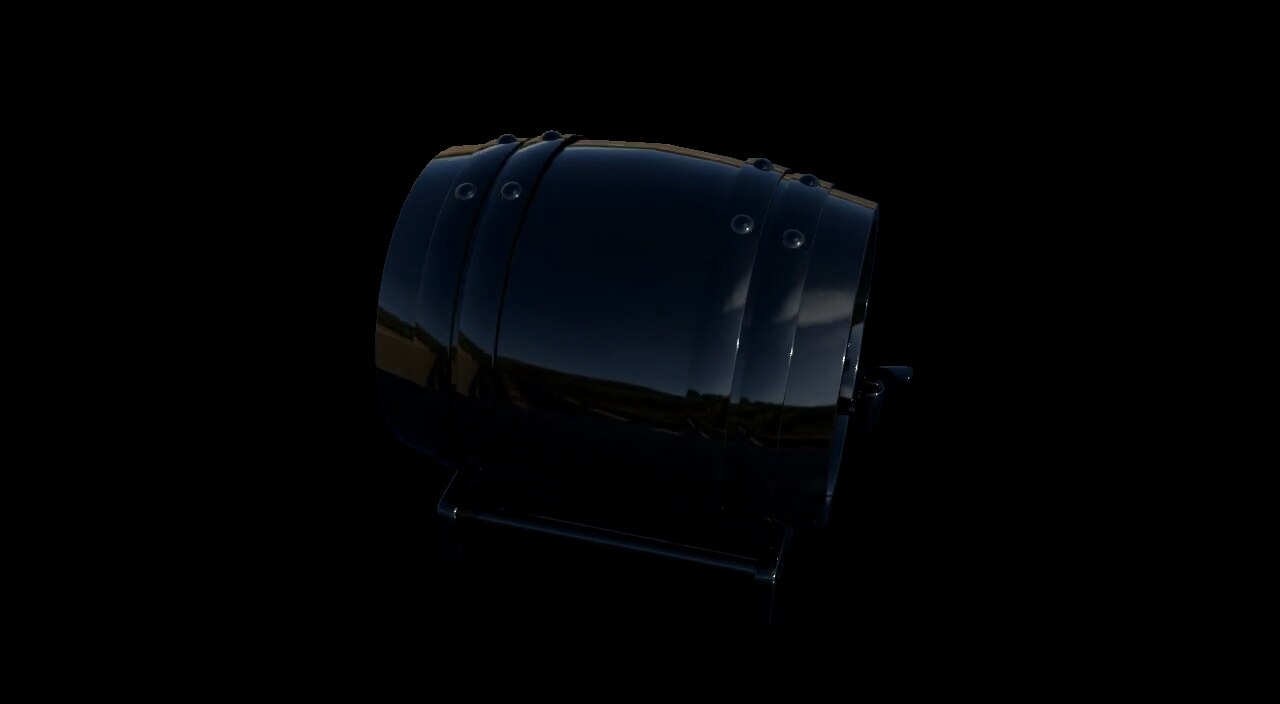} & 
\includegraphics[width=\lw]{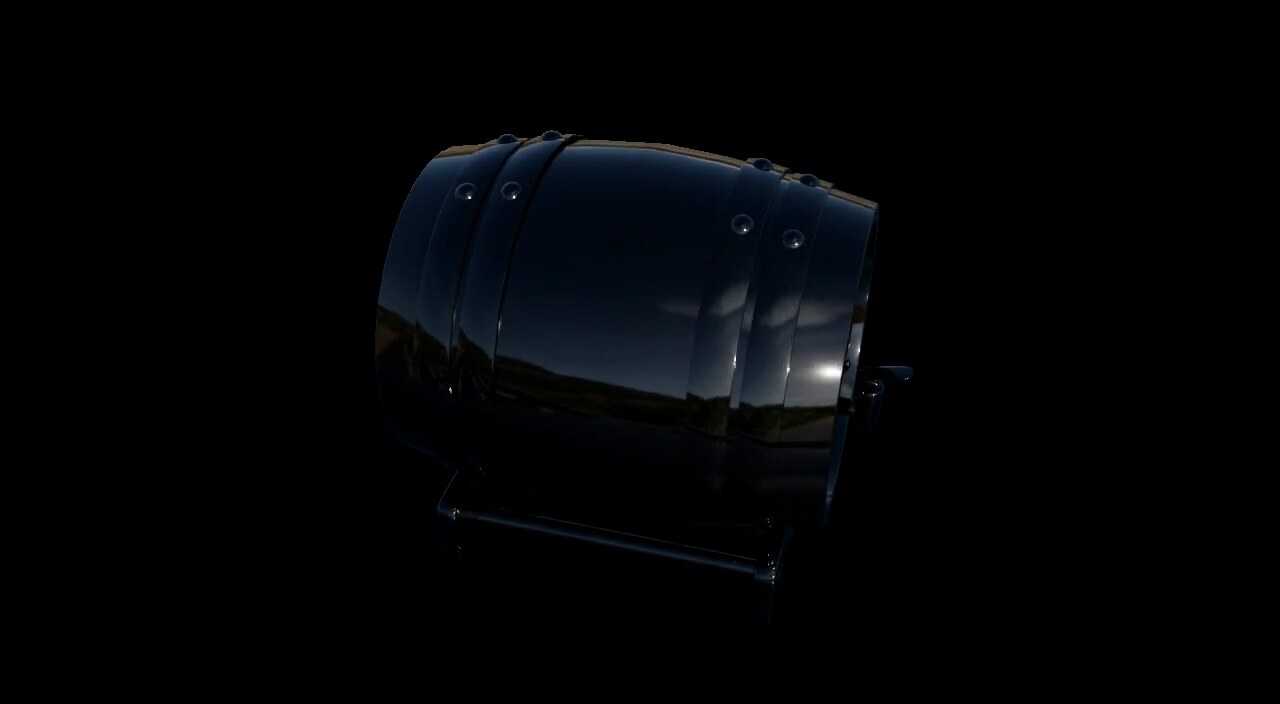}  \\
\rotatebox{90}{\hspace{2mm}\footnotesize{Output}}\hspace{1mm} & 
\includegraphics[width=\lw]{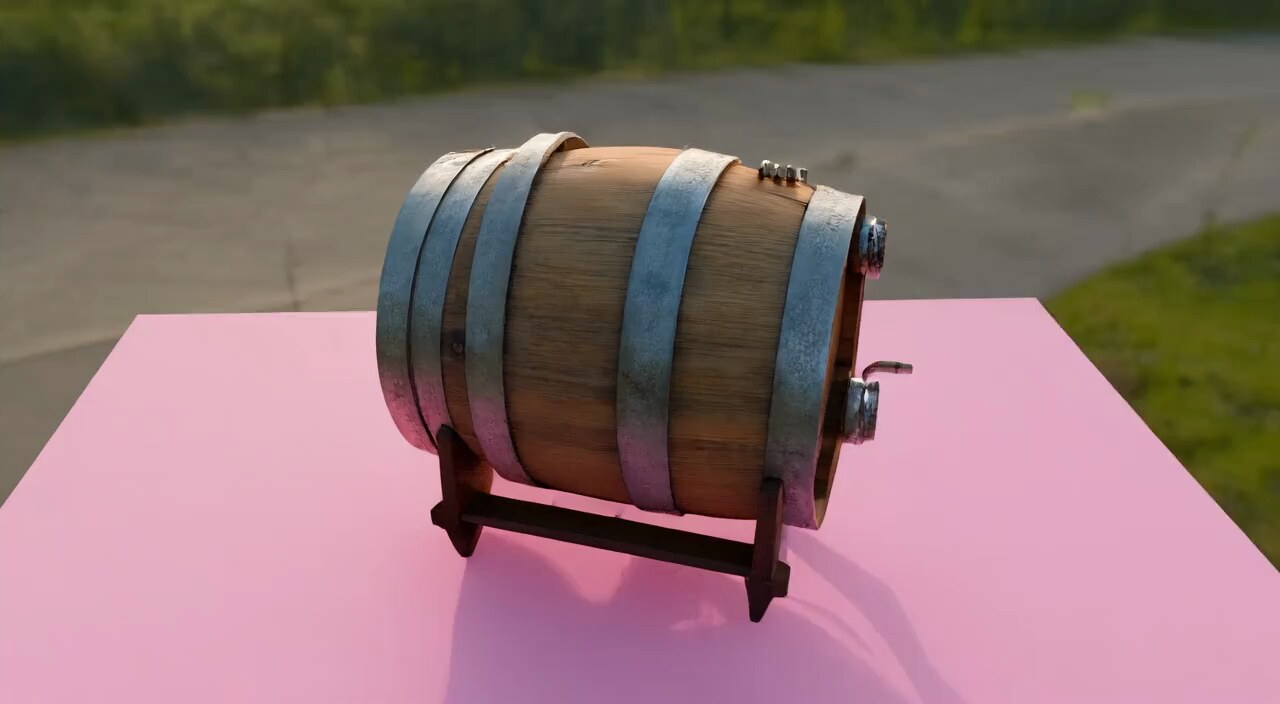} & 
\includegraphics[width=\lw]{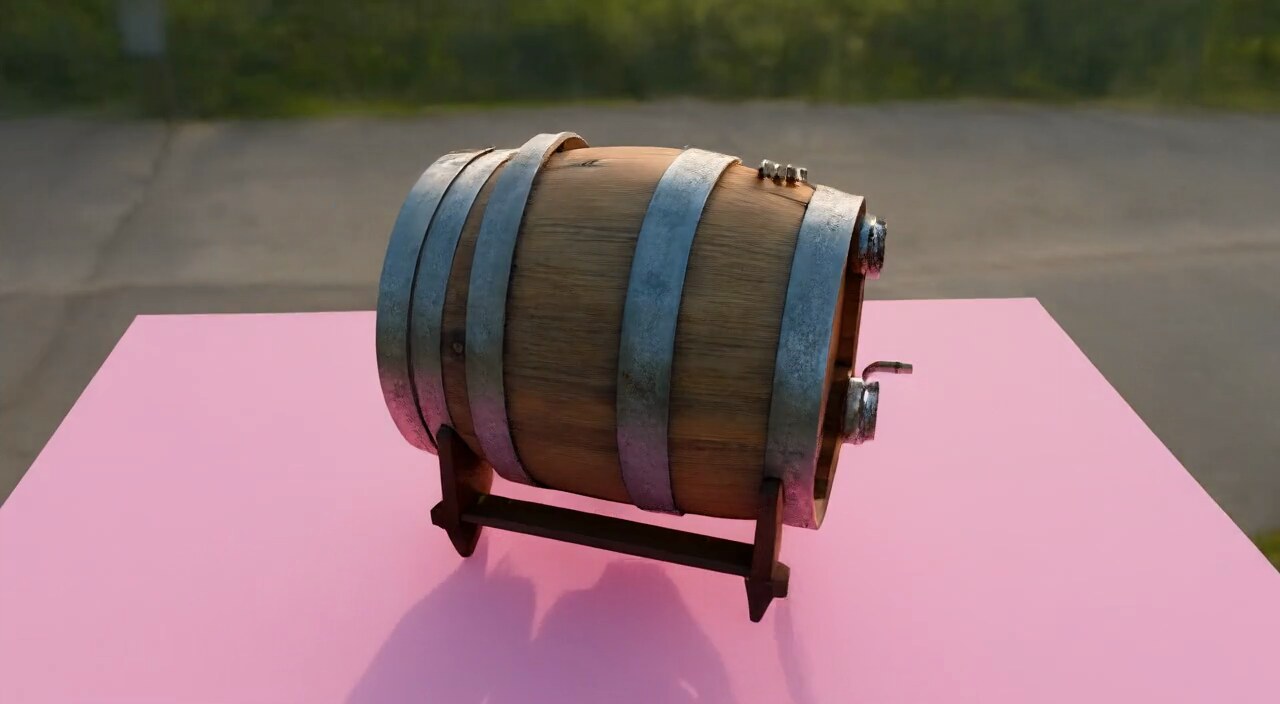} & 
\includegraphics[width=\lw]{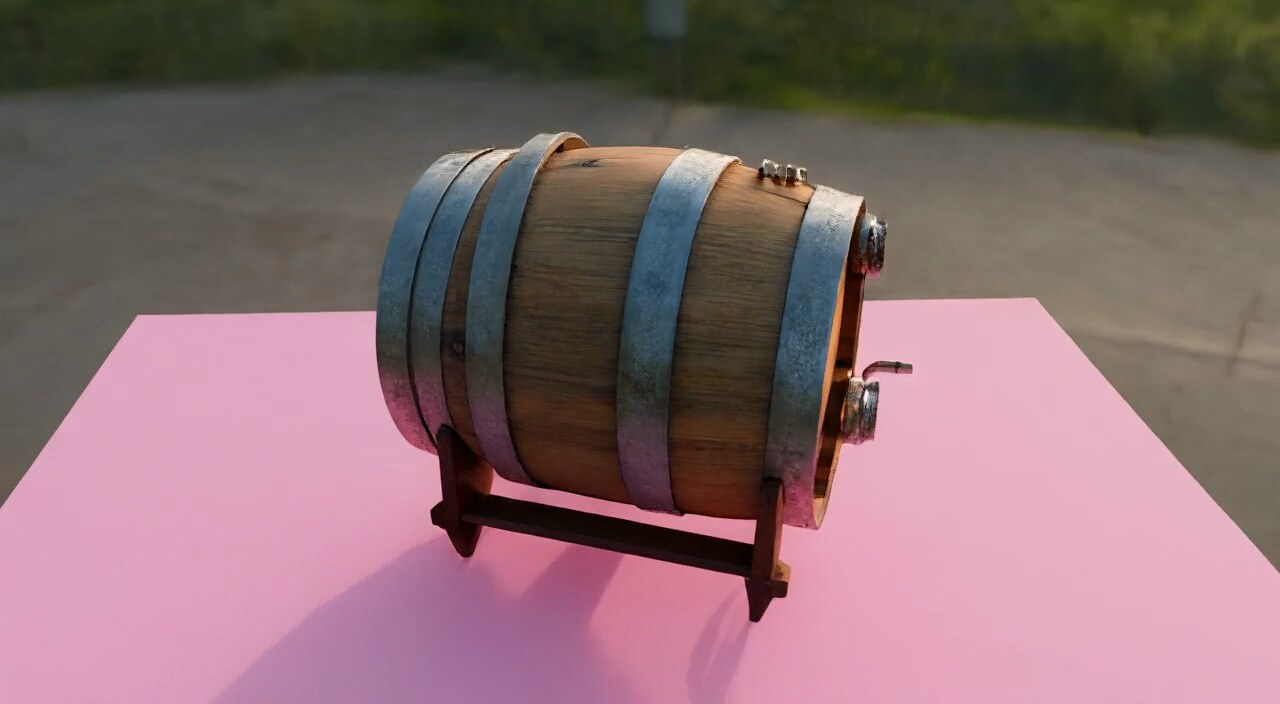} & 
\includegraphics[width=\lw]{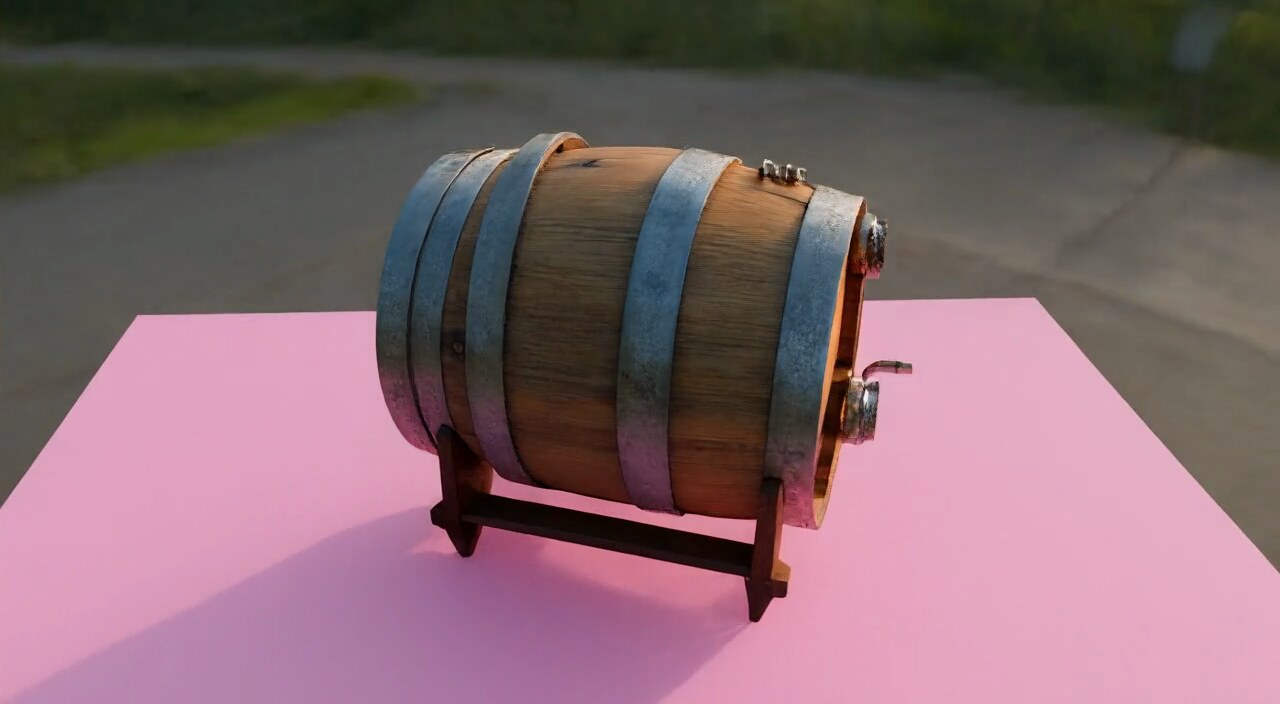} & 
\includegraphics[width=\lw]{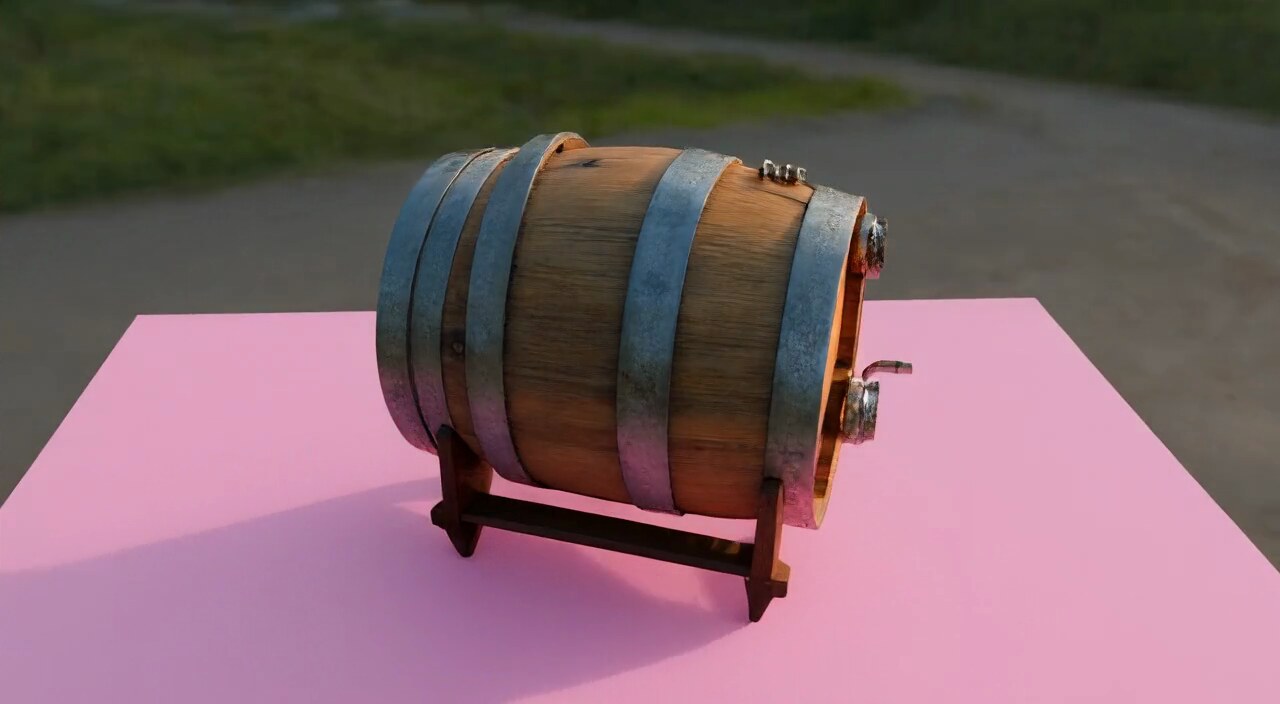} & 
\includegraphics[width=\lw]{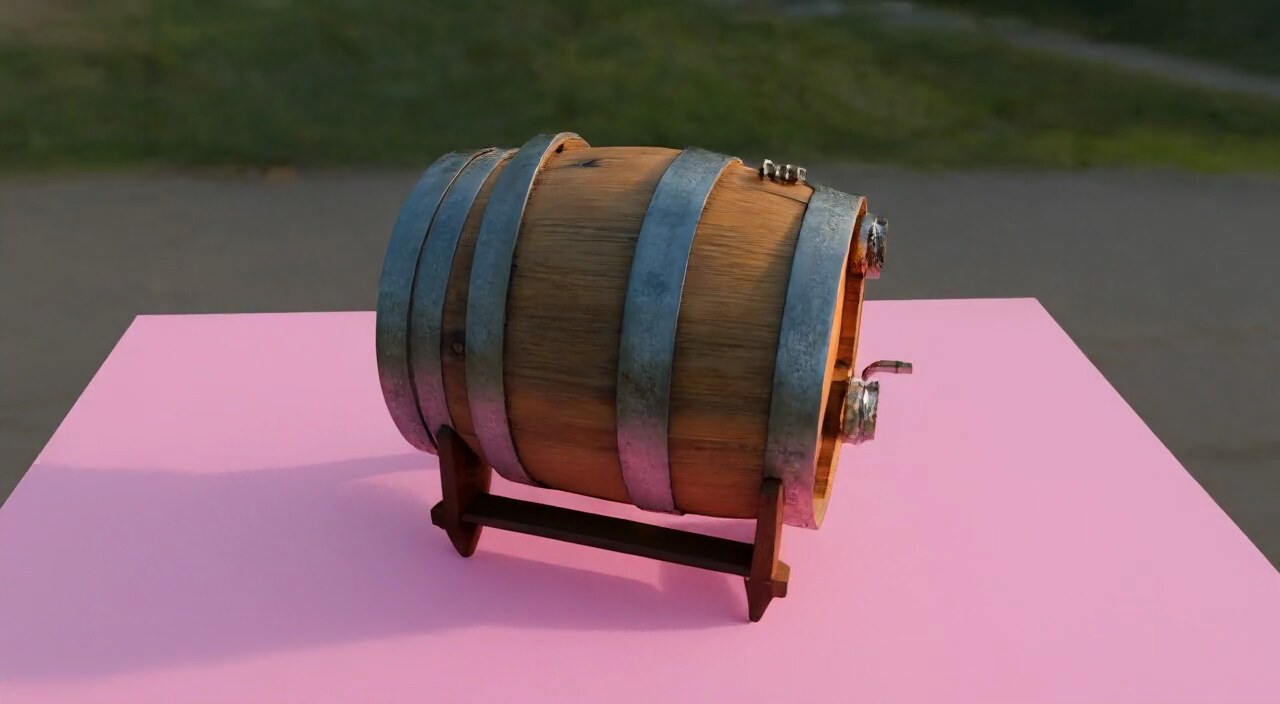} &
\includegraphics[width=\lw]{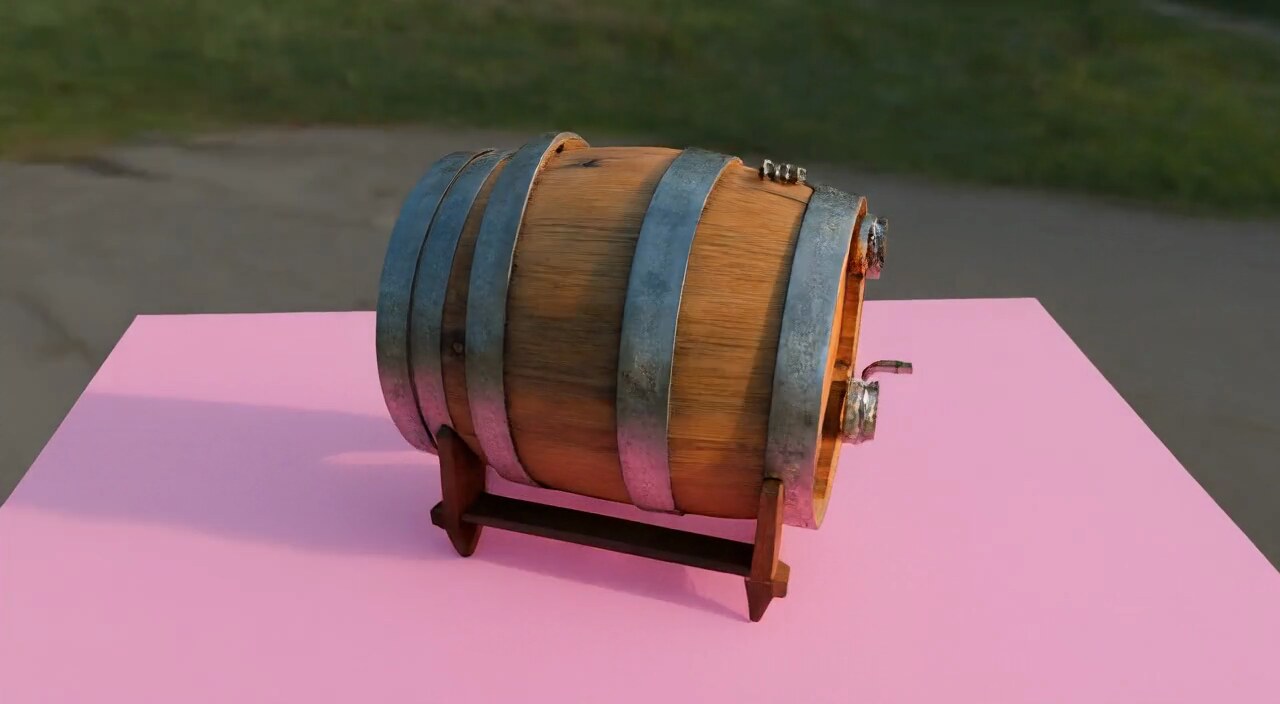} & 
\includegraphics[width=\lw]{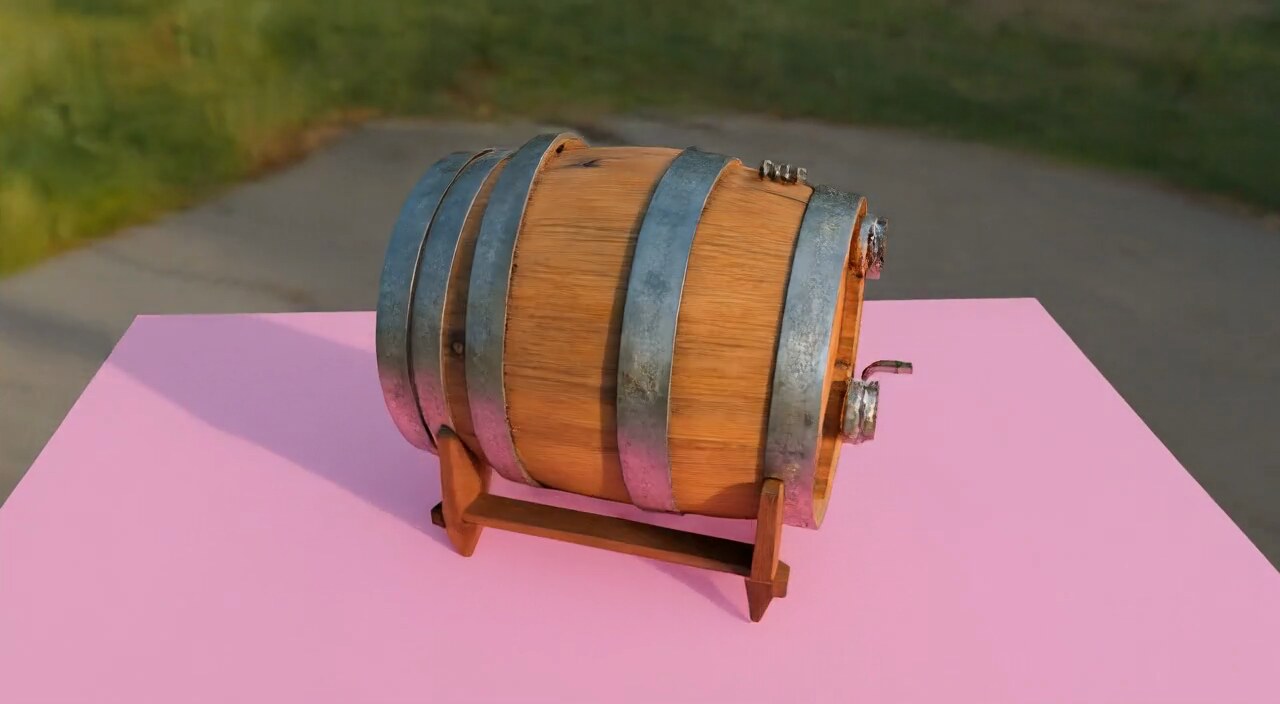} & 
\includegraphics[width=\lw]{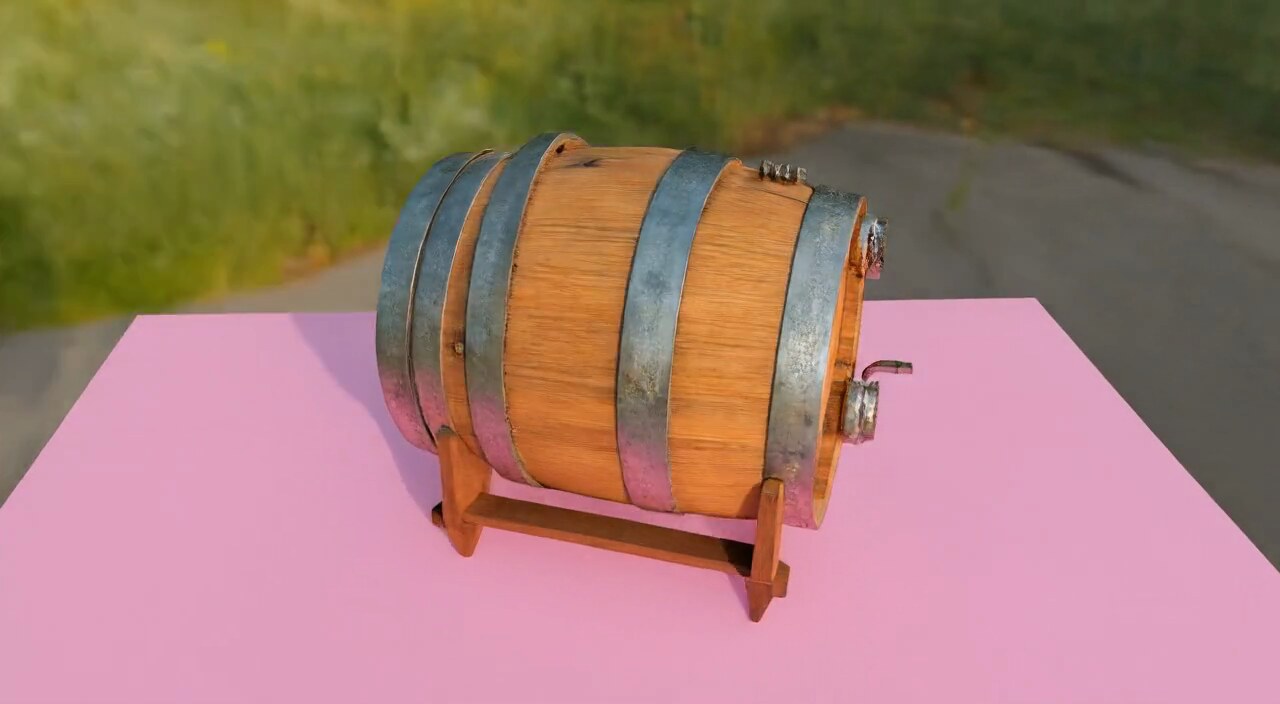} \\
\end{tabular}
\caption{By manipulating the HDR environment map, our model produces continuous and physically consistent lighting variations. We show the diffuse, glossy, and rough GGX components of the scene proxy (top three rows) and the corresponding synthesized outputs (bottom). Lighting changes are reflected in shading and reflections while geometry and materials remain stable.}
\label{fig:controlability}

\end{figure*}

\subsection{Qualitative Evaluation}

We present a visual comparison of our method with the baseline approaches in \cref{fig:qualitative_comparison}. As shown, our method achieves more realistic lighting effects as well as more precise layout and camera control, resulting in outputs that most closely match the reference video (Ref).

\begin{figure}[tb] 
\def\lw{0.25\linewidth}
\def\lh{0.125\linewidth}
\def\ftsz{\normalsize}
\renewcommand\tabcolsep{0.0pt}
\renewcommand{\arraystretch}{0}
\centering \small
\begin{tabular}{cccc}
\includegraphics[width=\lw]{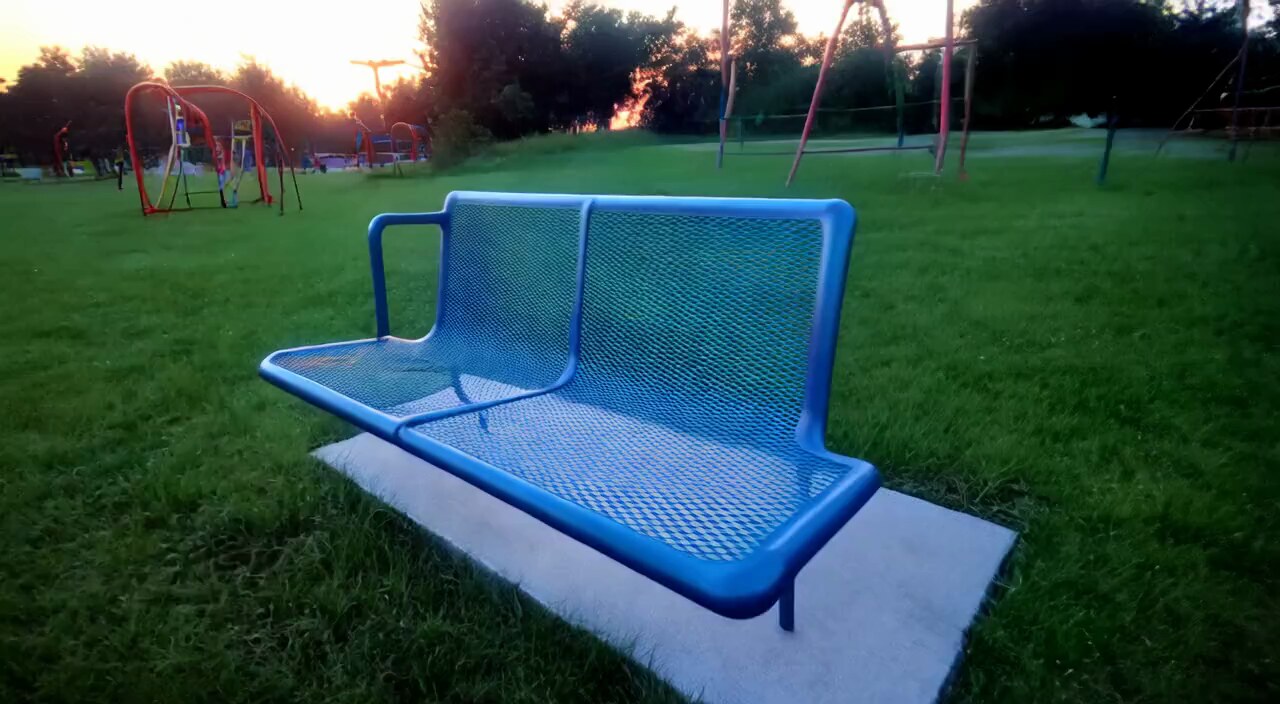} &
\includegraphics[width=\lw]{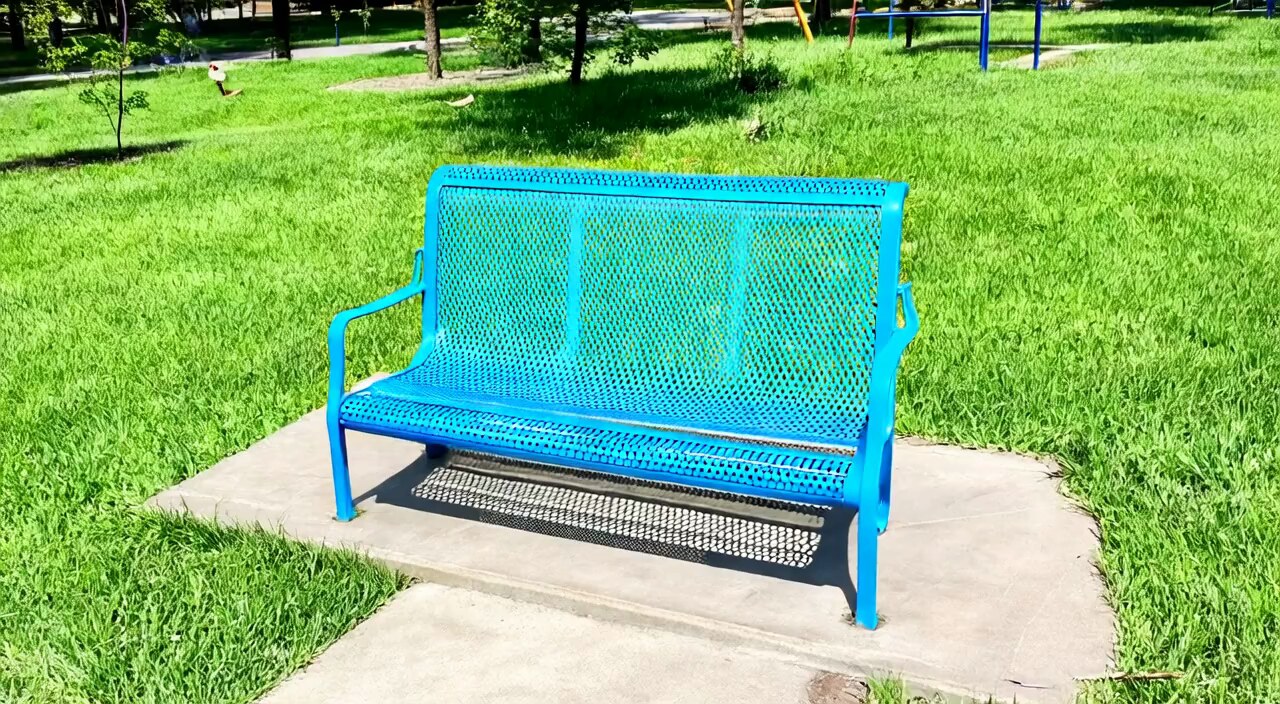} &
\includegraphics[width=\lw]{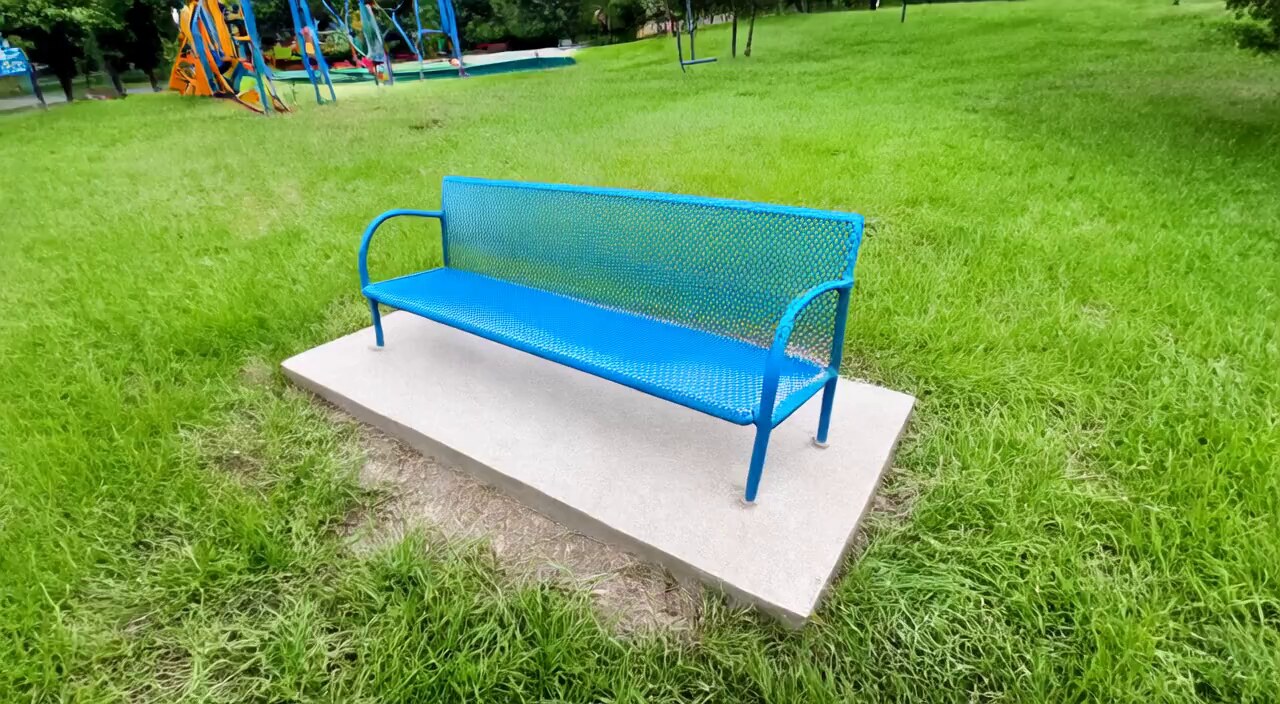} &
\includegraphics[width=\lw]{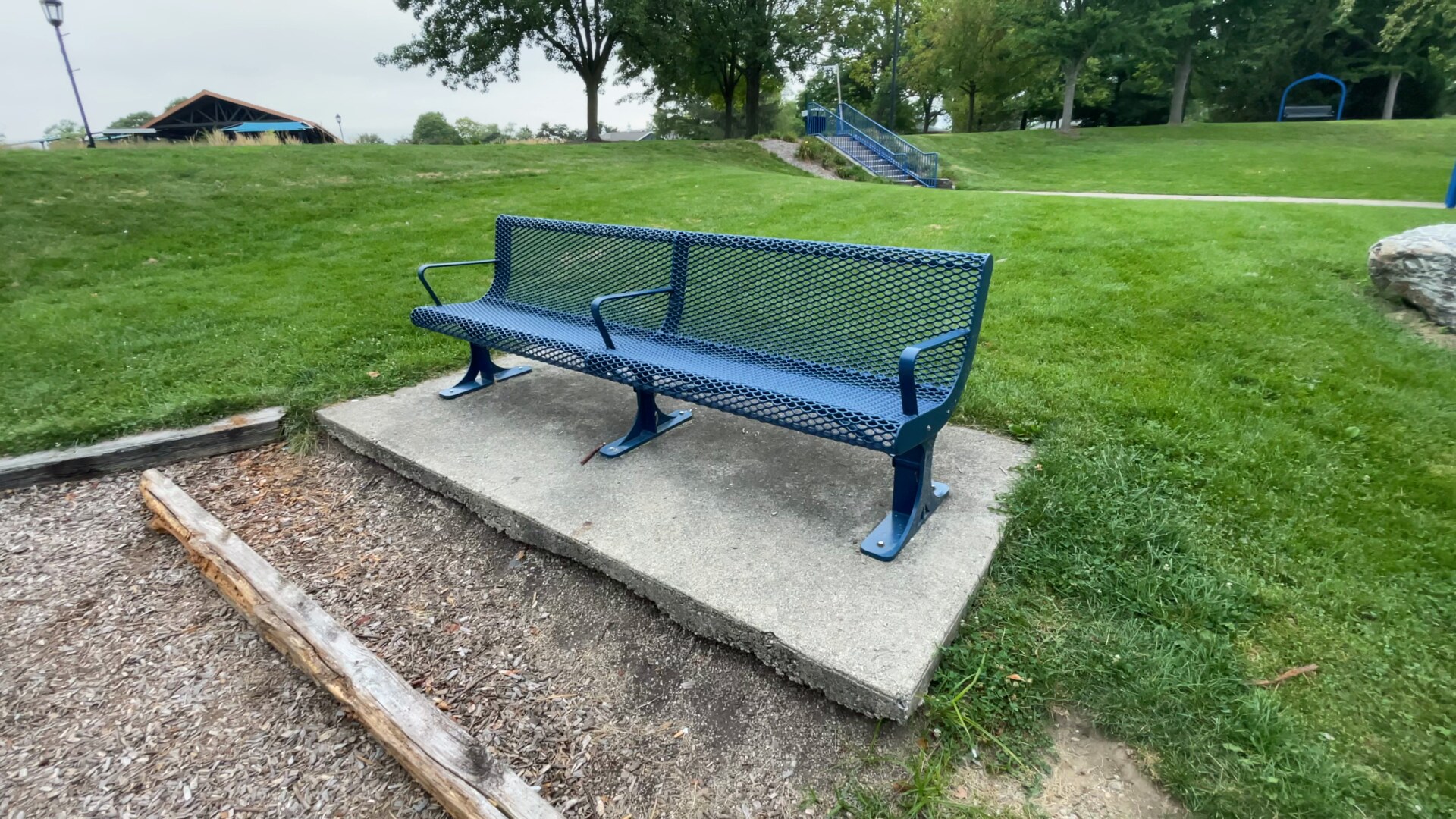} \\

\includegraphics[width=\lw]{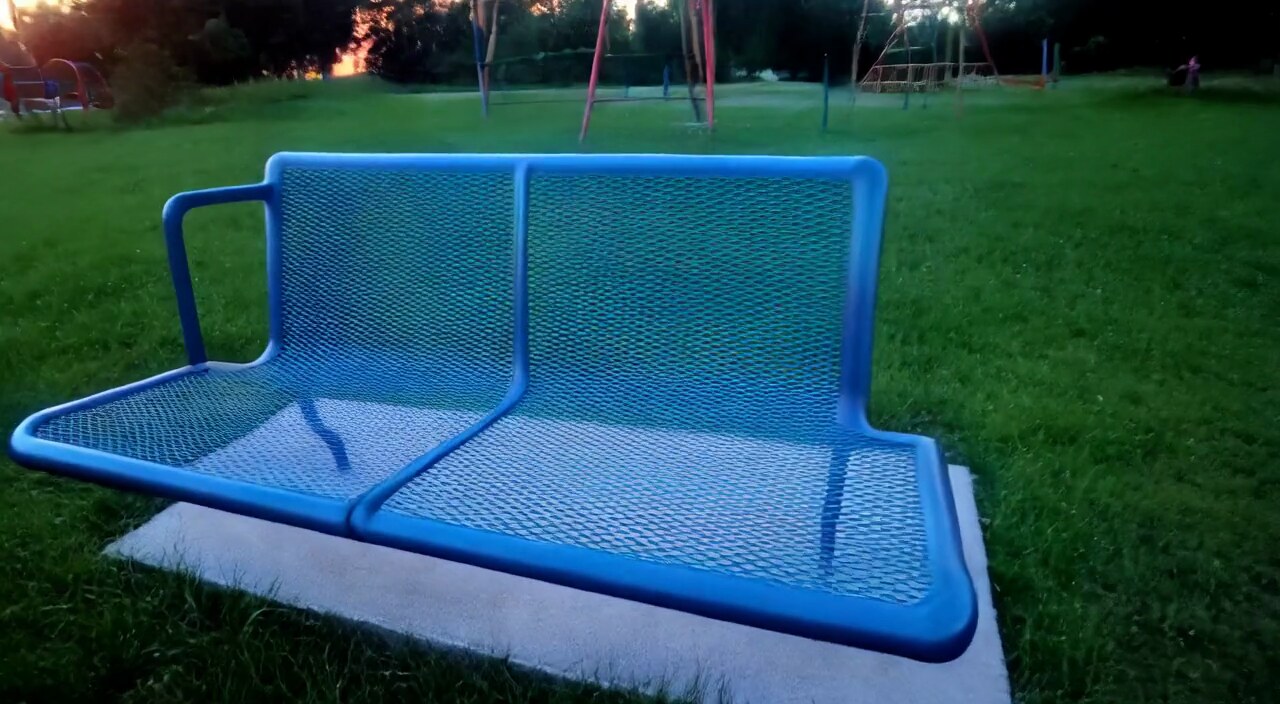} &
\includegraphics[width=\lw]{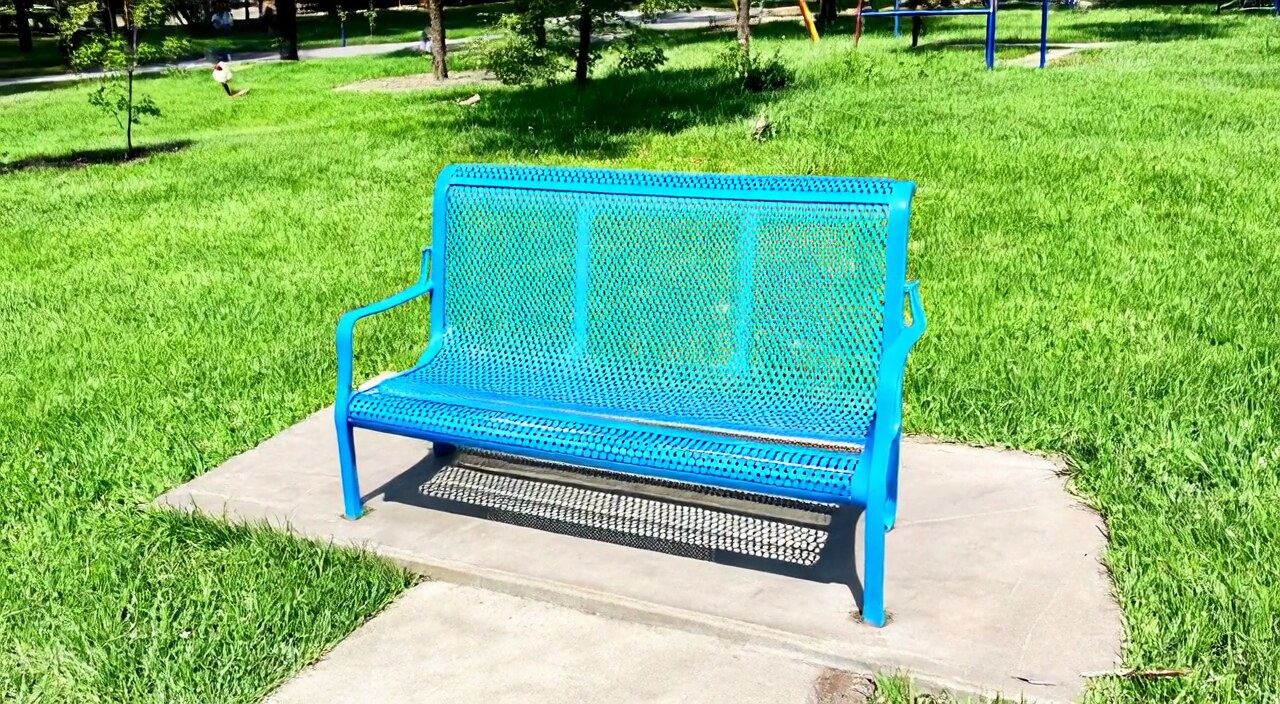} &
\includegraphics[width=\lw]{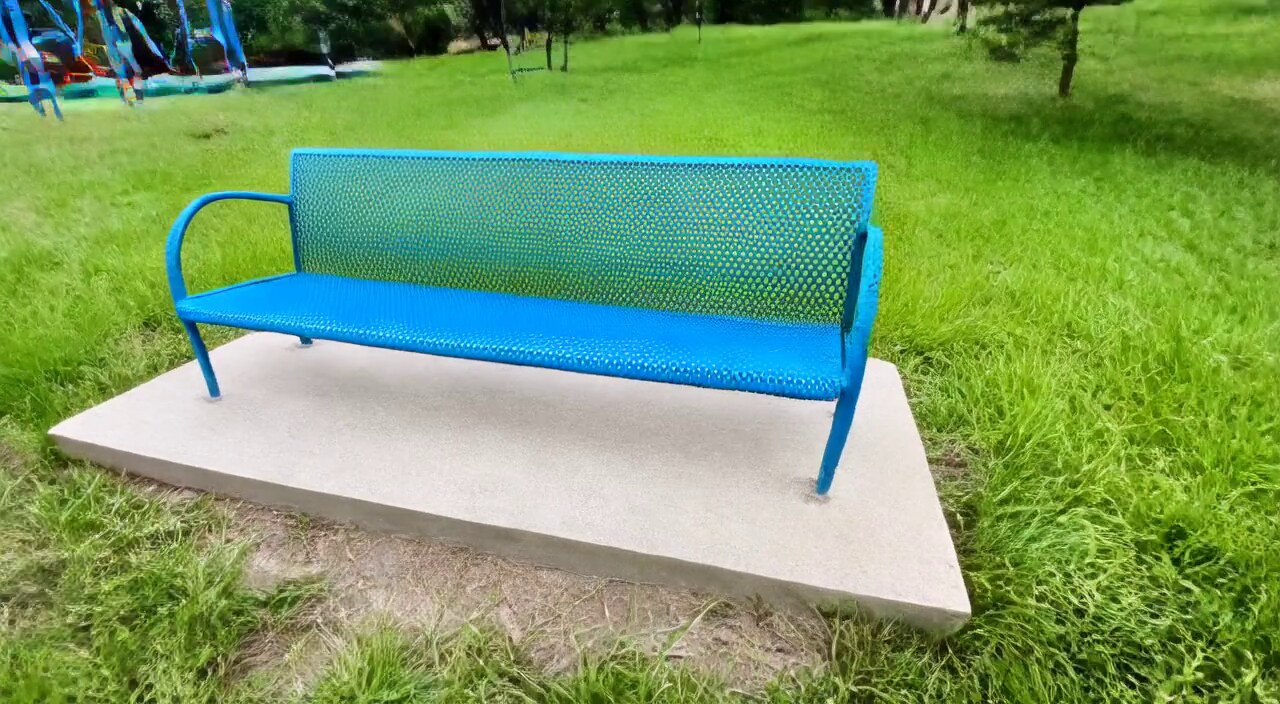} &
\includegraphics[width=\lw]{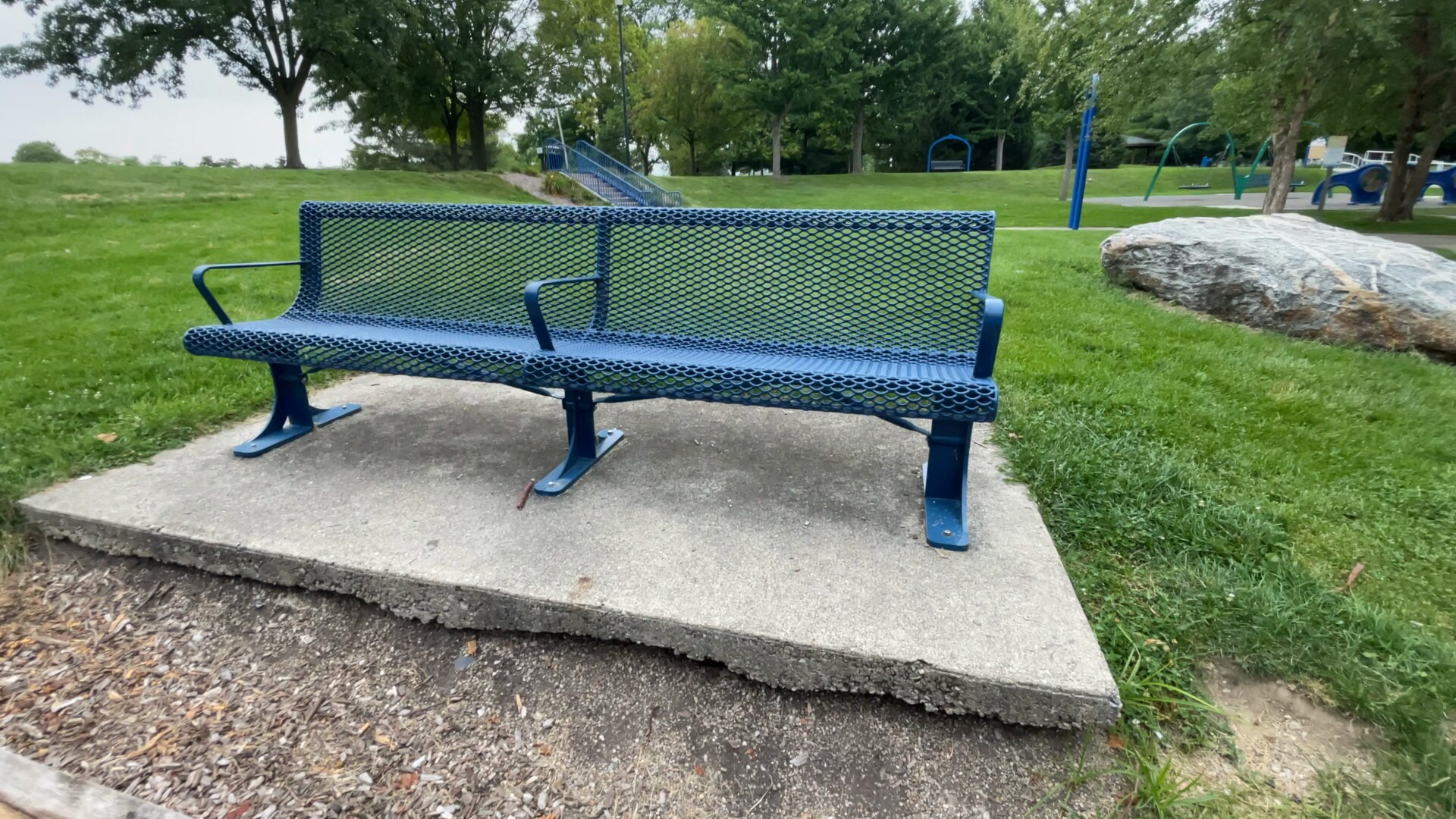} \\

\includegraphics[width=\lw]{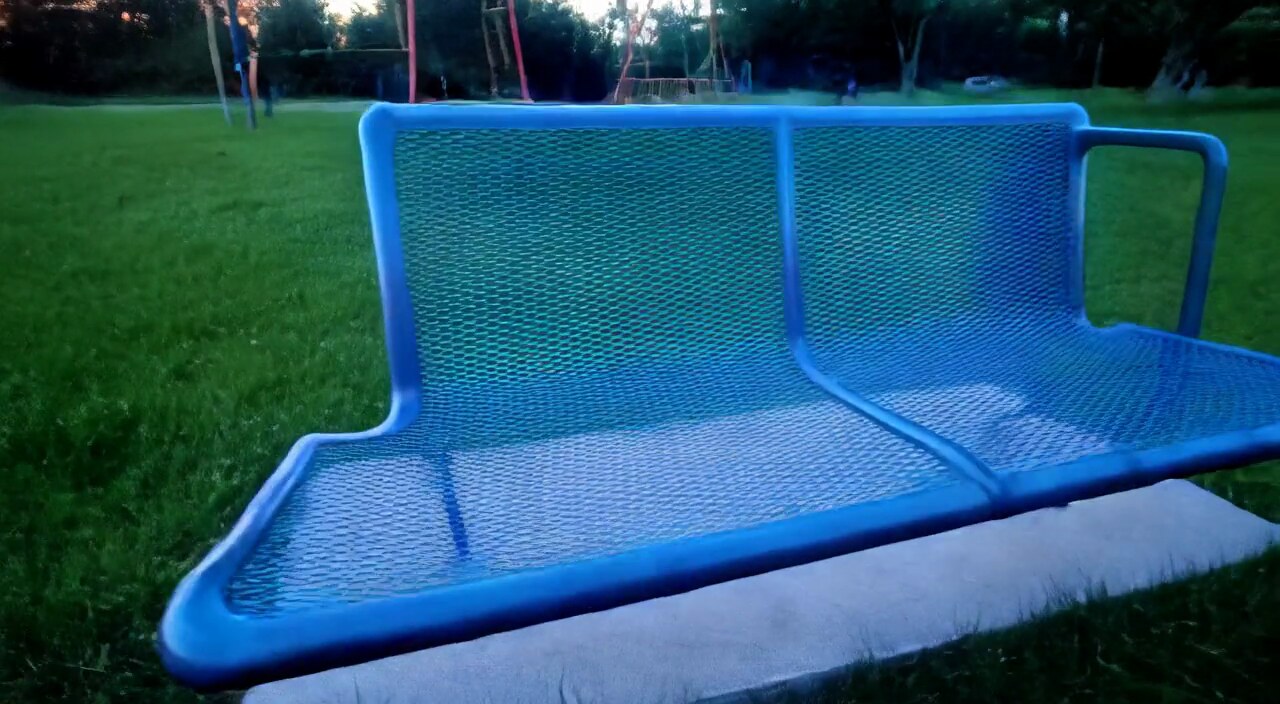} &
\includegraphics[width=\lw]{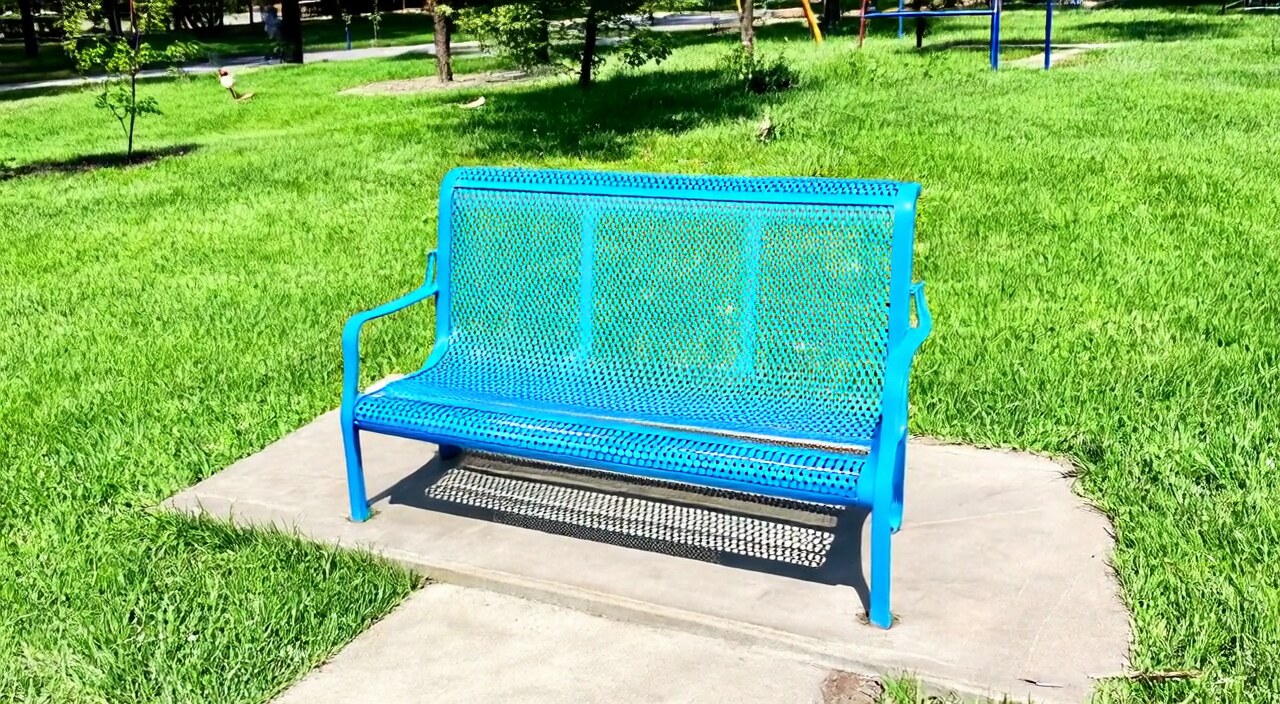} &
\includegraphics[width=\lw]{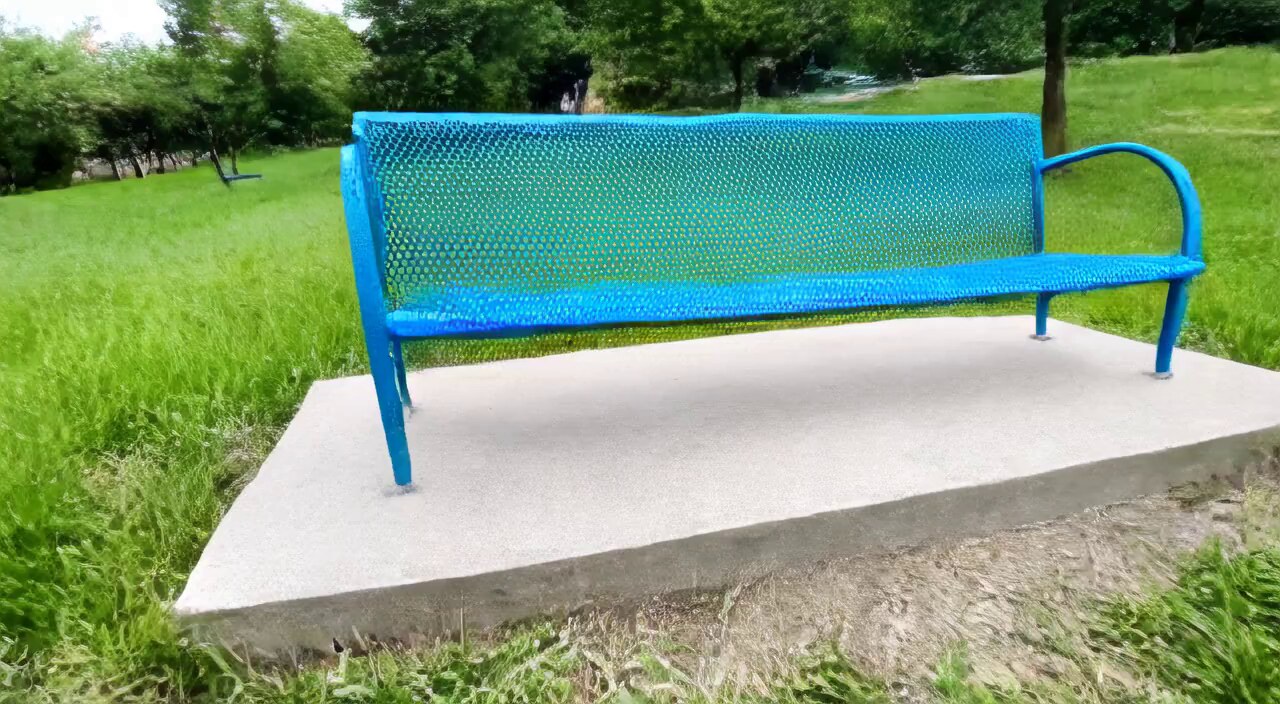} &
\includegraphics[width=\lw]{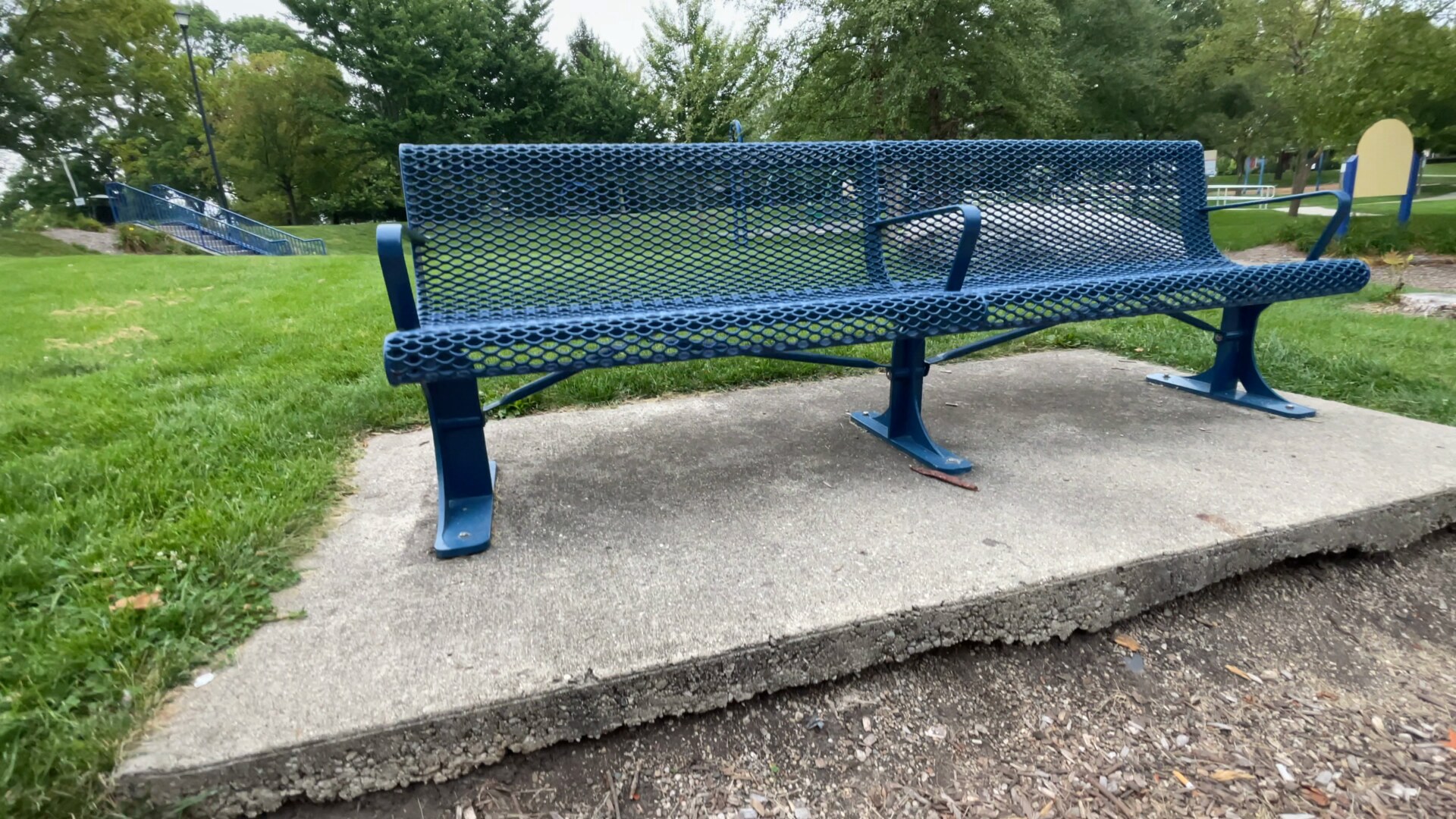} \\

\includegraphics[width=\lw]{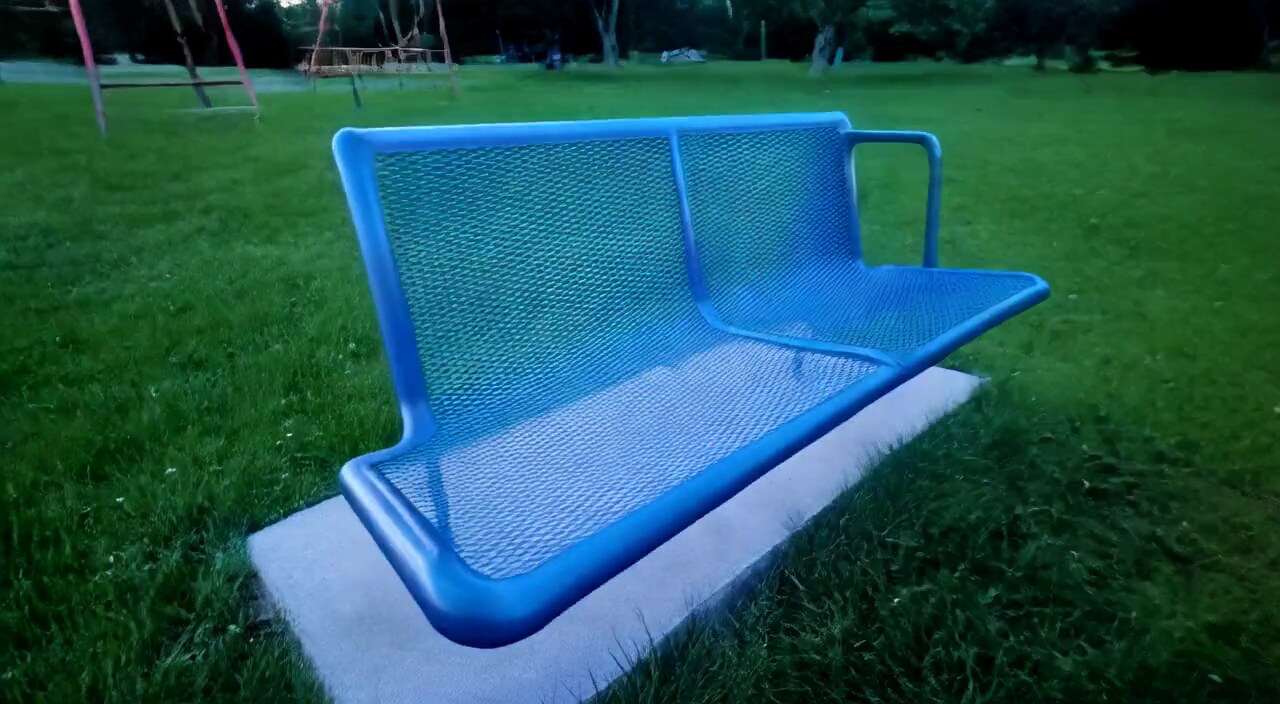} &
\includegraphics[width=\lw]{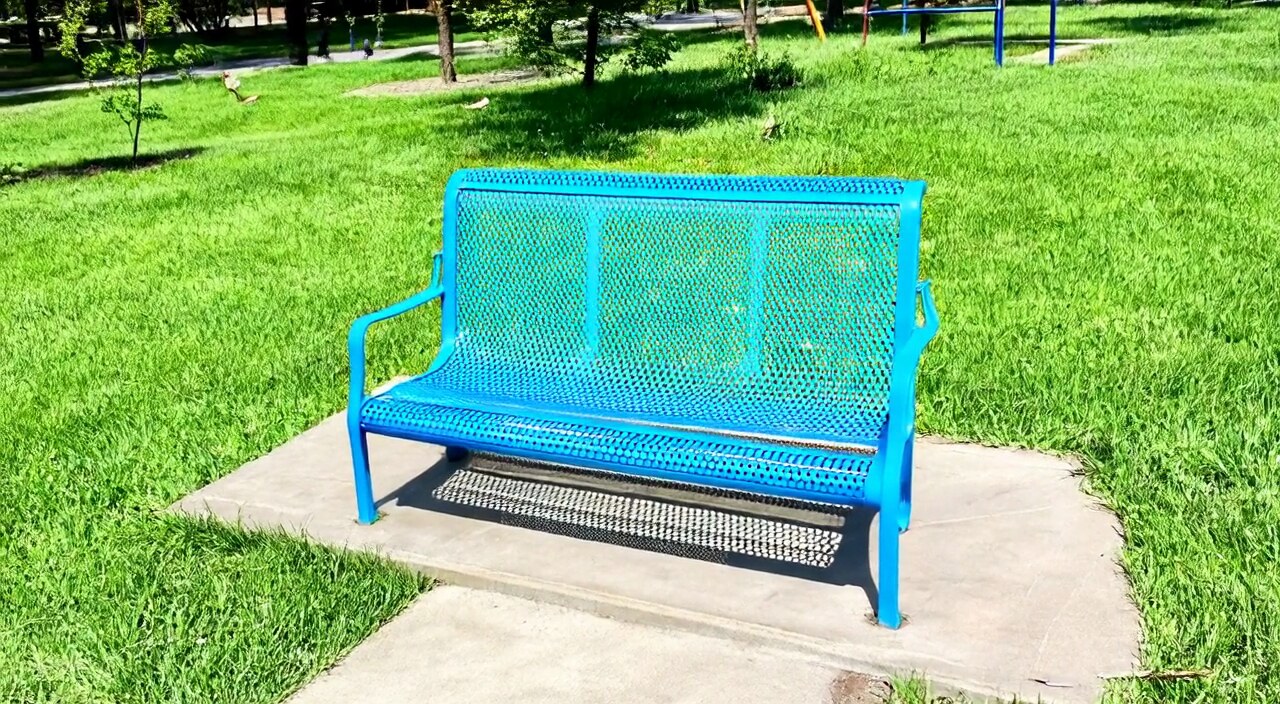} &
\includegraphics[width=\lw]{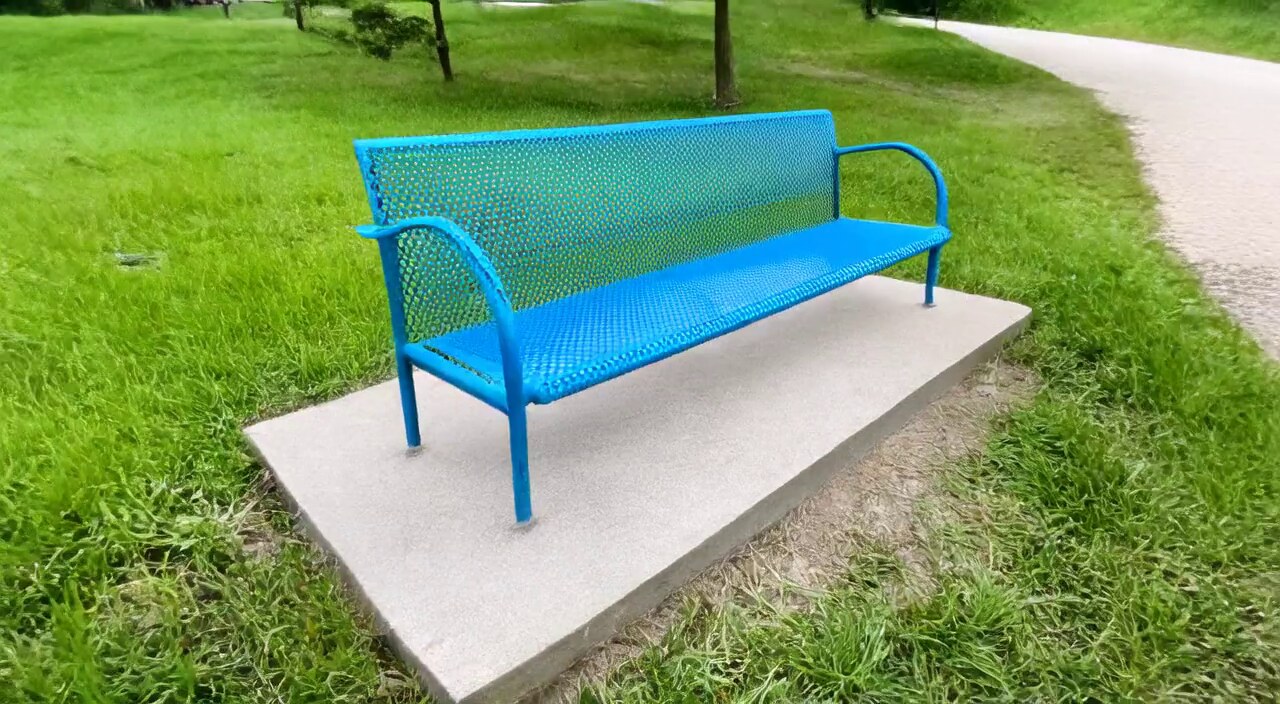} &
\includegraphics[width=\lw]{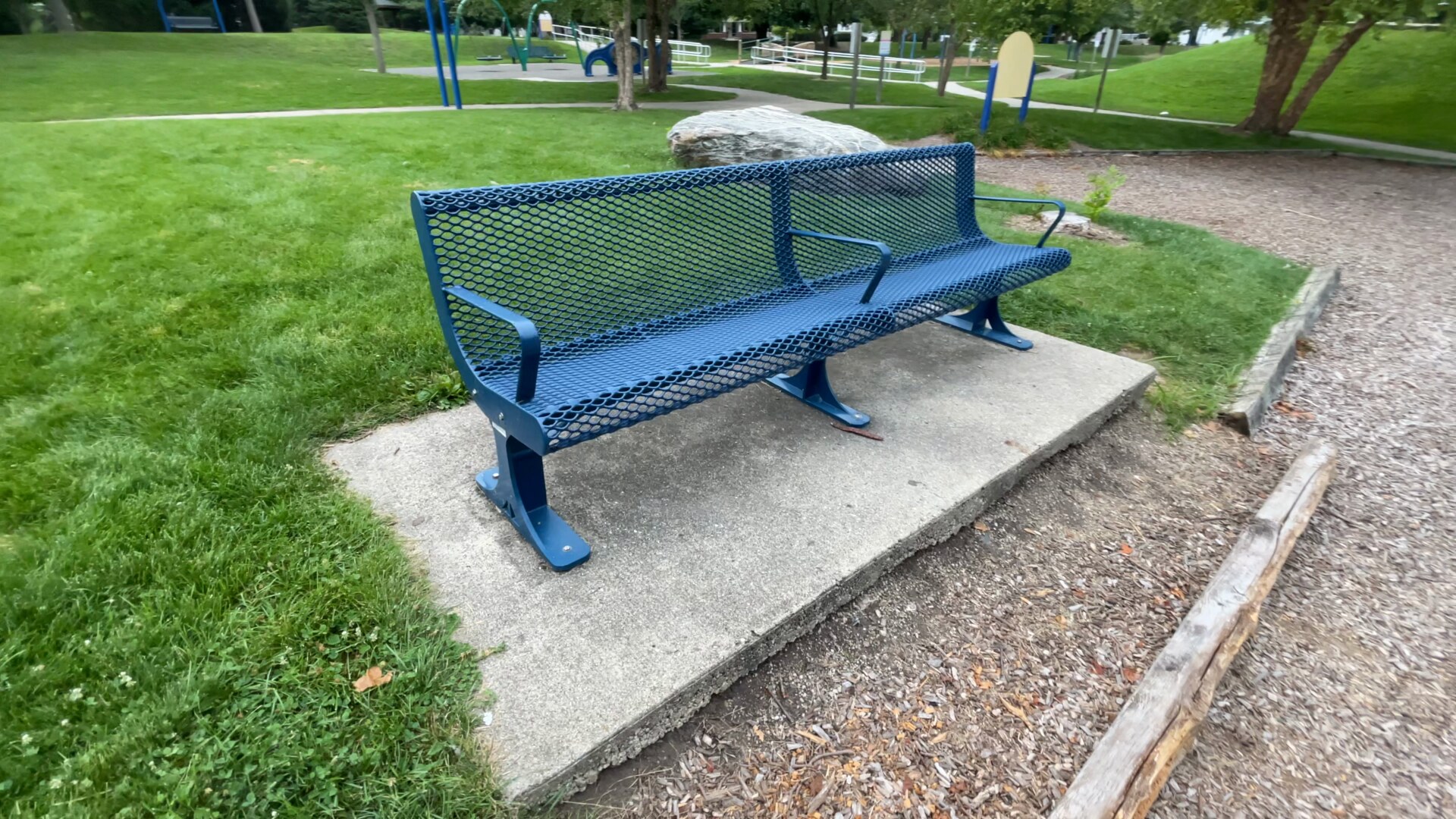} \\[2pt]
\footnotesize{w/o SynData} & \footnotesize{w/o staged} & \footnotesize{Ours} & \footnotesize{GT}
\end{tabular}
\caption{Qualitative results of our ablation study.}
\label{fig:ablation}
\end{figure}

\subsection{Ablation Study}

\paragraph{Impact of Synthetic Data.}
To validate the importance of our synthetic dataset, we trained a variant of our model exclusively on real-world videos. As shown in \cref{fig:ablation} (first column), this model produces wrong and mostly even illumination, failing to reproduce correct lighting effects. This result confirms that the diverse and dynamic illumination in our LiVER-Syn data is crucial for enabling lighting control and preventing the model from overfitting to the less varied lighting patterns typical of real-world footage.
  
\paragraph{Importance of the Staged Training Scheme.}
We also evaluated our multi-stage training strategy by training a variant end-to-end on the full dataset from scratch. As visualized in \cref{fig:ablation} (second column), this joint training approach leads to nearly still output and a notable degradation in performance. The model's ability to precisely follow specified scene conditions is diminished. This suggests simultaneously learning control signals while adapting a large-scale pretrained model presents a more challenging optimization problem. Our staged approach, which progressively introduces the conditioning, proves essential for ensuring stable convergence and effectively integrating our control module without corrupting the generative priors.

\subsection{User Study}

To further assess the perceptual quality and consistency of our results, we conduct a user study involving 25 participants.
Each participant is shown 20 sets of videos, where each set contains results from four competing methods applied to the same underlying scene. 
For every set, participants are asked to select, for each evaluation dimension, 
the method they prefer the most: \textit{video quality (VQ)}, \textit{scene control (SC)}, \textit{camera trajectory control (CC)}, 
and \textit{lighting control (LC)}. 
We count, for each method and each dimension, the percentage of samples in which it is chosen as the preferred solution.
As summarized in Table~\ref{tab:user-study}, our method outperforms across all four dimensions.

\begin{table}[t]
\centering
\caption{Percentage of samples in which each method is selected as the most preferred solution.}
\label{tab:user-study}
\setlength\tabcolsep{16pt}
\resizebox{\linewidth}{!}{
\begin{tabular}{lcccc}
\toprule
Method & VQ $\uparrow$ & SC $\uparrow$ & CC $\uparrow$ & LC $\uparrow$ \\
\midrule
CameraCtrl~\cite{he2025cameractrl} & 4.3\% & 4.1\% & 2.2\% & 6.0\% \\
MotionCtrl~\cite{motionctrl} & 3.6\% & 3.6\% & 1.6\% & 5.7\% \\
VideoFrom3D~\cite{kim2025videofrom3d} & 8.7\% & 8.9\% & 24.1\% & 29.0\% \\
Ours & \textbf{83.4\%} & \textbf{83.3\%} & \textbf{72.1\%} & \textbf{59.3\%} \\
\bottomrule
\end{tabular}
}
\end{table}

\subsection{Controllability}

\paragraph{Lighting Control.} Our method provides fine-grained control over illumination. Shown in \cref{fig:controlability}, by manipulating the HDR environment map, our model generates videos with dynamic and continuous lighting changes, such as rotating lighting. This is achieved while maintaining the consistency of the scene's geometry and material properties, showcasing a good disentanglement of lighting from scene structure.

\paragraph{Layout and Camera Control.} The use of an explicit 3D scene proxy as conditioning ensures geometrically precise control over both object layout and camera motion. As qualitatively and quantitatively validated in \cref{fig:qualitative_comparison} and \cref{tab:quantitative_results}, our method exhibits superior performance in accurately placing objects and following user-defined camera trajectories compared to methods relying on 2D-based proxies.

\paragraph{Flexible Editing Workflow.} A key design of our method is to empower users with a flexible and intuitive workflow. We introduce a renderer-based agent that automatically generates an initial 3D proxy, simplifying the creation of conditioning signals. Crucially, this proxy is not a fixed input but a fully editable starting point. Users can import it into standard 3D software to perform traditional edits: they can add, delete, or move geometry, refine the lighting conditions, and design new camera trajectory. This hybrid approach uniquely combines automated scene setup with the creative freedom of established CGI pipelines.

\section{Conclusion}

We present LiVER, a novel diffusion-based framework for controllable video generation that jointly models layout, lighting, and camera. Unlike prior work that focuses solely on text prompts or global scene cues, LiVER offers explicit, fine-grained control over the spatiotemporal composition of generated content. Our approach bridges the gap between generative quality and scene-level controllability, paving the way for practical deployment in creative media, virtual cinematography, and immersive content production.

\paragraph{Limitations.} As our initial 3D reconstruction of the scene geometry is coarse, the model relies on the text description to synthesize fine-grained geometric and material details. This makes the final output quality (\eg, geometric consistency) sensitive to the user-provided prompts. We will explore improving the agent's scene interpretation through more sophisticated prompt engineering in our future work to mitigate this issue.

\vspace{-1mm}
\section*{Acknowledgment}
This work is supported by National Natural Science Foundation of China (Grant No. 62136001) and Beijing Major Science and Technology Project (Grant No. Z251100008125009). PKU-affiliated authors thank \url{openbayes.com} for providing computing resources.

{
    \small
    \bibliographystyle{ieeenat_fullname}
    \bibliography{main}
}

\end{document}